# A Probabilistic Approach in Historical Linguistics

# Word Order Change in Infinitival Clauses: from Latin to Old French

Olga B. Scrivner

Submitted to the faculty of the University Graduate School
in partial fulfillment of the requirements
for the degree
Doctor of Philosophy
in the Department of French & Italian
and the Department of Linguistics
Indiana University

June 2015

Accepted by the Graduate Faculty, Indiana University, in partial fulfillment of the requirements for the degree of Doctor of Philosophy

Doctoral Commitee

Committee Co-Chair, Barbara Vance, PhD

Committee Co-Chair, Sandra Kübler, PhD

Julie Auger, PhD

Markus Dickinson, PhD

Marco Passarotti, PhD

May 6, 2015

Olga Scrivner

A Probabilistic Approach in Historical Linguistics
Word Order Change in Infinitival Clauses: from Latin to Old French

This thesis investigates word order change in infinitival clauses from Object-Verb (OV) to Verb-Object (VO) in the history of Latin and Old French. By applying a variationist approach, I examine a synchronic word order variation in each stage of language change, from which I infer the character, periodization and constraints of diachronic variation. I also show that in discourse-configurational languages, such as Latin and Early Old French, it is possible to identify pragmatically neutral contexts by using information structure annotation. I further argue that by mapping pragmatic categories into a syntactic structure, we can detect how word order change unfolds. For this investigation, the data are extracted from annotated corpora spanning several centuries of Latin and Old French and from additional resources created by using computational linguistic methods. The data are then further codified for various pragmatic, semantic, syntactic and sociolinguistic factors. This study also evaluates previous factors proposed to account for word order alternation and change. I show how information structure and syntactic constraints change over time and propose a method that allows researchers to differentiate a stable word order alternation from alternation indicating a change. Finally, I present a three-stage probabilistic model of word order change, which also conforms to traditional language change patterns.

Committee Co-Chair, Barbara Vance, PhD

Committee Co-Chair, Sandra Kübler, PhD

Julie Auger, PhD

Markus Dickinson, PhD

Marco Passarotti, PhD

## Table of Contents

## List of Tables

## List of Figures

# Chapter 1

# Introduction

*Pour reüssir en la recherche des origines de nostre Langue,*
*il faudroit avoir une parfaite connoissance de la Langue Latine*
*dont elle est venuë, et particulierement de la basse latinité*[1]
*(Ménage, 1650, 526)*

Sapir (1921) once said: 'Everyone knows that language is variable'. Studying language variation is a cornerstone of variationist linguistics, or sociolinguistics. In this approach, linguistic variation is an inherent part of any language system, showing a meaningful correlation with other components of languages, e.g., social, stylistic, syntactic, pragmatic. Being related to social science and sociology, the sociolinguistic field adopts their quantitative methods to measure variation probability in given contexts as well as the significance of these contexts. Variation may remain stable all the time or it may lead to a language change, when some variants become more frequent and entirely replace the other variants. This change, however, does not happen in one week. Language change is a long-lasting process that may span centuries. The study of language change is a nucleus of historical linguistics. Not long ago, historical linguistics adhered strictly to a categorical approach, not allowing for gradience and heterogeneity in language evolution. Sociolinguistic studies have, however, offered a new way of looking at diachronic change. They have shown that language change and language variation are just two dimensions of variation, namely synchronic and diachronic, and that synchronic variation provides a key to understanding language change (Labov, 1966; Milroy, 1992).

[1]Translation: In order to shed light on the origin of our language, we would need to have a great understanding of Latin, its ancestor, and in particular, the period of Late Latin.

As a branch of historical linguistics, historical sociolinguistics mainly focuses on written records from different chronological periods. In the past, such research was carried out manually, by identifying linguistic phenomena through the investigation of a given text. With the advent of corpus linguistics, researchers have gained access to syntactically annotated historical corpora, which provide an effective alternative method for collecting data. The combination of quantitative and corpus linguistic tools not only offers opportunity for the extensive quantitative data mining of historical documents, but also allows for data visualization methods, which make it possible to observe patterns otherwise hidden from a researcher using traditional methods.

## 1.1 Word Order Variation and Change

The main focus of this thesis is word order variation and change in Latin and Old French. It has been traditionally assumed that there is a gradual shift from *Object-Verb* (OV) in Latin to *Verb-Object* (VO) in French. The first attestation of VO order goes as far back as archaic Latin (around 200 BC) (Panhuis, 1984; Adams, 1976; Bauer, 1995). While this change has been studied extensively, the timing of the OV/VO shift still remains controversial: some scholars insist that Latin is an OV language throughout (Linde, 1923; Vincent, 1976; Watkins, 1964; Spevak, 2005); others argue that Late Latin is already a VO language (Adams, 1977) and that the actuation of the shift from OV to VO had occurred before the period of Plautus (around 254-184 BC) (Adams, 1976). Nevertheless, the evidence for OV/VO variation is striking, even among Classical authors. The examples in (1) illustrate the co-occurrence of OV and VO order in the work of the same Classical author - Livy.[2] Such variation is traditionally attributed to the author's need to express various stylistic and discourse nuances.

[2] The examples are taken from Devine and Stephens (2006, 126).

(1) a.

| | | | |
|---|---|---|---|
| loco | edito | **castra** | posuit |
| place-abl | elevated-abl | camp-acc.obj | pitched-3p.sg |

‘He pitched camp on high ground’ (Livy 25.13.4)

b.

| | | | | | |
|---|---|---|---|---|---|
| Inter | Neapolim | et | Tycham... | posuit | **castra** |
| between | Neapolis | and | Tycha... | pitched-3p.sg | camp-acc.obj |

‘He pitched camp between Neapolis and Tycha’ (Livy 25.25.5)

Verb-medial order has often been claimed to constitute a transitional stage in the passage from OV to VO. This stage is described as a verb second order (V2), in which the initial position is reserved for a pragmatically salient element. It has been traditionally argued that Old French is a V2 language (Foulet, 1928). Some also argue that Old French displays characteristics of an OV language (Zaring, 2011; Mathieu, 2009). In addition, there is no consensus on the periodization of word order change. For example, some suggest that VO order is fixed by the 13th century (Marchello-Nizia, 2007; Rouquier and Marchello-Nizia, 2012). Nonetheless, evidence of word order variation is abundant, and is often ascribed to various contextual factors, such as sentence type, object length and emphasis, illustrated here by example (2), extracted from the *Roland* poem written in the 11th century.[3] In this example, both VO order (*ai ost*) and OV order (*bataille dunne*) co-occur in the same sentence.

(2)

| | | | | | | | |
|---|---|---|---|---|---|---|---|
| Je | nen | ai | **ost** | qui | **bataille** | li | dunne |
| I | not | have-1p.sg | army-acc.obj | who | battle-acc.obj | him-dat | give-3p.sg |

‘I don’t have an army that can combat him’ (Roland, v.18)

In recent years, there has been increasing interest in statistical variationist approaches to the word order phenomenon, as it has become apparent that word order depends on various conditions, internal and external. For example, in (2) VO order occurs in a main clause, whereas OV is found in a subordinate clause. In addition, an increasing number of large annotated historical corpora, state-of-the-art methods of corpus linguistics for data retrieval and advances in statistical tools have made the investigation of word order variation more accessible and comprehensive. With access to larger datasets, historical sociolinguistics is now confronted with a different issue. The contemporary researcher may not always be aware of certain sociolinguistic or stylistic nuances of a given period, which constitute synchronic variations. As a result, it is often challenging to tease apart whether a word order pattern manifests word order variation or signals word order change. This type of investigation requires some mechanisms for detecting a neutral word order. This neutral word order represents the most common, pragmatically unmarked word order, from which we can infer the basic word order of a given period. The following examples illustrate two word order patterns in English. If, hypothetically, we did not know that the basic word order in English is Subject-Verb-Object, we would deduce that word order in (3a) is not a neutral order, based on the expression *as for*, which indicates a topicalisation, or emphasis on *beans*. On the other hand, the example in (3b) provides no explicit discourse clue. However, before jumping to the conclusion that the example in (3b) reflects a basic word order, we need first to consider other contextual factors that may also influence this order, for example, its frequency, heaviness (length), type of clause, genre, etc.

[3]The example is extracted from the MCVF corpus (Martineau et al., 2007).

(3) a. As for beans, I like them.
    b. I like beans.

Finally, not having access to native speakers and with limited historical records, it is more appropriate to conduct a probabilistic analysis for this investigation and present results as a probability or likelihood of the event, as compared to making categorical statements.

## 1.2 The Aim of this Study

The goal of this thesis is to investigate word order change from a comparative variationist perspective by incorporating data from both Latin and Old French. Such a comparative approach has been shown to be fruitful in historical linguistics. For example, Benincà (2004) incorporates evidence from several Medieval Romance languages in order to build a complex structure of left functional periphery, whereas Cruschina (2011) uses some modern Romance varieties to analyze data from Medieval Romance. Polo (2005) compares word order patterns from the Classical Latin *Satyricon* with patterns from the modern Italian translation, and Marchello-Nizia (1995) and Zaring (2010) compare word order patterns between different periods of Old French. In contrast to previous studies that have looked at Latin and Old French separately, the present study examines corpus data on a continuous scale. I argue that the data drawn from various stages of language development, such as Classical Latin, Imperial Latin, Late Latin, Early Old French and Old French, can help us construct a more representative image of the actuation and diffusion of word order change, enhancing our understanding of language change.

The methodological aim of this thesis concerns both data collection and data analysis. My first goal is to propose a novel approach by merging computational linguistics and corpus linguistics methods. Although several annotated corpora of Latin and Old French have become available in recent years, they are restricted to a limited number of historical records and periods. Such chronological lacunae in these corpora present challenges for diachronic studies, as they hinder how the change unfolded. Computational linguistics, on the other hand, offers state-of-the-art methods for automatic annotation. This study shows that the methods traditionally used for annotating modern languages can also be applied to Latin and Old French, thus providing additional annotated resources.

My second goal is to incorporate advances in statistical analyses and probabilistic approaches to language change. Most previous approaches to word order change are based on a simple frequency analysis. This type of analysis is subject to great variation, and outliers resulting from the unbalanced nature of historical data, e.g., small size of corpus, sparse evidence. In addition, each factor is traditionally viewed in isolation, failing to encompass all external and internal factors. In contrast, the present analysis is based on a probabilistic approach, that is, the results show the likelihood of a given word order pattern's association with a given period and in a given context. This view can unveil some aspects of language variation and change that may not be visible through traditional approaches. While most diachronic studies still rely on the examination of patterns in a single context, e.g. in main or subordinate clauses, this approach allows for the examination of word order in conjunction with all available contextual factors, e.g., pragmatic, semantic, syntactic.

My third goal is to demonstrate that information structure and infinitival clauses are, in fact, useful methodological devices for tracing the actuation and diffusion of VO order. First, it is well known that in both languages under consideration, pragmatic functions such as focalization and topicalization trigger deviations from a common word order. Thus, we need to focus on an *unmarked* context, namely an order without any contrast, emphasis, etc., to be able to track the VO diffusion from Latin to Old French. Second, this study focuses exclusively on infinitival clauses with nominal direct objects. It has been shown that the structural ambiguity and complexity of word order patterns that obscure analysis can be overcome through the examination of non-finite clauses (Zaring, 2010, 2011; Kroch, 2001). Since infinitival clauses are traditionally viewed as reduced clausal structures, their reduced character minimizes possible ambiguity. Finally, word order in infinitival clauses and nominal objects remain relatively understudied, as compared to word order in main clauses and pronominal objects. Thus, the purpose of their examination is twofold: First, reduced clauses are used as methodological devices for obtaining a less ambiguous context for identifying a basic word order. Second, this study examines changes that occur in the infinitival structures over time.

These methodological approaches will enable this study to identify discourse and syntactic factors motivating word order variation as well as trace the spread of VO order from Classical Latin to Old French in pragmatically neutral contexts. The results will also show that OV and VO not only become equally frequent by Late Latin (4th-6th centuries), but that there is also a change in discourse structure: pragmatically unmarked nominal objects from the preverbal position are moved to the postverbal position. The equal rate of OV and VO orders continues from Late Latin to Old French until the 13th century, where VO order, finally, becomes dominant. Furthermore, this study will be able to trace three changes in infinitival clauses, indirectly related to word order change: i) the decline of *accusativus cum infinitivo*, a very common infinitival structure in Latin, allowing for a subject in the accusative case, mood and tense markers; ii) the rise of prepositional infinitives in Old French and iii) the restricted use of preposed infinitives in Old French. These three contexts, namely AcI, prepositional infinitives and preposed infinitives, favor OV order.

## 1.3 Establishing the Background

It is generally recognized that the passage from Latin to Romance languages involves fundamental changes in the areas of nominal group, verbal group and sentence. This thesis focuses on the third area, namely a sentence. In order to explain why OV or VO is chosen in any given sentence and any given historical period, two possible situations should be considered: i) Synchronic word order variation and ii) Diachronic word order change. In the first scenario, word order alternation represents a speaker's choice governed by specific constraints, e.g., stylistic, pragmatic or syntactic. In this case, the interplay between word order variation and these constraints remains constant across time. For example, NP heaviness triggers VO order regardless of period. In the second scenario, word order alternation reflects a language change. In this view, the interaction between word order and constraints changes its direction, e.g. prediction changes from OV to VO, or this interaction discontinues its influence. For instance, pragmatically neutral sentences in Latin are more likely to have OV order, whereas in Old French VO is considered a pragmatically neutral order, that is, the direction of prediction has changed from OV to VO. Not knowing which factors are stable and which reflect language change, it is essential to examine the interplay of various factors, namely syntactic, pragmatic, semantic and sociolinguistic. As Waltereit and Detges (2008, 14) point out, 'syntactic change is often caused by semantic and pragmatic reasons, but sometimes it is also motivated by genuinely syntactic factors'. In addition, there might be certain internal or external contexts that lead or favor syntactic change, that is, the diffusion of a new form begins more robustly in these contexts, in comparison with the others.

Finally, as mentioned earlier, there is no agreement on word order shift in Latin and Old French. As a consequence, this study starts with no prior assumption with respect to basic word order, namely OV or VO, for each language. In order to overcome this methodological limitation, this study proposes a novel method, combining i) basic word order criteria, ii) information structure and iii) infinitival clauses. These three components are described in the following sections.

### 1.3.1 Basic Word Order

In this thesis, the notion of word order refers to a surface linearization of various constituents in a sentence. For example, OV indicates a linear placement of the nominal object before the verb, and VO indicates a placement of the nominal object after the verb on a surface. As mentioned earlier, this study concentrates on word order alternation, namely alternation between OV and VO orders. It is traditionally assumed that word order variation is a common language characteristic, as speakers fulfill their particular communicative needs by way of word order variation (Hinterhölzl and Petrova, 2009, 1). We have seen in (3) that word order pattern can express certain pragmatic or stylistic nuances in the speech, e.g. emphasis, contrast, focus. Furthermore, it is widely recognized that all languages have a basic word order, namely 'a surface ordering of subject, object and verb relative to one another that is at least more common than other possible orders' (Steele, 1978, 587). And many would agree that the basic word order is, in fact, an unmarked order. Although the definition of basic word order remains a controversial empirical issue, use of the concept in this thesis is strictly methodological. That is, this order does not signal an underlying representation of the sentence. In this study, basic word order refers to a linear word order with unmarked prosodic and pragmatic patterns in a given language. In this view, basic word order is defined in terms of pragmatically neutral word order. This approach is especially useful when dealing with a discourse configurational language, e.g. Latin, in which a linear positioning of constituents is defined by pragmatic functions. Let us briefly review several methodological criteria that have been proposed to define basic word order. The very first systematic definition of basic word order came from typological studies, for which the main goal was to establish a universal word order typology, namely correlations between word order patterns and language characteristics. It was recognized that languages have a primary word order from which it is possible to infer language characteristics. One of the first criteria used to identify a primary word order was 'the relative order of subject, verb, and object in declarative sentences with nominal subject and object' (Greenberg, 1963, 76-77). The other typological criteria include the following characteristics: a) simple clause, b) least marked environment (least phonologically marked (Steele, 1978), least grammatically marked (Hawkins, 1983), least morphologically marked (Hawkins, 1983; Brody, 1984) and least pragmatically marked (Dryer, 1995)) and c) frequency (Hawkins, 1983). In the typological approach, main declarative clauses are thought of as a simple clause, as opposed to subordinate, interrogative or imperative clauses. In addition, the restriction to main clauses is considered a methodological device for ensuring consistency and eliminating variations. Second, the marked environment is determined by certain features, such as stress, emphasis, mood and specificity. For example, if a direct object is stressed, the word order pattern is more likely to represent a marked pattern. Finally, only the most frequent patterns are considered basic word order (Greenberg, 1963).

More recent approaches define basic word order as a given language's unmarked prosodic pattern (Hinterhölzl, 2010). In this view, an unmarked pattern is determined by information structure categories. These categories are described in the next section.

### 1.3.2 Information Structure and Syntax

The term *information structure* refers to the relations between a sentence, its surrounding environment or context, and speaker's or hearer's knowledge about what has been said. For example, the arrangement of words in a given sentence may convey which information is new and which is old for the speaker. It has been recognized that information structure is expressed through three main modules: semantics, prosody and syntax. Since diachronic studies lack prosodic information, the identification of pragmatic or discourse functions must rely only on syntax and discourse semantics. In a recent approach, information structure is based on a multi-layered model with three levels (Petrova and Solf, 2009).[4] Such a multi-layered schema allows for the examination of each layer independently as well as for the evaluation of their interplay with word order configuration (Petrova and Solf, 2009, 132-133). In what follows, I will briefly review the three layers of the information structure schema.

The first layer concerns the information status of linguistic constructions. The current approach recognizes a tripartite classification: 1) *new* - a referent that is introduced to the context for the first time (see (4a)), 2) *accessible* - a referent that has not been mentioned previously but is 'in a certain relation of relevance either to communicative situation as a whole or to entities already established in discourse', e.g., proper names, coreferential nouns or world knowledge objects (see (4b)) and 3) *given* - a referent that is explicitly mentioned earlier in the context (see (4c)) (Petrova and Solf, 2009, 145-146).

(4) a. In the garden we saw **Peter**.

b. **The sun** set.

c. **He** was happy.

This type of information, namely referential information, can usually be determined from the context. Several textual cues help us interpret this information, such as definite or indefinite marking (Lambrecht, 1994, 74-113). Using these clues and other contextual information, we can determine whether the information has been mentioned earlier or can be inferred from the context.

The second layer represents a predicational bipartite structure of the utterance: *topic* and *comment*. The identification of *topic-comment* structure incorporates a blend of various components: 1) givenness or accessibility, 2) aboutness (*A says about X that X ...*) (see (5)), 3) definiteness and 4) syntactic constructions of topicality, e.g. *as for* (see (3b)) (Petrova and Solf, 2009, 146-148).

(5) **Cats** are snooty.

[4] The tri-dimensional structure was first put forward by Molnár (1993) and continued in Krifka (2007a).

In the information structure annotation guidelines, Götze et al. (2007) suggest the following cues to help identify *topic*: 1) referential expressions, e.g., proper names and definite NPs, 2) indefinite expressions with generic or specific interpretations and indefinite NPs in adverbially quantified sentences and 3) bare plural with generic interpretation and bare plural in adverbially quantified sentences (Götze et al., 2007, 163-164).[5]

The last layer identifies informational relevance - *focus-background.* This layer determines which expression provides the most relevant information (Götze et al., 2007). The current information structure approach recognizes two contextual situations allowing for two mutually exclusive *focus* types (Petrova and Solf, 2009): 1) *new-information focus* - new information or information which carries the discourse forward (see (6a)) and 2) *contrastive focus* - information that is contrasted with other constituents in a given discourse (see (6b))(Götze et al., 2007, 171-172).

(6) a. She is reading **a book**.

b. We do not **export** but **import** goods.

It is widely accepted that information structure is mapped into syntactic structure. Consider our earlier example *As for beans, I like them* in (3a), where the author talks about beans, as compared to the example *I like beans* in (3b), where the author talks about his or her likes. In (3a), we know that the topic of the sentence is *beans* and that it is emphasized by means of the expression *as for.* However, if we try to move this expression to the end of the sentence *I like them, as for beans*, the sentence becomes ungrammatical. That is, there is a special position in the sentence in the left periphery that corresponds to emphasized topics. Furthermore, the empirical evidence for ordering of information structure elements in the left periphery of the sentence further demonstrates that it is possible to incorporate information structure into syntax, where special fields (syntactic projections) are assigned to constituents bearing information structure features, such as *Topic* and *Focus.* This approach, known as a cartographic approach (Rizzi, 1997), includes three traditional layers, following the generativist tradition: *VP*, *IP* and *CP*. The *VP* layer is the core level, on which semantic selection occurs. The *IP* level hosts inflection, while the third, leftmost layer is related to discourse information. The *CP* layer links the previous two layers into discourse. In the cartographic approach, the *CP* domain is further decomposed into discourse functional projections, where *TopP* is a Topic projection hosting topicalized constituents, *FocP* is a Focus projection hosting focalized elements, *ForceP* is related to the clause type and *FinP* concerns complementizers and modality (Rizzi, 1997). The *CP* layer is illustrated in (7):

(7) $[ForceP[TopP * [FocP[TopP * [FinP]]]]]$

[5]For annotation procedure and test for *topic*, see (Götze et al., 2007, 165).

In recent years, there has also been growing interest in the influence of information structure on word order change. While it is still not clear whether syntax triggers change in information structure or information structure causes changes in word order, their relationship is well established in historical syntax.

### 1.3.3 Infinitival Clauses

This study is concerned with constructions in which the verb form is not inflected. In order to integrate infinitival clauses into a word order study, it is necessary to discuss the characteristics of these clauses in Latin and Old French. It is usually assumed that Indo-European infinitives have their origin in action nouns that can bear two cases, dative or locative (Brugmann, 1888; Disterheft, 1981; Ross, 2005). These nouns undergo a long evolutionary transformation that involves

> "not only an expansion in the number of clause types that use infinitive as embedded predicate but also change in the structure of the infinitive clause itself (...), change in object case, and word order shift" (Disterheft, 1981, 3).

Vincent (1999) suggests that the infinitive first emerges as a nominal form, then acquires verbal properties in Latin and finally develops clausal properties in its passage into Romance. On the other hand, Schulte (2007) proposes that the Latin infinitive has already developed clausal characteristics, as it allows for tense, aspect, voice and subject. Nevertheless, there appears to be a general consensus on the reduced verbal status of infinitival clauses. As a consequence, few authors have studied OV/VO distributions in such clauses in Latin and Old Romance languages (Zaring, 2010, 2011; Poletto, 2014). Several cross-linguistic studies, however, show that the study of non-finite clauses not only contribute to OV/VO research, but also circumvent the structural ambiguity often found in main finite clauses (Pintzuk, 1996; Kroch and Taylor, 2000; Taylor and Pintzuk, 2012b).

#### 1.3.3.1 Latin Infinitival Clauses

One of the most common infinitive constructions in Classical Latin is *Accusativus cum Infinitivo* (henceforth AcI). The notable features of these infinitival clauses are the lack of the complementizer *that* and the presence of a subject bearing the accusative case (as compared to finite clauses, where subjects are marked with the nominative case). AcI clauses are commonly used to complement verbs of discourse in indirect speech, e.g., *dico* 'I say', and verbs of mental or physical perception, e.g., *puto* 'I think', *video* 'I see' (Morin and St-Amour, 1977; Bolkestein, 1999; Bartoněk, 2006). An example with a declarative verb *dicitur* 'it is said (that)' is illustrated in (8):

| | | | | | | |
|---|---|---|---|---|---|---|
| (8) | dicitur | eo | tempore | matrem | Pausaniae | vixisse |
| | said-3p.sg | same-abl | time-abl | mother-acc.obj | Pausanias-gen | live-inf.past |

It is said that the mother of Pausanias lived at the same time (Nepos 4.5.5 (Ferraresi and Goldbach, 2003))

AcI constructions can also be used independently without a governing verb. The independent constructions are often described as historical infinitives or infinitives of narration, as illustrated in (9), where two infinitives *arcessere* 'replace' and *providere* 'provide' express quick successions of events in the narration:

| | | | | | | |
|---|---|---|---|---|---|---|
| (9) | integros | pro | sauciis | arcessere, | omnia | providere |
| | fresh-acc.obj | for | wounded-abl | summon-inf | all-acc.obj | provide-inf |

'he summoned fresh troops to replace the wounded, had an eye to everything' (Sallust, *Catilina* 60.4)

As both subject and direct object are marked with the accusative case, non-finite clauses can be syntactically ambiguous. For example, in (10) both subject and object bear the accusative case, and the sentence is ambiguous between two readings: i) he is praising her or ii) she is praising him. Only through context is it possible to determine that *eam* 'her' is an object in this case.[6]

(10) Dicunt **eum** laudare **eam**
say-3p.pl him-acc praise-inf her-acc
'They say that **he** is praising **her**' (Cecchetto and Oniga, 2002)

Another distinctive feature of AcI constructions concerns morphological tense and voice markers, e.g. *amaviss-e* 'to have loved' and *amar-i* 'to be loved'. For example, in (11a) the infinitival verb is in the present tense and in (11b) the verb is in the past tense. Person and number specifications, however, are not attested. The morphological paradigm is presented in Table 1.1 (Ferraresi and Goldbach, 2003, 242).

(11) a. Ei dicunt me venire
he-dat say-3p.pl me-acc.sbj come-inf
'They tell him that I am coming' (Cecchetto and Oniga, 2002)

b. Dico te venisse
say-1p.sg you-acc.subj come-inf.past
'I say that you have come'

Table 1.1: Morphological Paradigm for the Latin Infinitive **amare** 'to love'

| | Active | Passive |
|---|---|---|
| Present | amar-e | amar-i |
| Perfect | amaviss-e | amatum esse |
| Future | amaturum esse | amatum iri |

In addition to tense and voice, AcI clauses are able to express mood, aspect and modality, albeit only periphrastically (Ferraresi and Goldbach, 2003). For example, the subjunctive mood (the irrealis modality) is expressed by a combination of the participle future active and the infinitive perfect active of the verb *esse* 'to be': *amaturus fuisse* 'to will love'.

[6]Herman (1989), however, disagrees and states that this double interpretation is a figment of grammarians' imaginations. He argues that Latin authors usually employ various contextual devices to aid in disambiguation.

From the 1st century AD onward there is a continuous but slow increase in the new competing construction after verbs of discourse and verbs of mental or physical perception, namely a finite *that*-clause, introduced by the complementizers *quod*, *quia* and *ut*. At the same time, the frequency of AcI constructions becomes less frequent and there is a noticeable weakening in the morphological paradigm, namely the disappearance of *irrealis* mood and *future passive* voice. Some argue that there is a relation between the decrease in AcI and the emergence of subordination (Perrochat, 1926; Herman, 1989); others suggest that the new *quod*-construction is a feature of a different register in Latin, a non-literary genre (Wirth-Poelchau, 1977, 22). On the other hand, the use of infinitival constructions seems to expand to other contexts. For example, in Classical Latin adverbial meaning is typically expressed by supine (12a) or gerund (12b) verb forms (Schulte, 2004, 2007; Wanner, 2001). However, in Late Latin, there is an emergence of prepositional infinitival constructions in the contexts, where gerund and supine are used.

(12) a. **Spectatum** veniunt, veniunt spectentur ut ipsae
watch-supine come-3p.pl, come-3p.pl watch-3p.pl that themselves
'They come in order to watch, and they come to be watched themselves' (Ovid, 1.99 (Schulte, 2007, 520))

b. Homo ad **agendum** est natus
man-nom.subj to act-gerund is born
'Man is born to act' (Schulte, 2007, 520)

Another type of infinitival clause includes infinitives that act as a complement of a main verb, as shown in (13). It is argued that these infinitival clauses do not have a lexical subject or a morphological tense (Cecchetto and Oniga, 2002).

(13) Rhenum transire decreverat
Rhenum-acc.obj cross-inf decide-3p.sg.past
'He decided to cross the Rhine' (Caesar, *De Bello Gallico* 4,17,1)

#### 1.3.3.2 Old French Infinitival Clauses

Among the various transformations that Old French undergoes in its transitional stage from Latin, several developments need to be highlighted. First, its case system is reduced to two cases: nominative and accusative. Second, there is a decrease in the morphological paradigm for infinitival clauses: only three morphological features are found, namely present active, present passive and perfect active (Goldbach, 2008), as compared to a richer morphological paradigm in Latin (see Table 1.1). Third, the infinitival complements become widespread, and AcI almost disappears (Bauer, 1999). While the infinitive still occurs in indirect speech acts, there is no evidence of AcI in assertive speech acts, which are common in Latin, e.g. *dicere* 'to say' and *confirmare* 'to confirm'; its use is now limited to directive speech acts, e.g. *comander* 'order'. In addition, in these constructions, the subject is usually lacking or is placed in the matrix clause as a direct or indirect object, as illustrated in (14):

(14) vos comant cest cheinse changier
you-obj order-1p.sg this shirt-obj change-inf
'I order you to change this shirt' (Erec 1617)

However, Bauer signals that 'the rule for subject identity - according to which the infinitive is automatically used in case of subject identity - did not yet exist' (Bauer, 1999, 77). For example, the common use of infinitives in Modern French with the same subject for main and infinitival clauses (15a) can be still rendered in Old French as a subjunctive finite complement (15b) or infinitival non-finite complement (15c) (Bauer, 1999, 77):

(15) a. je ne sai quoi faire
I not know-1p.sg what do-inf
'I do not know what to do'

b. ne sai que face
not know-1p.sg what do-1p.sg

c. ne sai que faire
not know-1p.sg what do-inf

In this thesis, I will focus on Latin and Old French infinitival constructions, which include AcI, prepositional clauses and infinitival complement clauses.

## 1.4 Corpora

In diachronic studies one relies on written records, from which the textual evidence is collected. In the past, the manual collection of textual evidence was the only available method to a historical linguist. From the middle of the 20th century we see the emergence of corpora, or digital text collections. In fact, the first electronic corpus was created in the 1940s for Medieval Latin by father Roberto Busa S. J. (*Index Thomisticus*). Such digital corpora offer an alternative method for data collection. In such corpora, reading has been replaced by a keyword search, e.g., word form, syntactic structure, making the process of data collection more effective.

The material for this study is taken from several syntactically annotated corpora, spanning from Classical Latin (i-ii) to Old French (iii-v): i) PROIEL;[7] ii) LDT (Bamman and Crane, 2011); iii) MCVF (Martineau et al., 2007), iv) Nouveau Corpus d'Amsterdam NCA (Stein, 2011) and v) Base de Français Médiéval BFM.[8] Further details on these corpora will be presented in section 5.1.1. Such corpora are valuable tools in diachronic word order studies, as they allow for the exhaustive search and retrieval of annotated data. Using query searches, the researcher is able to search for specific contexts, for example, main clauses or infinitival clauses. In addition, query tools allow researchers to extract obtained data and output it in a text format. Nevertheless, each corpus is limited to a certain chronological period and provides access to limited textual material. For example, Latin resources are represented by the following periods: 1st BC, 1st AD and 4th century, whereas Old French resources cover texts starting from the 12th century.[9] Such a substantial chronological gap between periods makes it difficult to draw generalizations about word order change.

Recent achievements in computational linguistics provide methods for the automatic annotation of additional textual material that exists in Latin and Early Old French. Computational linguistics is a field that is concerned with language modeling and language pro-cessing (Jurafsky and Martin, 2000). This field offers methods for learning language models from pre-existing resources, e.g. an annotated corpus, and for using the learned model to annotate new texts. In order to annotate additional resources for Late Latin and Early Old French, I made use of existing annotated corpora in Latin and Old French: i) Perseus Latin Treebank (Bamman and Crane, 2011) and ii) MCVF Treebank (Martineau et al., 2007). These corpora were used as a training model for TnT, a statistical part-of-speech tagger (Brants, 2000b). The learned model was further employed to annotate additional texts in Imperial and Late Latin as well as Early Old French. The method will be described in section 5.1.2.

[7] http://www.hf.uio.no/ifikk/english/research/projects/proiel/

[8] http://bfm.ens-lyon.fr/

[9] The recent Penn Supplement containing annotated texts for Early Old French (Kroch et al. 2012) was not available at the time of the corpus compilation.

## 1.5 Probabilistic Approach in Linguistics

Quantitative linguistics is an emerging field in linguistics. While the probabilistic approach has certainly become a norm in psycholinguistics, natural language processing, cognitive science and sociolinguistic studies, the categorical approach is still dominant in historical linguistics.[10] In the categorical approach the linguistic phenomenon consists of well-defined discrete categories. However, a large body of psycholinguistic and sociolinguistic studies has shown that the linguistic phenomenon represents a gradient continuum that can show a different degree of distribution at a given time. Furthermore, with recent advances in quantitative studies, the shortcomings of previous techniques applied to historical linguistics have become obvious. First, frequency tables cannot rule out the possibility that the observed frequency distribution is due to a random chance (Gorman and Johnson, 2013, 214). For example, the lower frequency of a given token may result from the small size of the corpus. Second, it is well acknowledged that raw frequencies of the data may vary in different contexts, as these frequencies are conditioned by internal or external factors (Kroch, 1989c). Thus, the examination of contextual internal and social external environment remains essential for a full understanding of data distribution. In addition, some of the factors, or explanatory parameters, may have interaction, which would make it necessary to analyze the linguistic variable under investigation in contexts with each factor and contexts without it. Yet, all these factors considered, the effects will rarely be determined categorically. What they will tell us is the likelihood of hearing 'different forms in different contexts and with different speakers' (Meyerhoff, 2006, 10). In other words, the results are probabilistic. Finally, this approach makes it possible to handle chronological gaps in the data by providing a probabilistic estimation of change.

[10]For an overview of probability modeling in psycholinguistics, see Jurafsky (2003); in natural language processing, see Abney (1996) and Manning and Schütze (1999); in cognitive science, see Ward (2002) and in sociolinguistics see Tagliamonte (2011).

One of the earliest linguistic applications of probabilities can be found in Greenberg's typological language universals (Greenberg, 1963). Greenberg identifies two types of language universal: those that concern *existence* and those that concern *probabilities*. The latter universals are labelled *statistical*. The main tenet of statistical universals is the following:

> "when languages change over time, these historical processes are influenced by a large array of factors and because these factors are in highly dynamic competition, universal effects manifest themselves only statistically, never categorically" (Bickel, 2014, 12).

Furthermore, Greenberg's cross-linguistic comparison provided evidence for the role of frequency in language structures. His findings challenged the generative school, in which 'any focus on the frequency of use of the patterns or items of language is considered irrelevant' (Bybee, 2006, 6). However, it was the sociolinguistic field that introduced the ideas of directionality, rate of change, heterogeneity and variation in language (Labov, 1966).[11] While generative linguistics 'is concerned with an ideal speaker-listener, in a completely homogeneous speech-community' (Chomsky 1965:3-5), sociolinguistics shows that 'a model of language which accommodates the facts of variable usage (...) leads to more adequate descriptions of linguistic competence' (Weinreich et al., 1968, 99). Based on the principle that language variation behaves systematically and is 'integrated into a larger system of functioning units' (Labov 1972, 8), sociolinguistics demonstrates that linguistic phenomena can be statistically modeled (Labov 1963, 1969). Despite the opposition to and skepticism about syntactic variation (Lavandera, 1978), various studies have provided justification for the extension of probabilistic applications from phonology to syntax (Sankoff, 1973; Romaine, 1984). Furthermore, Manning (2003, 297), in his article on probabilistic syntax, illustrates two shortcomings of previous approaches to syntax:

(1) "Categorical linguistic theories claim too much. They place a hard categorical boundary of grammaticality where really there is a fuzzy edge, determined by many conflicting constraints and issues of conventionality versus human creativity."

(2) "Categorical linguistic theories explain too little. They say nothing at all about the soft constraints that explain how people choose to say things (or how they choose to understand them)."

[11] Direction of change as a pertinent characteristic of the system was mentioned first in Fries and Pike (1949, 42).

In contrast, in a probabilistic approach, each syntactic category is associated with a probability function. Moreover, such a data-driven language model is able to express a probability for the occurrence of a syntactic category given some contextual conditions, namely a conditional probability (Manning, 2003, 307).[12] To calculate such complex models, previous statistical tools, e.g. *the Pearson chi-square test* ($X^2$), have been replaced by more sophisticated linear model tools. The first application of a multiple linear regression model can be found in the pioneering sociolinguistic work on variable rules by Labov (1969). This model was replaced by a logistic regression model (Rousseau and Sankoff, 1978; Sankoff, 1988), which became the basis of the VARBRUL and Rbrul sociolinguistic toolkits (see section 4.3). In addition, the logistic function is used to model language change because its underlying function is S-shaped, which corresponds to the traditionally assumed S-shape model of language change (see section 4.1). With the transformation of this function, it is possible to identify the rate of language change (Kroch, 1989c) (see section 4.2). Finally, advances in the field of statistical analysis offer novel methods of examining variables in the data and for validating given language theories. While raw frequencies of data fluctuate, it has been shown that language-internal constraints persist (Meyerhoff and Walker 2007, 346), which enables statistical models to build a probabilistic, data-driven language model.

[12] Conditional Probability is represented as $P(form|meaning, context)$, where probability of form is calculated when the form is conditioned by its meaning and its context.

In historical linguistics, the researcher is constrained by the availability of written records. This often implies a small corpus size, e.g. 50 lines of verse, and unbalanced data, e.g. the number of texts is unequal in each period. With such unbalanced data, it would be inappropriate to rely on raw frequencies and percentages based on these frequencies. In fact, even simple statistical tests, such as *chi-square*, are sensitive to corpus size. Furthermore, word order patterns are often affected by internal and external contexts. For example, it is well known that lengthy constituents are often postverbal and that focused and topicalized constituents are preverbal. Not including these factors in the model would make it hard to draw conclusions, as in each case raw frequencies would reflect an order that is affected by these factors. In contrast, a multi-factorial analysis not only examines the significance of all these factors on word order, it also evaluates how factors interact and predict word order patterns.

## 1.6 Research Questions

As pointed out earlier, this thesis uses a corpus-based variationist methodology with the purpose of making a statistical inference as to the word order variation and change in infinitival clauses. The following questions are addressed in this thesis:

(1) When does the word order change take place? And can it be broken into separate stages?

(2) What role does information structure play in word order variation and word order change?

(3) What are the differences (if any) in word order constraints at different chronological stages?

(4) What do these results indicate about the rise of VO and the nature of word order change?

(5) Can we map our statistical model of OV/VO change onto a cartographic syntactic model?

In order to explain why OV or VO is chosen in any given historical document, three questions should be considered, namely variation, change and context. First, the alternation may represent a speaker choice governed by specific constraints, e.g., stylistic, pragmatic or syntactic. In this scenario, the influence of constraints should remain the same across time. Second, the alternation may be related to a language change. In this view, there will be a change in the significance of some constraints or even a change in their direction. For example, if nouns with old information are more likely to be preverbal in one period and postverbal in another chronological period, this factor will indicate a language change. Third, there might be certain contexts that favor syntactic change, that is, the diffusion of VO begins earlier in these contexts, in comparison with the others. Finally, contrary to previous analyses, where variables are studied in isolation and without a robust statistical analysis, in this study I will present a novel multi-factorial approach to word order change. Not only will various factors be analyzed simultaneously as one system, but individual variation by author and word will also be part of the analysis.

## 1.7 Structure of the Thesis

After this introductory chapter, chapter 2 will provide a general overview of word order variation, specifically the role of various factors influencing word order alternation in Latin and Old French. Chapter 3 will discuss the previous research on word order change. Chapter 4 will introduce probabilistic models of language change. Chapter 5 will present the corpora and methodology used in this study. Chapters 6 and 7 will illustrate results and analysis for Latin and Old French; their cross-examination and the final language change model will be presented in chapter 8. Chapter 9 will provide discussion and conclusion.

# Chapter 2

# Word Order Variation

*Everyone knows that language is variable.*

(Sapir, 1921, 147)

Word order variation and change is a complex topic that embraces a vast range of methodological issues. First, the complexity of this topic emerges from methodological limitations, such as chronological gaps in the written record, scarcity of non-literary resources, lack of prosodic cues and unbalanced representation of various genres and styles. For example, the disadvantage of studying the Early Old French period is well known. This period is essentially represented in a verse format, and its word order is often subject to certain rearrangement of constituents, in order to accommodate rhyme or meter (Foulet, 1923; Hirschbühler, 1990; Labelle, 2007). In a similar fashion, Latin prose also reflects an imposed rhythmical structure. Oberhelman and Hall (1984, 114) indicate that 'Latin prose authors commonly employed rhythmical units, or clausulae, to round off and embellish their sentences'. That is, we cannot ignore that the observed syntactic construction may not be representative for that language, as it could be a product of poetic style, translation influence or some literary convention (Petrova and Solf, 2009, 122-123). In addition, the contemporary researcher may not always be aware of certain sociolinguistic or stylistic variations for a given period. Second, this topic has been discussed and conceptualized through various theoretical frameworks, e.g., typological, discourse-functional, generative, cognitive and psycholinguistic, among others. The disputed concepts touch upon many aspects of word order, namely structural configuration, linear representation, basic word order, synchronic variation and diachronic change.[1]

In this chapter, I will present a sociolinguistic view on word order variation, where OV and VO orders are viewed as alternative inherent variants of the language. I will further introduce cross-linguistic generalizations with respect to the interplay between variation and factors influencing speaker's choices. Subsequently, I will review linguistic and sociolinguistic factors that have been proposed to account for OV/VO variation in Latin and Old French.

## 2.1 Word Order Variation

Many sociolinguistic studies have demonstrated that variation is 'a universal and functional design feature of language' (Foulkes 2006). This variation is not a random or optional event; on the contrary, it is structural, systematic and predictable (Labov, 1969). This means that word order variation is also a systematic, inherent structure of language. In this view, when there are two or more variants, each choice is more likely to be influenced or constrained by certain factors. In word order studies, these conditional factors can be categorized into the following two domains: 1) the *Syntactic* domain describes word patterns in terms of grammatical relations, hierarchical structures, head-dependent relations and syntactic categories and 2) the *Cognitive-pragmatic* domain examines the relationship between order and speaker-hearer interaction (Payne, 1992, 2-3). Many recent cross-linguistic studies further show that word order variation 'is situated at the crossroads of syntax, prosodic structure, semantics, and discourse pragmatics' (Lenerz, 2001, 249). Even languages with fixed word order, such as Modern English or Modern French, demonstrate a great variation of word patterns, as speakers attain their particular communicative needs by way of word order variation (Hinterhölzl and Petrova, 2009, 1). For example, the sentence in (16a) represents a canonical English order, namely Subject-Verb-Object, whereas in the sentence (16b) the object is placed initially, communicating the need of a speaker to emphasize or topicalize the object.

[1]For comprehensive evaluation of word order research in various theoretical frameworks, see Song (2012).

(16) a. I put the books on the shelf.

b. The BOOKS I put on the shelf.

Furthermore, there exists a relative cross-linguistic homogeneity in the interplay between linearization hierarchies and word order (Siewierska, 1993). These hierarchies are commonly classified into three groups: i) the *Formal* hierarchy, ii) the *Dominance* hierarchy and iii) the *Familiarity* hierarchy (Allan, 1987). The *Formal* hierarchy measures a linear dependency between length, syntactic complexity and word order. This category establishes the linear precedence of simple elements over more complex elements (17a) and short elements over heavy ones (17b).

(17) a. Structurally simple > Structurally complex

b. Short > Long

The *Dominance* hierarchy reflects how salience is perceived by speakers. The term *dominance* simply implies that one element dominates or is more salient than others, according to a speaker's perception. This category consists of personal hierarchy and semantic roles:

(18) a. 1st person > 2nd person > 3rd person

b. Human > Animals > Other organisms > Inorganic matter > Abstracts

c. Agent > Patient > Recipient > Benefactive > Instrumental > Spatial > Temporal

Finally, the *Familiarity* group deals with speakers' knowledge and their involvement with discourse topics. This category reflects familiarity, or 'closeness to the speaker's cognitive field' (Ertel, 1977). Traditionally, several notions have been related to this concept, namely familiarity, givenness, definiteness, topicality and referentiality, as illustrated in (19) (Siewierska, 1993, 830-831). Among these subcategories, familiarity can also be regarded as a superordinate category (Allan, 1987). Indeed, elements that are familiar in the speaker's world tend to be referential definite NPs expressing topics and old information.

(19) a. More familiar > Less familiar

b. Topic > Comment

c. Given > New

d. Definite > Indefinite

e. Referential > Nonreferential

Most of these linear hierarchies operate on a speaker's cognitive level. For example, it is often suggested that processing light simple information requires less cognitive effort, while heavy materials require more effort (Bock, 1982; Allan, 1987). Thus, the variability in word order seems to closely correlate with cognitive processing and communicative needs. In fact, it is cross-linguistically recognized that the influence of discourse factors tends to be of greater significance than purely semantic and syntactic considerations (Siewierska, 1993, 840). As Petrova and Solf point out:

> "the statistical evaluation of a corpus enriched with the features of the proposed information-structural scheme is of enormous value for detecting some ordering principles in an apparently unordered system" (Petrova and Solf, 2009, 154).

In addition to Payne's three-way classification, namely syntactic, cognitive and pragmatic domains, there exists another category that often seems to be excluded from conventional word order studies, namely a sociolinguistic domain (Currie, 2000). It is well known that historical documents vary by genre, register and time, e.g. prose vs. verse, historical treaty vs. hagiography, Classical Latin vs. Late Latin. Variability can also be observed on the individual writer level, as writers can shift their styles within the same work. Finally, we must not forget that historical texts may also be externally influenced by other languages, e.g. translation from Greek or Latin (Rizzi and Molinelli, 1994; Sornicola, 2006a; Lühr, 2009).

Having described the essential components of word order variation, I will turn to word order studies in Latin and Old French. I will start each section by identifying types of possible word order patterns, following by the description of an unmarked word order in each language. Subsequently, I will present various marked word order patterns and introduce various factors affecting word order: 1) pragmatic, 2) syntactic, 3) semantic and 4) sociolinguistic.

## 2.2 Latin Word Order

Latin exhibits great variation with respect to word order. It is often described as a 'free-word order' language, as the positioning of constituents is not determined by their syntactic function (Lehmann, 1992). The syntactic category can be clearly identified by inflectional endings, regardless of their sentential position. Thus, in Latin we can find all six theoretically possible combinations of a verb, its subject and object in a main declarative clause:[2]

(20) a. SOV

| | | | |
|---|---|---|---|
| Caesar | eius | dextram | prendit |
| Caesar-nom.sbj | his-gen | hand-acc.obj | took-3p.sg |

'Caesar took his right hand' (Caesar, *Gallic Wars* 1.20)

b. SVO

| | | |
|---|---|---|
| Ambracienses... | aperuerunt | portas |
| Ambraciots-nom.sbj | took-3p.pl | gates-acc.obj |

'The Ambraciots opened their gates' (Livy 38.9.9)

c. VSO

| | | | |
|---|---|---|---|
| Avertit | hic | casus | vaginam |
| turns-3p.sg | this | accident-nom.sbj | scabbard-acc.obj |

'This accident turns aside his scabbard' (Caesar, *Gallic Wars* 5.44)

d. OSV

| | | |
|---|---|---|
| Caesarem | Brutus | occidit |
| caesar-acc.obj | Brutus-nom.sbj | killed-3p.sg |

'Brutus killed Caesar'

e. VOS

[2]Examples are taken from Devine and Stephens (2006).

| | | |
|---|---|---|
| Peragit | concilium | Caesar |
| finished-3p.sg | conference-acc.obj | Caesar-nom.sbj |

‘Caesar finished the conference’ (Caesar, *Gallic Wars* 6.4)

f. OVS

| | | |
|---|---|---|
| Patrem | occidit | Sex.Roscius |
| father-acc.obj | killed-3p.sg | Sex.Roscius-nom.sbj |

‘Sex.Roscius killed his father’ (Cicero, *Oration* 39)

Any of the six word orders could be used to express the same truth-conditional meaning; however, the pragmatic values of the components vary with the different word orders. For example, the statement in (20f) can be interpreted as ‘Who Sex.Roscius killed was his father’, while the statement in (20d) is equivalent to ‘What happened to Caesar was that Brutus killed him’ (Devine and Stephens, 2006, 3-4). The dependence of word order variation on discourse considerations motivates the generalization that Latin is a discourse-functional language in which word order is defined by pragmatic functions (Panhuis, 1982; Pinkster, 1990; Lehmann, 1992; Spevak, 2005). In this view, word order patterns are conditioned upon *theme-rheme* structure and *focus-background* structure (Lehmann, 1992, 398). However, the frequency of each pattern is not the same, and SOV order is claimed to be dominant (Richter, 1903; Linde, 1923; Marouzeau, 1938; Watkins, 1964; Lehmann, 1992). Hence, Latin word order is traditionally thought to take the form of SOV, or verb-final order:

> “Die L.W. für den einfachen Aussagesatz ist: das Subjekt an der Spitze, das Akkusativ-Objekt vor dem Verb, das Übrige in der Mitte” (Richter, 1903, 7). [trans: Latin Word Order in simple declarative sentences is the following: Subject is at the head, Accusative object is before the verb, and the rest is in the middle]

The tendency for verb-final order is common among several Early-Indo-European languages, such as Hittite, Vedic, Greek, and Latin (Watkins, 1964). It is traditionally argued that this order was inherited from Proto-Indo-European, where verb-final position is considered unmarked and verb-initial position is marked (Brugmann and Delbrück, 1900; Lehmann, 1974; Watkins, 1964). Similarly, in Latin when the final position is occupied by a verb, it is considered the most unmarked position, whereas the initial verb is treated as marked (Brugmann and Delbrück, 1900). For example, verb-initial constructions may carry emphasis often indicating surprise or an unexpected event (21), stress a succession of events (22), take part in contrastive structures (23) or be used with verbs that express emotions, wishes, certainty or mental state (24) (Devine and Stephens, 2006; Bauer, 1995; Dettweiler, 1905). On the other hand, Spevak (2010) argues that verbs in the initial position are not always salient and that initial verbs in sentences providing explanations or presentative sentences do not bear any special pragmatic saliency.

(21) Ei! **Video** uxorem
Ei! see-1p.sg wife-acc.obj
'Dear me! I see my wife' (Terence, 797, (Bauer, 1995, 94))

(22) **Instruit** deinde aciem
drew-3p.sg next army-acc.obj
'Next he drew up his battle line.' (Livy, (Devine and Stephens, 2006, 159))

(23) **vicid** pudorem libido, timorem audacia
triumphed-3p.sg shyness-acc.obj lust-nom.sbj fear-acc.obj impudence-nom.sbj
'Lust triumphed over shyness, impudence over fear' (Cicero, *Pro Cluentio* 6.15, (Marouzeau, 1938, 53))

(24) **Moverat** plebem oratio consulis
moved-3p.sg plebs-acc.obj speech-abl consul-nom.sbj
'Consul speech had moved the plebs' (Livy 3.20.1)

Furthermore, the initial marked position can also be occupied by nominal constituents: i) topic constituents (25); ii) focus constituents (26) or iii) constituents with an emphasized focus, especially negative quantifiers and negated nominal objects (27) (Pinkster, 1990; Halla-aho, 2008):

(25) **Quintum** **fratrem** cotidie expectamus
Quintus-acc.obj brother-acc.obj everyday expect-1.pl
'We are expecting brother Quintus back any day.' (Cicero, *Atticus*1.5.8)

(26) (...) Scipio ... **Gracchum** ... interfecit ? **Catilinam**
(...) Scipio-nom.sbj ... Gracchus-acc.obj ... killed-3p.sg ? Catiline-acc.obj
nos consules perferemus ?
we-nom.sbj consuls-nom.sbj will-tolerate-3p.pl ?
'Shall Scipio have killed Gracchus, and shall we, consuls, put up with Catiline?' (Cicero, *Catilina* 1.3)

(27) (...) **ne mentionem** mihi fecit
(...) no mention-acc.obj me-dat made-3p.sg
'He has not even mentioned it to me' (*Tabulae Vindolandenses* II 343 (Halla-aho, 2008, 147))

The first type, *topic*, needs to be further differentiated into i) *Sentence topic* and ii) *Discourse Topic*. *Sentence topic* acts as a 'message foundation' and can include any entity or temporal and local settings (Weil, 1844:25). As Spevak points out:

> "When an entity is chosen as Sentence Topic, the speaker or author will say something about it; when a sentence starts with a temporal or local setting, the speaker or author will inform us about what happened then or there" (Spevak, 2010, 284).

In contrast, *Discourse Topic* refers to a person well present in the discourse and usually comes after *Sentence Topic*. Both types are, however, always placed initially. The second type, *Focus*, is often defined as an entity that conveys salient information. The placement of *Focus*, namely the most informative element in a sentence, is not always straightforward. Spevak states that in Latin *Focus* shows great mobility and is not confined to just one position, as salient constituents may occur in sentence-final or sentence-internal positions (Spevak, 2010, 283). She suggests using contextual and situational dependency to identify *Focus*, since contextually independent constituents are more likely to function as a focus. Furthermore, Spevak points out that *Focus* should not be confused with *Contrast*, which, according to her, is just a pragmatic feature that can be applied to both *Topic* and *Focus*. The following example illustrates contrastive Topics, *Caesar* and *Pompey*, and contrastive Foci, *secretly by night* and *openly by day* (Spevak, 2010, 46):

(28) (...) eorum (...) exercitum educunt: Pompeius clam et noctu,
(...) their (...) army-acc.obj lead-3p.pl: Pompey-nom.sbj secretly and night-abl,
Caesar palam atque interdiu
Caesar-nom.sbj openly and day-abl

'... and the two of them lead their armies (...): Pompey secretly by night, Caesar openly by day' (Caesar, *Civ.* 3.30.1-3)

Contrast can also be marked by various focus particles, such as *etiam* 'even', *quoque* 'too', *non ... sed, non ... verum* 'not ... but', *non solum ... sed (etiam)'* 'not only .. but (also)', *nisi* 'except', *solum, tantum* 'only'. While *Contrast Focus* is not determined by constituent position, *Verum Focus* is usually identified according to the initial position of the verb (Bolkestein, 1996, 17). The *Verum Focus* is a type of Contrast Focus that refers to actions. It is often marked with particles, such as *mehercule* 'really' and *igitur* 'then' (Spevak, 2010):

(29) Evolve diligenter eius eum librum, qui est de animo (...) –
read carefully his-gen that book-acc.obj, which is about soul-abl (...) –
Feci **mehercule**, et quidem saepius
did-1p.sg by-Hercules, and so repeatedly

'Read carefully that book of his, which is about the soul (...) – I have done so, I assure you, and done so many times' (Cicer, *Tusc.* 1.24)

The third type, *Emphasis*, is another pragmatic feature relevant to Latin constituent order, as it allows a writer to add personal evaluation (de Jong, 1989, 528). Emphasis is usually found with evaluation expressions, e.g., *tantus, talis* 'such', *multus* 'numerous' and *magnus* 'big', as well as negative words, such as *nullus* 'not any, none' and *nemo* 'nobody'. These emphatic expressions are often found in initial or final position.

Likewise, it is well recognized that word order linearization is also subject to certain syntactic and semantic conditions besides the pragmatic ones.[3] First, verb-initial order is often observed in imperative and main clauses. However, Spevak (2010) notes that imperative sentences are verb-initial only when they lack *Topic.* In contrast, verb-final order has a higher frequency rate in subordinate clauses. Table 2.1 shows that verb-final order occurs more frequently in subordinate clauses than in main clauses.

[3]It should be noted that most Latin word order studies focus on a verb position, namely initial, medial or final (Linde, 1923; Bauer, 1995). While this does not refer directly to object position, we could infer VO position from the verb initial structure and OV position from the verb final structure.

Table 2.1: Verb Final Structure in Main and Subordinate Clauses (Linde, 1923, 154-156)

| Period | Author | Genre | Main (%) | Subordinate (%) |
|---|---|---|---|---|
| Pre-classical | Cato (ch.1-27) | Science | 70 | 86 |
| Classical 1st BC | Caesar (Book II) | History | 84 | 93 |
| | Sallust (ch.1-36) | History | 75 | 87 |
| | Tacitus (ch.1-37) | History | 64 | 86 |
| | Livius (Book XXX 30-45) | History | 63 | 79 |
| | Gaius (I 1-38 and IV 160-187) | Science | 65 | 80 |
| Classical 1st AC | Seneca (1-9) | Letters | 58 | 66 |
| | Cicero (de Inventione 1-22) | Letters | 50 | 68 |
| | Cicero (de re Publica 1-32) | Letters | 35 | 61 |
| 2nd AC | Gaius (1-20, 45-69) | Science | 58 | 62 |
| | Vitensis (Book I) | History | 37 | 63 |
| Late 4th AC | Aetheria (1-20) | Ecclesiastic | 25 | 37 |

Furthermore, lengthy constituents, e.g., long attributes, enumerations, appositions and relative clauses, appear more often postverbally (Haida, 1928; Bauer, 1995). For example, in (30) the postverbal object *ilum locum* 'that place' is a heavy constituent, as it is combined with its relative clause.

(30) (...) et leget **illum locum qui scriptus est in evangelio**.
(...) and reads-3p.sg that-acc passage-acc.obj which written is in Gospel-loc
'... and he will read the passage that is written in Gospel.' (*Peregrinatio*)

It has also been shown that semantics and verb voice play a role in word order (Adams, 1976; Russo, 2000). For example, verb-initial position is common with motion verbs; aux-iliary verbs; perception verbs, e.g. *videre* 'to see' and causative and ditransitive verbs, e.g. *dare* 'to give' (Linde, 1923). Finally, a recent study by Devine and Stephens (2006) shows that semantics is also interwoven with word order patterns in Latin by means of abstractness and animacy features. For example, VO order seems to be preferred with abstract direct objects, e.g. 82% in *Livy*, while OV is preferred for inanimate nouns in contrast to animate entities that exhibit more mobility in their positioning (Devine and Stephens, 2006). They found that in Classical Latin non-presuppositional new referents are hosted in an immediate preverbal position which constitutes the neutral word order. These NPs become definite, and they move to a higher position when they are mentioned again. Based on their in-depth studies of Classical Latin prose, Devine and Stephens (2006, 79) suggest the following schema of a neutral word order:

(31) Subj DO IO/Obl Adj Goal/Source Nonref-DO V

where the subject is followed by a direct object, indirect object or oblique argument; adjunct, goal or source argument; non-referential non-specific abstract direct object and verb. In a similar fashion, Spevak (2010) observes that animate, agentive and individuated entities exhibit more mobility, whereas inanimate entities are often seen as a pragmatic unit with their verbs, resulting in high frequencies of OV in Classical Latin historical narratives (Spevak, 2010, 284-285).

While some argue that the unmarked SOV order remains predominant for a long time (Linde, 1923; Marouzeau, 1938; Watkins, 1964), the examination of its frequency across authors, texts and centuries shows that it is not constant (Linde, 1923). In fact, Lehmann (1974, 245) argues that 'the main clause pattern of Late Latin was VO already at the time of Saint Augustine'.[4] In fact, the oldest record with a considerable amount of VO dates as far back as the 2nd century AD. This source consists of letters written by a soldier named Terentianus, which were extensively examined by Adams (1977). While Greek influence cannot be ruled out in Terentianus's writing, Adams found many features that show that Terentianus was 'a genuine bilingual rather than one who had learnt Latin late' (Adams,

[4]Saint Augustine - 4th century AD.

1977, 75). The results from Adams's study show that VO is preferred with the ratio 2:1 and that VO order is a characteristic of 'spoken Latin of the informal varieties' that gained high frequency by the 3rd century in semi-literate Latin (Väänänen, 1967; Adams, 1976).[5] Table 2.2 illustrates the high frequency of VO order in semi-literate work in the 2nd and 4th centuries (Pinkster, 1990):

Table 2.2: Semi-literate Latin Work

| Period | Author | OV (%) | VO (%) |
|---|---|---|---|
| 2nd AC | Terentianus | 22 | 78 |
| 4th AC | Peregrinatio | 37 | 73 |

This section has reviewed various factors proposed to account for word order patterns in Latin. While SOV is traditionally seen as an unmarked neutral word order in Latin, some studies have shown that its status is not dominant in Late Latin (4th century). That is, it is not clear what to consider as an unmarked word order in Late Latin. While we have a better understanding of Classical Latin prose, 'our understanding of Late Latin order is still fragmentary', as József Herman has noted (Herman, 2000, 82)

## 2.3 Old French Word Order

Unlike Latin, Old French (texts from 842 - ca. 1350) is traditionally described as a verb-second (V2) language (Thurneysen, 1892; Skarup, 1975). One of the main characteristics of this stage is the obligatory placement of the finite verb into the second position in the main sentence; hence the label - V2 language (Adams 1987, Roberts 1993, Vance 1997). In addition, as a result of changes in the case system, Old French retains only two distinctive cases: nominative and accusative, as illustrated in (32) (Marchello-Nizia, 1999, 38):

(32) Li amiralz la sue gent apele
the emir-nom.sbj the his people-acc.obj sermons-3p.sg

[5] Väänänen (1967, 39-49) defines the following sources for semi-literate Latin: informal letters, technical writing, informal inscriptions, christian writing, glosses, literary writing reflecting popular speech (e.g. comedy) and writing in the late Roman period.

‘The emir sermons his people’ (*Roland*, 3400)

However, despite the loss of the Latin declension system, Old French provides evidence for all six possible word order patterns (Foulet, 1923):

(33) a. SOV

| | | | | | |
|---|---|---|---|---|---|
| Li | quens | Rollant | Gualter | de l’Hum | apelet |
| the | count-nom.sbj | Rolland | Walter-acc.obj | de l’Hum | calls-3p.sg |

‘Count Roland calls over Walter de l’Hum’ (*Roland*, 803)

b. SVO

| | | | | |
|---|---|---|---|---|
| Li | rois | apele | un | escuier |
| the | king-acc.nom | calls-3p.sg | a | horseman-acc.obj |

‘The King is calling a hourseman’ (Béroul, *Tristan*, 1483)

c. VSO

| | | | | | | | |
|---|---|---|---|---|---|---|---|
| Dunc | perdreit | Carles | le | destre | bras | del | cors |
| then | would-lose-3p.sg | Charles-nom.sbj | the | right | arm-acc.obj | of-the | body |

‘Then Charles would lose his right arm from his body’ (*Roland*, 597)

d. OSV

| | | | | | | | | |
|---|---|---|---|---|---|---|---|---|
| Un | grail | entre | ses | ii. | mains | une | dameisele | tenoit |
| a | grail-acc.obj | between | her | two | hands | a | demoiselle-sbj | held-3p.sg |

‘A demoiselle was coming forward, holding a grail in her two hands’ (*Perceval*, 3208)

e. VOS

| | | | | | | |
|---|---|---|---|---|---|---|
| Mult | a | grand | doel | Carlemagnes | li | reis |
| much | has-3p.sg | great | sorrow-acc.obj | Charlemagne-nom.sbj | the | king |

‘King Charlemagne has great sorrow’ (*Roland*, 3451)

f. OVS

| | | | |
|---|---|---|---|
| Ceste | parole | dist | Salemons |
| this | word-acc.obj | said-3p.sg | Solomon-nom.sbj |

‘Solomon said this proverb’ (*Queste*, 220)

Similarly to Latin, the first position in a sentence is often reserved for a topicalized or focused constituent (Thurneysen, 1892; Adams, 1987; Roberts, 1993; Vance, 1997; Labelle and Hirschbühler, 2005; Marchello-Nizia, 2007). Thus, any preverbal object is expected to bear some pragmatically marked features. For example, in (33f) the nominal object *parole* 'proverb' is a marked theme or topicalized object of the sentence. Furthermore, several studies have examined the spectrum of pragmatic functions on preverbal objects (Rickard, 1962; Marchello-Nizia, 1995; Rinke and Meisel, 2009; Zaring, 2010; Labelle and Hirschbühler, 2012). The findings show that preverbal objects can also be pragmatically unmarked and that the range of pragmatic features influencing OV order does not remain stable. Moreover, this range seems to be decreasing (Marchello-Nizia, 1995; Zaring, 2010). For the 12th century, the following features of preverbal direct objects have been identified: 1) *thematic* (marked and unmarked), 2) *echo*, 3) *archaism* and 4) *rhematic* (marked and unmarked).[6] The first type, *thematic*, refers to a noun that is closely tied to the discourse. In this case, the object is often modified by a definite article, a demonstrative or possessive determiner, or the adjectives *autre* 'other' or *tel* 'such'. The preverbal object can also be a proper noun or an ordinal number. The second type, *echo*, concerns a contextual repetition 'echoing' the previous information. In (34), the nominal object *escu* 'shield' is repeated information that can also bear some emphasis (Marchello-Nizia, 1995, 93).

(34) Or ne me faut mes fors **escu**(...) - Beau sire, fet
Now NEG me-dat must-3p.sg more except shield-acc.obj - fair sire, says-3p.sg
li rois, **escu** vos envoiera Dieux...
the king-nom.sbj, shield-acc.obj you-dat will-send-3p.sg God-nom.sbj...
'All I need now is a shield... - Fair sir, said the king, God will send you a shield'
(*Queste*, 12)

The *archaism* type refers to non-thematic, fixed expressions, which are complements of 'helping verbs',[7] e.g. *prendre conseil* 'to take advice'. Finally, the *rhematic* type describes an object that introduces new information that can also be emphatic. Thus, preverbal objects

[6]This classification is based on works by Haruku (1981) and Rickard (1962).
[7]See Marchello-Nizia (1995, 95-96) for a complete list of 'helping verbs'.

can be marked or unmarked and function as a theme or rheme. In contrast, by the 13th century the preverbal object only carries marked pragmatic features, such as marked theme and marked rheme.[8] Labelle and Hirschbühler (2012) have further statistically confirmed that the rate of preverbal objects carrying informational focus (rheme) is high until the beginning of the 13th century, where there is a considerable increase in preverbal topicalized objects. In their study, Labelle and Hirschbühler analyze preverbal objects in terms of *I-Focus* (Information Focus), which represents new information, and *I-Topic* which refers to a given or accessible entity. In this representation, *I-Focus* and *I-Topic* correlate with new/given information. Using the tripartite classification, namely *I-Focus*, *I-Topic* and *Unclear*, the authors have examined 19 texts between 980 and 1309 from the annotated corpus MCVF (Martineau et al., 2007). Due to the nature of the texts, the collected data fall into two different genres: 1) verse - between 10th-13th centuries and 2) prose - 13th century. Their findings show that preverbal objects tend to display *I-Focus* until 1205 and *I-Topic* after 1225 (see Figure 2.1):

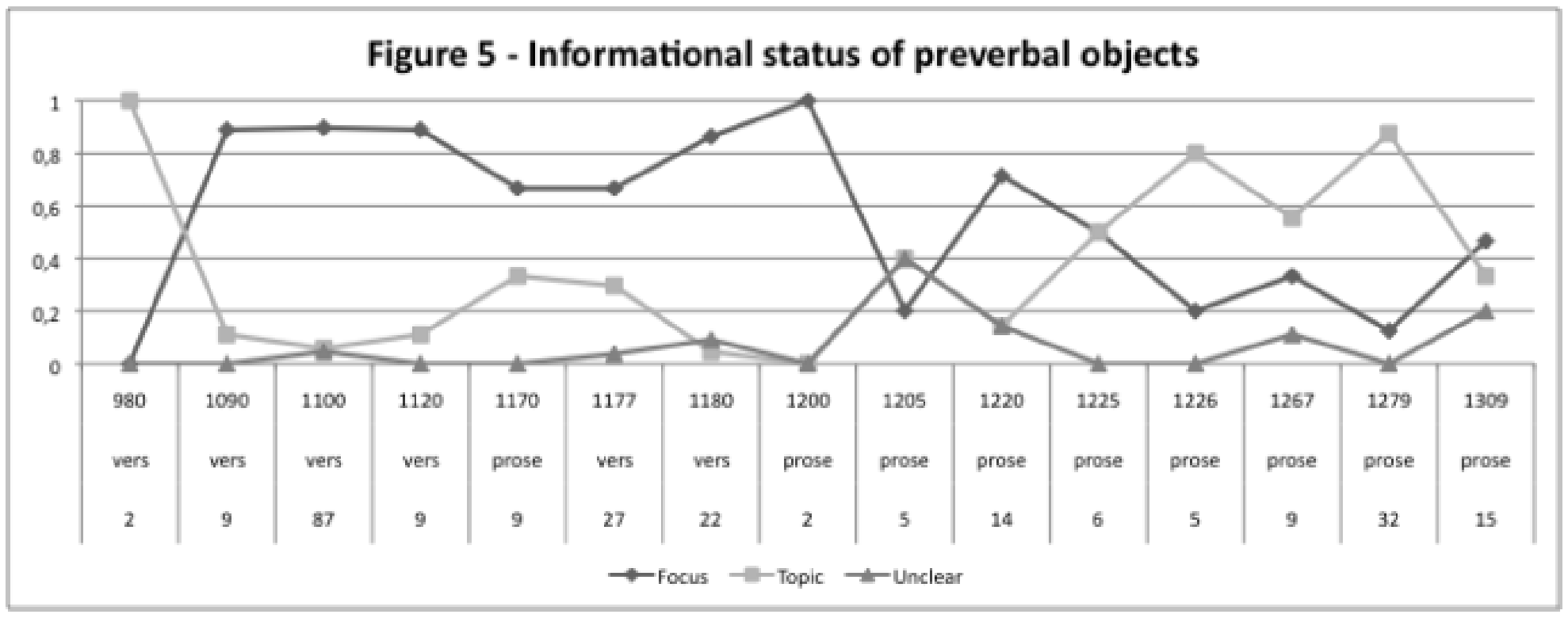

Figure 2.1: Information Status of Preverbal Nominal Objects (Labelle and Hirschbühler, 2012, 20)

Let us now examine other factors that have been observed in the literature with respect to word order in Old French. In addition to pragmatic factors, several syntactic factors play a role in word order. Similar to Latin, we find sentence type, the length of constituents and the presence of a subject. Recall that Latin demonstrates a persistence of OV order in subordinate clauses (Adams, 1976; Ledgeway, 2012a). In Old French SOV also occurs more frequently in subordinate clauses, mainly relative clauses, whereas main clauses show a high rate of OV/VO alternation (Rickard, 1962; Marchello-Nizia, 1995). Similar to Latin, heavy constituents influence VO order in Old French (Pearce, 1990, 246). In addition, it has been noted that OV/VO variation persists with null subject, whereas VO order is dominant with explicit subjects in main clauses (Marchello-Nizia, 1995, 82). On the other hand, it has been observed that verb form (finite and non-finite) appears to be significant for word order in Old French (Zaring, 2010). OV order with finite verbs is shown to be progressively decreasing. OV order appears 68% of the time in the early 11th century (*Passion*), 34% in the early 12th century (*Roland*) and only 3% in the early 13th century (*Queste*) (see Table 2.3).[9]

[8]Marked theme refers to a topicalized constituent carrying old information and marked rheme refers to a focalized constituent carrying new information.

Table 2.3: Frequency of OV/VO Order in Finite Main Clauses (adapted from Marchello-Nizia (1995), Zaring (2010) and Rouquier and Marchello-Nizia (2012))

| Period | Text | Genre | OV | VO |
|---|---|---|---|---|
| ca.1000 | La Passion de Clermont | Verse | 68% (78) | 32% (37) |
| 1040 | La Vie de saint Alexis | Verse | 34% (54) | 67% (107) |
| 1100 | La Chanson de Roland | Verse | 34% (366) | 66% (724) |
| ca.1190 | Le Roman de Perceval | Verse | 36% | 64% |
| 1205 | La Conqueste de Constantinople | Prose | 3% | 97% |
| 1220 | La Queste del Saint Graal | Prose | 3% (56) | 97% (1671) |

A similar tendency can be observed in subordinate clauses. The data from Early Old French (11th century) demonstrate that OV order is more frequent in *Passion* than in *Alexis*, as illustrated in table (2.4):

[9] In this table, I combined OV(S), OXV, SOV and XOV together to form an OV category. While it may be argued that OV(S) and SOV are two different phenomena, I am looking at the word order as a linear string. In addition, a caveat should be taken here with respect to raw frequencies, as the numbers are based on one work per period.

Table 2.4: OV/VO in Early Old French in Finite Subordinate Clauses (Rouquier and Marchello-Nizia, 2012, 139)

| Period | Text | Order | *Non-relatives* | *Relatives* |
|---|---|---|---|---|
| ca.1000 | Passion | OV | 14 (61%) | 5 (63%) |
| | | VO | 9 (39%) | 3 (38%) |
| 1040 | Alexis | OV | 11 (41%) | 2 (40%) |
| | | VO | 16 (59%) | 3 (60%) |

In contrast, non-finite verb forms exhibit a different tendency. OV order appears very frequently with infinitives even in the 13th century:[10] 60% in *Perceval* and 47% in *Conqueste* (see Table 2.5). Similarly, but somewhat less frequently, this tendency occurs with past participles: 43% in *Perceval* and 28% in *Conqueste*, as illustrated in Table 2.5.

Table 2.5: Frequency of OV Order for Non-finite Forms (Zaring, 2010, 8)

| Period | Text | Infinitives | Past Participles |
|---|---|---|---|
| 12th century | *Perceval* | 60% | 43% |
| 13th century | *Conqueste* | 47% | 28% |

In recent decades, there has also been growing interest in analyzing the influence of metrics on linear word orders in Old Romance languages (Hirschbühler, 1990; Rainsford et al., 2012; Rainsford and Scrivner, 2014). Prose and verse present two structurally different genres and it is recognized that word order patterns need to be examined separately for prose and verse in Old French.[11] For example, Rainsford et al. (2012) observe that in verses the verb is frequently placed at the end of a verse line, thus affecting the linear structure of the sentence. They also notice that in the 13th century the number of preverbal objects is still frequent in texts written in verse in contrast to the very small frequencies in prose cited by Marchello-Nizia (1995). Thus, it is essential to include metrical analysis in order to determine its influence on the linearization of word order patterns. This information can be inferred from syntactic, lexical and phonological properties of the texts and is fully described in Rainsford (2011).

[10] Zaring (2010) treated all types of infinitival clauses together: complements of modal, perception, causative verbs, prepositional clauses (*à*, *de*) and adjunct clauses (*por*, *sanz*). Several studies, however, have revealed that the clausal architecture for complements differs for restructuring verbs and lexical verbs (Pearce, 1990; Cinque, 2006)

[11] "Il est indispensable d'examiner séparément les textes en prose et les textes en vers" (Hirschbühler, 1990, 53).

## 2.4 Summary

This chapter has shown that word order variation is an inherent language characteristic and is deeply interwoven with all levels of language, namely pragmatics, semantics and syntax. A second purpose has been to provide methodological grounds by systematizing the knowledge about various determiners of word order linearization in Latin and Old French. The examination of various studies has revealed a sensitivity of word order patterns to information structures as well as to various semantic and syntactic factors. Finally, this review has also shown that some factors are consistent across Latin and Old French. For example, lengthy constituents tend to be placed postverbally, while topicalized and focalized constituents are mostly preverbal. That is, these factors reveal common cross-linguistic tendencies. In chapter 8 I will cross-examine the interplay of all these factors and their influence based on the data from the present study.

# Chapter 3

# Word Order Change

*There is no shortage of proposals in the literature*
*explaining why individual word orders may change.*
(Hawkins, 1979, 644)

It is a well-known fact that there is a progressive shift from OV to VO already at work in Latin itself, which finally results in the VO order, typical of (most of the) Romance languages. The central task of a historical linguist is thus to determine the principles and mechanisms of this change. When and why does the change occur? How does one language transition from one state into another, bringing about the microscopic variations that eventually alter the whole language system? How fast does the change progress and what are the constraints to such progress? To answer these questions, we inevitably confront many challenges and serious methodological limitations imposed by the nature of historical resources. First, the written evidence is often not available and is also unequally represented in the available historical documents. Second, the nature of written documents poses a reliability question. We cannot exclude the possibility that the observed construction may not be representative for the language under study, as it could be a product of poetic style, translation influence or some literary convention (Petrova and Solf, 2009, 122-123). On the other hand, 'not all variability and heterogeneity in language structure involves change' (Weinreich et al., 1968, 188), and some variation may remain stable across time (Sornicola, 2000, 102). There also exists a close relation between synchronic variation and diachronic change. In fact, both synchronic variation and diachronic change are just two dimensions of language variation. Finally, language change is a complex phenomenon that may have 'multiple causes and cannot be reduced to a single shift' (Ringe, 2013, 212) (also see Longobardi (2003)).

## 3.1 Language Change

In her chapter on *Variationist Approaches to Syntactic Change*, Pintzuk (2003) makes three important generalizations stemming from quantitative studies of language change. The first generalization concerns the distinction between stable variation and language change. Stable variation 'does not necessarily lead to or play a direct role in syntactic change' (Pintzuk, 2003, 509). Stable synchronic variations are usually governed by information structure or prosody and are frequently found in modern and old languages. Consider, for example, the common postposition of heavy constituents in Latin and Old French (Haida, 1928; Pearce, 1990; Bauer, 1995). This type of diachronic variation remains stable across time, i.e., the rates of usage are fairly consistent, and is also common cross-linguistically, e.g. the heavy NP shift in Old Icelandic (Hróarsdóttir, 2000). In contrast, when syntactic variation is unstable, there exists a competition between the old and new forms, leading to the consequent replacement of the old form. In reality, it is not always apparent whether observed variation is stable or dynamic. As Wolfram (2006) points out, many psycholinguistic and sociolinguistic factors must be considered prior to any conclusion. The second generalization refers to the gradual character of the change. The new syntactic form gradually becomes more and more frequent until the old form is completely replaced. As the language moves gradually through time, the passage from one stage into another inevitably involves the following components of change: i) *actuation*, ii) *transition*, iii) *embedding*, iv) *constraints* and v) *evaluation* (Weinreich et al., 1968). The *actuation* refers to an actuation of the change - when and why does this change occur? Needless to say, the 'why' question poses the most challenges, as the innovation occurs in speaker's (or group of speakers) present (Joseph, 1992), to which we do not have direct access in historical data. The *transition* concept makes reference to a process of transition from one language stage to another.

The *embedding* concept examines correlations between the advancement of language change in each stage and changes in language system. The effect of correlation is established when the process of change is accompanied by a change of other linguistic or non-linguistic factors 'in a predictable direction' (Labov, 1972, 163). The *constraints* concept identifies constraints on language change. Finally, the *evaluation* concept deals with evaluation of change - how this change is perceived in the society. The examination of these components in detail helps researchers understand the process of diffusion of new forms, their navigation through various linguistic and social constraints and the ways in which they are perceived in societies. The third generalization makes reference to the rate of language change. The mechanism behind the rate of change will be discussed in more detail in Chapter 4.

## 3.2 Interplay of Syntax and Information Structure

In the last two decades, we have seen considerable progress in our understanding of word order change, e.g. the theories of verbal position shift from verb-final to verb-medial (van Kemenade, 1987; Pintzuk, 1999; Taylor and Kroch, 1997) and verb second phenomenon (Adams, 1987; Roberts, 1993). This progress has been made possible in part by advances in corpus linguistics, specifically the compilation of The Helsinki Corpus of English Texts[1] and the ARCHER Corpus (Biber et al., 1994), both spanning from 750 to the 20th century in the history of English. In recent years there has also been growing interest in the interplay between syntax and cognitive-pragmatic domains, namely information structure. Several research projects have been launched with the specific goals of investigating the effects of information structure in language change: PROIEL: *Pragmatic Resources in Old Indo-European Languages*,[2] SFB 632 Project B4: *The role of information structure in language change*,[3] *Information structure in Welsh and its implications for diachronic*

[1] The Helsinki Corpus of English Texts. 1991. Department of Modern Languages, University of Helsinki.

[2] http://www.hf.uio.no/ifikk/english/research/projects/proiel/

[3] http://www.sfb632.uni-potsdam.de/en/cprojects/b4.html

*syntactic change*,[4] ISWOC: *Information Structure and Word Order Change in Germanic and Romance Languages*[5] and The Prague Dependency Treebank (Buráňová et al., 2000), along with many published books and monographs dedicated to this field of research (see Hinterhölzl and Petrova (2009), Ferraresi and Lühr (2010), Breul and Göbbel (2010) and Meurman-Solin et al. (2012)).

The interplay of syntax and the cognitive-pragmatic domain is one of the most discussed and most controversial topics in the literature. The problematic nature of this topic is due to a great variation in terminology and even variable definitions of the same terms (Mithun, 1992; Sornicola, 2006b). For example, the two main concepts are referred to in various ways: as *theme* and *rheme* (Mathesius 1939, Firbas 1964), *topic* and *comment* (Gundel 1974), *open proposition* and *focus* (Ward 1985) and *ground* and *focus* (Vallduvi, 1993). In addition, there are different definitions of *Topic* and *Focus*. Some argue that syntactic *Topic* and *Focus* do not necessarily coincide with their pragmatic and prosodic counterparts (Benincà, 2006; Bech and Eide, 2014), whereas others state that they are closely related (Steedman, 1991; Batllori and Hernanz, 2011). In the past two decades, there have been many attempts to consolidate this diversity of terminology and methodological application (Mithun, 1992; Lambrecht, 1994; Kiss, 1998; Krifka, 2007a; Götze et al., 2007).

The second challenge concerns methodological issues. First of all, such research consists of the simultaneous examination of the formal and discourse aspects of language, which entails a complex multidimensional relationship 'between linguistic form and the mental states of speakers and hearers' (Lambrecht, 1994, 1-2). Secondly, diachronic studies of discourse structure can offer only 'little cues to the prosodic realization of the utterance' via written form as compared to the synchronic studies that can rely on various intonational cues to identify pragmatic categories (Petrova and Solf, 2009, 132). Furthermore, some aspects of information structure are not yet fully represented in diachronic studies. For example, we can find plenty of work on *relational* structures, e.g. the relation between topic and focus. On the other hand, there is much less work on *referential* structures, e.g. the relation between the context and referents, such as old and new information (Bech and Eide, 2014, 1). There are also certain limitations to this approach, as it does not address issues related to psychological phenomena that are not expressed grammatically (Prince, 1981):

> "We are, therefore NOT concerned with what one individual may know or hypothesize about another individual's belief-state EXCEPT insofar as that knowledge and those hypotheses affect the forms and understanding of LINGUISTIC productions." (Prince, 1981, 233)

[4] http://www.research.leiden.edu/research-profiles/language-diversity/research/welsh.html

[5] http://www.hf.uio.no/ilos/english/research/projects/iswoc/

The investigation of information structure and its impact on word order change is still at the early stage of research. It is not clear whether word order is motivated by discourse structure first, which is then followed by this information structure becoming *syntacticized*, or whether word order is syntactically motivated first and then the discourse meaning is acquired (Los et al., 2012). However, several interesting generalizations have emerged from various cross-linguistic studies. The first generalization makes reference to the range of information structure features in the preverbal objects. Taylor and Pintzuk (2012a) demonstrate that in Old English the decrease of OV order is related to a decrease in the range of information structure features. They observe that objects bearing new information status disappear from the preverbal position, while at the same time word order becomes more fixed. We find a similar observation in Old French in Marchello-Nizia (1995) and Zaring (2010). Their data show that the range of pragmatic features on preverbal objects is decreasing, whereas the proportion of postverbal objects is increasing. The second generalization concerns new information focus. While the position of discourse-given constituents (topic) and the position of contrastive constituents seem to be the same across time and cross-linguistically, the position for new information focus undergoes some changes throughout time. In Old High German, for example, Petrova (2009) observes two different syntactic positions for contrastive and new information foci in subordinate clauses: 1) contrastive focus is left-adjacent to the finite verb and 2) new information focus is postposed. Bies (1996) finds a similar tendency in Early New Old German; in Modern German, however, the postposition of new information is unavailable. Similarly, Sitaridou (2011) notices that in Old Spanish starting in the 13th century the informational focus is no longer preverbal, while contrastive focus continues to be encoded in the left periphery.

## 3.3 Word Order Change from Latin to Romance

### 3.3.1 From Latin to Early Romance

This section concentrates on the literature that reflects word order shift from Latin to Old French. The diachronic research has been carried out in various frameworks, e.g., typological frameworks (Adams, 1977; Bauer, 1995) and generative frameworks (Salvi, 2004, 2005; Polo, 2005; Salvi, 2010; Ledgeway, 2012a,b). In addition, various studies have looked at the more narrow period of Old and Medieval Romance. For example, Marchello-Nizia (1995), Zaring (2010), Labelle and Hirschbühler (2012) have examined OV/VO change in Old French during the 12-13th centuries and Sitaridou (2011) has analyzed Old Spanish in the 12th-13th centuries. This shift remains one of the most debatable issues in diachronic studies; and there is no consensus on the time of its actuation or its triggers. Claims about time of shift actuation range from the time of *Plautus* (around 254-184 BC) (Adams, 1976) to Late Latin (4th century AD) (Grandgent, 1907; Bauer, 1995). For example, we read from Adams (1976, 88) that 'Latin is basically VO in type from the earliest texts' and from Grandgent (1907:31): 'Classic Latin may be said to represent an intermediate stage (...). By the fourth century the new order prevailed'. On the other hand, it is argued that Late Latin is not a VO language (Spevak, 2005).[6] Similarly, some argue that Old French is not a VO language and displays characteristics of an OV language (Zaring, 2011; Mathieu, 2009).

Furthermore, the shift is attributed to various causes: a) the evolution from a non-configurational syntax, where the grammatical relationship is marked by case and agreement, to a configurational syntax, where it is marked by a fixed functional position (Vincent, 1988); b) the change from head-final to head-initial parameters (Bauer, 1995; Ledgeway, 2012a); c) the grammaticalization of 'information-structurally marked forms' (Vennemann, 1974; Marchello-Nizia, 2007; Hinterhölzl, 2009) and d) the change in the left periphery of the clause (Salvi, 2005; Ledgeway, 2012a). Traditionally, the basic tenet of word order change is expressed through the shift of the finite verb from its final position to an intermediate position, followed by the grammaticalization of the subject in its initial position (Vennemann, 1974):

(35) SOV >TVX > SVO

where T in the intermediate stage TVX represents any topical element that is determined by a discourse function, and X refers to any non-topical constituent. The TVX stage is claimed to be a transitional stage between the Latin SOV and the Romance SVO order and is also identified as Verb Second (V2) (Ledgeway, 2012b, 64-65). In fact, Bauer (1995) statistically confirms that verb-medial constructions emerge in Classical Latin, suggesting that this construction is an innovation in Classical Latin. By the 4th century, the verb-medial order becomes predominant. As Herman (2000) points out, 'the characteristic feature of late Latin texts seems to be to have the verb between the two noun phrases (...), either SVO or OVS' (Herman, 2000, 86). On this account, Clackson and Horrocks (2007, 292) suggest that the underlying order in Late Latin is 'with verb occupying the first position in the sentence, with an optional focus slot before it, which may be filled by a verbal argument (subject as the default) or adverbial phrase':

(36) (Focus) Verb Subject Object

[6]"La place postverbale de l'objet-Actant 2 n'est pas requise par des contraintes syntaxiques, mais correspond à la fonction pragmatique de focus. (...) L'Itinerarium ne montre pas un ordre des constituants figé: il s'agit bien d'un ordre latin" [The postverbal position of object-second argument is not forced by syntactic constraints, it rather corresponds to discourse function of focus. (...) Itinerarium does not display a fixed constituent order. It is still a Latin word order. (translation mine)] (Spevak, 2005, 260).

Indeed, examples with a verb in the medial position preceded by focused elements and verb-initial sentences have been attested since the 2nd century AD (Salvi, 2004). Furthermore, in a recent study, Salvi (2004) observes that only certain verbs can appear initially in Classical Latin. We have already seen in section 2.2 that in Classical Latin specific types of verbs can be preposed, namely motion verbs and causative verbs as well as verbs in the imperative mood. In addition, verb-initial structures constitute a marked word order in Classical Latin (Dettweiler, 1905; Bauer, 1995; Devine and Stephens, 2006; Ledgeway, 2012b). According to Salvi's analysis, these preposed verbs appear to be in complementary distribution with focused constituents, as there is no attested evidence of these verbs being preceded by a focused constituent in Classical Latin. Thus, Salvi (2004) concludes that these verbs and focused constituents must be raised to the Focus projection. He further suggests that the verbs are raised to the head of the Focus projection, and that the specifier position is occupied by abstract operators (*Op*), e.g. jussive, assertive, optative, interrogative or concessive (see Figure 3.1a). On the other hand, when the constituent is focused, it is moved to the specifier position of Focus projection, and the head position remains empty (see Figure 3.1b) (Salvi, 2004, 43-44).

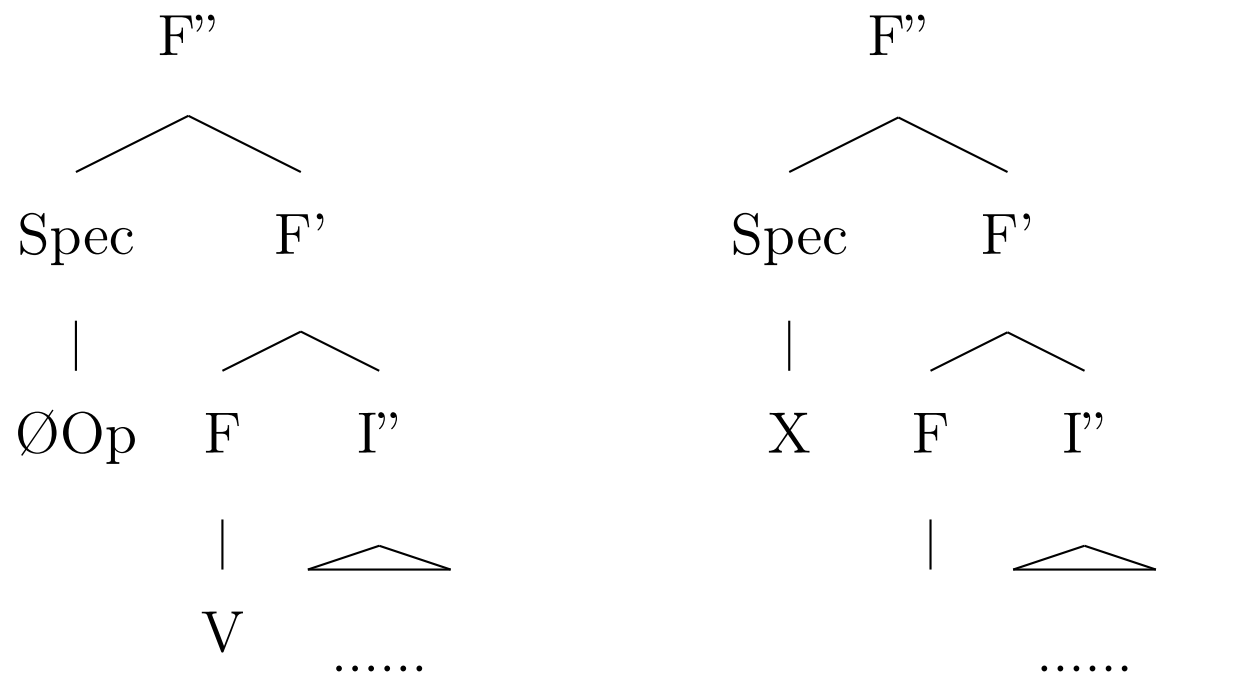

(a) Raised Verb (b) Focused Constituent

Figure 3.1: Focus Projection in Classical Latin (Salvi, 2004)

In Late Latin, the range of verb types in the initial position seems to be expanding. Salvi notices that verb preposing increases not only in main but also in subordinate clauses. Table 3.1 shows an increase of preposed verbs in main and subordinate clauses.

Table 3.1: Verb-Initial Structure in Main and Subordinate Clauses (Salvi, 2004, 102)

| Period | Author | Main Clause | Subordinate Clause |
|---|---|---|---|
| 1st BC | Cicerone | 19/181 (10%) | 9/192 (5%) |
| 2nd AD | Petronio | 35/167 (21%) | 4/83 (5%) |
| | Apuleio | 32/187 (17%) | 7/96 (7%) |
| | SHA | 43/238 (18%) | 8/147 (5%) |
| 4th AD | Tertulliano | 26/140 (19%) | 12/135 (9%) |
| 5th AD | Atti Cartagine | 41/224 (18%) | 24/194 (12%) |
| Vulgar Latin | | | |
| 2nd AD | Terenziano | 54/101 (53%) | 18/49 (37%) |
| 4th AD | Itinerarium | 85/250 (25%) | 57/169 (34%) |
| 4th AD | Acta/ Gesta | 101/259 (39%) | 41/180 (23%) |
| 6th AD | Theodericiana | 59/223 (26%) | 20/116 (17%) |

Furthermore, Salvi notices that in Late Latin the complementary distribution between a focused constituent and a raised verb is no longer a strict condition. Finally, by the 6th century, the author observes that thematic constituents start alternating with focused constituents. This alternation suggests a reanalysis of thematic constituents. Thus, Salvi (2005) defines a three-phase passage from Latin to Romance word order:

(37) **Latin** Left periphery | (Focus) (V) [SOXV] | Right periphery

**Phase 1** Left periphery | (Focus) **V** [SOX] | Right periphery

**Phase 2** Left periphery | (**Theme**/Focus) **V** [SOX] | Right periphery[7]

At first, fronted focused constituents and raised verbs are in a strict complementary distribution. The passage from Latin to Phase 1 is characterized by an expansion of all verbs in the initial position, yielding an unmarked verb-initial order where verb preposing becomes a norm. In this new phase fronted focused constituents and verbs are not in a complementary distribution. While in Classic Latin and Late Latin the left periphery of the sentence is reserved for thematic constituents, frame-settings, dislocated phrases or subordinate sentences, in Phase 2 thematic constituents are no longer associated with the left periphery and become reanalyzed as a part of the internal sentence, alternating with focus elements.

[7]Alexiadou (2000, 121) noticed that in Modern Spanish preverbal Focus and subject seem to compete for the same position, similar to Phase 2 (ex. **LAS ESPINACAS Pedro trajo* 'SPINACH, Pedro brought'), while in Modern Italian both focus and subject can be preverbal (ex. *QUESTO Gianni ti dira* 'THIS John will tell you').

In a similar fashion, Ledgeway (2012b) states that verb raising is 'an unmistakable precursor to the full-fledged V2 syntax of late Latin/early Romance' (Ledgeway, 2012b, 154). However, Ledgeway argues that the passage from Latin to Romance is best characterized as two parallel changes: 1) the change from head-final to head-initial parameters and 2) the change from specifier syntax (Latin XP-type) to head syntax (Romance X-type). First, Ledgeway provides arguments against the non-configurational approach to Latin, in which verbal and nominal constituents are represented as a *flat* non-hierarchical structure and the sentential structure is determined through morphological form in contrast to a dedicated syntactic position as, for example, in Modern French. In the non-configurational approach, the passage from Latin to Romance is viewed as the development of full-fledged nominal and verbal hierarchical phrases, where grammatical relations are marked by a fixed syntactic position. It is argued that the development of configurationality is triggered by the weakening of morphological inflection, which leads to a more fixed word order (Ledgeway, 2012a, 428). Ledgeway argues against the non-configurationality and offers evidence for the ubiquity of functional and configurational structures since early periods of Latin (for a detailed discussion, see Ledgeway (2012b)). Similar to Bauer (1995), Ledgeway views the change from Latin to Romance as 'a progressive reversal of the directionality parameter' (Ledgeway, 2012a, 436). The shift progresses from a rigid head-final order in archaic Latin (38) via a *mixed* intermediate order, e.g., head-final in (39a) and head-initial in (39b), to a head-initial order (Ledgeway, 2012a, 436):

| | | | | | |
|---|---|---|---|---|---|
| (38) | quoius | forma | uirtutei | parisuma | **fuit** |
| | whose | beauty-nom. | valour-dat. | most-equal-nom. | was |

'whose beauty was fully equal to his valour' (Archaic Latin, *Scipio Inscriptions* 1.7)

(39) a. ut constantibus hominibus par **erat**
that resolute-abl.pl men-abl. equal-nom. was
'that it was equal to that of men of strong character' (Classic Latin, Cicero, *Letters* 349 XI.28)

b. illa erat uita (...) libertate **esse** parem ceteris
that-nom was life-nom (...) freedom-abl be-inf equal-acc rest-dat.pl
'What he considered life (...) was the being equal to the rest of the citizens in freedom' (Classic Latin, Cicero, *Philippics* 1.34)

In addition to the change in directionality, namely the shift from head-final to head-initial parameters, Ledgeway also posits a second parallel change, namely the change from specifier syntax (Latin XP-type) to head syntax (Romance X-type). This change concerns the availability of pragmatic fronting in the left periphery of the functional projection. Latin shows extensive use of pragmatic fronting to the specifier of functional projection in Latin (Ledgeway, 2012a, 438-439). This fronting refers to a topic or focus fronting to the left edge of the functional category (*Topic* and *Focus* Specifier position), allowing for discontinuous structures, as illustrated in (40), where the discontinuous focused modifier *summo* 'highest' is fronted to the left edge of nominal phrase DP.

(40) [$_{DP}$[$_{Spec}$**summo**][$_{D'}$ Ø[$_{NP}$ homo [~~summo~~ ingenio]]]]
highest-abl man-nom talent-abl
'a man of the highest ability' (Cicero *De or.* 1.104)

In contrast, in Romance languages this pragmatic fronting seems to be blocked and such discontinuous structures are not found. Ledgeway suggests that the new functional categories, such as determiners and auxiliaries, block this fronting. These categories are traditionally viewed as being lexicalized in the head position of the functional projection. It seems that the presence of this lexicalized head blocks access to the left-edge fronting. Latin, on the other hand, does not have these categories. Thus, Ledgeway (2012a) concludes that there is a strict complementary distribution in the functional projection between 'pragmatic edge-fronting' and the presence of functional categories, namely non-empty head categories, such as determiners, auxiliaries etc. According to his analysis, this distribution reflects an ongoing change from Latin XP-type (specifier syntax) to Romance X-type (head syntax). In this view, XP syntax in Classical Latin entails access to a full phrasal specifier position of the functional projection (*Focus* or *Topic* Specifier), whereas Romance X syntax is restricted to only the head position in a functional projection. Compare the following examples (Ledgeway, 2012a, 438):

(41) a. $[_{IP}[_{Spec}\mathbf{XP}_{Topic/Focus}][_{I'}\emptyset[_{VP}V\,\sout{\mathrm{XP}}]]]$ Latin

b. $[_{IP}[_{Spec}\emptyset]\quad [_{I'}V[_{VP}\sout{\mathrm{V}}XP]]]$ Romance

In (41a) we see that the focused constituent/element is moved to the specifier position of a functional projection (left edge) in Latin. Note that the head position is empty (Ø).[8] In contrast, Romance languages, e.g. Spanish, allow for a head movement. This overt lexicalization of the head blocks a fronting movement for focused constituents (Ledgeway, 2012b, 277-291). This results in a strict complementary distribution between focused constituent movement and head movement in the same functional projection.

Similarly, Devine and Stephens (2006) claim that Classical Latin displays the characteristics of a specifier syntax. In their view, however, the word order shift results from the emergence of V-bar syntax in Classical Latin. According to their proposal, there are two types of Latin syntax involved: i) a specifier syntax and ii) 'VO-leakage'. Figure 3.2 shows the discourse-configurational structure of a sentence in Latin, as proposed by Devine and Stephens (2006).

[8] Latin lacks head movement in this approach (Ledgeway, 2012a, 438).

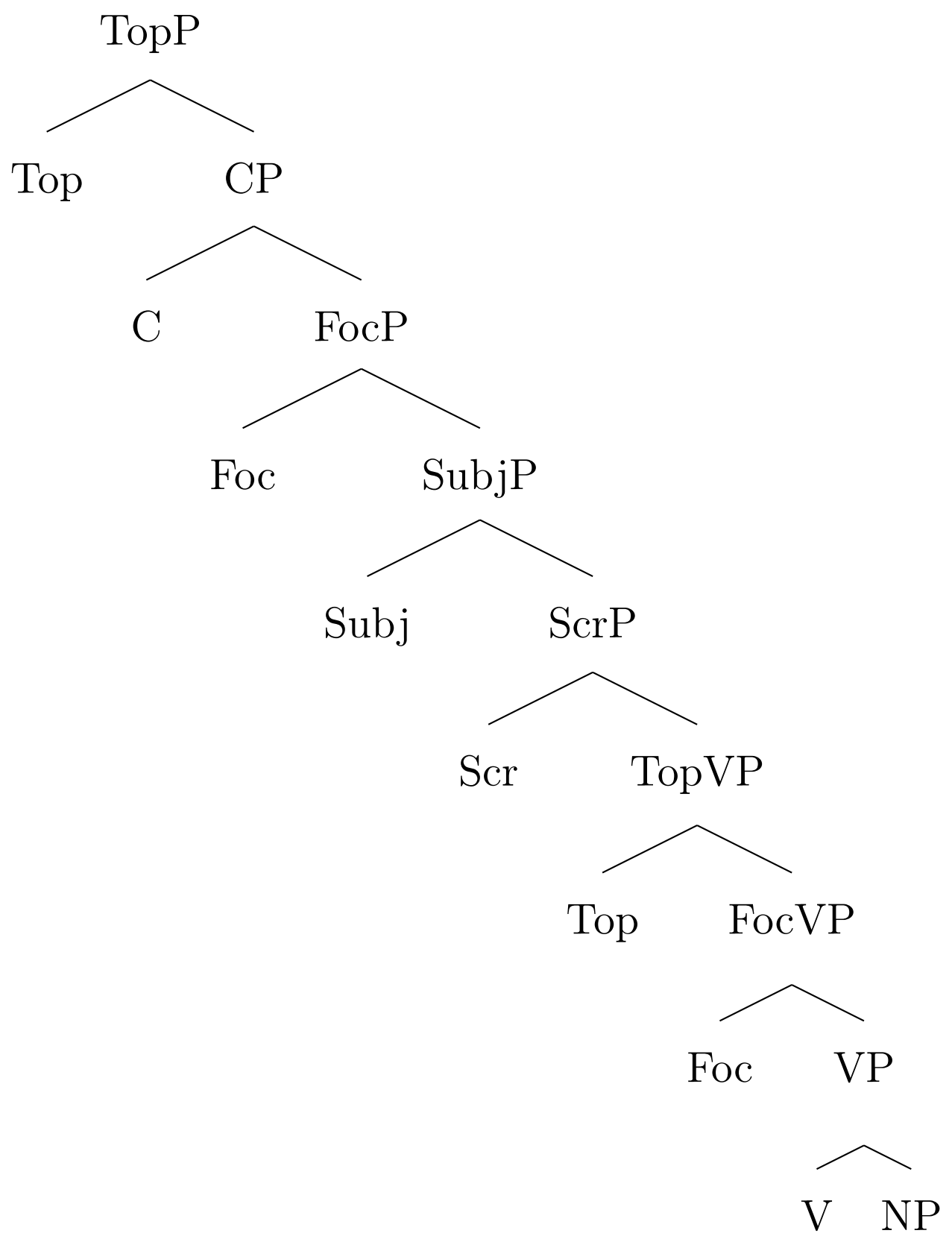

Figure 3.2: Three-layered Architecture of a Latin Sentence (Devine and Stephens, 2006, 28)

In their view, the top layer, named *CP*, contains strong topics (*TopP*), interrogatives and complementizers (*CP*) as well as strong focus (*FocP*). The second layer, named *IP*, hosts subjects (*SubjP*) and phrases scrambled out of the third layer (*ScrP*). The *IP* layer functions as a mere link between events and discourse, namely between the 1st and the 3rd layers. The last layer, named *VP*, contains 'nuclear assertion' or 'bare event'. This layer also holds verbal inflection and tense. In addition, the *VP* layer includes two projections, namely *FocVP* and *TopVP*. *FocVP* is projected to host non-presuppositional new referents in the neutral word order, whereas *TopVP* hosts referents that have been already mentioned. These projections should not be confused with the pragmatic strong topic *TopP* (topicalisation) and focus *FocP* (focalization) hosted in the *CP* layer.

According to Devine and Stephens, in the specifier syntax, nominal objects are moved from *VP* to the projection *FocVP*, which corresponds to information that is discourse-new. If the noun phrase is an old referent, it moves to the *TopVP*, which hosts old information referents. This transformation generates a linear neutral word order SOV. V-bar syntax, in contrast, is an innovation and allows an object to remain stranded in *VP* instead of moving to a discourse-new *FocVP* or discourse-old *TopVP*, as shown earlier. This postverbal position is restricted to non-referential abstract nouns or tail nouns, which are not used as discourse referents (Devine and Stephens, 2006, 135). An example of a non-referential object is illustrated in (42), where *legatos* 'envoys' refers to a non-referential entity. It should be noted, however, that Devine and Stephens base their analysis only on two authors from the Classic Latin, namely Caesar and Livy.

(42) (...) ut ad regem mitterent **legatos**
(...) that to king send-3p.pl envoys-acc.obj
'(...) that they send envoys to the king' (*Livy*)

### 3.3.2 From Early Romance to Medieval Romance

From the late 19th century it has been recognized that the verb-medial or verb second order is one of the main characteristics of Old French (Thurneysen, 1892). This verb-medial order is often viewed as a transitional stage between Latin SOV and Romance SVO. Marchello-Nizia (2007) further suggests that the transition to SVO in Old French must be split into two separate stages. First, the VO order becomes prevailing by the middle of the 12th century (Marchello-Nizia, 1999, 40). Second, there is an increase in the overt subject, which leads to the grammaticalization of the thematic (fronted) subject as a syntactic subject in the 14th century (Marchello-Nizia, 2007, 110). Similarly, several recent studies reveal a notable difference between Early Old French (11th -12th centuries) and Late Old French (13th century) (Marchello-Nizia, 1995; Labelle and Hirschbühler, 2005; Labelle, 2007; Zaring, 2011).[9] First, the difference is observed in embedded clauses, namely the non-restricted distribution of null subject and verb-initial and verb-third structures in Early Old French and their restriction in Late Old French (Hirschbühler, 1990; Roberts, 1993; Labelle, 2007). On this account, Labelle states that 'embedded clauses came to be strictly SVO in late Old French, with overt subject being norm' (Labelle, 2007, 314). Second, Labelle and Hirschbühler (2005) and Zaring (2011) suggest that Early Old French seems to have a richer *IP* structure, whereas Late Old French has a reduced *IP* structure. Furthermore, the change is observed on an information structure level, namely a decrease in the pragmatic range for preverbal objects. Until the 13th century, preverbal objects bear features of topic, contrastive focus and information focus (Marchello-Nizia, 1995; Zaring, 2010; Labelle and Hirschbühler, 2012). In 13th-century prose texts, information focus seems to shift from preverbal to postverbal position. It is even suggested that in the 13th century 'information structure does not seem to be a major factor in the ordering of constituents; word order is probably primarily determined by syntactic constraints' (Bech and Salvesen, 2014, 257).

[9] Zaring (2011) mentions that there is no clear definition of Early and Late Old French periods. Roughly speaking, the Early Old French boundary is placed at the end of the 12th century (Marchello-Nizia, 1995), but Labelle and Hirschbühler (2005) and Labelle (2007) also include the 13th century verse texts.

A similar shift from the preverbal to postverbal information focus is observed in other Romance varieties, e.g., Old Spanish and Old Italian (Sitaridou, 2011; Cruschina, 2011). Cruschina suggests that the change is triggered by the activation of two *Focus* projections, a higher left peripheral Focus and a lower clause-internal Focus. In fact, it has been widely recognized that contrastive focus and information focus are two different types. Recent studies also show that these two foci target two different syntactic positions.[10] Benincà (2004) proposes to encode these two foci in the left periphery for Medieval Romance varieties based on the data from the 12th to the 14th centuries. In the schema (43), the lowest field is the *Focus* field, which has two dedicated positions, namely *Focus I* for contrastive focus and *Focus II* for information focus (Benincà, 2004, 288):

(43) $[ForceP[FrameP[TopP[FocP[IFocus][IIFocus][FinP]]]]]$

Finally, some studies interpret the observed correlation between the decrease of preverbal pragmatic features and the increase of VO order as a result of prosodic change

[10] Cruschina (2011) indicates that despite the fact that Contrastive Focus and Information Focus do not co-occur, the evidence from multiple studies suggests that they occupy two different syntactic positions.

(Marchello-Nizia, 1995; Zaring, 2010). This prosodic change reflects a shift from a strong lexical stress to a phrase-based stress, where stress is defined by the syntactic phrase. Since the phrase-based stress targets a right edge, namely the last syllable of the phrase, postverbal objects become more frequent, as part of the verbal phrase. The evidence from the versification system shows that the new stress is first attested in the 11th century and is not fully developed until the mid 13th century (Rainsford, 2011).

## 3.4 Summary

In this chapter, I have discussed the change from Latin SOV to Romance SVO word order, in particular the interplay of syntax and information structure. I have also outlined the inevitable methodological challenges associated with diachronic studies. Subsequently, I have examined various proposals reflecting the shift's timing and causation. In spite of some theoretical differences, these studies show that word order shift is a complex phenomenon involving changes on syntactic, pragmatic and semantic levels. We have seen that verb fronting in Classical Latin, a marked structure, extends to the larger class of verbs. Simultaneously, there are changes in the left periphery, such as reanalysis of thematic constituents and a decreasing range of pragmatic features on preverbal constituents. It is also argued that the change proceeds in stages, namely SOV (archaic Latin), verb-medial (Late Latin and Old French) and SVO (Medieval French). However, as Bech and Salvesen (2014, 236) point out:

> "In spite of decades of research, neither the exact nature of the change nor the causes of the change have been fully understood".

It should be noted that the majority of word order studies have drawn conclusions based on their use of qualitative and quantitative methods. It is well known, however, that raw frequency in historical documents can be particularly noisy, as we do not know whether we are dealing with random outliers or a representative sample. In contrast, statistics and probability allow researchers to measure and define statistically significant features and detect outliers. In the next chapter I will review various probabilistic approaches that have been applied to language change models, which will shape the remainder of the thesis.

# Chapter 4

## Probabilistic Approaches to Language Change

*Jedes Stadium der Sprache ist ein Uebergangsstadium,*
*ein jedes ebenso normal wie irgend ein anderes*[1]
*(Schuchardt, 1885, 21)*

In this chapter, I will review several models of language change that are based on variation in language use. Since data in historical linguistics are represented as a quantity, namely the sum of all occurrences of a specific phenomenon in a given text or period, this quantity can be statistically modeled with respect to a timeline. Various models have been developed to reflect and construct a model of language change, including the *S-shape* model (Osgood and Sebeok, 1954; Weinreich et al., 1968; Bailey, 1973), the *Variable Rule* model (Labov, 1969; Cedergren and Sankoff, 1974; Sankoff and Labov, 1979; Gorman and Johnson, 2013), the *Constant Rate Effect* model (Kroch, 1989c) and the *Usage-Based* model (Bod, 2006; Bybee, 2013). Although all statistical models assume variation in language use, they diverge, however, in their underlying beliefs. The generative model of the *Constant Rate Effect* considers language variation as a reflection of different parameter settings in grammar, allowing for syntactic diglossia among speakers of the same generation. This assumption is based on a categorical view of parameters (Kroch, 2001, 721). Moreover, in this approach, 'once a community becomes diglossic with respect to a given parameter setting, every speaker will learn both settings' (Kroch, 2001, 722). In contrast, the sociolinguistic *Variable Rule* approach asserts that language variation is inherent (Labov, 1969, 728) and that 'language change is one type of linguistic variation, with particular social properties' (Chambers, 2013, 297). In this view, linguistic variables are systematic in language and the choices between alternative variants are influenced by many independent factors (Labov, 1963, 1969). Thus, to understand a language change, one must account for all linguistic and social factors (Tagliamonte, 2011, 3), in contrast to a generative model, where these factors play no role in language change. Furthermore, the tendency of speakers to select between alternative forms constitutes the rate of language change for the speech community. While the sociolinguistic model integrates these probabilities into a grammar, the generative model opposes them (**?**, 722).[2] On the other hand, there is evidence from usage-based studies that 'frequency and probability matching are part of a generalized cognitive capacity to learn and manipulate the environment' (Sabino, 2012, 412). In this approach, the *Usage-Based* model views language 'not as a grammar but as a statistical ensemble of experiences that slightly changes every time a new utterance is perceived or produced' (Bod, 2006, 318).

[1]'Every stage of a language is a transitional stage, each as normal as any other...' (Schuchardt, 1972, 53).

## 4.1 The S-Curve Model

It is generally assumed that 'the time course of the propagation of a language change typically follows an S-curve' (Croft, 2000, 183), as illustrated in Figure 4.1. Language change is conceived as a 'gradual replacement of one form by another in the language of speakers [...] over the course of many generations' (Pintzuk, 2003, 512).

[2]"[T]he pattern of favoring and disfavoring contexts does not reflect the forces pushing the change forward" (Kroch, 1989c, 241).

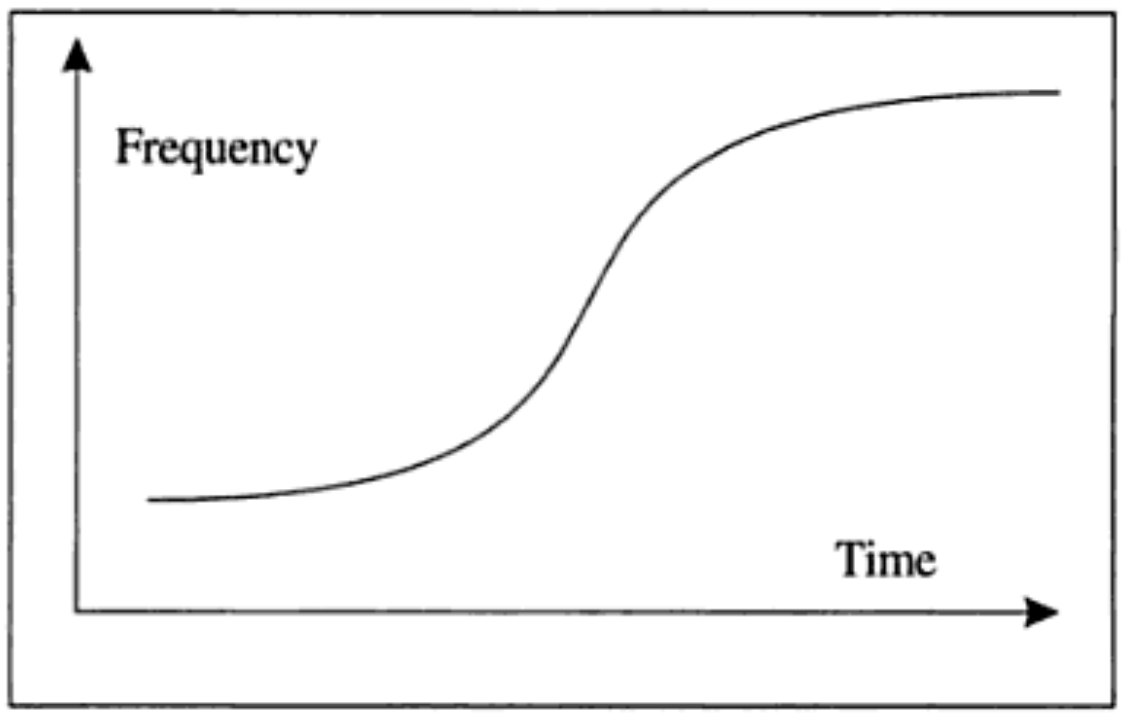

Figure 4.1: The S-Shaped Model of Language Change (Hróarsdóttir, 2000, 32)

This idea was first introduced in Osgood and Sebeok (1954) (see 44a), Weinreich et al. (1968) (see 44b) and Bailey (1973) (see 44c):

(44) a. 'The process of change in the community would most probably be represented by an S-curve' (Osgood and Sebeok, 1954, 155)

b. '[T]he progress of a language change through a community follows a lawful course, as S-curve from minority to majority to totality' (Weinreich et al., 1968, 113)

c. 'The result is an S-curve: the statistical differences among isolects in the middle relative times of the change will be greater than the statistical differences among the early and late isolects' (Bailey, 1973, 77)

This model is commonly characterized by three stages of change (Weinreich et al., 1968, 184):

(45) a. A speaker learns an alternate form

b. The two forms exist in contact within speaker's competence

c. One of the forms becomes obsolete

According to this model, the initial diffusion of a new form (45a) proceeds very slowly and in limited linguistic environments (Bailey, 1973, 13). The first stage is typically represented from zero to 20% of occurrence for an innovation form (see Figure 4.2).

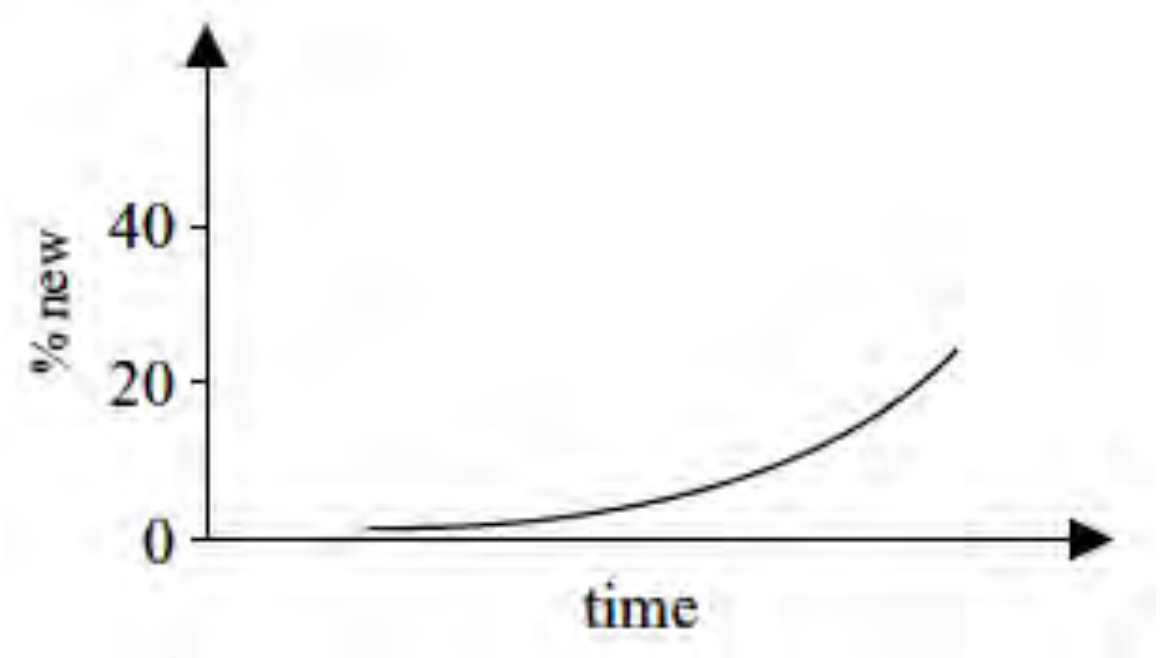

Figure 4.2: The beginning of a change (Denison, 2003, 55)

During the second stage (45b), the replacement rate is drastically accelerated and the change is spread 'from more restricted to more general linguistic environments' (Bailey, 1973, 13). This stage is commonly placed on the interval between 20% and 80% of the new form (Chambers, 1992, 695). Finally, the change decelerates at the third stage (45c), when the old form becomes rare and eventually disappears.[3] The reasoning behind this S-shaped model, namely its 'slow, slow, quick, slow' pattern, is offered in Weinreich et al. (1968). First, Weinreich et al. (1968, 188) show that any change 'involves variability and heterogeneity'. Second, the variation must be initiated. That is, the variation, namely alternative ways of saying the same thing, does not necessarily lead to a change - it can be stable (i.e. not undergoing change). To initiate change, one of the variants must present some slight advantages in its use, such as social, stylistic or structural. Initially, the new form is rare, and there is no strong pressure to choose this new form - this fact is reflected in the small frequencies of new cases. In contrast, the middle stage provides strong competition between the two forms; thus, the selective choice is greater, which is illustrated by a greater rate in S-curve (Denison, 2003, 58). Lastly, the final stage is represented by a weak selective choice for the old form, which becomes rare, yielding a smaller rate of change on an S-curve.

Bailey (1973, 43) further suggests that 'the change begins variably and spreads across the social barriers of age, sex, class, space, and the like in waves'. He reformulates the 19th-

[3]It is possible that the change may never be fully completed (Denison, 2003, 56) Some instances of old form may resist change and remain as residue (Chambers, 1992, 695).

century wave theory as an effort 'to build into generative theory both social and historical variation' (Le Page, 1998, 21). The wave representation can be conceptualized as follows: 'new features of a language spread from a central point outwards like waves just as when a stone is thrown into a body of water' (Tagliamonte, 2011, 56). As Le Page (1998, 21) notes, this was one of the attempts to represent data as a linear continuum showing different degrees of relationship between the two ends of the continuum (see also DeCamp and Bickerton 1971, 1975). This model also assumes that the actuation of the change begins at different time point for different linguistic contexts. That is, 'what is quantitatively less is slower and later; what is more is earlier and faster' (Bailey, 1973, 82). This correspondence is illustrated in Table 4.1, which shows the process of change from the categorical use of the form $A$ to a new form $B$ in two different environments, $E_1$ and $E_2$:

Table 4.1: The Wave Model of Change by Bailey (1973), as exemplified in Wolfram (2006)

| **Stage of Change** | $\mathbf{E_1}$ | $\mathbf{E_2}$ |
|---|---|---|
| Categorical status | A | A |
| Early stage, restricted environment | A/B | A |
| Change in full progress, the use of new form in $\mathrm{E_1}$ and in $\mathrm{E_2}$ | A/B | A/B |
| Completion of change in $\mathrm{E_1}$ | B | A/B |

As we can see from Table 4.1, in some contexts the change progresses early ($E_1$ - the most preferable context) or late ($E_2$ - the least preferable context). Thus, actuation time and diffusion speed may vary based on context. As a result, the observed difference in frequencies is attributable to the different time of actuation. While at the microlevel the new form[4] follows an S-shape pattern, the spread of this form through social space is described as a wave pattern, illustrated in Figure 4.3. Figure 4.3 shows that at any given time, the propagation exhibits a higher frequency rate at the earliest context, e.g. $a > b > c$.

[4]Bailey describes change in terms of rules. As the main object of his study is phonetic change, the contexts are phonetic environments and the rules are composed of a set of phonetic features that are reordered based on their markedness, weight and hierarchy (see Bailey (1970, 1973) for more detail).

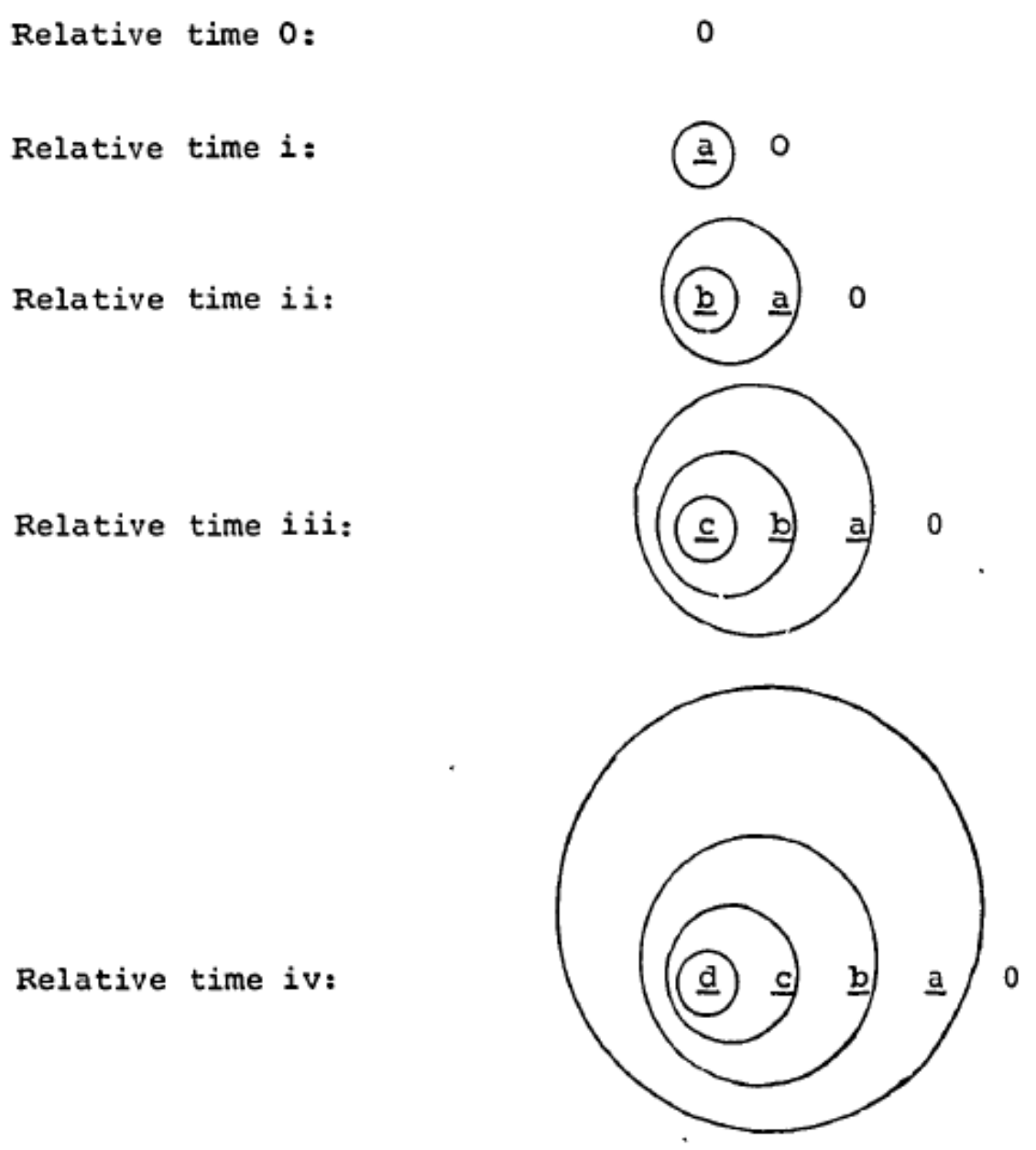

Figure 4.3: The Wave Model through Social Space and Time (Bailey, 1973, 68)

It is important to understand that at its early stage the language model hypothesis was conceptualized but not supported empirically. Only two decades later, several studies proposed specific mathematical functions to fit the S-curve model (Piotrovskaja and Piotrovskij, 1974; Altmann, 1983; Kroch, 1989c; Labov, 1994). Their findings had great impact on historical linguistics, enabling comparisons between different datasets and their parameters, which allowed researchers to draw conclusion about language change (Pintzuk, 2003, 512). Piotrovskaja and Piotrovskij have proposed an inverse tangent function to represent the increase of new forms over time; both Altmann and Kroch have looked at the logistic function, and Labov has suggested cumulative frequencies of the binomial distribution (see Section 4.2 and Section 4.3). Using the logistic function to fit the S-shaped model has been shown to have several advantages. First, the logistic function is traditionally used in cases where the predicted value (dependent variable) is binary, e.g. success and failure, which can easily be translated in the presence or absence of new forms in linguistic terms. Thus, the logistic function can express the rate of replacement of an old linguistic form by a new linguistic form. Second, the logistic model has been successfully tested on various morpho-syntactic datasets, providing evidence for validity of this model in morphological and syntactic language changes (Altmann, 1983; Kroch, 1989a,c).

## 4.2 The Constant Rate Effect Model

Kroch (1989c) formulates a hypothesis known as the *Constant Rate Effect* (CRE). This hypothesis challenges previous approaches by saying that 'change proceeds at the same rate in all contexts' (Kroch, 1989c, 36) as compared to, for example, Bailey's (1973) approach stating that the rate is different in different contexts. Using data from the study of English *do*-support (Ellegård, 1952), Kroch (1989a) first shows that the change follows a gradual pattern. The periphrastic *do* is first attested in the 13th century, but it becomes categorical only in modern English. Example (46) exhibits one of the earliest illustrations of the use of *do* in a declarative sentence:

| | | | | | |
|---|---|---|---|---|---|
| (46) | his sclauyn | he | dude | dun | legge |
| | his pilgrim cloak | he | did | down | lay |

'He laid down his pilgrim's cloak' (Ellegård, 1952, 56)

Kroch (1989a) observes that the change proceeds gradually in various syntactic contexts, namely affirmative and negative questions, and affirmative and negative declaratives, as illustrated in Figure 4.4 (Kroch, 1989c, 22).

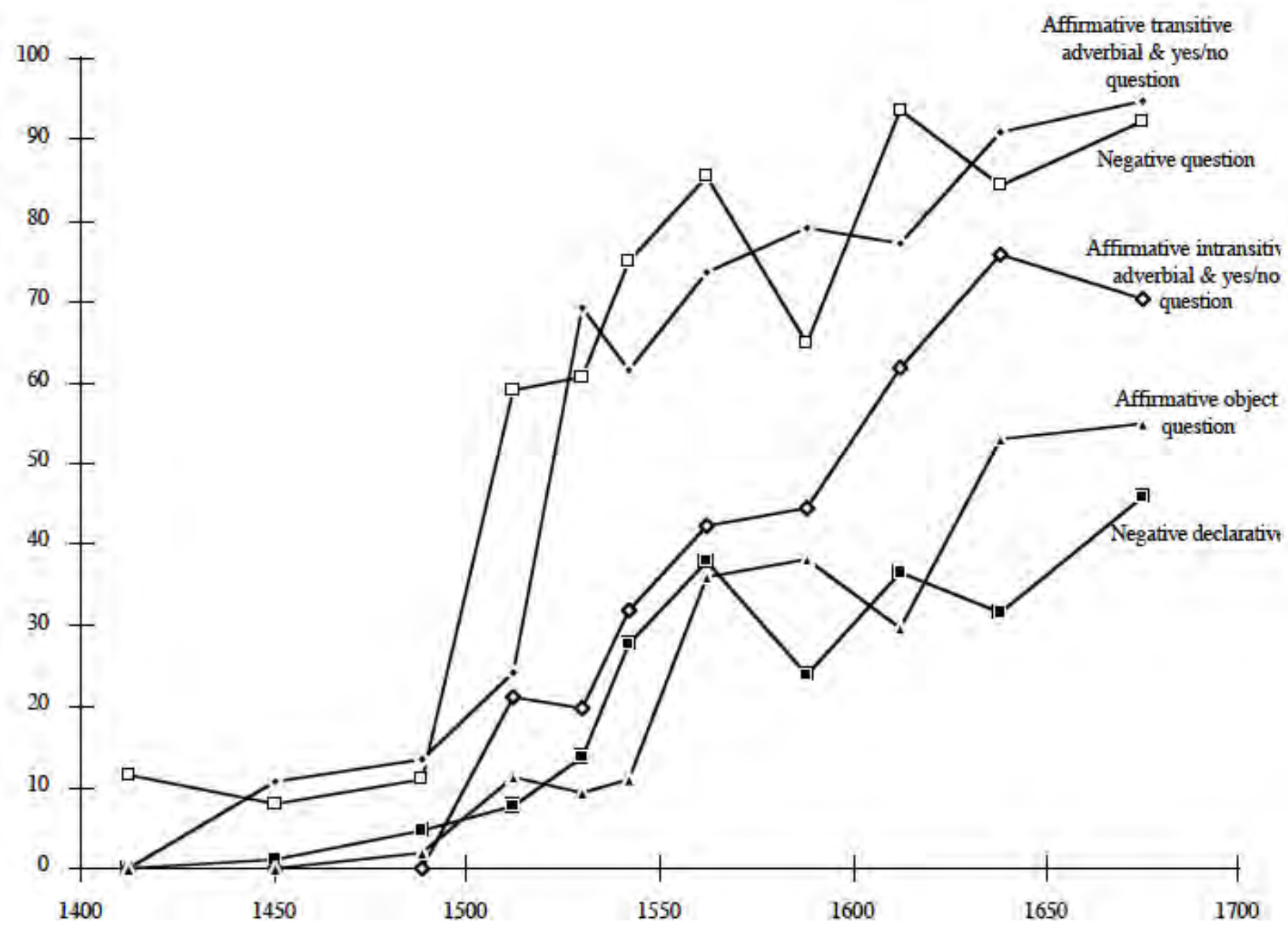

Figure 4.4: Frequency Data from Ellegård (1952, 166)

Figure 4.4 shows that individual contexts exhibit variation in use and illustrate that the frequency increases 'more rapidly in the more favored contexts', e.g., *affirmative transitive adverbial* and *yes/no questions* (Kroch, 1989a, 141). Thus, he suggests that while 'the relative frequencies of the competing forms slowly change [...] the advantage that one form has over the other may vary from one context to another' (Kroch, 1989a, 138). Kroch (1989a) further applies a statistical method to determine a rate of change for each context in order to compare them. This method, 'logistic transform', is not new; it is commonly used in mathematics and statistics to illustrate the rate of a given change. In math the average rate of change is rendered as a slope, where 'slope is the difference of the dependent variable divided by the difference of the independent variables' (Hill, 2011). The slope can further be used to identify statistical trends. For example, the steeper the slope, the faster the value is rising. However, the notion of slope was not used in historical linguistics until the introduction of the *Constant Rate Hypothesis* (Kroch, 1989c). First of all, Kroch translated the logistic function from (47a) into a logistic transform, a linear representation of an S-curve, illustrated in (47b):

(47)

$$p = \frac{1}{1 + e^{k+st}}$$

$$ln\frac{p}{1-p} = k + st$$

where $p$ is the probability of use of the specific linguistic form, $t$ is a variable that represents time, $k$ is the intercept (initial value) and $s$ is the slope (steepness) (Zuraw, 2003). The intercept provides a measurement of frequency for a new form when $t = 0$. While actual linguistic change has a beginning and end to its course, this statistical model can only approximate the actuation and the end of change - 'there is no time $t$ for which *p=0*, nor any for which *p=1*' (Kroch, 1989c, 204). In contrast, this formula models a diffusion of the new form over time - the slope signals the rate of change from one form into another (Kroch, 1989c).[5] Using this logistic transform, the *Constant Rate Effect* model shows that the new form is spread at the same constant rate in all contexts (Kroch, 1989c). Table 4.2 presents the results obtained from this method:

[5] Agresti (2007) describes an alternative formula:

$$log\frac{\mu}{t} = a + \beta x$$

where a dependent variable $Y$ has an index (time) equal to $t$, the rate is $Y/t$, the expected value of dependent variable Y is $\mu = E(Y)$, and the expected value of the rate (its probability distribution) is $\mu/t$ (Agresti, 2007, 82). This model can include other explanatory (independent) variables in constructing the rate of change, compared to Kroch's model that exclude such factors based on the theoretical assumptions of the *Constant Rate Effect* model.

| Negative declaratives | | Negative questions | | Affirmative trans. adv. and yes/no q. | | Affirmative intrans. adv. and yes/no q. | | Affirmative wh- object q. | |
|---|---|---|---|---|---|---|---|---|---|
| slope | intercept | slope | intercept | slope | intercept | slope | intercept | slope | intercept |
| 3.74 | -8.33 | 3.45 | -5.57 | 3.62 | -6.58 | 3.77 | -8.08 | 4.01 | -9.26 |

Table 4.2: Rate of Change (Slope) and Actuation of Change (Intercept) for Periphrastic *do* (Kroch, 1989c, 24)

The small variation between slopes in Table 4.2 shows no significant difference; therefore, these results provide evidence for the same rate of change in each context. That is, the syntactic contexts 'are merely surface manifestation of a single underlying change in grammar' (Kroch, 1989c, 199). Furthermore, Kroch analyzes various independent factors from the study by Oliveira e Silva (1982). Her data show that an increase of definite articles in possessive nouns in Portuguese is favored in several contexts, namely discourse factor (unique referent), morpho-syntactic factor (3rd person and object of a preposition) and semantic factor (kinship). The statistical significance of these factors was determined by the sociolinguistic program VARBRUL (the implementation and logistics behind this program will be discussed in more detail in section 4.3). First, Kroch fits the data into the logistic transform function. Figure 4.5 shows the frequency increase of definite articles in possessive nouns over time (white diamonds) and their statistical rate of change (black diamonds).

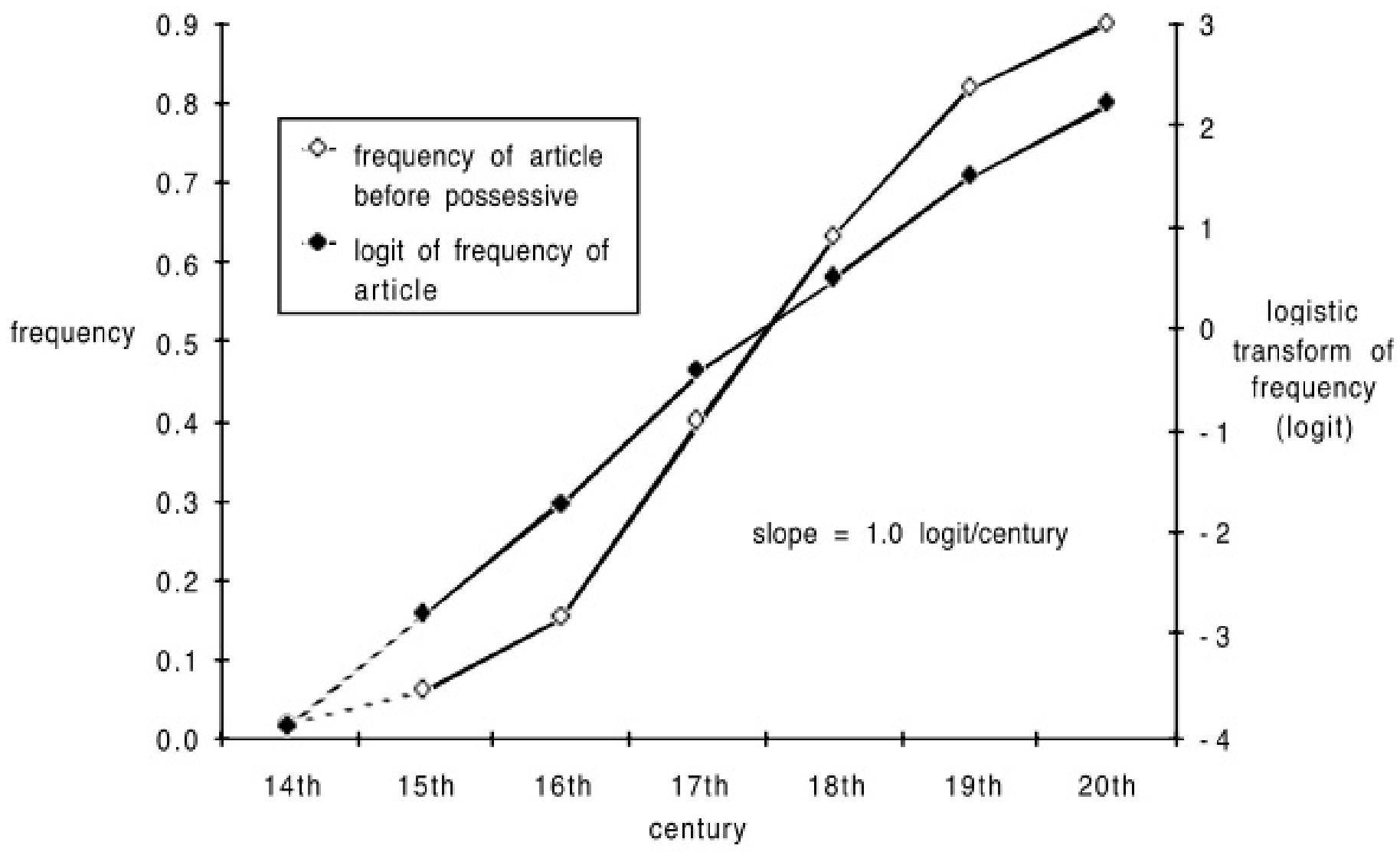

Figure 4.5: Rise of Definite Articles before Possessives in Continental Portuguese from Oliveira e Silva (1982)

Kroch further plots linear regression lines for each significant factor and demonstrates that the effect of these factors remains stable with slopes 'very close to zero', showing no change in time, as illustrated in Figure 4.6.

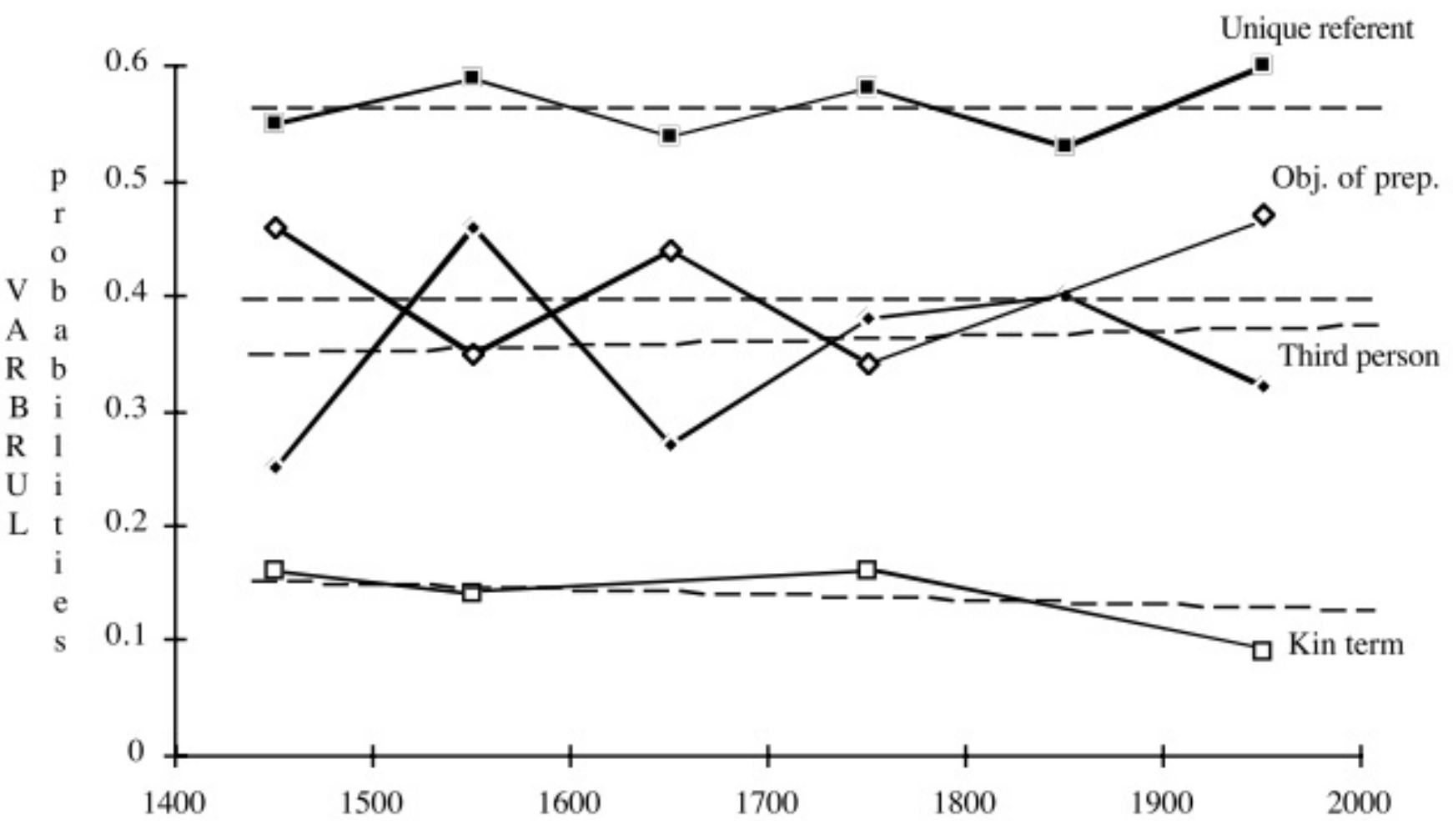

Figure 4.6: The Stable Effect of Factors over Time in the Use of Definite Articles in Continental Portuguese from Oliveira e Silva (1982)

From Figure 4.6 one can see that the slopes of dashed lines for each significant factor remain flat, i.e. their effect on the increase of definitive article does not change over time. That is, the definite article increases over time at the same rate across all these independent factors. Thus, Kroch formulates a hypothesis known as the *Constant Rate Effect*: 'change proceeds at the same rate in all contexts, and that, as far as one can tell, disfavoring contexts acquire new forms no later than favoring ones, though at lower initial frequencies' (Kroch, 1989c, 36). He argues that

a. "the pattern of favoring and disfavoring contexts does not reflect the forces pushing the change forward. Rather, it reflects functional effects, discourse and processing, on the choices speakers make among the alternatives available to them in the language as they know it; and the strength of these effects remains constant as the change proceeds. [...] None of these effects would have any privileged causal status" (Kroch, 1989c, 36).

The *Constant Rate Effect* has been confirmed in several recent studies. For example, Kallel (2005) examines the negative concord phenomenon in Early Modern English. Using the logistic function, Kallel shows that the rate of change (slope) is the same in different syntactic contexts, namely coordinated and non-coordinated constructions, as illustrated in Figure 4.7:

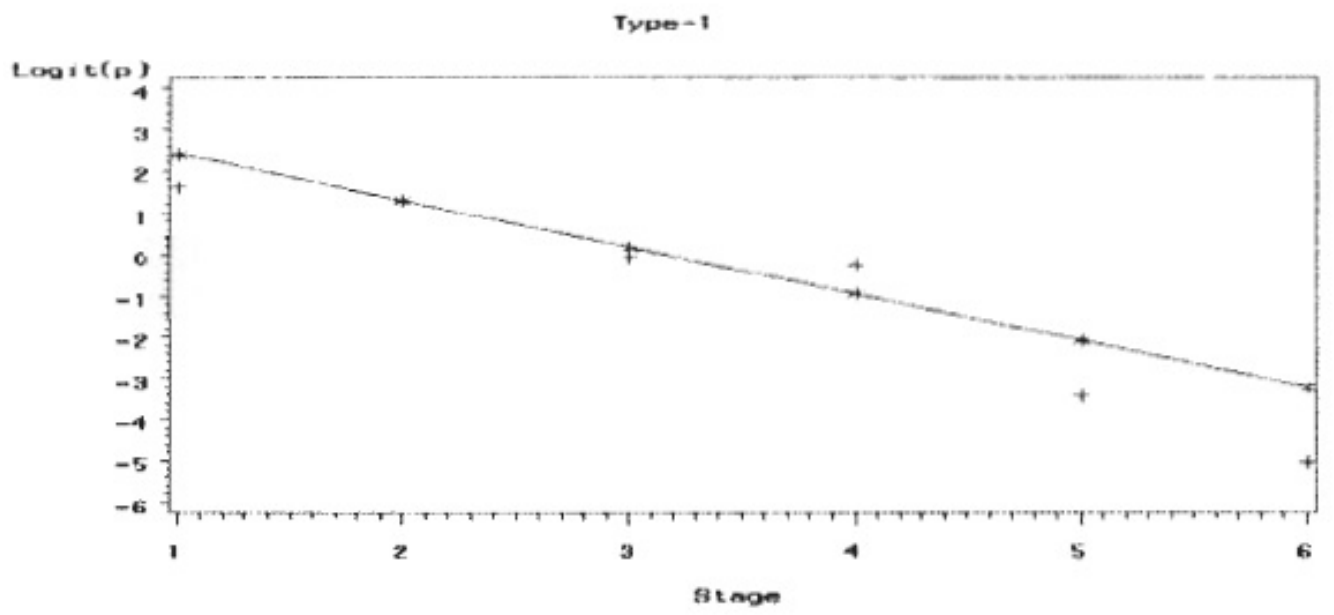

(a) Negative Concord in Non-coordinated Constructions

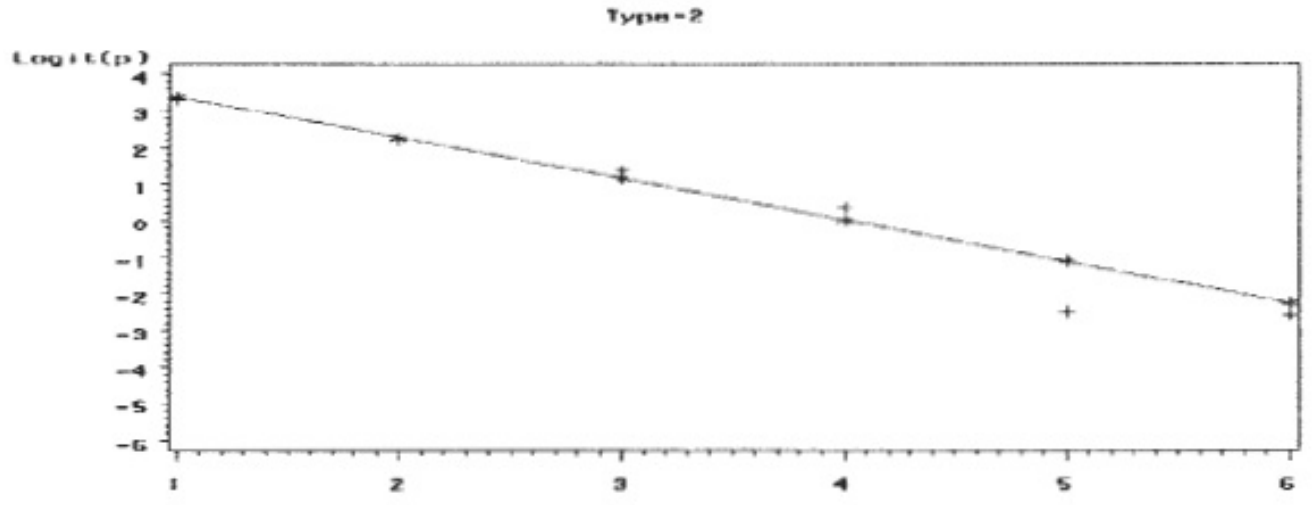

(b) Negative Concord in Coordinated Constructions

Figure 4.7: The Logit Transform of the Decline of Negative Concord (Kallel, 2005, 134)

Similarly, the *Constant Rate Effect* has been successfully tested in other languages: the change from verb-final (Infl-final) to verb-medial (Infl-medial) in Yiddish (Santorini, 1994), the change from OV to VO in Greek (Taylor, 1994) and Icelandic (Ingason, Sigurdsson and Wallenberg, 2012) as well as the loss of verb-second in Middle French (Kroch, 1989c). These studies demonstrate that different syntactic changes reflect the same underlying grammatical change. For example, Kroch (1989c) shows that three syntactic contexts, namely the loss of subject-verb inversion, the rise of left dislocation and the loss of null subjects, proceed at the same rate (slope), as illustrated in Figure 4.8. He suggests that these events describe the same grammatical change - the loss of verb-second constraint (see Hudson (1997) for a different analysis of this change[6]). In a similar fashion, Taylor (1994) shows that the rate of preposed and postposed NPs, weak pronouns and PP complements provides evidence for the underlying grammatical change from verb-final to verb-medial.

[6]Hudson (1997) compares two approaches: parameter resetting (Kroch, 1989c) and word-class change (Warner, 1993).

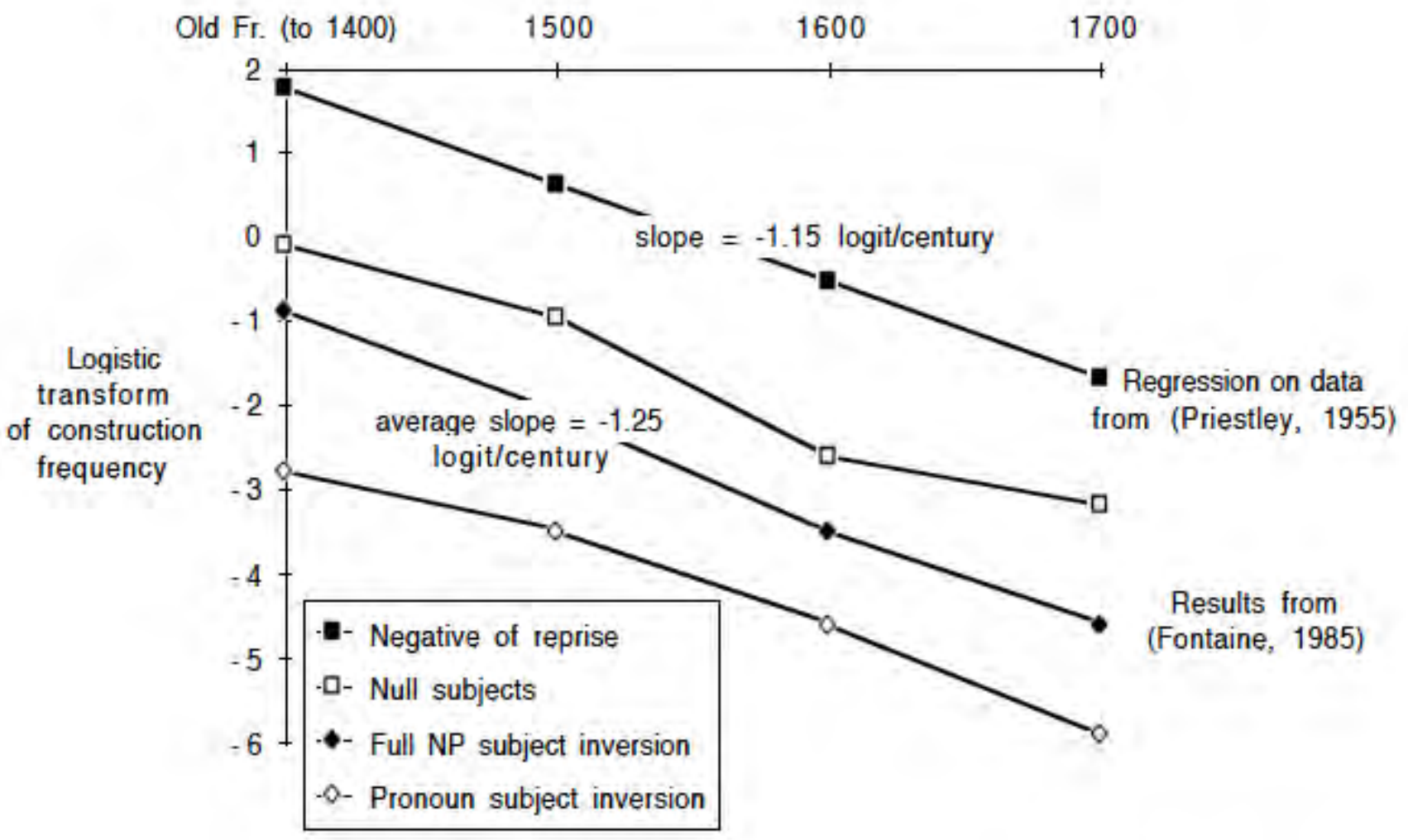

Figure 4.8: *Constant Rate Effect* in Middle French: Rate of Change in the Loss of V2 (Kroch, 1989c, 15)

In contrast, Pintzuk (2003) shows that in some cases of syntactic variation, the *Constant Rate Effect* does not hold. She states that the *Constant Rate Effect* 'holds only for contexts in which the surface forms are reflexes of the same underlying grammatical alternation' (Pintzuk, 2003, 515). It is generally accepted that an 'identical surface may be derived by different grammatical processes in different contexts' (Pintzuk, 2003, 515). Therefore, it is unlikely that we would observe the same rate for these unrelated contexts. One of such cases is the change of negation in infinitival clauses in French (Hirschbühler and Labelle, 1994). Hirschbühler and Labelle analyze the change from *ne*[7]+ infinitive + *pas/point*[8] (see 48a) to *ne pas* + infinitive (see 48b).

(48) a. Pour ce, (. . . ) est il bon de ne se haster point
for that, (...) is it good to not Refl. hasten-inf. neg.adverb
'For that, (. . . ) is it good not to hasten' (14th century, Hirschbühler and Labelle (1994, 6))

[7]Negative Particle
[8]Negative Adverb

b. ils estoient délibérés de ne point rendre le prisonnier
they were determined to not neg.adverb render-inf. the prisoner
'they were determined not to surrender the prisoner' (16th century, Hirschbühler and Labelle (1994, 17))

This change is examined in three contexts: lexical verbs, modals and auxiliaries. Figure 4.9 shows the results of logistic transform.

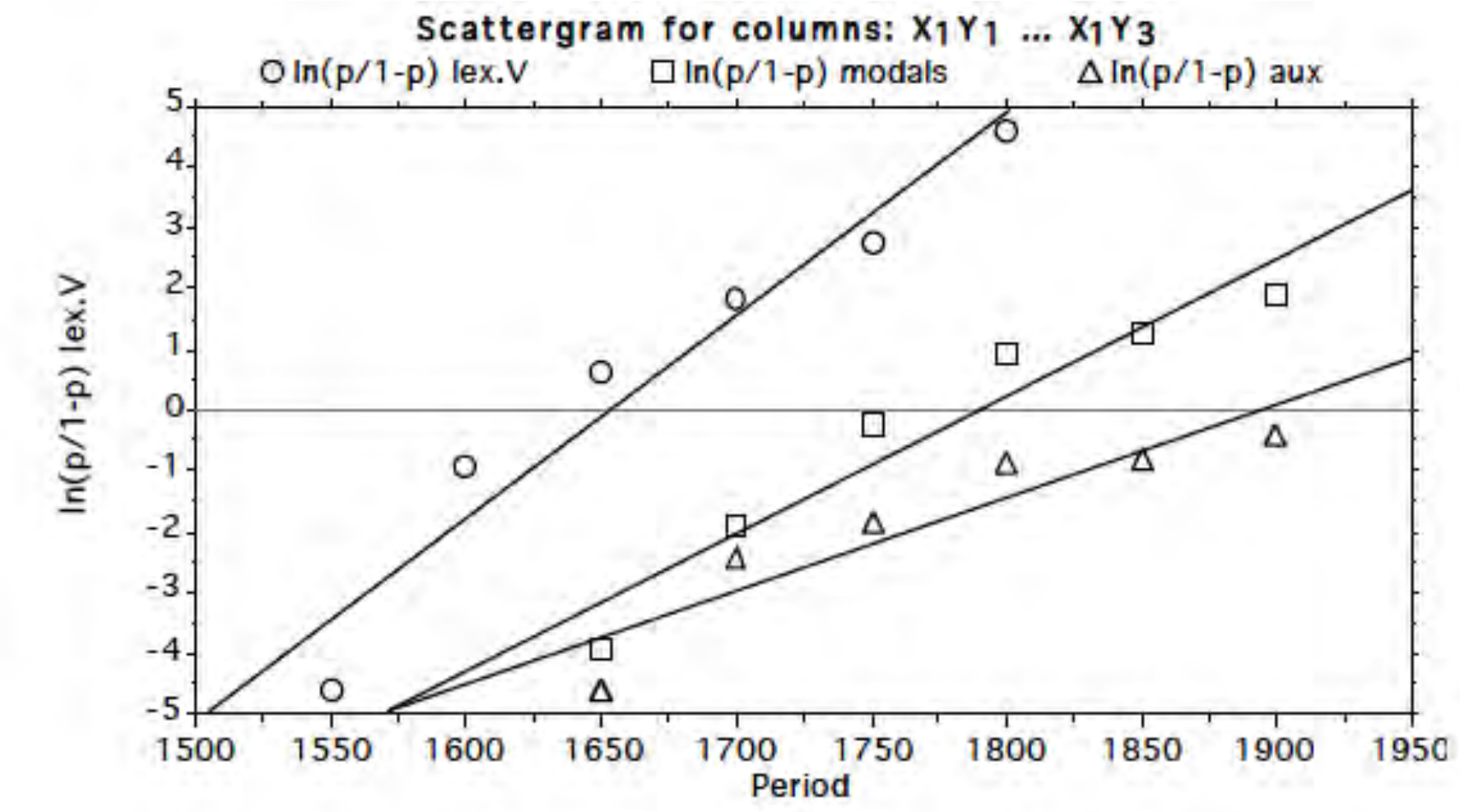

Figure 4.9: Logistic Transform of *ne pas* + Infinitive in Three Contexts (Hirschbühler and Labelle, 1994, 11)

From Figure 4.9 it is clear that the rate of change is different for lexical, modal and auxiliary verbs. This reveals that 'what appears to be a single grammatical change in three different contexts actually represents two separate changes: a change in the position of *pas*, and a loss of verb movement to T' (Pintzuk, 2003, 515). Furthermore, the *Constant Rate Effect* is related only to the frequency of the new form and 'does not consider the multiple constraints that may be operating on the variation' (Tagliamonte, 2011, 84). One of the main assumptions of this model is the existence of competing grammars. That is, speakers are able to acquire and produce two different grammars. Santorini (1994) displays evidence of two grammars in Yiddish, where she finds examples of verb-final (Infl-final) and verb-medial

(Infl-medial) in the same corpus.[9]

## 4.3 The Variable Rule Model

The foundation of the first multifactorial probabilistic model in linguistics was introduced in Labov (1969). Labov in his classic study on copula deletion observes that language variation is *inherent* and *systematic*. The observed linguistic variation is denoted as linguistic variable, namely 'two or more ways of saying the same thing' (Labov, 1972). Furthermore, this linguistic variable, e.g. the absence of *be* (see the example in 49), is influenced by a number of internal constraints: a) type of subject, b) type of predicate and c) type of preceding phonetic environment. These findings challenge previous accounts of language variation in which the variation was considered *free* or *optional* (Cedergren and Sankoff, 1974, 333).

(49) I know, but he wild, though (Labov, 1969, 717)

Labov observes that each constraint, or context, operates independently from other constraints and has a *fixed effect* that depends on the presence or absence of a given feature. Furthermore, the relative *weight* of each constraint contributes to 'the applicability of the rule' (Cedergren and Sankoff, 1974, 335). That is, the *Variable Rule* model is based not just on the proportion of a new form but on the combination of all environments in which the form occurs (Labov, 1982, 75), in contrast with the *Constant Rate Effect*, where the contexts do not play a role in the language shift. In addition, the sociolinguistic approach to language change offers 'a microscopic view of the grammar from which we can infer the structure (and possible interaction) of different grammars' (Tagliamonte, 2011, 9).

Thus, the key concept of the *Variable Rule* model is to incorporate linguistic variation into language systems through variable rules.[10] According to this model, 'the predicted relative frequency of a rule's operation is, in effect, an integral part of its structural description' (Cedergren and Sankoff, 1974, 334). These rules are interpreted in terms of 'quantitative re-lations which are the form of the grammar itself' (Labov, 1969, 759). The first probabilistic application of these rules was developed by Cedergren and Sankoff (1974), who showed that 'the presence of a given feature or subcategory tends to affect rule frequency in a probabilistic uniform way' (Cedergren and Sankoff, 1974, 336). Given that the features are independent, this can be expressed mathematically, as an additive formula (50a) or multiplicative formula (50b):

(50) a.

$$p = p_0 + p_i + p_j + ...$$

b.

[9]See Roberts (2007, 319-331) for arguments against competing grammars.

[10]It should be noted that this probabilistic model was met with resistance: 'the essence of this argument is the belief that the human mind cannot handle probabilities' (Kay and McDaniel, 1979, 154) (see also Cedergren and Sankoff (1974); Sankoff (1978)).

$$p = p_0 * p_i * p_j * ...$$

where $p_0$ denotes the input probability, and $p_i$ and $p_j$ represent the effects of the features $i$ and $j$ (Cedergren and Sankoff, 1974). These features may designate a condition, phonological or syntactic, or a tendency of a speaker or social group (Rousseau and Sankoff, 1978, 57). The additive approach (see 50a) posits that 'the total effect of an environment is the sum of the individual effects of the parts' (Naro, 1981, 68), and the multiplicative model (see 50b) multiplies these effects. The additive (50a) and multiplicative (50b) formulae have been criticized for several shortcomings,[11] which resulted in the development of the logistic model that became a core of the sociolinguistic statistical toolkit. The logistic model was developed by Sankoff (1975) as a necessary step to accommodate specifically linguistic data for which traditional statistical tools could not be applied (Sankoff, 1978; Rousseau and Sankoff, 1978; Sankoff, 1988; Mendoza-Denton et al., 2003).[12] The logistic model is a case of Generalized Linear Models (GLM) for binary dependent variable, illustrated in (51) (Agresti, 2007, 66):

[11]With a large number of factors, the models 'could produce percentages in excess of 100 per cent and below 0 per cent ' (Tagliamonte, 2006, 132). For more details see also Sankoff (1978); Sankoff and Labov (1979); Manning (2003).

[12]In its modern use, *variable rule analysis* 'pertains specifically to the probabilistic modeling', compared to the original usage, which implies rules (Sankoff, 1988, 984).

(51)

$$g(\mu) = a + \beta_1 x_1 + ... + \beta_k x_k$$

The right-hand side of the model defines *the linear predictor*, where $x_1 ... x_k$ are *explanatory variables*, or independent variables, e.g., time, speaker and type of clause. The $a$ value denotes the likelihood of the *application value*, namely the dependent variable that we are interested in, e.g. the presence or absence of the phenomenon (Roy, 2013, 262). The expected value (mean) of the dependent binary variable $Y$ is represented as $\mu$: $E(Y) = \mu$. Finally, the function $g$ (*link function*) links the expected value of $Y$ (its *probability distribution*) with the *linear predictor*. While for the continuous variable the *link function* $g(\mu) = \mu$, namely its mean, the binary variable requires a log of the mean: $g(\mu) = log(\mu)$. The formula (51) is rewritten as a loglinear model in (52) (Agresti, 2007, 67):

(52)

$$log(\mu) = a + \beta_1 x_1 + ... + \beta_k x_k$$

When the mean $\mu$ denotes the probability distribution, its value must be between 0 and 1. Therefore, the *link function* requires the log of odds (ratio of application and non-application values): $g(\mu) = log[\mu/(1 - \mu)]$. This *link function* is called a *logit link*, and the linear model is referred to as a *logistic regression model* (Agresti, 2007, 67):

(53)

$$log \frac{\mu}{1 - \mu} = a + \beta_1 x_1 + ... + \beta_k x_k$$

This logistic regression model was implemented in the *Variable Rule* program VARBRUL, which was consequently replaced by an improved version Goldvarb X (Sankoff, 1975; Sankoff et al., 2005). The *Variable Rule* model entails the independence of factors and tokens (Sankoff et al., 2005; Roy, 2013). However, it has been argued that linguistic variables are rarely independent and that 'many potential predictors are in a nesting relationship with speaker or word' (Johnson, forthcoming). In addition, the categories of the *Variable Rule* model must be discrete, and factor groups with 100% or 0% must be excluded.[13] Thus, to improve the sociolinguistic variable model and to allow for non-discrete predictors, the mixed-effects model has recently been introduced into the sociolinguistic field. It has been shown that the mixed-effect model 'returns more accurate *p*-values compared to a fixed-effects model that ignores nesting' (Gorman and Johnson, 2013, 223). Several recent sociolinguistic studies have sought to extend the traditional variable model Goldvarb with fixed effects to the models allowing for mixed and random effects, e.g., Rbrul (Johnson, 2009) and the *glm* in R and Stata (Agresti, 2007, 67).[14]

Finally, there has been growing interest in using new statistical tools and methods such as random forests, conditional inference trees and correspondence analysis to enhance the *Variable Rule* model and improve its limitations (Tagliamonte and Baayen, 2012; Greenacre, 2007). These methods will be described in more detail in Chapter 5.

## 4.4 The Usage-Based Model

With the growing number of large corpora and the availability of statistical tools, frequency and its role in the language use, suggested first in Zipf (1935) and Greenberg (1963), has been revisited. Corpus studies 'reveal quantitative patterns that are not available to introspection but that are likely to be important to the understanding of how speakers store and access units of language' (Bybee, 2006, 7). In the *Usage-Based* model, the memory stores what speakers hear and produce. Stored elements are further organized with various *linkages* among themselves. These linkages display a different strength and also change over time (Hopper, 1987; Bybee, 2007; Sabino, 2012). In fact, recent psycholinguistic studies (Ju-rafsky, 2003; Tomasello, 2000) provide evidence that 'language users store virtually every linguistic token they encounter' (Bod, 2006, 292). In this view, these linguistic tokens are mapped into *exemplars*, namely clusters for categorization, classification or analysis (Bod, 2006, 293). This model also distinguishes between two types of frequency: *token* and *type*. *Token* frequency refers to 'the number of times a unit appears in running text' (Bybee, 2007, 9). Any particular sound, word, phrase or sentence can be counted as a unit. In contrast, *type* frequency refers only to patterns of language (Bybee, 2007, 9). For example, the tokens *grew, knew* can be associated with the type of *past tense* patterns or the tokens *sell, give* can refer to the type of *ditransitive* patterns (Bybee, 2006; Goldberg, 1995). In fact, Smith (2001) compares both *token* and *type* frequency in his diachronic study of English anterior aspect. He examines the distribution of auxiliaries, e.g. *I am gone* and *I have gone*. While their *token* frequencies are similar, they differ in *type* frequencies - the auxiliary *have* shows a higher *type* frequency. Smith suggests that *type* frequency is a better predictor to identify a winning form in a competition between two forms than *token* frequency (Bybee 2001, 6). While early usage-based studies were interested in phonological and morphological changes, recent studies have shown the validity of usage-based approaches for grammatical variation and change. For example, Waltereit and Detges (2008) argue that the origin of syntactic change is in language usage. Furthermore, Langacker (2000) points out that constructions are seen as dynamic symbolic conventions and they are constantly updated by language use. In this view, grammar 'includes knowledge of probabilities of syntactic structures' (Gahl and Garnsey, 2004). That is, there is a bi-directional relation between language use and language grammar: 1) the grammar provides the knowledge for language use and 2) the language use 'redefines the language system in a dynamic way' (Tummers et al., 2005, 228).

[13]For the calculation of factor weights for each independent factor, the Goldvarb program implements *Iterative Proportional Fitting (IPF)*. This algorithm treats data as a contingency table, and therefore the cells are required to have non-zero values (Roy, 2013, 264).

[14]For more details on mixed effects and random effects and comparison between Goldvarb and statistical logistic regression see Johnson (2009); Gorman and Johnson (2013) and Roy (2013).

In recent years we have seen many advances in the field of usage-based models. Early methods relied only on raw frequency and token percentages. While many agree that the memory and cognitive processes involved in storage and production are not *linear*, the use of frequency implies a linear relation between the frequency of *tokens* in a *context* and the total frequency of *tokens* (Gries, 2012, 488). Furthermore, the traditional usage-based *bivariate analysis* consisting of only two factors (one dependent and one independent) simplifies linguistic reality (Tummers et al., 2005, 243). Recently, more studies have turned to *multivariate* and *multidimensional* approaches, (for example, Jenset (2010), McGillivray (2010) and (Claes, 2014)). That is, 'linguistic/constructional knowledge is conceived of as a high-dimensional space with formal (phonetic, phonological, morphological, syntactic, ...) and functional (semantic, pragmatic, discoursal, contextual, ...) dimensions' (Gries, 2012, 504). This model constructs high-density multidimensional exemplars. To accommodate these complex models, new statistical methods have been introduced into usage-based linguistics, namely multivariate analysis, collostructional analysis and clustering analysis. For multivariate analysis, such as linear and logistic models, see section 4.3, which describes these methods in detail. In this section, I will focus on collostructional analysis and cluster analysis.

The collostructional analysis is a 'quantitative approach to the syntax-lexis interface' (Gries, 2012, 477). Consider for example the following paradigm (Gries and Stefanowitsch, 2004, 97-98):

(54) a. John sent Mary the book

b. John sent the book to Mary

The example in (54a) illustrates a *ditransitive* expression and the example in (54b) demonstrates a *to+dative* construction. The two sentences in (54) are semantically equivalent expressions or *alternating pairs* (Gries and Stefanowitsch, 2004, 97). Such constructions are traditionally described in terms of semantic, functional or formal similarities. The traditional methods, however, are not able to analyze statistically the linguistic context and its interactions. Gries and Stefanowitsch (2004) have developed a statistical method that measures the association strength between constructions based on their context. To demonstrate the usefulness of this approach in the field of corpus and usage-based linguistics,[15] Gries and Stefanowitsch examine the *ditransitive* construction (54a) and *to+dative* construction (54b).

[15] The collostructional analysis was met with some criticism (see Bybee 2010). See also the reply to the criticism in Gries (2012).

While these constructions may seem to be semantically equivalent, there exist certain restrictions on verbs and NPs. The calculation is based on four components: 1) the lemma frequency of a token in construction A, 2) the lemma frequency of a token in construction B, 3) the frequency of construction A with other tokens and 4) the frequency of construction B with other tokens (Gries and Stefanowitsch, 2004, 102). Table 4.3 provides an example of this calculation for the verb *give*:

| | **give** | **Other verbs** | **Total** |
|---|---|---|---|
| **Ditransitive** | 461 (213) | 574 (822) | 1,035 |
| **To-Dative** | 146 (394) | 1,773 (1,525) | 1,919 |
| **Total** | 607 | 2,347 | 2,954 |

Table 4.3: The Distribution of *give* in Two Constructions (Gries and Stefanowitsch, 2004, 102)

The numbers in parenthesis denote expected frequencies, which show that the actual frequency of *give* in ditransitive construction is twice higher than in to-dative construction. The distinctiveness index is then calculated using the $p$-value of the *Fisher-Yates exact test.* This distinctiveness index is calculated for all the verbs in ditransitive and to-dative constructions in the corpus. As a result, the authors observe that ditransitive constructions demonstrate a preference for verbs 'encoding a direct contact between agent and recipient', such as *give*, *show*, *tell*, whereas to-dative constructions prefer verbs involving 'some distance between agent and recipient', such as *bring*, *take*, *pass.* Gries and Stefanowitsch further demonstrate this method with active-passive alternating pairs and verb-particle constructions, as in *John picked up the book* or *John picked the book up.* Several recent studies have also used this approach to consider various diachronic syntactic variations and language changes. For example, using the collostructional analysis, Hilpert (2006) identifies three stages of diachronic development for *shall* in English: 1) 1500-1640 - a strong association with *obligation*, 2) 1640-1780 - a shift toward *intention* and 3) 1780-1920 - a change toward *promise* or *willingless to comply* (Hilpert, 2006, 254-255)

The second innovative method associated with the usage-based approach is cluster analysis. The cluster approach allows researchers to address issues of subjectivity and overgeneralization (Gries and Hilpert, 2008). One of the key issues in corpus linguistics is internal and external variability in corpora (Gries, 2006, 110). The evaluation of variability can be a very hard task, as the variability might occur not only between hierarchical level, e.g. registers, but also within the same level. In addition, there might be a various degree of variability on every level, which makes it difficult to categorize data into groups. The cluster analysis provides the researcher with an 'data-driven and objective categorization of the groups' (Gries, 2006, 129). In addition, this method considers both the homogeneity and the variation of data. Consider for example the diachronic study of *shall* in English (Hilpert, 2006). The data from the Early and Late Modern English corpora are traditionally grouped into three periods: 1500-1640, 1640-1780 and 1780-1920. By applying a cluster analysis, the periods are grouped as tree branches according to their similarity or differences. The dendrogram representation (tree branching graph) in Figure 4.10 shows that the data are grouped into two 180-year periods. This cluster analysis represents a hierarchical relationship between periods.

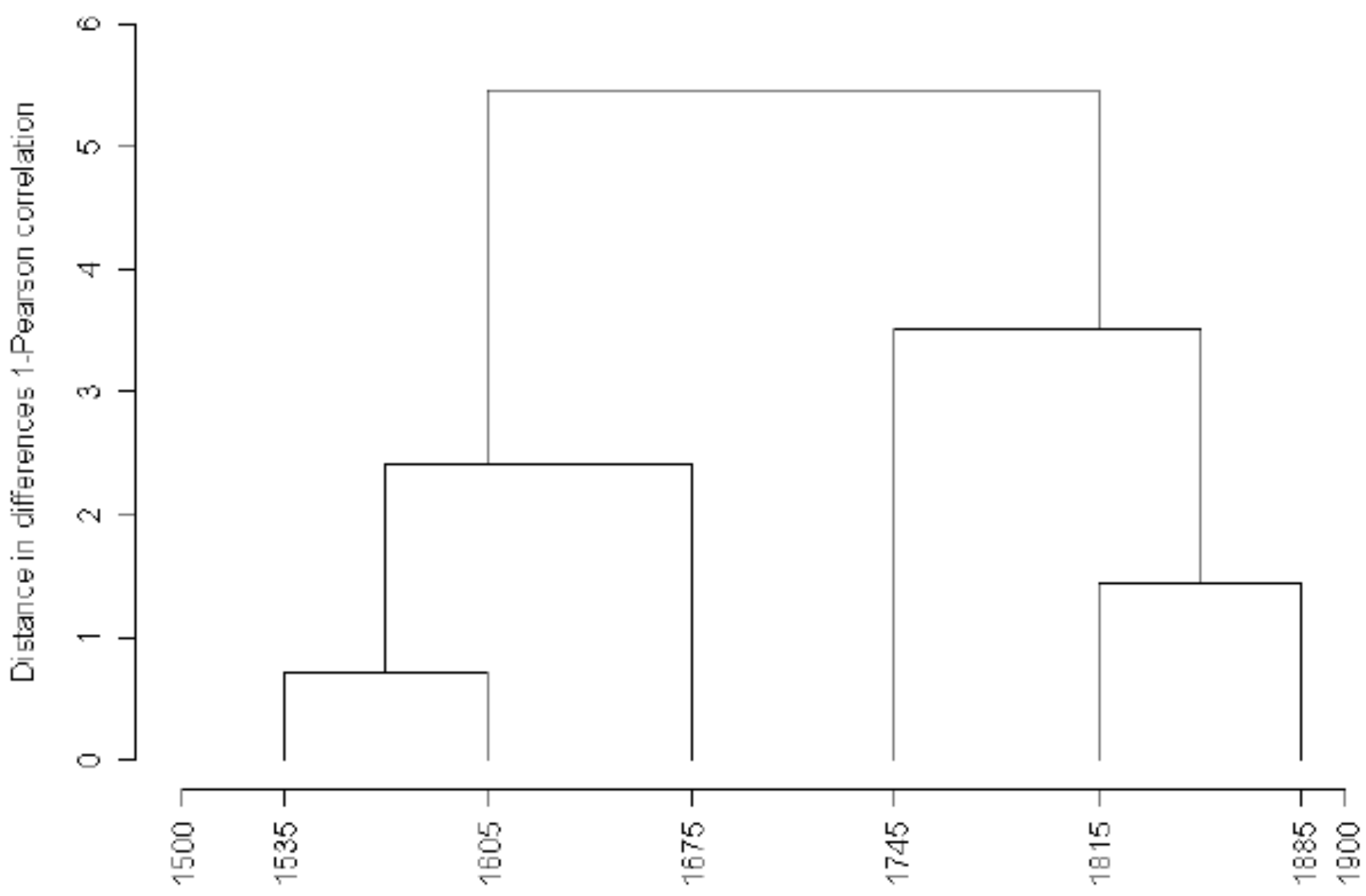

Figure 4.10: The Dendrogram of Six Periods in Hilpert's Study (Gries and Hilpert, 2008, 69)

Figure 4.10 clearly illustrates two groupings instead of three: 1500-1710 and 1710-1920.[16] Finally, the cluster analysis can display data on a multidimensional level. For example, Passarotti et al. (2013) examine word-order patterns in Latin and identify interaction between patterns and different Latin authors:

[16] 1710 is a half point between 1675 and 1745.

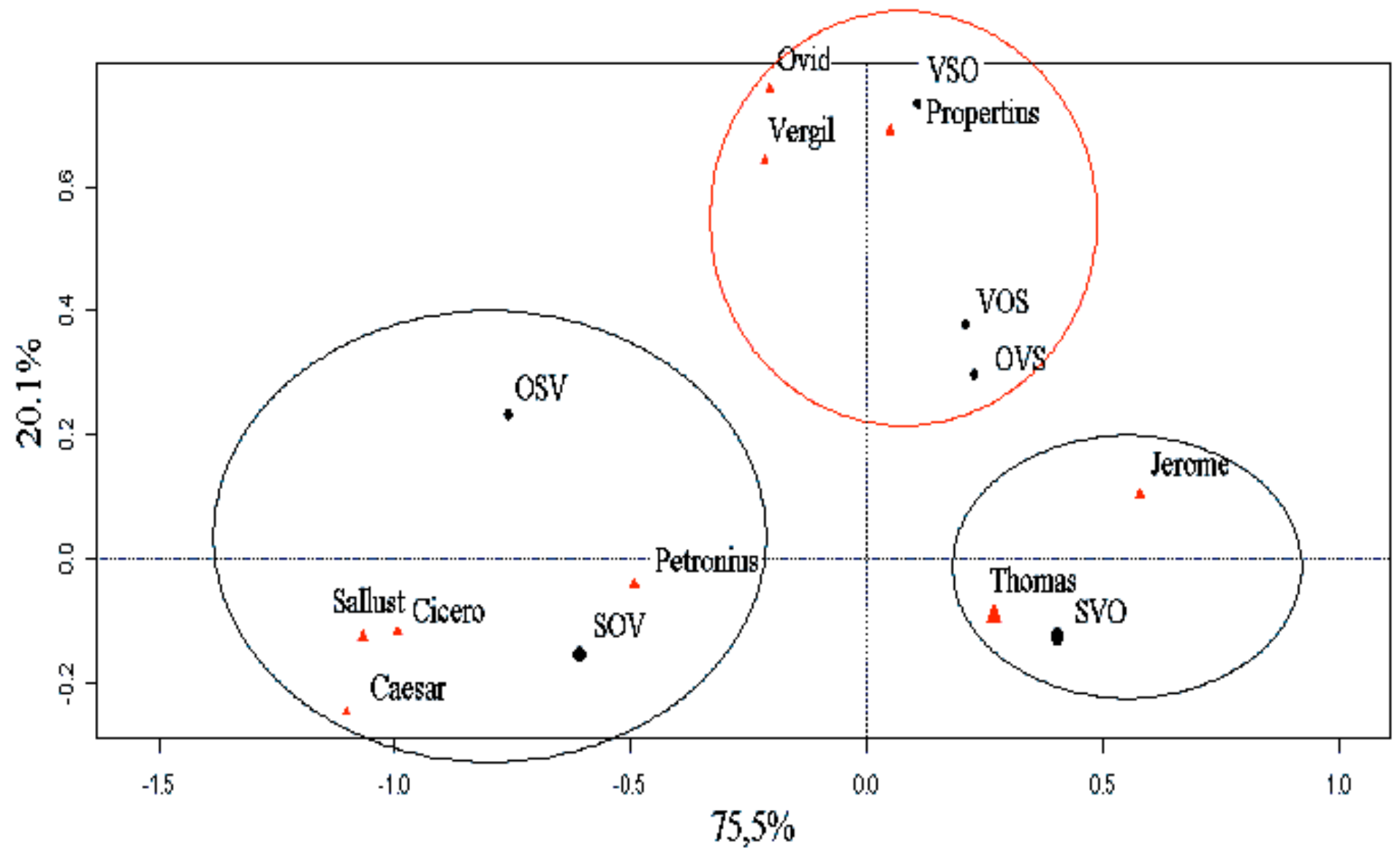

Figure 4.11: Correspondence Cluster Analysis of Six Word-Order Patterns and Latin Authors (Passarotti et al., 2013, 347)

Figure 4.11 shows three clusters: 1) Caesar, Cicero and Sallust; 2) Ovid, Propertius and Vergil and 3) Jerome and Thomas. First, there is a clear association between clusters and different genres - the first group is prose and the second group is poets. Petronius represents both types of writing and is positioned slightly away from the central tendency. Finally, there is a clear grouping by period - Late Latin (Jerome and Thomas) is clustered separately from Classical Latin (Caesar, Cicero, Sallust, Ovid, Propertius and Vergil). In addition, Jerome and Thomas show preference toward an SVO order, as compared to other clusters.

## 4.5 Summary

In the last two decades, probabilistic modeling and language processing have emerged in various areas, such as cognitive science, psycholinguistics and computational linguistic. My intention in this chapter has been to cross-examine various probabilistic models and their application to language change. While each model seems to be bound to a certain theoretical framework, it has been shown that the synthesis between usage-based and sociolinguistic approaches is suitable for phonological variation and change (Clark, 2009). My goal is to show the usability of joint probabilistic methodological application in the study of a diachronic syntactic variation, allowing for a bird's-eye view and a micro-perspective. In the initial explorative stage, the overall picture of the data can be obtained by means of the rate of change (section 4.1) and cluster analysis (section 4.4). The rate of change will show us how the change is spread over time, whereas cluster analysis can be used to identify similarities or differences between word order patterns and chronological periods, genres or authors, as well as the relations between lexical frequencies and grammatical constructions. In the second stage, I will examine how the linguistic change is influenced by contextual internal and social external environments. First, I will obtain visual summaries of variable distributions. Second, by using the logistic mixed-effect model (section 4.3), I will examine the interplay between century, word order and various pragmatic, semantic and syntactic factors. That is, the combination of various methods will provide a multifaceted view of the phenomenon of word order change.

# Chapter 5

# Data and Methodology

*Nothing is more thrilling than*

*to dip into a corpus and go exploring.*

*(Tagliamonte, 2011, 116)*

In recent years, several annotated corpora of Latin and Old French have become available. While the existing corpora are a valuable resource in word order studies, the chronological lacunae in these corpora present challenges for diachronic studies. One of the contribution of this study is to provide a novel resource-light method for a corpus linguistic study by merging existing corpora with additional non-annotated resources. While any corpus can be annotated manually, corpus linguistics offers various state-of-the-art methods for semi-automatic annotation. These methods are presented in section 5.1. Furthermore, this chapter considers various issues with respect to research design, such as text selection, search tools, corpus size and statistical modeling. First, text selection for corpus study depends to a great extent on the availability of annotated data, such as treebanks or plain texts that could be processed. While a lot of texts can be found in scanned-image format, such data cannot be utilized in machine processing.[1] Second, data formats are not uniform across annotated corpora, and each corpus requires a specific tool or conversion as well as a specific query language. Only a few recent projects, e.g., HamleDT (Zeman et al., 2012) and

[1]It is well recognized that the OCR conversion produces errors, especially with Old languages, and thus, such texts first require manual correction.

Universal Dependency Treebank (McDonald et al., 2013), The Sustainable Interoperability for Language Technology (SILT),[2] as well as CLARIN infrastructure,[3] are able to overcome this limitation. The third question is related to the notion of data representativeness. Given data availability, what would be the ideal number of syntactic tokens for each period to build a representative sample of the language? On the other hand, it has been recognized that no corpus would be an absolute representation of the language we are investigating. This uncertainty in the data will inevitably obstruct the generalization about the language, unless we apply statistical modeling. As McGillivray (2010, 8) points out:

> “we can choose the statistical techniques in such a way that uncertainty is maximally reduced in our data and it is as safe as possible to generalize the results to the whole population”.

Finally, while the aim of this study is to investigate the change from OV to VO in infinitival clauses, it is equally important to examine the various contextual and pragmatic factors that may influence the data or play a role in this change.

This chapter is structured as follows. Section 5.1 discusses the existing annotated resources for Latin and Old French. Subsequently, this section elaborates on corpus linguistics methods to compile the present corpus from existing resources and to build additional resources via a resource-light approach. Furthermore, this section reviews query tools used to extract data, namely CorpusSearch 2.0, Tiger Search and ANNIS, and introduces the present Latin and Old French corpus. Section 5.2 describes dependent and independent variables used for statistical analysis. Finally, section 5.3 provides an overview of analytical procedures applied to the data analysis.

[2] `http://www.anc.org/LAPPS/SILT/index.html`

[3] `http://clarin.eu/`

## 5.1 Corpus

### 5.1.1 Existing Treebanks

There exists a large body of open-access digital editions for Latin and Old French texts, including the Latin Library,[4] Perseus Digital Library,[5] Biblioteka Augustana,[6] and many others. Such non-annotated corpora are limited to search for words or word sequences, which are valuable in lexical studies. However, for syntactic word order studies, it is more important to have access to morpho-syntactically annotated resources, allowing for identification and comparison of word order patterns. In recent years, there has been a continuous increase in the development of annotated corpora and more sophisticated search tools. Despite the advances in Natural Language Processing (NLP), Latin and Old French have remained behind, as compared to modern languages (for a review of digital resources available for Latin see Passarotti (2010) and for Old/Medieval French see Guillot et al. (2008)). At present, there exist three open-access Latin treebanks: i) Latin Dependency Treebank LDT (Bamman and Crane, 2011), ii) PROIEL[7] (Haug and Jøhndal, 2008) and iii) Index Thomisticus IT-TB[8] (Passarotti, 2009). The size of these treebanks is around 50 000 tokens for LDT, 100 000 for PROIEL and 80 000 tokens for IT-TB. The first two treebanks include several texts from Classical Latin and two texts from Late Latin, whereas IT-TB focuses on Medieval Latin. Since the present study concentrates on Classical and Late Latin, only two treebanks, LDT and PROIEL, will be used to collect data. In addition, there is another annotated resource, Opera Latina (LASLA);[9] it is, however, limited to token search by lemma and morpho-syntactic tag. With respect to Old French, three treebanks are available for diachronic studies: i) MCVF (Martineau et al., 2007), ii) Nouveau Corpus d'Amsterdam NCA (Stein, 2011) and iii) Base de Français Médiéval BFM.[10] Tables 5.1 and 5.2 provide a summary of texts and treebanks used in the present work. Verse texts are excluded from the Latin corpus, as it is well known that 'word order in poetry is often distorted' (Viti, 2010, 39) and that discontinuous word order is very common in Latin verse.

[4] http://www.thelatinlibrary.com/
[5] http://www.perseus.tufts.edu/hopper/
[6] http://www.hs-augsburg.de/~harsch/augustana.html
[7] http://www.hf.uio.no/ifikk/english/research/projects/proiel/
[8] http://itreebank.marginalia.it/
[9] http://www.cipl.ulg.ac.be/Lasla/descriptionop.html
[10] http://bfm.ens-lyon.fr/

Table 5.1: Latin Corpus

| Period | Text | Corpus |
|---|---|---|
| 50 BC | Caesar (Bello Gallico) | LDT |
| 50 BC | Sallust (Catilina) | LDT |
| 60AC | Satyricon | LDT |
| 384AC | Egeria | LDT/PROIEL |
| 390AC | Vulgate | LDT/PROIEL |

Table 5.2: Old French Corpus

| Period | Text | Corpus |
|---|---|---|
| 1090/1100 | Chanson de Roland | MCVF |
| 1170 | Tristan et Iseult | NCA |
| 1177 | Chrétien de Troyes | NCA |
| 1180 | Marie de France | MCVF |
| 1220 | La Queste de Saint Graal | MCVF |
| 1283 | Roisin | MCVF |
| 1309 | Jean Joinville | MCVF |
| 1370 | La prise d'Alexandrie | MCVF |

While all these treebanks provide a syntactic layer, their format and labeling are different. In addition, each treebank requires a different search tool, specific to the format. The MCVF corpus is a constituency treebank, which follows a Penn-Treebank format. An example from the MCVF corpus is illustrated in (55):

(55) a. `((IP-MAT (NP-SBJ (D Li)`

```
                    (NCS reis)
                    (NP-PRN (NPRS Marsilie)))
          (VJ esteit)
          (PP (P en)
                    (NP (NPRS Sarraguce)))
          (PONFP .)) (ID ROLAND,2.12))
```

b. Li reis Marsilie esteit en Sarraguce
the king-nom.sbj Marsile was-3p.sg in Saragossa
'King Marsile was in Saragossa' (Roland, MCVF ID ROLAND,2.12)

The MCVF format is designed to be used with *CorpusSearch* 2.0.[11] This command line program can search through tree nodes by means of a specific query file. The following example illustrates a query that searches all IP nodes for infinitival clauses (IP-INF) that immediately dominate direct objects (NP-ACC). These direct objects do not dominate pronouns (PRO); that is, direct objects must be a nominal phrase.

(56)
```
node: IP-MAT*
query:  (IP-INF* iDominates NP-ACC*)
                AND (NP-ACC* iDominates !PRO*)
```

The remaining corpora, namely NCA, PROIEL and LDT, are all available in TIGER-XML format. An example of the sentence in (57a) is illustrated as a TIGER-XML format in (57b) and as a tree graph in Figure 5.1:

(57) a. ostendebantur iuxta scriptura
pointed-3p.pl according scriptures-acc
'they were pointing according to scriptures' (Egeria, PROIEL s57394)

b.
```
<nt lemma="ostendo" pos="V-" person_number="3p"
tense_mood_voice="iip"
case_number="-p" gender="-" degree="-" strength="-" inflection="i"
word="ostendebantur" id="p766469">
```

[11]http://corpussearch.sourceforge.net/

```
            <edge idref="w766469" label="--"/>
            <edge idref="p766470" label="adv"/>
</nt>
<nt lemma="iuxta" pos="R-" person_number="--"
tense_mood_voice="---"
case_number="--" gender="-" degree="-" strength="-" inflection="n"
word="iuxta" id="p766470">
            <edge idref="w766470" label="--"/>
            <edge idref="p766471" label="obl"/>
</nt>
<nt lemma="scriptura" pos="Nb" person_number="-p"
tense_mood_voice="---"
case_number="ap" gender="f" degree="-" strength="-" inflection="i"
            word="scripturas" id="p766471">
            <edge idref="w766471" label="--"/>
</nt>
```

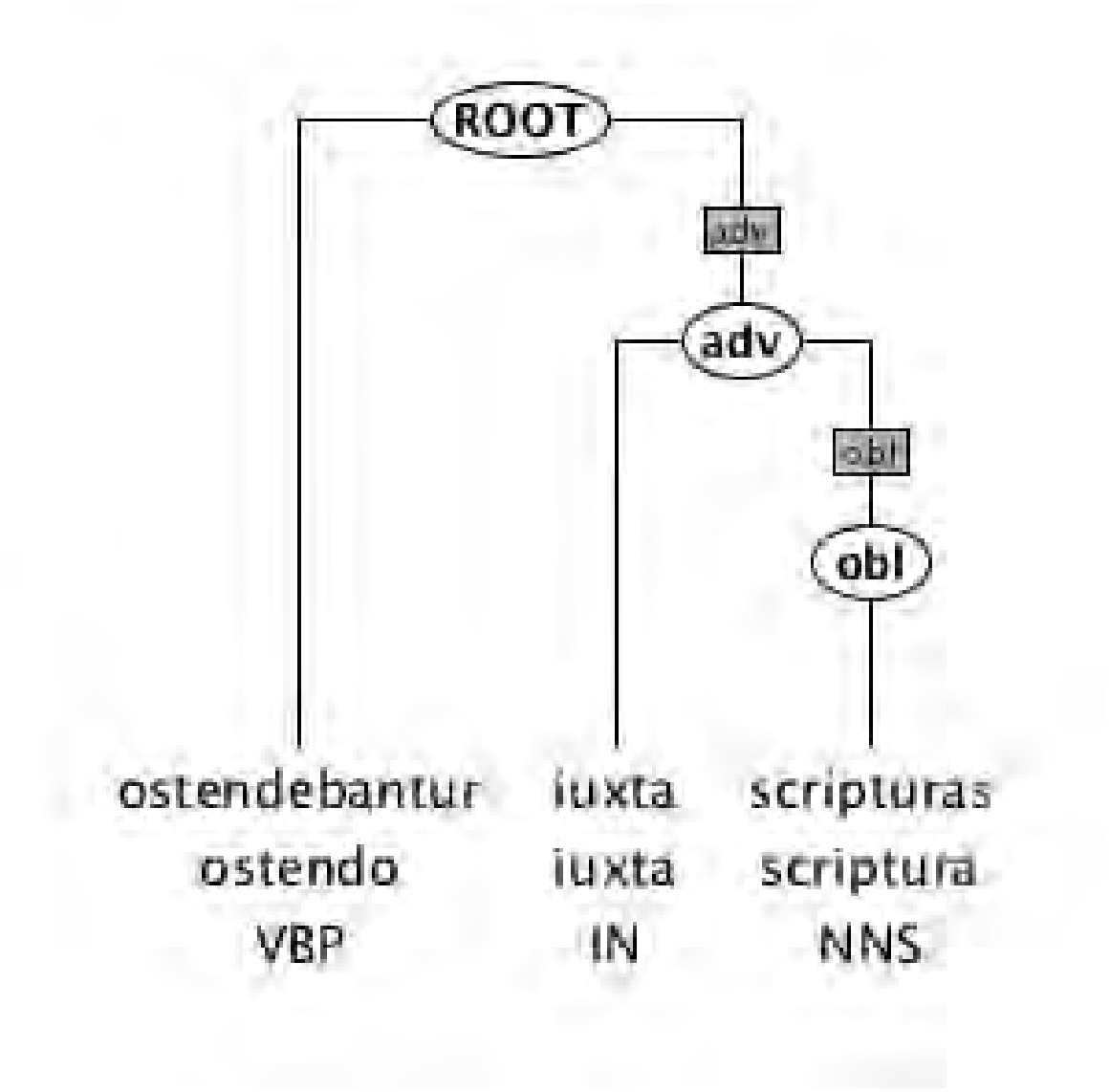

Figure 5.1: Tree Graph in TigerSearch

This format is supported by the *TigerSearch* tool (König and Lezius, 2003). The query from *CorpusSearch* (56) can be rewritten in *TigerSearch* as follows.[12]

(58)
```
#n0:[cat="IP"]>INF #n1:[cat="IP"] &
#n1>ACC #n2:[cat="NP"]&
#n2>[pos!="PRO"]
```

In (58), the query is looking for any IP node that dominates an infinitival clause (INF). The INF must dominate direct objects (ACC), where the direct object is not a pronoun (PRO).

### 5.1.2 Additional Resources

While any corpus can be annotated manually, corpus linguistics offers different methods of automatic annotation. One such methodology consists of using a pre-existing annotated corpus. In this approach, a morpho-syntactically annotated corpus is used as a training corpus to build a model for automatic annotation. Since the main focus of this study is infinitival clauses, it is essential to be able to recognize infinitive from other constituents, for example, noun or matrix verb. This type of annotation is traditionally performed by a part-of-speech (POS) tagger. The POS tagger, a type of software designed for learning and assigning POS tags, calculates probabilities for each tag with respect to the position of other tags in the corpus, for instance, the likelihood that D (determiner) is followed by N (noun) or D is followed by V (verb). The tagger further extracts these probabilities by analyzing a tagged corpus. It is important to train a tagger model on the data similar to the one that one must annotate in order to minimize tagging errors. Finally, before the application of the tagger model the raw data must be pre-processed by splitting tokens and placing them one token per line. While there are many available POS taggers, TnT seems to be a very efficient statistical tagger, showing a high performance as compared to other taggers (Brants, 2000a). In what follows, I will describe TnT model training and evaluation for each language separately.

[12] While the labels and part-of-speech tags are different in each corpus, the structure of the query is similar.

For Old French, I use a TnT model trained on 28 265 sentences from the Medieval French section of the MCVF corpus (Scrivner and Kübler, 2012). The trained model is then used to POS tag three Early Old French texts,[13] namely *Saint-Léger*, *Passion de Clermont* and *La vie de Saint Alexis* and one Early Old Provençal text, namely *Boece*.[14] The following example illustrates how a sentence from *Saint-Léger* (59a) is annotated with POS tags (59b):

(59) a. Domine Deu devemps lauder,
Lord-acc.obj God-acc.obj must-1p.pl praise-inf
'We must praise Our Lord God,' (*Saint Leger*, 1.1)

b. Domine NPRS *(proper noun singular)*

Deu NPRS *(proper noun singular)*

devemps VJ *(main verb conjugated)*

lauder VX *(verb infinitive)*

, PON *(punctiation)*

For the performance evaluation, I have randomly selected 50 tagged sentences (636 tokens) from an Early Old French text *Saint Alexis* and manually checked and corrected infinitival tags. I have compared these sentence with an output produced by the trained model. Table 5.3 presents the results for precision and recall in the tnt output. Precision evaluates the proportion of the tagged infinitives that are correct (*Precision=correct extracted infinitives/all extracted tokens*), and recall measures the proportion of correct infinitives that are selected (*Recall=correctly extracted infinitives/all actual infinitives*). In addition, the overall measure $F$ is calculated, which evaluates both recall and precision in a single measure. The POS tagger performed with a precision of 75%, recall of 92% and F-score of 83%.

| Precision | Recall | F |
|---|---|---|
| 75.00% | 92.31% | 82.63% |

Table 5.3: POS Evaluation of Old French Model

[13]The recent Penn Supplement containing annotated texts for Early Old French (Kroch et al. 2012) was not available at the time of the corpus compilation.

[14]The other texts were not available in digital format needed for text-processing.

A manual analysis of these sentences shows that most errors result from TnT algorithm for handling unknown words by means of a suffix trie. That is, unknown words are assigned to an ambiguity class depending on their suffix. For example, in (60) the word *moyler* 'wife' is a noun; the TnT algorithm recognizes this unknown word as an infinitive (VX) based on its ending 'er', a common infinitival ending in the Old French corpus. Similarly, the remaining words, such as *pedre* 'father' and *fare* 'affair' are labeled incorrectly as infinitives based on their ending 're', a common infinitival ending.

(60) Or volt que prenget moyler
Now wants-3p.sg that takes-3p.sg wife-obj
'Now he wants him to get married' (*Saint Alexis*)

TnT output: ADV VJ CONJS VJ**VX**

Correct tags: ADV VJ CONJS VJ **NCS**

While the Old French Treebank contains a relatively small tagset (55 tags), the Latin treebanks employ more fine-grained tagsets, including additional morphological information, such as person, tense, mood, voice, gender and case. For example, the Latin word *alium* 'other' is represented in LDT as $a - s - - - ma-$ (adjective singular masculine accusative). The recently developed Classical Language Toolkit (CLTK) provides a trained TnT model based on this tagset (et al., 2014 2015). Despite its high performance with known words (98.7% accuracy), this model fails with unknown words. In order to optimize TNT performance for this study, I have trained a second TnT model on the prose texts from the LDT treebank with a reduced tagset, which is based on the Penn Treebank tagset. For example, the $a - s - - - ma-$ tag for *alium* has been mapped to $JJ$ (adjective). To evaluate the accuracy of the new trained model, I have selected 50 sentences from *Gregory of Tours*, a text that represents a different chronological period than the trained model. These sentences were manually corrected and compared with the TnT output. The results are presented in Table 5.4. The POS tagger performed with a precision of 97%, recall of 100% and F-score of 99%. These high results show that it is feasible to use pre-existing annotated corpora in Latin from one period to annotate data from a chronologically different period.

| Precision | Recall | F |
|---|---|---|
| 97.30% | 100.00% | 98.48% |

Table 5.4: POS Evaluation of Latin Model

Each model demonstrates certain deficiencies: i) the tnt model does not distinguish between active and passive infinitives and ii) the CLTK model does not process unknown words correctly. However, the combination of both models compensates for the individual limitations of each, namely low accuracy for unknown words in the 1st model and loss of rich morphological information in the 2nd. Thus, both models are applied to the present corpus. The example in (61) illustrates the TnT output (61b) produced by the two models for the sentence in (61a). In (61b), the second column is a tagged output from the reduced tagset model and the third column is a tagged output from the fine-grained model:

(61) a. auctus Antoni beneficio est
raised-p.p. Antoni-gen favor-abl is
'He was advanced by the favor of Antoni' (Cicero, *Letter to Brutus*, 13)

b.

| Token | TnT | CLTK |
|---|---|---|
| auctus | VBN (*verb past participle*) | *Unk* (*unknown*) |
| Antoni | NN (*noun singular*) | *Unk* (*unknown*) |
| beneficio | NN (*noun singular*) | $N-S---NB-$ (*neuter ablative*) |
| est | VBP (*verb singular present*) | $V3SPIA---$ (*3p indicative active*) |

Compared to the high accuracy of POS tagging and relatively small training corpus needed for a tagger training, automatic syntactic parsing requires a large number of syntactically annotated trees (sentences), and the syntactic label accuracy on average does not exceed 85.77% (Gesmundo et al., 2009). As the main focus of this study is non-finite clauses with simple transitive infinitives and direct nominal objects, it is expected that the frequency of such clauses would be very small compared to, for example, main clauses. Thus, it is important to exhaustively extract all infinitival clauses given their relatively small number in any given text. Not having a large training model for parsing annotation, I use only POS an-notation. POS tag search is performed with the use of ANNIS, an open-source search engine for multi-layered annotations (Zeldes et al., 2009). The TnT output is first converted to the ANNIS relation database format using the SaltNPepper converter[15] and then imported to ANNIS. A sample of the query is illustrated in (62), where the query searches through two POS layers, namely reduced POS and fine-grained POS (see an example in 61b). In (62), the query first looks for all infinitives which are tagged as *VB* in the first model. This tag is assigned to all infinitives, passive and active. By using the second layer (pos2), a fine-grained model, I can specify for an active form of infinitive with the following tag - $V--.NA---$ (verb infinitive active voice).[16] Recall, however, that the second model performs well with known words but very poorly with unknown words, as compared to the first model, which employs a special algorithm for unknown words. Therefore, I add an optional query for the tag *Unk* for unknown tokens, in case if the second model did not recognize a given token. Finally, all extracted tokens are manually checked and only infinitival transitive clauses with nominal objects are selected.

(62)
```
pos="VB" &
pos2=/V--.NA---|(Unk)/&
#1 _=_ #2
```

Furthermore, the corpus needs to be structured to enable a comparison of different periods and text forms. In this respect, several methodological issues need to be addressed. First, the Early Old French texts (10th-11th centuries) are scarce, and some of them are not available in the plain format necessary for automatic processing. Given their relatively small size and their importance for the study, it is necessary to add them to the corpus and then search them manually. Second, the Old French database in BFM and MCVF is composed of various dialects, such as Norman, Anglo-Norman, Champenois, Picard, Walloon and Franco-Occitan. Thus, the inclusion of another dialect into the corpus, namely Early Old Provençal, seems to be acceptable given the scarcity of data for the very early period.[17]

[15] http://korpling.german.hu-berlin.de/saltnpepper/

[16] The tense selection is not important here, which is marked by (.).

[17] The included texts are *Boece*, *Testament of Saint John* (2 chapters), *two sermons* and *Saint*

Third, it would be ideal to compare various text styles, such as verse and prose, in both Latin and Old French. It is well known, however, that Early Old French texts (10-12th centuries) are written in verse form; prose mainly becomes available in the 13th century. Several Early Old Provençal texts are, however, written in prose, allowing for a comparison between prose and verse style. In contrast, there is a large body of prose and verse in Latin, but Latin verses present a great challenge for linear word order studies, as their patterns are often discontinuous and distorted (Viti, 2010). On the other hand, some Latin writers use a stylistic metric system in prose that is based on the quantity of syllables. Thus, instead of comparing prose and verse texts in Latin, I will compare metric and non-metric prose. While the metric systems differ in Latin and Old French, such a corpus would at least allow us to examine how word patterns may change according to style.

I will conclude the section with a summary of all the texts used in the present study. Table 5.5 describes the Latin corpus and provides either the manuscript date or the author's date of birth and death. This summary includes the texts from LDT and PROIEL and my additional texts. These additional resources were available for download in XML format from the Perseus Digital Library and HTML format from the Latin Library. Table 5.6 summarizes the Old French Corpus, providing the manuscript data and total number of words for each annotated text; in the case of image-scanned texts, only the number of lines is indicated. The table includes the texts from MCVF and NCA and my additional resources, which were obtained from the Biblioteka Augustana and Chrestomathie Provençale (Bartsch, 1892).

---

*Foy.*

Table 5.5: Latin Reference Corpus

| **Period** | Date | **Text/Author** | **N tokens** |
|---|---|---|---|
| Classical | 106BC-43BC | Cicero | 17671 |
| | 100BC-44BC | Caesar | 1488 |
| | 85BC-25BC | Sallust | 12311 |
| | 80BC-25BC | Vitruvis Pollio | 19685 |
| | 26AD-66AD | Satyricon | 12474 |
| Late Imperial | 61AD-113AD | Pliny the Younger | 23248 |
| | 125AD-180AD | Apuleus | 15291 |
| Late Antiquity | 347AD-420AD | Saint Jerome | 24972 |
| | 381AD-384AD | Egeria | 17519 |
| | 325AD-390AD | Ammianus Marcelinus | 25264 |
| | 390AD | Valesianus I | 1821 |
| | 480AD-524AD | Boethius | 26447 |
| | 484AD-585AD | Cassiodorius | 7278 |
| | 538AD-594AD | Gregory of Tours | 9018 |
| | 540AD-550AD | Anonymous Valesianus II | 3586 |
| Total Tokens: | | | **218 073** |

Table 5.6: Old French Corpus (Old Gallo-Romance Corpus)

| **Period** | Date | **Text/Author** | **N tokens/lines** |
|---|---|---|---|
| 10th | 900 | Eulalia | 29 lines |
| | 980 | Saint-Léger | 1426 |
| | 980 | La Passion de Clermont | 2978 |
| 11th | 1010 | Évangile de Saint Jean | 156 lines |
| | 1010 | Deux Sermons | 122 lines |
| | 1010 | Le Martyre de Saint Étienne | 68 lines |
| | 1030 | Boece | 1912 |
| | 1050 | Chanson de Sainte Foy | 593 lines |
| | 1050 | Gormond et Isembart | 661 lines |
| | 1090 | La vie de saint Alexis | 3999 |
| | 1090/1100 | Chanson de Roland | 4004 |
| 12th | 1170 | Tristan et Iseult | 29962 |
| | 1177 | Chrétien de Troyes | 59750 |
| | 1180 | Marie de France | 61294 |
| 13th | 1220 | La Queste de Saint Graal | 43045 |
| | 1283 | Roisin | 7227 |
| 14th | 1309 | Joinville | 94542 |
| | 1380 | La prise d'Alexandrie | 67718 |
| Total Tokens: | | | **367 857** |

## 5.2 Defining Linguistic Variables

In this section I will outline criteria for defining and coding dependent and independent variables in this corpus. The notion of a linguistic variable is one of the key elements of variationist studies. It was first introduced in early sociolinguistic studies by Labov (1966) (see section 4.3). The linguistic variable can be defined as 'a structural unit that includes a set of fluctuating variants showing meaningful co-variation with an independent set of variables' (Wolfram, 2006, 334). In the sociolinguistic field it is common to refer to the linguistic variable as a dependent variable, while all other contextual, stylistic or sociolinguistic factors are named independent factors or independent variables. In statistical analysis, the dependent variable is referred to as a predicted variable, while other factors are predictors.

First, I will begin with the identification of the *predicted* variable, namely the dependent variable. This variable incorporates the variation between Object-Verb (OV) and Verb-Object (VO) in infinitival transitive clauses. Adopting the methodology proposed by Kroch and Santorini (2014), empty categories and pronouns are excluded, since their positions are fixed or undeterminable. All instances of direct nominal objects scrambled outside the clause are also discarded, as their position relative to the non-finite verb is not recoverable. In the cases where the noun and its adjective or modifier are split, I follow the methodology implemented in Viti's (2010) study of Vedic, a Latin sister language. Viti assigns OV or VO based on the position of a noun, the head of a nominal phrase. For example, I annotate the sentence in (63) with the split noun *partes* 'parts' and adjective *oblitas* 'forgotten' as VO based on the position of the noun.

(63) Vt seruatis queat **oblitas** addere **partes**
for preserved-dat be-able-3p.sg forgotten-acc.adj add-inf parts-acc.obj
'so that one could add the forgotten parts to the ones that are preserved' (Boethius, M5.3)

Following the methodology presented above, a total of 2 488 infinitival transitive clauses with direct nominal objects have been extracted from the corpus of over 500 000 words.

In what follows, I will describe independent variables that are grouped into five categories: i) sociolinguistic, ii) syntactic, iii) semantic, iv) pragmatic factors and v) frequency.

### 5.2.1 Sociolinguistic Variables

These variables describe the date of composition, domain, genre and format of texts. To determine these values for each text, I have consulted The Latin Library,[18] Biblioteka Augustana[19] and Perseus Digital Library[20] for Latin and BMF,[21] MCVF (Martineau et al., 2007) and Biblioteka Augustana for Old French.

#### 5.2.1.1 Chronological Periods

Chronological classification is a central question for diachronic studies. Chronological coding allows us to trace the diffusion or decline of linguistic phenomena across time. According to historical chronological classification, Latin is divided into 5 periods (Väänänen, 1967, 11-13), which are illustrated in (64), while Old French is divided into three (Lodge, 1993, 10), as illustrated in (65):

(64)
1. **Archaic Latin** - until 200 BC (end)
2. **Pre-classic Latin** - 200 BC - 100 BC (middle)
3. **Classical Latin** - 100 BC (middle) - 14 BC (Death of Augustus)
4. **Post-classic Latin** - 14 BC - 200 AD
5. **Late Latin** - 200 AD - Romance Languages

(65)
1. **Proto-French** - AD 500-842
2. **Early Old French** - 842-1100
3. **Classical Old French** 1100-1350

The data in the present study span several centuries, from the 1st century BC to the 14th century, namely from Classical Latin to Classical Old French. In Latin, each text is described in terms of the author's dates of birth and death. This approach is chosen due to the fact that there is rarely a reliable, precise date of composition for historical documents in Latin. A median is further calculated for each interval. One approach to data periodization would be to split data by centuries, e.g. 1BC, 1AD, 2AD. The second approach is to apply the historical periodization described in (64). In order to evaluate which approach fits better the present data, the variability-based neighbor clustering method (Gries and Hilpert, 2008) is used (see section 5.3.2). This method allows for a visual examination of chronological clusters. The resulting temporal clusters for Latin are illustrated in Figure 5.2.

[18] `http://www.thelatinlibrary.com/`
[19] `https://www.hs-augsburg.de/~harsch/augustana.html`
[20] `http://www.perseus.tufts.edu/hopper/`
[21] `http://bfm.ens-lyon.fr/`

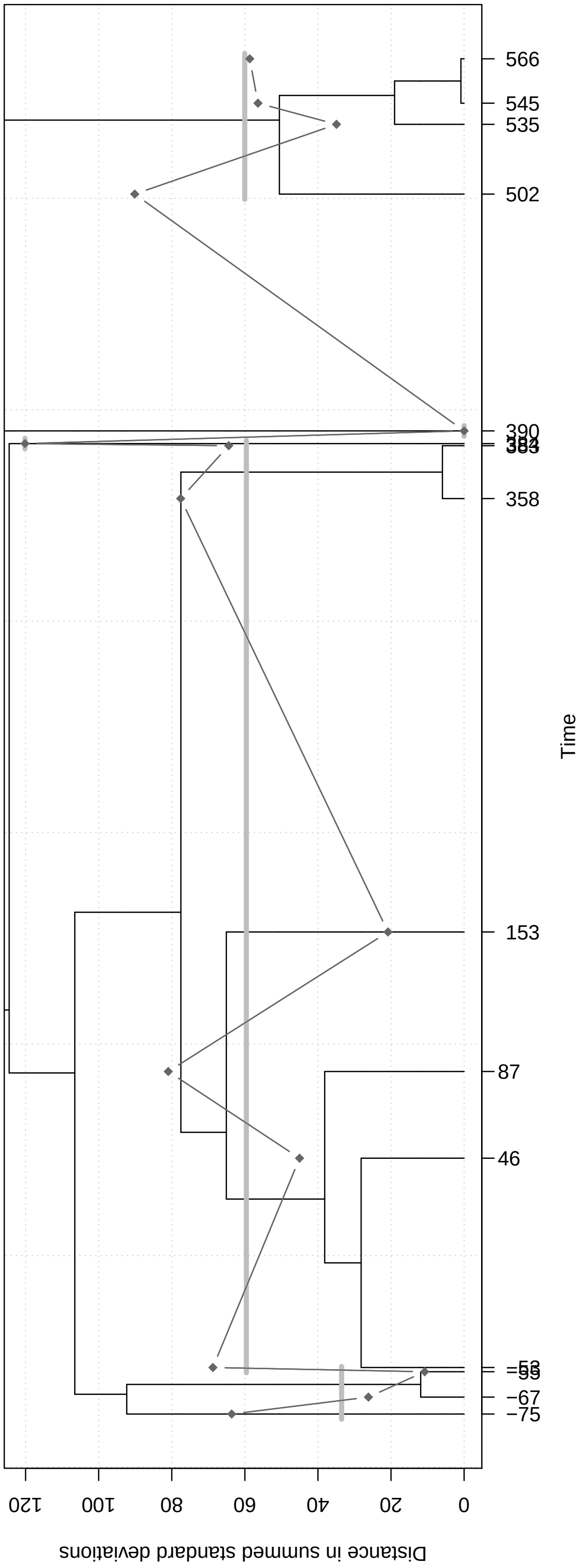

Figure 5.2: VNC Results for Latin

From Figure 5.2, we can identify the first cluster, *Classical Latin*, ranging from 75BC to 55BC (see also Figure 5.3 for Zoomed-in View). Interestingly, the year 53BC is treated as a part of the second cluster. However, if we look at the Latin text from this year, its scientific genre (*Vitruvis*) is more likely to cause the deviation. Since the data are coded for genre, it is safe to conflate this text with the first cluster.

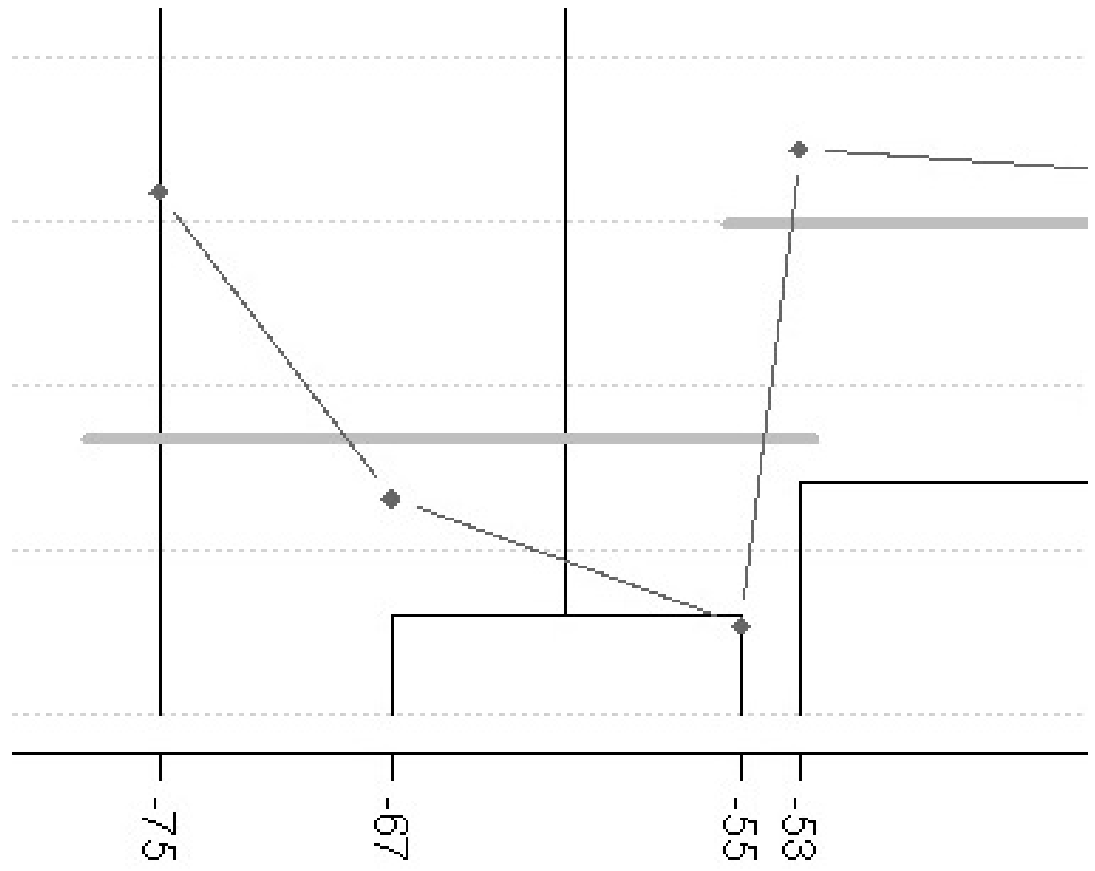

Figure 5.3: Zoomed-in View: Classical Latin

In the next cluster, ranging from 46AD until 383AD, we can detect two subclusters, *Late Imperial* and *Late Latin*, namely 46-153 and 358-383. Two outliers are observed at the temporal points 385 and 390, shown in a zoomed-in view in Figure 5.4. Careful analysis of the texts reveals that in the first case we have an instance of a specific genre (Jerome's letters and the Vulgar Bible), whereas the second case (Valesianus I) is represented by only two tokens. Therefore, these two instances are merged into the preceding cluster. Finally, there is a clear cluster spanning from 502 until 566. Although this partitioning corresponds to the historical periodization described in (64), the data shows that Late Latin group calls for two subclusters, namely 358-390 and 502 and 566. Thus, 4 clusters are adopted for Latin in this study: i) Classical (75BC-53BC), ii) Post-Classical (Late Imperial) (46-153), iii) Early Late Latin (358-390) and iv) Late Latin (502-566).

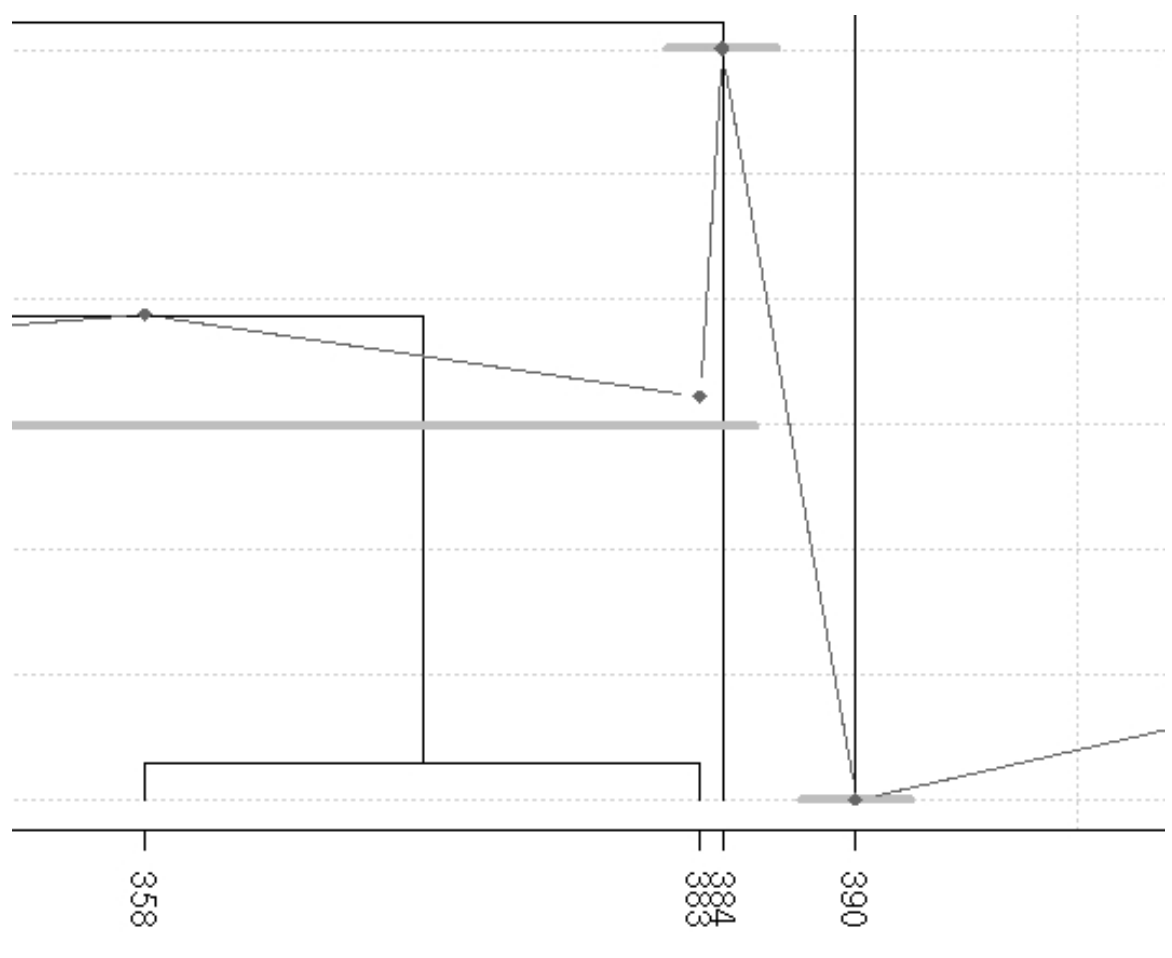

Figure 5.4: Zoomed-in View: Late Latin

Next, Figure 5.5 shows the VNC periodization for Old French. First, Figure 5.5 shows that the years 900 and 980 are combined into one cluster, followed by an outlier, a single temporal point for the year 1010. This year is represented by texts written in prose, as compared to verse texts in the first cluster. Thus, the year 1010 can be merged with the first stage. The next cluster spans from 1030 to 1283. A detailed analysis of the subclusters shows that the texts from the 11th-century texts (1170-1180) are conflated with the year 1090, which is represented by *Roland*. In fact, the date of composition for this epic poem is not clear and is listed as 1090 or 1100. The remaining years, 1030 and 1050, are clustered together. Since the very early period is sparsely populated, I combine the years 900, 980, 1010, 1030 and 1050 into the first cluster, namely the 10-11th centuries, followed by the second cluster, which represents the 12th century. Furthermore, there is a cluster for 1220 and 1283, which represents the 13th century. Finally, there are two individual temporal points for 1309 and 1370. Since both texts reflect different genres, namely *Joinville* - prose and *Alexandrie* - verse, these two periods are combined into one cluster representing the 14th century. As a result of the VNC method, 4 periods are established: i) 10-11th centuries, ii) 12th century, iii) 13th century and iv) 14th century.

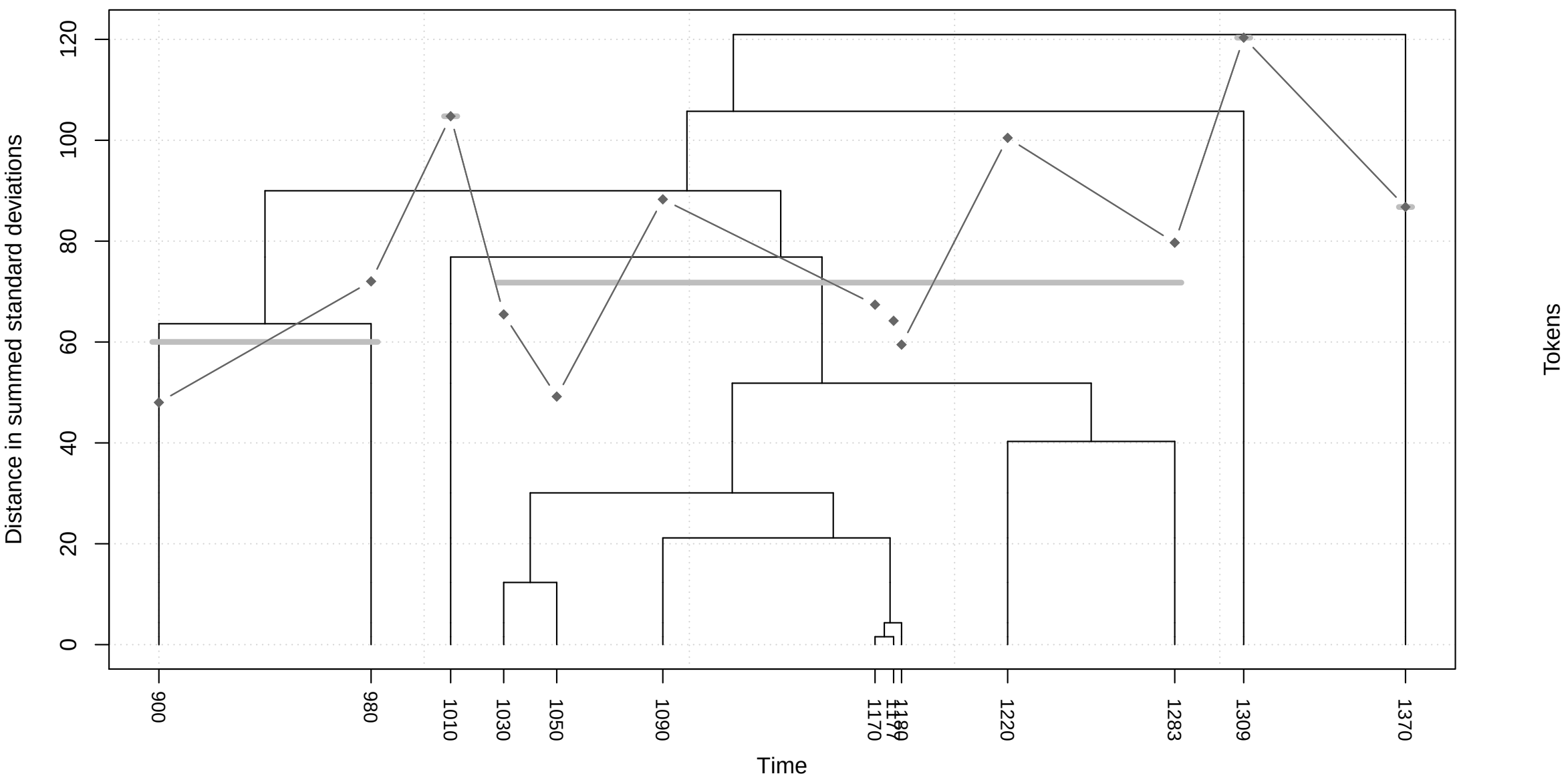

Figure 5.5: VNC Results for Old French

#### 5.2.1.2 Genre

This sociolinguistic factor is divided into three subgroups: i) metrics, ii) theme and iii) genre. First, each text is codified according to its metric form, namely metric or non-metric. For Latin data, metric format refers to metric prose, while for Old French it refers to verse.[22] The theme factor categorizes data by topics and subject matter, namely history, religion and literature. In contrast, the genre factor classifies texts according to their 'formal arrangement' and 'mode of address' (Montgomery et al., 2007, 41-44). Following this approach, the following genres are established: i) epistolography (letters), ii) public speech, iii) narrative ('stories involving a sequence of related events'), iv) hagiography (biography of saints) and v) treatise (scientific, historical or philosophical reflections). The following table summarizes the codification schema in the present corpus:

[22]Metric style in Latin is defined by the quantity of syllables, whereas in Old French it is defined by accent and rhyme.

Table 5.7: Genres

| Domain | Genre | Metric/Non-Metric | Texts |
|---|---|---|---|
| history | narrative | non-metric | Caesar (de Bello) |
| | | non-metric | Anonymous Valesianus I |
| | | non-metric | Anonymous Valesianus II |
| | | non-metric | Cassidorius |
| | | non-metric | Joinville (Memoires) |
| | | metric | Alexandrie (Take of Alexandia) |
| | treatise | non-metric | Vitruvis (de Architectura) |
| | | non-metric | Sallust (Catilina) |
| | | non-metric | Roisin (Le livre de Roisin) |
| | | metric | Ammianus |
| religion | epistolography | non-metric | St.Jerome (Letters) |
| | | non-metric | Gregory The Great |
| | | non-metric | Egeria |
| | narrative | non-metric | St.Jerome (Vulgata) |
| | | non-metric | Gregory of Tours |
| | | non-metric | Saint Jean |
| | hagiography | metric | Passion |
| | | metric | St. Leger |
| | | metric | Eulalia |
| | | metric | St.Etienne |
| | | metric | Alexis |
| | | metric | St.Foy |
| | speech | non-metric | 2 Sermons |
| literary | epistolography | metric | Cicero |
| | | non-metric | Pliny The Younger |
| | narrative | non-metric | Petronius (Satyricon) |
| | | metric | Apuleus (Metamorphose) |
| | | metric | Gormon |
| | | metric | Roland |
| | | metric | Tristan |
| | | metric | Troy |
| | | metric | Marie de France |
| | | non-metric | Queste de Saint Graal |
| | treatise | metric | Boethius |
| | | metric | Boece |
| | speech | metric | Cicero (Oratione) |

### 5.2.2 Syntactic Variables

#### 5.2.2.1 Presence of Subject

This factor studies the role of the explicit subject in word order. The subject can be present in AcI constructions, where it bears an accusative case, as illustrated in (66).

| (66) | Dico | te | venisse |
|---|---|---|---|
| | say-1p.sg | you-acc.sbj | come-inf.past |

‘I say that you have come’ (Bolkestein, 1979, 20)

Traditionally, AcI constructions function as complements for verbs of discourse, verbs of mental or physical perception as well as independent historical infinitives. Some insist that AcI constructions should also include the verbs of commands, such as *to order*, *to force*, as shown in (67); whereas others argue that this is a case of a pseudo AcI, where the subject of the infinitive *te* is structurally an object of the matrix clause (Morin and St-Amour, 1977; Bolkestein, 1979).

| (67) | Cogo | te | venire |
|---|---|---|---|
| | force-1p.sg | you-acc.sbj | come-inf |

‘I force that you come/I force you to come’ (Bolkestein, 1979, 20)

In the present study the subjects of both types (66 and 67), genuine AcI and pseudo AcI, are coded as *explicit subject*. In contrast, the following cases are treated as *absent subject*: i) when the main verb is a ditransitive verb, such as *mittere* ‘to send’ (68a), where the logical subject of the infinitive *vocare* ‘to call’ is a thematic argument (goal) of the main clause, e.g. *servos* ‘servants’ and ii) when the subject of the infinitive is outside of the infinitival clause (68b).

| (68) a. | misit | **servos** | **suos** | vocare | invitatos | ad | nuptias |
|---|---|---|---|---|---|---|---|
| | sent-3p.sg | servants-acc.obj | his-acc | call-inf | guests-acc.obj | to | marriage |

‘And he sent his servants to call them who were invited to the marriage’ (Vulgate, M22.3)

| b. | et | **choraulen** | **meum** | iussi | Latine | cantare |
|---|---|---|---|---|---|---|
| | and | conductor-acc.sbj | my-acc | ordered-1p.sg | Latin-acc.obj | sing-inf |

'I ordered my conductor to play Latin airs' (Satyricon, 53)

#### 5.2.2.2 Position of Infinitival Clause

This factor group investigates the influence of the position of infinitival clauses on word order choice. This variable includes the following values: i) anteposition with respect to the matrix verb (69a), ii) postposition (69b) with respect to the matrix verb, iii) independent historical infinitive (69c) and iv) prepositional infinitive (69d).

(69) a. **Interrogare** ergo atriensem coepi
ask-inf then keeper-acc-obj began-1p.sg
'I began asking the hall-keeper' (Satyricon, 29)

b. coepere senatum criminando plebem **exagitare**
begin-inf senate-acc accusing-pp people-acc.obj excite-inf
'they thereupon began to excite the commons by attacking the senate' (Sallust, 38.1)

c. neque modum neque modestiam victores **habere**
nor moderation-acc no restrain-acc victors-acc have-inf
'the victors showed neither moderation nor restraint' (Sallust, 11.4)

d. A **celer** bien un suen consel
to hide-inf well a his advice-acc.obj
'To conceal well one of his secrets' (Tristan 1315)

#### 5.2.2.3 Tense of the Infinitive

This factor group is only relevant to Latin, in which infinitives can bear tense markers: present, past and future. Since the main focus of the study is an active simple infinitive, the data are coded for i) simple present (70a) and ii) simple past (70b) (see Table 1.1 for the morphological paradigm for Latin infinitives).

(70) a. iam scies hoc ferrum fidem **habere**
now know-2p.sg this-acc iron-acc.sbj credit-acc.obj have-inf.pres
'you'll soon see this bit of iron commands some credit' (Satyricon, 58)

b. Iam scies patrem tuum mercedes **perdidisse**
now know-2p.sg farther-acc.sbj your-acc salary-acc.obj lose-inf.past
'You'll soon see your father wasted his money on you' (Satyricon, 58)

#### 5.2.2.4 Structure of Infinitival Phrase

As Wurmbrand (2003) points out, 'infinitival complements do not all have the same functional (i.e., syntactic) architecture above the V[erbal]P[hrase]' (Wurmbrand, 2003, 1-2). It is common to distinguish between at least the following five types (Iovino, 2010, 68):

(1) Accusativus cum Infinitivo: *Dico illum venire* 'I say that he comes'

(2) Raising structure: *Flamma videtur arder* 'The flame seems to burn'

(3) Control structure: *Galli intermiserant obsides Caesari dare* 'The Welsh stopped giving hostages to Caesar'

(4) Simple infinitive: *Ridiculum est currere* 'Running is ridiculous'

(5) Restructuring structure: *Posso dicere hoc* 'I can say this'

Three classes of infinitives can be identified among these five types: a) infinitives that manifest an independent full clausal structure (1), b) infinitives that do not exhibit clausal behavior - reduced infinitives (2-4) and c) restructuring verbs (5), where the infinitive forms a monoclausal structure with the main verb (Rizzi, 1976; Cinque, 1998, 2004; Wurmbrand, 2004; Iovino, 2010). Cross-linguistically, restructuring main verbs include modal, e.g., *to want*, *can*, aspectual, e.g., *to start*, *to continue*, and motion verbs, e.g. *to come*.[23] These verbs are classified as *functional*, as it is argued that they are generated in functional projections. In this view, the main verb does not form an independent clause that embeds an infinitive; the infinitive rather becomes a main predicate or a verbal complement (Cinque, 2004).

Based on the taxonomy suggested in Pearce (1990), Wurmbrand (2003), Cinque (2004) and Iovino (2010), infinitival structures are codified as follows: a) Accusativus cum Infinitivo

[23] Cinque (2006) also includes perception verbs.

(71a), b) Raising structure (71b), c) Control structure (71c), d) Simple infinitive (71d), e) Restructuring structure (71e) and f) Prepositional infinitives (71f). The AcI class consists of verbs of saying and knowing, verbs of mental and physical perception and independent infinitives. The Raising class includes the following matrix verbs: *to seem*, *to appear* and passive voice of *to see*. The Control class consists of subject and object control verbs, e.g., *to order* and *to permit*. Restructuring verbs, e.g., *to want*, *to begin*, are excluded from this class, as they will be members of a separate Restructuring class. Simple infinitive structure includes impersonal constructions. The Restructuring class comprises the following groups of verbs: i) Functional group - modal verbs and impersonal verbs of necessity (*loisir, estevoir* 'to be necessary'),[24] and ii) Lexical group - perception,[25] aspectual and motion verbs, causatives, verb *savoir* with the meaning of *be able* (Cinque, 2006) and several aspectual prepositional infinitives that follow under Pearce's category of reduced structures (*commencer à*, 'to begin').[26] Finally, the Prepositional class includes all prepositional infinitives, except those that are coded as Restructuring verbs.

```
(71) a. Non puto         illum          capillos       liberos  habere
        not  think-1p.sg him-acc-subj  hair-acc.obj  free-acc have-inf
```

'I don't suppose he has a hair on his head unmortgaged' (Satyricon, 39)

```
     b. vereor      ne        decepisse          Caesarem         videar
        fear-1p.sg  that-not  mislead-inf.past  caesar-acc.obj  seem-1p.sg
```

'I am afraid that I will be seen to have misled the Emperor' (Pliny, *Book* 2.9)

```
     c. duplicare      tormenta          iubet
        redouble-inf  tortures-acc.obj  orders-3p.sg
```

'he ordered to redouble her tortures' (Jerome, *Letter* 1, p.6)

```
     d. fas    esse        arbitror      uel  occuluisse        ueritatem        uel  concessisse
        right  to-be-inf  think-1p.sg  nor  hide-inf.past  truth-acc.obj  nor  concede-inf.past
        mendacium
        lie-acc.obj
```

'do not I think it is right either to have the truth concealed or to have a lie

[24] This condition applies only to Old French data.

[25] Latin perception verbs are considered AcI (Morin and St-Amour, 1977; Bolkestein, 1999).

[26] Pearce (1990) argues that the prepositional infinitival complements introduced by *à* and *de* lack CP structure.

conceded' (Boethius, 1.4)

e. coepit repetere arras
began-3p.sg ask-inf earnest-money-acc.obj

'he began to ask the return of the earnest-money' (Valesianus II, 2.62)

f. en lor mantiax anvelopes vindrent, por lor lermes covrir
in their mantles wrapped came-3p.pl, for their tears-acc.obj cover-inf

'They entered with their mantles wrapped about them to conceal their tears' (Yvain, 121.4202)

#### 5.2.2.5 Intervening Constituents

This factor analyzes the role of intervening elements between the direct nominal object and its infinitive. This variable is codified on a binary scale: presence or absence.[27] The following categories of intervenors have been included: i) NP (72a), ii) adverbial phrase (72b), iii) intensifier (72c), iv) prepositional phrase (72d) and v) pronoun (72e):

(72) a. magnam calamitatem **pulsos** accepisse
great-acc.obj lost-acc.obj defeat-abl undertake-inf.past

'they undertook great loss by defeat' (Caesar, 1.31)

b. Lors li vont son cheval **fors** treire
then him-dat go-3p.pl his horse-acc.obj outside bring-inf

'Then they go to get him his horse out' (YVAIN,127.4376)

c. A celer **bien** un suen consel
to hide-inf well a his advice-acc.obj

'To conceal well one of his secrets' (Tristan 1315)

d. E le doel **el chastel** mener
and the grief-acc.obj in-the castle lead-inf

'and to grieve in the castle' (Marie de France, 2369)

e. et nemo poterat respondere **ei** verbum
and nobody-nom.subj could answer-inf him-dat word-acc.obj

'And no man was able to answer him a word' (Vulgata, M22.46)

[27]Given the very small number of clauses with intervening constituents per period, the binary scale is chosen.

#### 5.2.2.6 Heaviness of NP

This factor group analyzes the role of the length of a nominal phrase on word order. It has been suggested that various ways of calculating length produce very similar results (Wasow, 1997). In the present study two different methods are tested: i) word count and ii) syntactic complexity. The first method is somewhat arbitrary, as length is calculated according to the number of words in a nominal phrase.[28] The second approach is based on syntactic length with a binary distinction between heavy and light (Bies, 1996; Taylor and Pintzuk, 2012b). Syntactically heavy constituents are coordinated NPs and NPs with postmodifiers, namely relative clause, PP or appositive.

### 5.2.3 Semantic Variables

#### 5.2.3.1 Animacy of NP

The factor group studies the impact of animate and inanimate NPs on word order. This category is often represented by the following hierarchy: human > animate (animals) > inanimate. Given that very few tokens belong to the category of animals, the data are coded using a binary distinction: human - non-human.

### 5.2.4 Pragmatic Variables

The key components of the *information structure* approach are i) *predicational structure*, ii) *information relevance* and iii) *information status* (see section 1.3.2). The *predicational* layer distinguishes between *topic* and *comment*. Recall that *topic* refers to what the sentence is about and is determined by *aboutness* (e.g. *A says about X that X ...*) and *topicalization*. Subjects and topicalized objects are more likely to be *topics* in a sentence. Given that in my corpus topicalized objects (i.e. objects outside of the infinitival clause) are excluded and that subjects are present only in a small subcorpus of *Accusativus cum Infinitivo* (see section 1.3.3.1), the *predicational* layer is not included in the present study.

[28] Coordination, determiners and prepositions are not counted as a word.

#### 5.2.4.1 Informational Relevance of NP

Following Kiss (1998) and (Götze et al., 2007) I distinguish between contrastive focus and information focus (see section 1.3.2). In this study I apply the following methodology: if the token (the nominal object or its verb)[29] carries a contrastive feature it is marked as a contrastive noun (73a) or contrastive verb (73b).

(73) a.

| | | | | | | | | |
|---|---|---|---|---|---|---|---|---|
| neque | illum | in | tanta | re | **gratiam** | **aut** | **inimicitias** | exercere; |
| not | he-acc.sbj | in | such | thing | favor-acc.obj | nor | enmity-acc.obj | show-inf |

‘that in a matter of such moment he showed neither favour nor enmity’ (Caesar, 51.16)

b.

| | | | | | | | | |
|---|---|---|---|---|---|---|---|---|
| nemo | dignus | inventus | est | **aperire** | librum | nec | **videre** | eum |
| nobody-nom.sbj | worthy | found | is | open-inf | book-acc.obj | nor | see-inf | it-acc.obj |

‘no man was found worthy to open the book, nor to see it.’ (Vulgata, 5.4)

I use various syntactic and morphological markers and operators for focus – such as *only*, *even* and *but* – as well as contrastive negation (Krifka, 2007b) to identify the contrastive property.[30] Following this methodology, each token is tagged as contrastive focus or information focus regardless of its cognitive status, namely old or new information.

#### 5.2.4.2 Information Status of NP

Recent studies have shown that many languages (e.g., Old English, Yiddish, Old German) strongly associate pre-verbal NPs with entities recoverable from context or from speakers’ shared knowledge (Prince, 1981; Hróarsdóttir, 2000; Petrova, 2009). Therefore, the data in this study have been coded for the information status of NPs. Following the methodology of Götze et al. (2007), all tokens are categorized in three ways: 1) given, 2) accessible and 3) new information. Tokens that can be evoked from previous context are tagged as *old*; token that can be inferred from common knowledge are tagged as *accessible* and tokens that are not recoverable are coded as *new*.

[29] Assuming the claim that the contrastive focus moves to a functional projection, I deem it necessary to mark verbs if they carry contrast, as this movement creates the VO order.

[30] While the above mentioned operators are pertinent to English, I use their equivalents in Latin and OF, e.g. *enim* ‘even’, *ne fait que*, and *mais*, among others.

### 5.2.5 Frequency

In section 4.4 I have reviewed the role of frequency in language change. To assess its role in the present study, tokens are coded for frequency. Given a relatively small corpus size (2 488), some tokens may occur only once due to the nature of this corpus. The examination of lemma, namely all word forms of a given token, would be more representative of an actual token use. However, not all texts in this study are annotated for lemma. On the other hand, there exist two on-line lexical databases: LASLA[31] for Latin and (NCA)[32] for Old French. Thus, lemma frequency calculation in Latin is performed by using the lemma search in Opera Latina (LASLA);[33] lemma frequency in Old French is assessed by using the on-line TWIC lemma search in NCA corpus.[34]

## 5.3 Applied Statistical Methods

In this section I will describe the statistical methods that will be used to analyze the corpus described above.[35] All the statistical techniques are performed using open source R software (R Development Core Team, 2008) and various R packages, e.g., *ggplot2*, *ca*.

### 5.3.1 Correspondence Analysis

Correspondence analysis is a useful non-parametric statistical technique that is well suited for categorical data (Greenacre, 2007). Correspondence analysis is based on the Chi-square statistics for the matrix, where the matrix is represented by a table with frequency

[31] `http://www.cipl.ulg.ac.be/Lasla/inscription.html`

[32] `http://www.uni-stuttgart.de/lingrom/stein/corpus/`

[33] 10 words are absent from the Opera Latina lexicon: for 9 of them I used Perseus library via token frequency search and one word (*traversare* 'to cross') is assigned its actual corpus frequency - 1, as it was not possible to find it in both libraries.

[34] For 15 tokens it was not possible to determine their lemma frequency and their actual corpus frequency was assigned instead.

[35] For a more detailed description of statistical methods for the corpus study see McGillivray (2010) and Jenset (2010).

distribution, namely a contingency table. Table 2.4 from Chapter 2 is modified below to illustrate the structure of a contingency table:

Table 5.8: Example of a Contingency Table

| Text | Non-relatives - OV | Relatives - OV | Total |
|---|---|---|---|
| Passion | 14 | 5 | 19 |
| Alexis | 11 | 2 | 13 |
| Total | 25 | 7 | 32 |

The traditional Chi-square measures dependence, or significance between rows or columns. Chi-square in Correspondence analysis measures the distance from a cell to its row mean (Greenacre, 2007). This distance is further displayed in a 2 dimensional plot, where the proximity between points indicates their association, and the closer the points are to the origin (0,0), the less association they exhibit. An example of correspondence analysis is shown in Figure 4.11, repeated here for illustration purposes. This plot illustrates a two-dimensional representation of word order patterns and authors. The proximity of a pattern to an author indicates their strong association. For example, there is a strong association between Jerome (Vulgata) and Thomas (Medieval Latin) with SVO order.

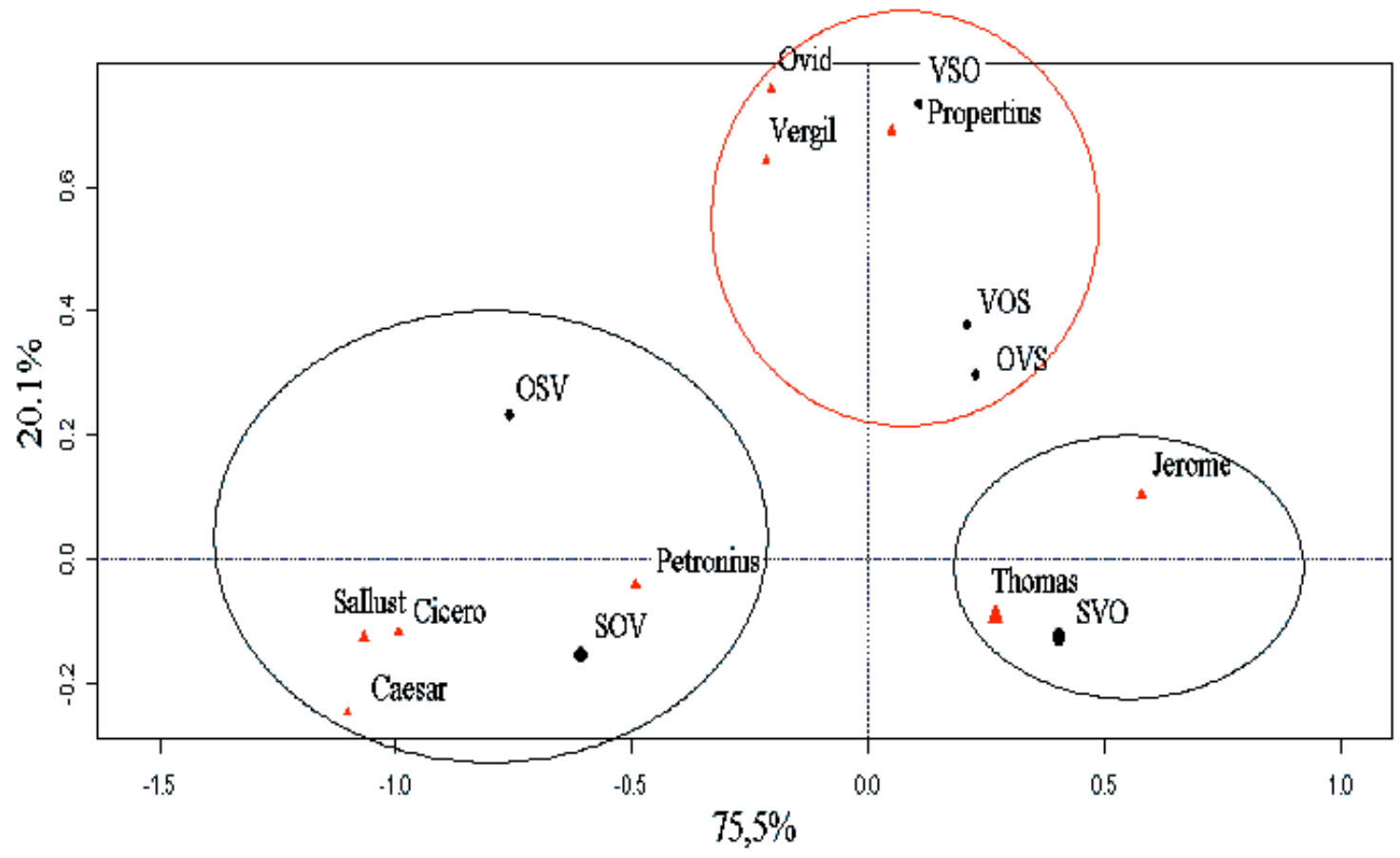

Figure 5.6: Illustration of Correspondence Analysis: Word-Order Patterns and Latin Authors (Passarotti et al., 2013, 347)

The example of R commands used in this study is illustrated in (74). First, the script creates a contingency table; then, it creates and displays associations between word order patterns, e.g., SOV, OVS, and periods.

(74)
```
>contingency_table <-prop.table(table(mydata$Period,
mydata$WordPattern),2)
>library("ca")
>fit<-ca(contingency_table)
>plot(fit, mass = TRUE, contrib = "symmetric",  map ="colgreen",
  arrows = c(FALSE, TRUE))
```

### 5.3.2 Cluster Analysis

While Correspondence analysis examines the relations between individuals (e.g., authors, texts) and variables, cluster analysis allows for the grouping of variables or individuals. Traditionally, groups are clustered into tree branches, and the visual representation is often referred to as a *dendrogram.* The R code for creating a dendrogram for VO order with different genres is illustrated in (75). First, the script creates a subset of data with VO order; then, it clusters different genres based on the use of VO order.

(75)
```
>mysubsetVO<-subset(mydata, mydata$OVVO=="VO")
>mysubsetVO$OVVO<-factor(mysubsetVO$OVVO)
>two_factors <-table(mysubset$Genre,mysubset$OVVO),2)
>library("stats")
>d<-dist(two_factors,method="euclidean")
>fit<-hclust(d,method="ward")
>plot(fit)
>rect.hclust(fit,k=3, border="red")
```

The clustering method can also be used for temporal ordering. Variability-based neighbor clustering (VCN) is an approach for diachronic studies developed by Gries and Hilpert (2008). As Gries and Hilpert (2008, 60) point out, traditional grouping methods in diachronic linguistics are often based on ‘eye-balling’ or subjective decisions on data grouping, often by 25, 50 or 100 years, depending on the availability of data. The VNC method is intended to help researchers by offering 'objective quantifiable suggestions' with respect to period classification. The VCN package and instructions are available upon request from Gries and Hilpert (2008). A sample of the VCN dendrogram is shown in Figure 4.10, repeated here for illustration:

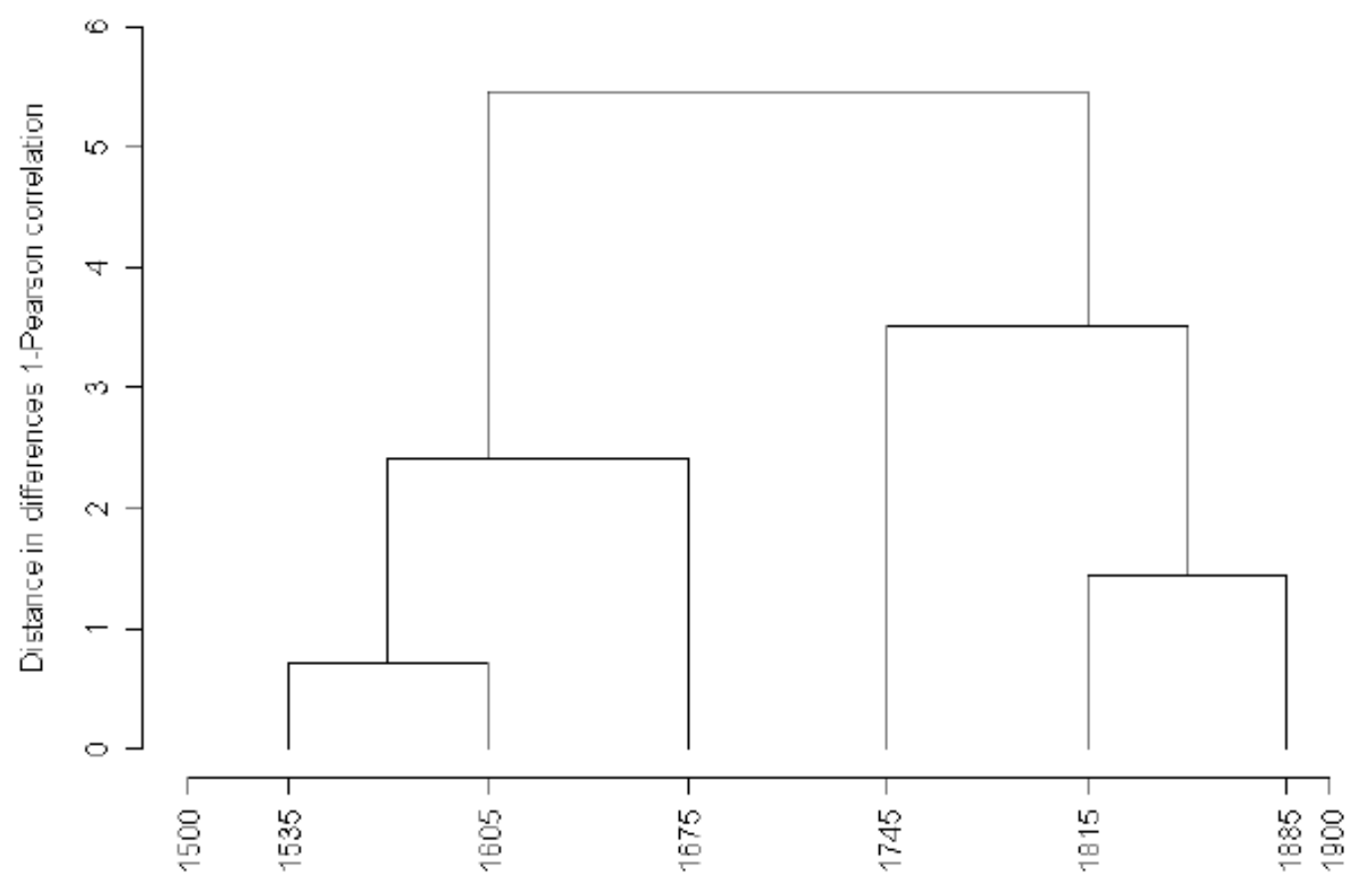

Figure 5.7: Illustration of Dendrogram: Grouping Data by Periods (Gries and Hilpert, 2008, 69)

### 5.3.3 Random Forest and Conditional Inference Tree

The traditional logistic regression model often becomes problematic when dealing with a small number of tokens and a large number of predictors. One of the methods used in machine learning, namely *random forest*, is able to deal with such issues (Breiman, 2001). In addition, it is shown that the random forest ranking 'proves to be more stable than stepwise variable selection approaches available for logistic regression' (Strobl et al., 2009, 324). To determine the best model, the algorithm splits data into multiple trees for each predictor and then selects the most probable predictors. Compared with a linear regression model, the random forest model is able to represent non-linear relations and associations with multiple partitioning in the same variable (Strobl et al., 2009, 325). Random forest can be modeled in several R packages, e.g., *rpart*, *partykit*. The use of the package *partykit* is illustrated in Figure 5.8, where the model includes several linguistic and sociolinguistics factors. The results in this model are presented in a sorted order by importance.

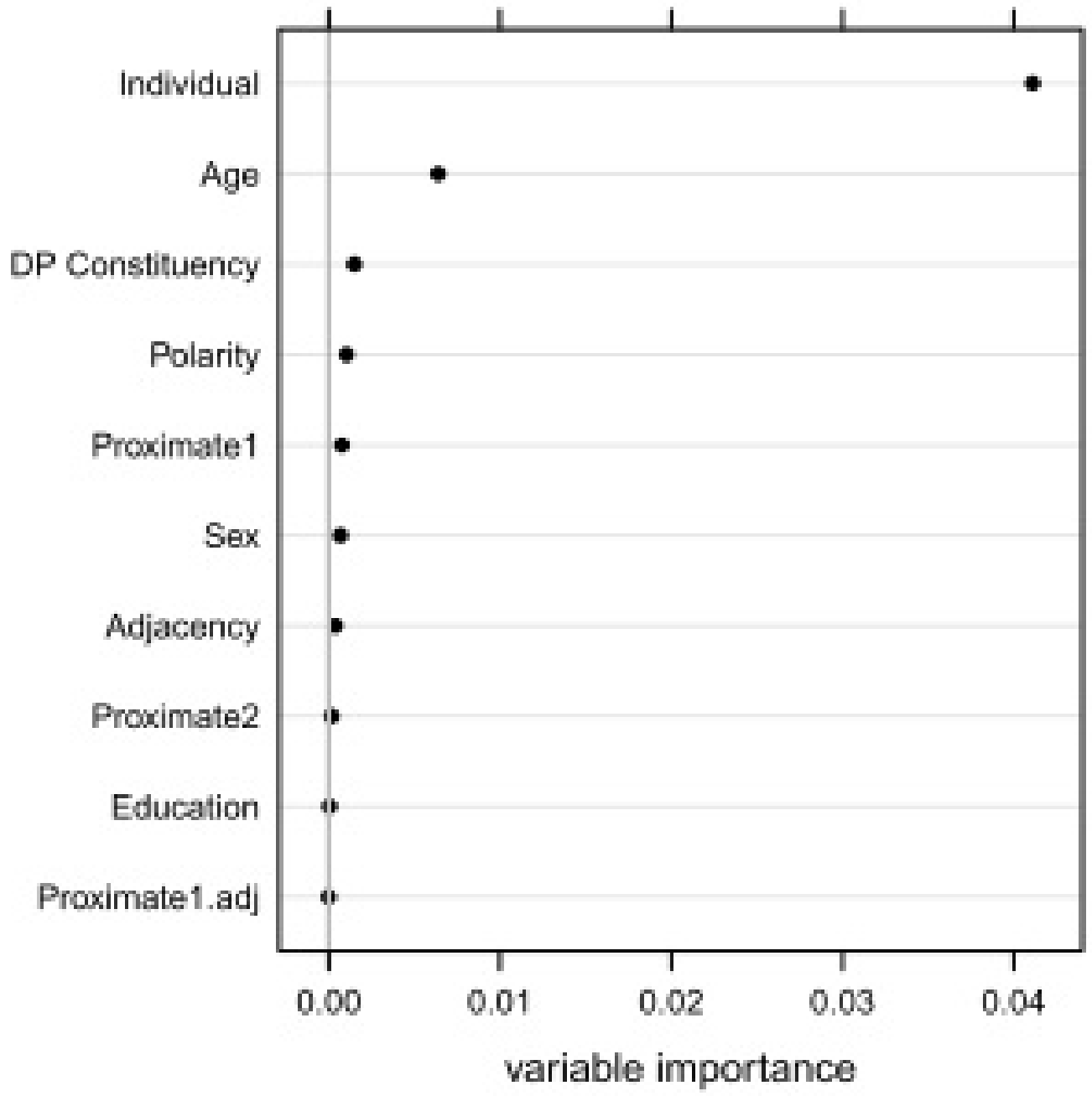

Figure 5.8: Illustration of Random Tree - *Was/Were* Case Study (Tagliamonte and Baayen, 2012, 25)

The R script for a random analysis is shown in (76), where *OVVO* is a dependent variable and *Period, OldNewInformation, Focus, Theme, Animacy, Genre* and *VerbType* are independent variables.

(76)
```
>library(partykit)
>mytree = cforest(OVVO ~ Period +  OldNewInformation + Focus +Theme
   +Animacy + Genre + VerbType, data=my_data)
>myforest.varimp <- varimp(myforest)
>dotchart(sort(myforesrt.varimp))
```

In order to visualize a hierarchical tree, a conditional tree method is used. This method assigns a p-value to each factor based on its significance and its relationship with other predictors (Tagliamonte, 2011, 152-153). In the conditional tree plot, each significant factor demonstrates how its values affect an independent variable. This method is shown in Figure 5.9, using an example from the study of *there was/there were* by Tagliamonte and Baayen (2012). In this plot we can see that the use of *there was* is more likely with affirmative polarity and with individuals ≤46 years old.[36]

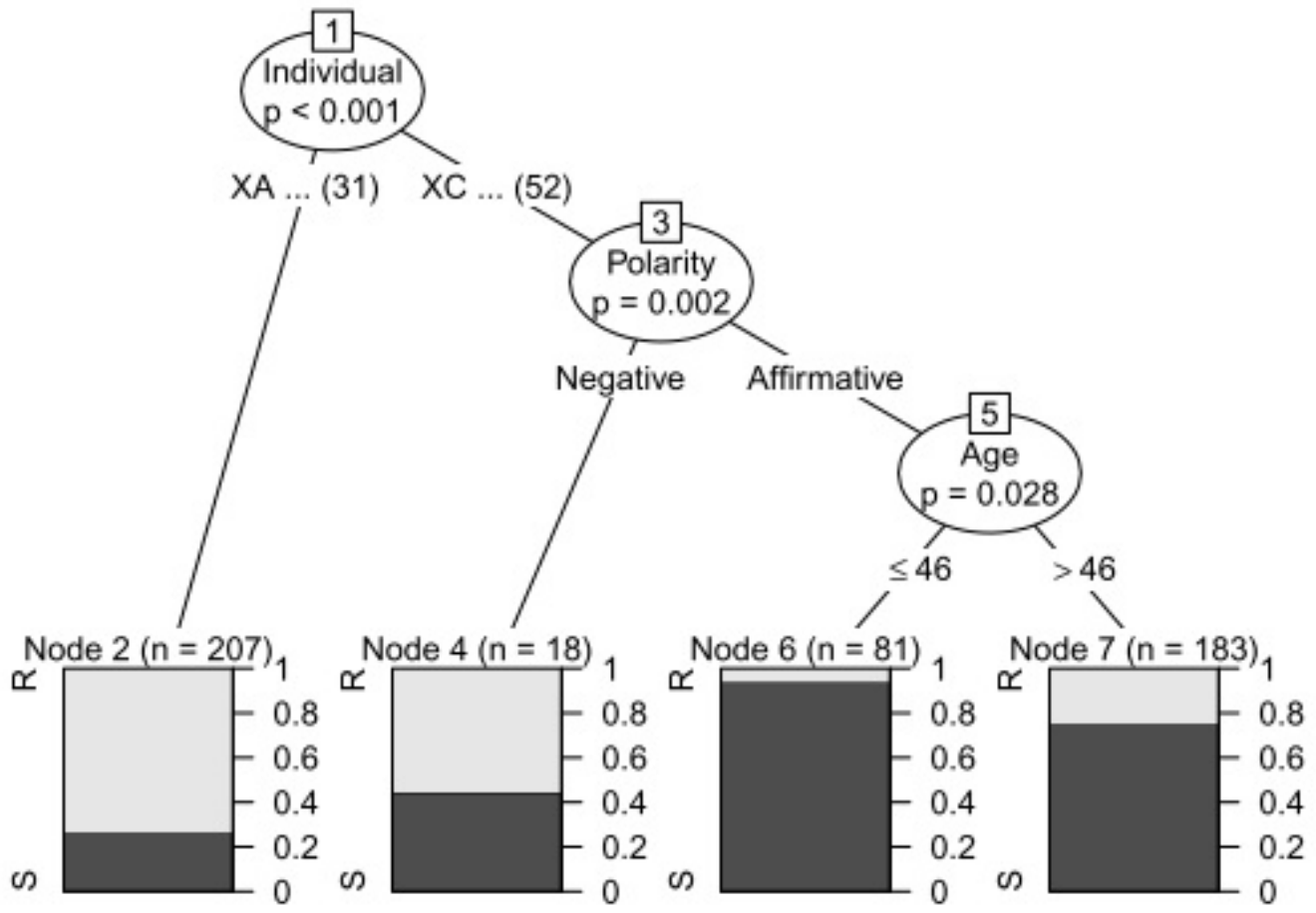

Figure 5.9: Illustration of Random Tree - *Was/Were* Case Study (Tagliamonte and Baayen, 2012, 26)

The following example illustrates the use of this method:

(77)
```
>library(partykit)
mytree<-ctree(OVVO ~ Period +  OldNewInformation + Focus +Theme
   +Animacy + Genre + VerbType, data=my_data)
>plot(mytree)
```

### 5.3.4 Logistic Model: Fixed Effect and Random Effect

Two of the most common functions for logistic regression are *lmer()* (package *lme4*) and *glm()* (package *stats*). These logistic models are used when the dependent variable is a binary or binomial, for example, yes-no, OV/VO, etc. The R formula for logistic regression is shown in (78):

(78)
```
>glm(OVVO ~ Period + VerbType, data=mydata, family = binomial)
```

[36]See also a sociolinguistic study of the pluralization of *haber* (Claes, 2014).

The fixed effect logistic model treats all the predictors as independent and constant. That is, the underlying assumption of this model is that there is no group-internal variation between speakers or tokens. This approach, however, excludes the possibility that some factors involve only a few speakers or a few frequent tokens. As a result, the fixed model will overestimate the influence of these factors for the whole population. In contrast, a mixed-effect model makes it possible to account for individual- and group-level variation (Johnson, 2009; McGillivray, 2010; Jenset, 2010; Tagliamonte, 2011; Johnson, forthcoming). For example, in the mixed model, we can specify a *grouping* factor, or *random effect*, for individual speakers and individual words. With this *random* factor, the regression model will evaluate *fixed effects* (remaining factors) and the variation among speakers or words of the same group. If the effect is stronger than the group variation, the *fixed effect* will be considered significant for the whole population. If the *random effect* is stronger, the *fixed effect* will not be significant. To illustrate an R script for the mixed model, the previous example in (78) is modified below by adding authors as individual variation (random effect) and using the function *lmer()*:

(79)
```
>lmer(OVVO ~ Period + VerbType + (1|Authors),
  data=mydata, family = binomial)
```

### 5.3.5 Null Hypothesis and Bayesian Degree of Belief

Null hypothesis testing is no doubt one of the most common forms of statistical analysis. Following a comparison of observed and expected values and calculating a p-value, the null hypothesis is accepted if there is no effect ($p > 0.05$) or rejected if the p-value is smaller. However, several issues have been recently discussed in the literature with respect to the null hypothesis method. Firstly, null hypothesis tests are influenced by the number of observations. That is, the larger the number of observations, the more likely the null hypothesis is to be rejected. Secondly, all observations are treated as independent from each other. Furthermore, the interpretation of p-value is, in fact, a binary decision: the analysis considers only whether its size is more or less than 0.05. Finally, null hypothesis analysis looks at the probability of observed data given the null hypothesis $P(data|H_0)$, whereas the research question is really about the probability of the hypothesis given the data $P(H_0|data)$ (Jenset, 2010; Kruschke, 2012).

Bayesian inference is an alternative statistical method that determines the probability of hypothesis given data $P(H_i|data)$.[37] That is, this method 'determines what can be inferred about parameter values given the actually observed data' (Kruschke et al., 2012, 724). The Bayesian inference is constructed as a product of a prior distribution and the likelihood (ratio) of data, as illustrated in (80),

(80)

$$p(\theta|D) = \frac{p(D|\theta)p(\theta)}{p(D)}$$

where $p(\theta|D)$ is a posterior distribution of parameter (belief) given data; $p(D|\theta)$ is a likelihood, or the probability of data given the parameter; $p(D)$ is the 'evidence', or the probability of the data and $p(\theta)$ is a prior probability, or the strength of the belief without data (Kruschke, 2011, 56-58). Compared to the null hypothesis, instead of a fixed p-value, each parameter is expressed in terms of a posterior distribution. That is, each factor is depicted as a high density interval (HDI) that assesses the significance of that factor (Hosmer et al., 2013). Moreover, this method allows for a hierarchical model, where observations are not treated as independent and each new observation relocates the credibility of parameters. This method is available in various R packages. For the present study I use the *jags* package and R scripts from the book *Doing Bayesian data analysis* (Kruschke, 2011). An example of the model is illustrated in (81).

(81)
```
model {
for (i in 1:N){
OVVO[i] ~ dbern(mu[i])
mu[i]<-1/(1+exp(-(b0 + b1*Period[i] + b2*OldNewInformation[i] +
```

[37]For more detail on Bayesian statistics see Kruschke (2011).

```
b3*Focus[i] + b4*VerbType[i] + b5*Animacy[i] + u[Author[i]])))
}
for (j in 1:M) {
u[j] ~ dnorm(0,tau)
}
#Priors
b0 ~ dnorm( 0 , 1.0E-12 )
b1 ~ dnorm( 0 , 1.0E-12 )
b2 ~ dnorm( 0 , 1.0E-12 )
b3 ~ dnorm( 0 , 1.0E-12 )
b4 ~ dnorm( 0 , 1.0E-12 )
b5 ~ dnorm( 0 , 1.0E-12 )
#Hyperprior
tau ~ dgamma(0.001,0.001)
}
```

First, the model assigns a *Bernoulli* distribution (dbern) to the dependent variable (OVVO), since the dependent variable is a binary response. Next, I use a mixed effect logistic model, where $b0$ is an intercept; $b1, b2, b3, b4, b5$ are slopes (regression coefficients) for each fixed parameter (*Period*, *OldNewInformation*, *Focus*, *VerbType* and *Animacy*) respectively; the subscript $[i]$ refers to the value for the *ith* observation and $N$ is the total number of observations. In this model, the parameter *Author* is treated as a *random effect* (see section 5.3.3). The subscript $[j]$ refers to the value for the *jth* group, and $M$ is the total number of groups (authors). The following expression $mu[i] < -1/(1 + exp(-(b0 + ...)))$ is an alternative way of writing a logistic regression in the formula (see the logit function in section 4.1) (Kruschke, 2011, 551-552). The second step is to define a prior distribution for the model. The prior distribution, namely the belief, is an uninformative, very small noncommittal coefficient for each variable. In the present study, the prior is simply uncertain until we start analyzing data. With each new observation, the prior distribution will be relocated.

Each parameter (*fixed effects*) is measured in terms of *posterior distribution*. The most credible values of the parameters (their posterior distribution) are inside the 95% HDI, and if zero-value is excluded from the HDI, we consider this parameter to be credible (Kruschke, 2012). Figure 5.10 demonstrates an example of posterior distribution:

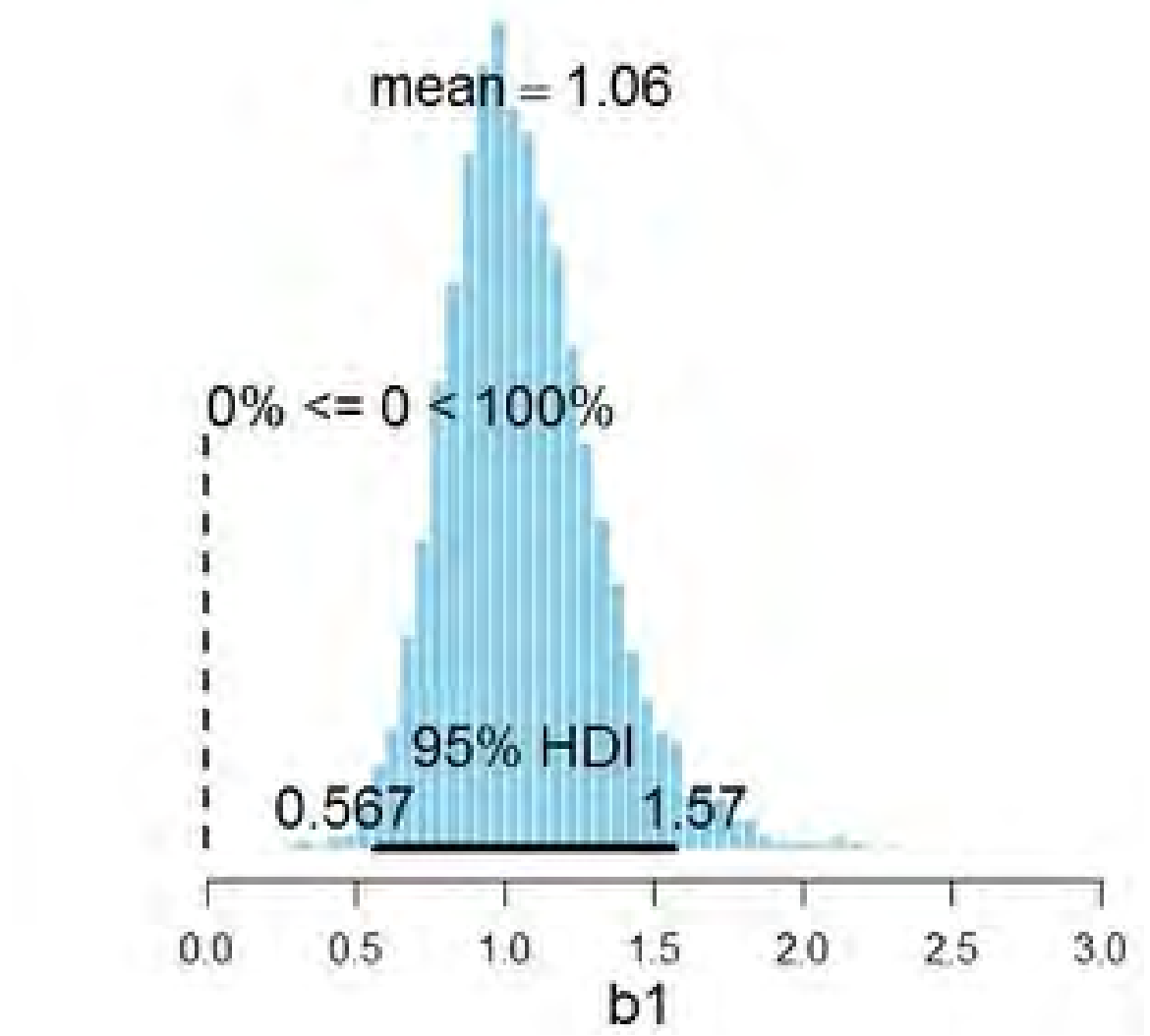

Figure 5.10: Posterior Distribution for a Parameter

Figure 5.10 shows that the values of this parameter on the x-axis are larger than zero and that 95% of the HDI excludes zero. Therefore, this parameter is considered *credible*, or significant. Furthermore, this parameter is expressed in terms of a *positive* or *negative* effect. Given that in Figure 5.10 the values are positive, i.e. above zero, this parameter shows a *positive* effect on a predicted variable. That is, this parameter (factor) favors the increase of the predicted variable. Recall that for a binary predicted variable we assign two values 0 and 1. For example, the *probability* of $OV$ is equal to 0, whereas the probability of $VO$ is equal to 1. In the case of positive HDI values for the parameter, we would say that this factor (parameter) favors the increase of VO. If the values are negative, the factor disfavors the increase of VO.

## 5.4 Summary

Three main methodological contributions of this study have been presented in this chapter. The first contribution concerns the resource-light methods for creating additional annotated resources. Syntactically annotated historical corpora are still very sparse as compared to modern annotated corpora. I have shown that the chronological gaps in the existing annotated corpora can be overcome by using resource-light methods, namely semi-automatic annotation.

The second novel approach is the incorporation of multiple sociolinguistic and linguistic factors into Latin and Old French data. While such a multi-factorial analysis is common in sociolinguistic studies, the application of this method to diachronic word order studies in Latin and Old French is new in comparison to traditional mono-factorial analyses.

Finally, I have introduced advanced statistical tools that are capable of modeling, ranking and clustering multiple factors, allowing for a more subtle understanding of language variation. In addition, their visual representation makes it possible to identify patterns, otherwise hidden in the traditional data representation.

# Chapter 6

# Quantitative Analysis: Latin Word Order

*The statistical analysis of empirical data, to be sure,*
*should not be considered a fancy gadget designed to overwhelm linguists*
*who are generally not really acquainted with statistical techniques.*
*Instead, statistical techniques constitute an essential part*
*of an empirical analysis based on corpus data.*
*(Tummers et al., 2005, 236)*

This chapter will focus on word order distribution in Latin and statistically evaluate previous statements presented in chapters 2 and 3. Particularly, I will investigate whether Late Latin can be considered a VO language. Furthermore, I will examine how the VO form is spread and which syntactic conditions advance the spread of this new form over time. The first section of this chapter will examine attested word order patterns in Latin. Particularly, I will focus on verb-initial and verb final constructions and examine their relation to information structure. Subsequently, section 2 will focus on the multi-factorial statistical analysis.

## 6.1 Descriptive Analysis

### 6.1.1 Word Order Patterns

Table 6.1 and Figure 6.1 show the overall distribution of OV-VO order in Latin infinitival clauses. The lowest rate of VO (20%) is observed during the Classical period of Latin (1st century BC), whereas the highest rate (57%) occurs during the Early Late period (4th century). Figure 6.1 suggests a steady but slow diffusion of the VO form, with an unexpected increase in the 4th century, which is more likely to be related to the nature of Vulgar Latin specimens in the data, namely Jerome's Vulgata.

Table 6.1: OV-VO Distribution in Infinitival Clauses: Latin

| Period (century) | OV | % | VO | % | Total |
|---|---|---|---|---|---|
| 1st BC | 210 | 80 | 54 | 20 | 264 |
| 1st AD | 43 | 75 | 14 | 25 | 57 |
| 2nd | 82 | 68 | 39 | 32 | 121 |
| 4th | 152 | 43 | 203 | 57 | 355 |
| 6th | 102 | 57 | 77 | 43 | 179 |
| Total | 589 | 60% | 387 | 40% | 976 |

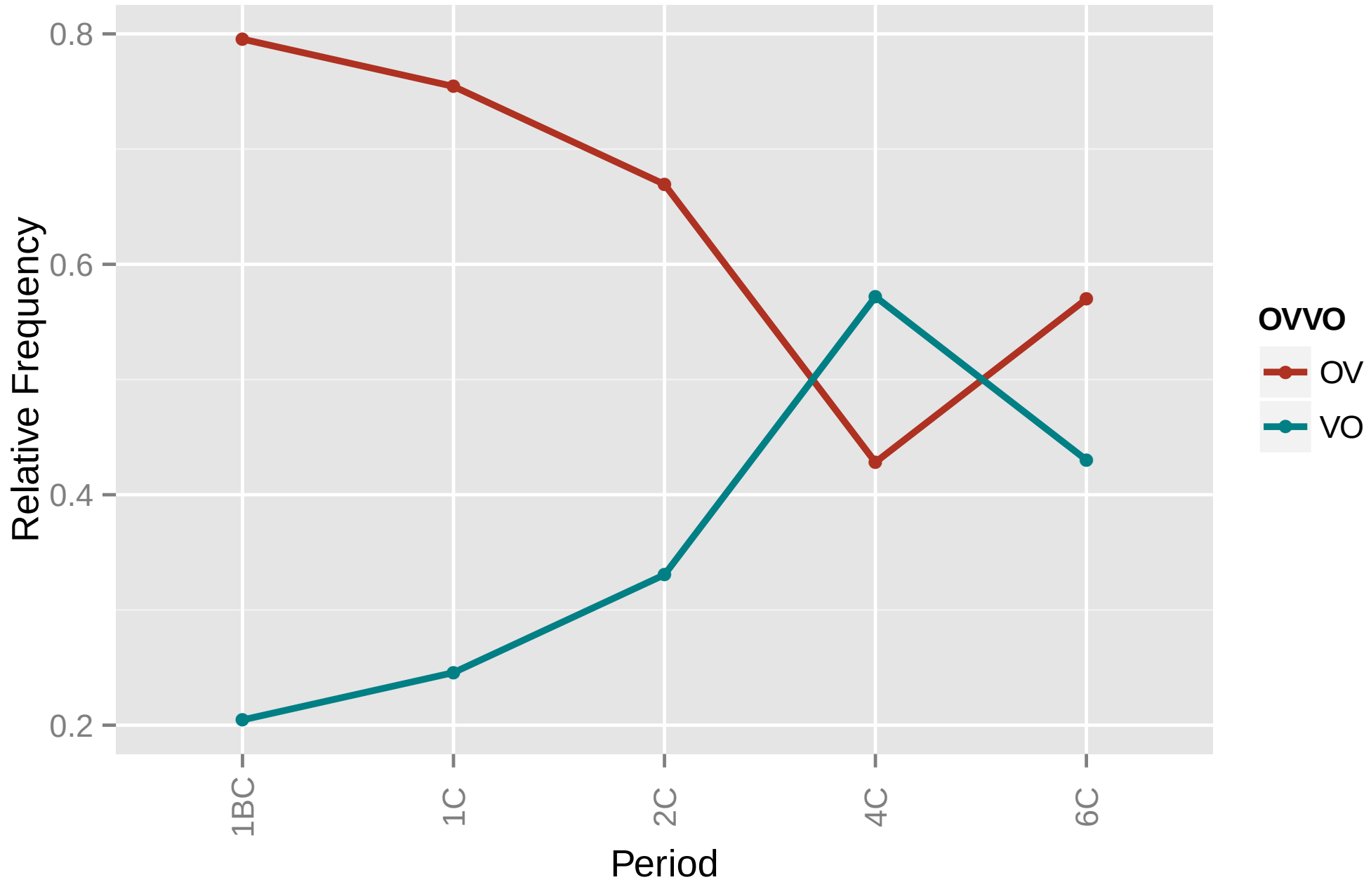

Figure 6.1: Frequency of Word Order in Infinitival Clauses: Latin

The rate of change is presented in Figure 6.2, using the logistic transform from the Constant

Rate Model (Kroch, 1989c). This transform allows us to identify the initial value (intercept) and the rate of change (slope) (see section 4.2). In Figure 6.2, the x-axis represents periods on a continuous scale and the red line is the logistic transform of the regression line. The red line has a slope equal to 1.6 ($exp(0.48)$) and an intercept equal to 0.2 ($exp(-1.64)$).[1] From this plot, it is clear that there is a statistically significant rise of VO order in Latin, as the slope is not equal to zero.

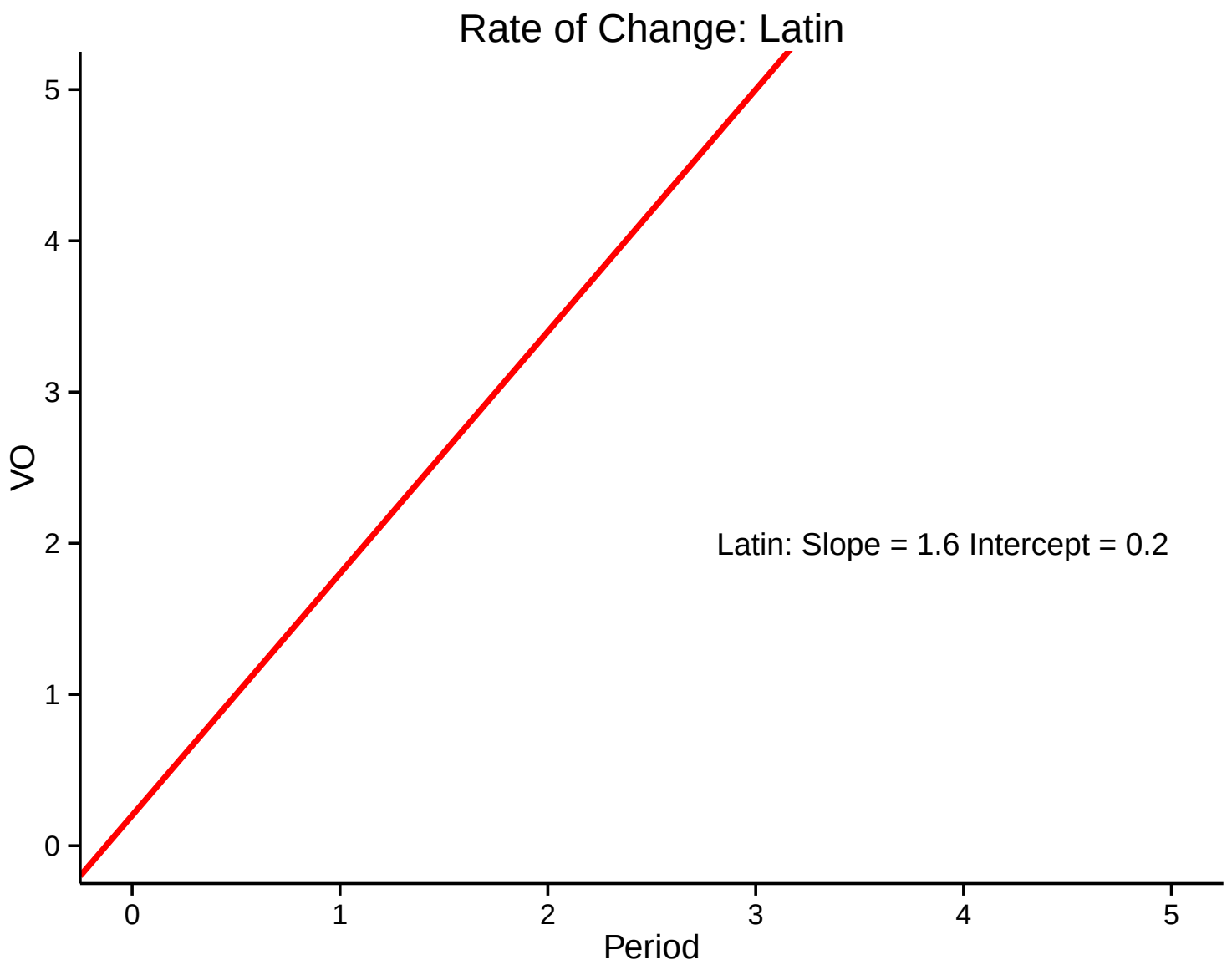

Figure 6.2: Rise of VO in Infinitival Clauses in Latin (Logit Transform)

The present data also exhibits a variety of word order patterns. Recall from section 2.2 that there are six possible word order patterns in Latin, namely SOV, SVO, VSO, OSV, VOS and OVS. In the present corpus all patterns except VOS are found, as shown in (82):

(82) a. SOV

| | | | | |
|---|---|---|---|---|
| Dolabellam | quinque | cohortis | misisse | in Chersonesum |
| Dolabella-nom.sbj | five | cohorts-acc.obj | send-inf.past | in Chersonese-acc |

'Dolabella has sent five cohort to the Chersonese' (Cicero, *Brutus1* 14)

b. SVO

[1] To obtain the measurement for slope and intercept, logodds units from the logit function have been converted using the *exp* function in R.

| | | | | | |
|---|---|---|---|---|---|
| Puto | mehercules | illum | reliquisse | solida | centum |
| believe-1p.sg | my-hercules | he-acc.sbj | leave-inf.past | solid-acc | hundred-acc.obj |

'Upon my word, I believe he left a round hundred million behind him' (Satyricon, 43)

c. VSO

| | | | | | | | |
|---|---|---|---|---|---|---|---|
| Indicavimus | senatui | ex | Norbano | didicisse | nos | publicam | causam |
| indicate-1p.pl | senate-dat | from | Norbanus-abl | learn-inf.past | we-acc.sbj | public | lawsuit-acc.obj |

'We acquainted the Senate, that as we had received our briefs in a public prosecution from Norbanus' (Pliny, Book 3 Letter9)

d. OSV

| | | | | | |
|---|---|---|---|---|---|
| neque | modum | neque | modestiam | victores | habere |
| no | moderation-acc.obj | no | restraint-acc.obj | victors-acc.sbj | have-inf |

'the victors showed neither moderation nor restraint' (Sallust, 11.4)

e. VOS Not attested

f. OVS

| | | | | | | | | | | | | | |
|---|---|---|---|---|---|---|---|---|---|---|---|---|---|
| Non | ideo | tamen | eximiam | gloriam | meruisse | me, | ut | ille | praedicat, | credo, | sed tantum | effugisse | flagitium |
| not | therefore | yet | extraordinary-acc | fame-acc.obj | merit-inf.past | I-acc.sbj | as | they-nom.sbj | declare-3p.pl, | think-1p.sg | but such-acc | flee-inf.past | misfortune-acc.obj |

'On this account, I do not think that I earned this great fame that my friend bestows on me, rather I think that I just escaped a disgrace' (Pliny, *Book 3 Letter1* 1)

Two patterns, namely OSV (82d) and OVS (82f), clearly demonstrate *Contrast* on nominal objects with focus particles *neque ... neque* 'nor ...nor' and *non ...sed* 'not .. but'. Another contrastive particle, *mehercule* 'really', is found in SVO (82b). This type of *Contrast* usually refers to actions, that is, the action of leaving (*reliquisse*) is focalized (see section 2.2). The interpretation of VSO in (82c) is not clear, as there is no explicit focus marking either on verb or nominal phrase. Finally, the SOV pattern in (82a) is more likely to be considered an unmarked order, as from the context of Cicero's letter, it is clear that Cicero simply shares this information with Brutus. As shown in section 2.2, Latin is a discourse-functional language, where pragmatic functions define word order patterns. The examples in (82) taken from infinitival clauses confirm this generalization about Latin, as we see that pragmatic features, such as *Contrastive Focus*, trigger the following linear orders: OSV, OVS and SVO. Previous studies also show that the frequency of each pattern is not the same and that SOV is the most frequent word order. To compare this statement with the present data, Table 6.2 illustrates the use of these patterns across chronological periods.[2]

Table 6.2: Word Order Patterns in Latin Infinitival Clauses: Verb, Object and Subject

| Period | OSV | OVS | SOV | SVO | VSO | Sum |
|---|---|---|---|---|---|---|
| Classical Latin | 5 | 0 | 33 | 3 | 0 | 41 |
| Late Imperial Latin | 1 | 2 | 22 | 7 | 2 | 34 |
| Early Late Latin | 3 | 2 | 20 | 19 | 1 | 45 |
| Late Latin | 2 | 0 | 12 | 7 | 1 | 22 |
| Sum | 11 | 4 | 87 | 36 | 4 | 142 |

Indeed, out of 5 available patterns with subject, verb and object, SOV is by far the most frequent order. However, there is a noticeable gradual decrease in SOV.[3] Furthermore, most of the patterns allow for both pronominal and nominal subjects. Table 6.3 illustrates the distribution of subjects, namely pronouns or nouns (NP), with these patterns.[4] SOV and SVO show an almost equal ratio between pronouns and nouns as subjects, which can also be observed from Figure 6.3.

[2]In this analysis only the following patterns are included: OSV, OVS, SOV, SVO and VSO. The remaining patterns, namely OV, VO, OVX, OXV, VOX, VXO, XOV, XOVX, XVO and XVOX, will be discussed later.

[3]The data in some tables will be mostly presented as raw frequencies to ensure the comparability of the data with future studies. In addition, given the small numbers and unbalanced data, the percentage will not always be reliable.

[4]No nominal subject is attested with VSO in the present data.

Table 6.3: Word Order Patterns in Infinitival Clauses: Subject Types

| Century | Type | NP | Pronoun | Type | NP | Pronoun | Type | Pronoun |
|---|---|---|---|---|---|---|---|---|
| Classical Latin | OSV | 3 | 2 | OVS | 0 | 0 | VSO | 0 |
| Late Imperial | | 1 | 0 | | 1 | 1 | | 2 |
| Early Late | | 2 | 1 | | 2 | 0 | | 1 |
| Late | | 0 | 1 | | 0 | 0 | | 1 |
| Classical Latin | SOV | 16 | 17 | SVO | 1 | 2 | | |
| Late Imperial | | 5 | 17 | | 3 | 4 | | |
| Early Late | | 11 | 8 | | 10 | 9 | | |
| Late | | 7 | 5 | | 2 | 5 | | |

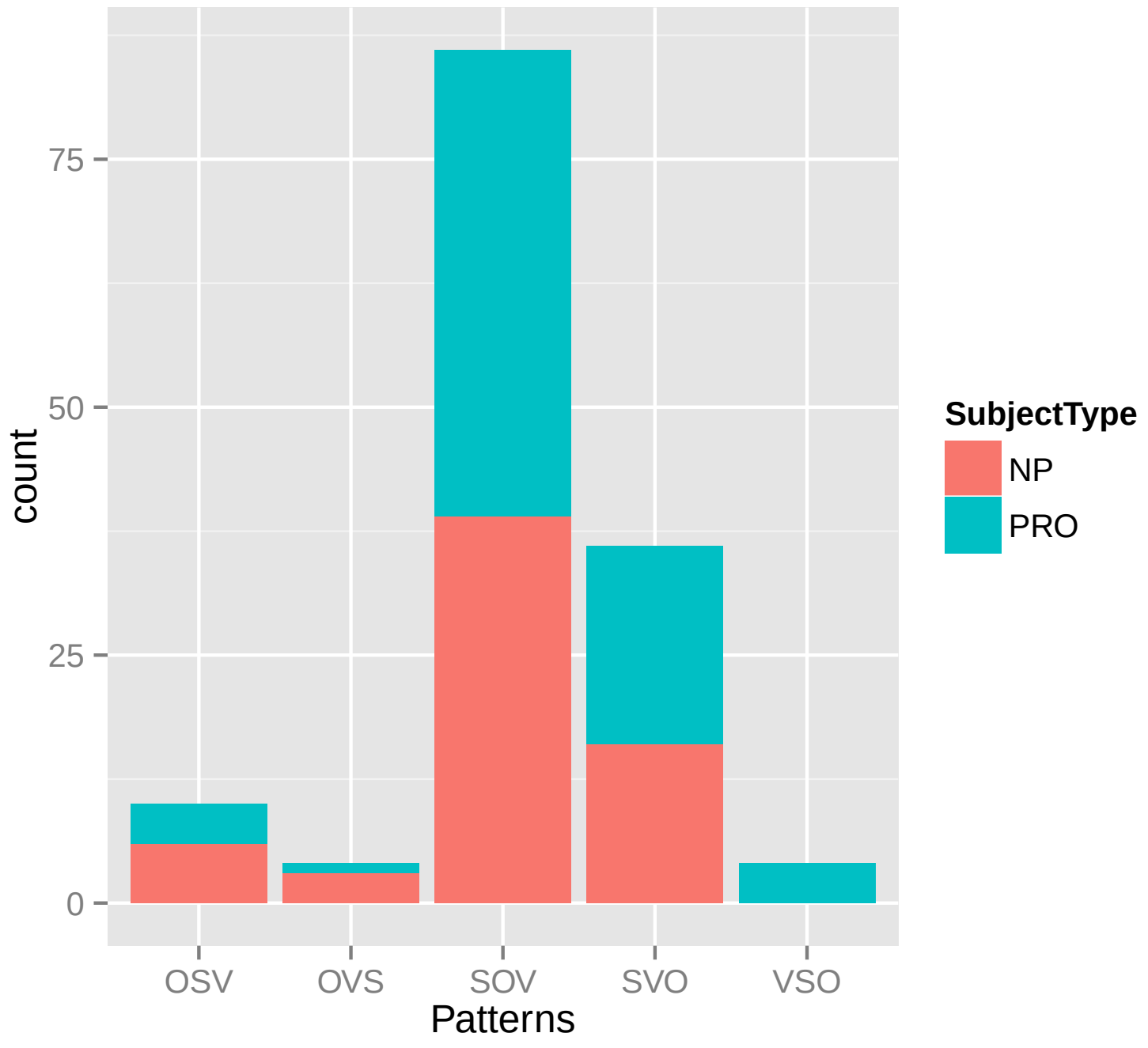

Figure 6.3: Subject Types in Infinitival Clauses

The overall distribution of subject pronouns and nouns over time is presented in Table 6.4. The overall decrease of subject pronouns and nouns is noticeable with preverbal objects (OV). However, the data do not reveal any clear effect of subject type on the choice of OV and VO orders.

Table 6.4: Subject Types in Latin Infinitival Clauses

| | NP | | Pronoun | |
|---|---|---|---|---|
| Period | OV | VO | OV | VO |
| Classical Latin | 19 | 1 | 19 | 2 |
| Late Imperial | 7 | 3 | 18 | 6 |
| Early Late | 15 | 10 | 9 | 10 |
| Late | 7 | 2 | 6 | 6 |
| Sum | 48 | 16 | 52 | 24 |

So far, we have looked at the raw frequencies; however, it is hard to infer any relationship between patterns and periods from these numbers. This type of relationship can be analyzed using a Correspondence Analysis (see section 5.3) (Greenacre, 2007), as shown in Figure 6.4. This method reduces multi-dimensional data into 2 major dimensions. Figure 6.4 displays a two-dimensional representation of word order patterns (SOV, SVO, OVS, OSV, VSO) by four periods of Latin, namely Classical, Late Imperial, Early Late and Late. The proximity of a pattern to a period indicates a strong association with that period. For example, the OSV pattern is strongly associated with Classical Latin. Both Early Late (4th century) and Late Latin (6th century) are clustered with SVO, whereas Late Imperial Latin shows an association with VSO and OVS. Finally, SOV order is associated both with Classical and Late Imperial periods. It should be noted that in this plot the horizontal axis represents 70% of the variation, and the vertical axis represents 24% of variation. Thus, this two-dimensional plot accounts for 94%, which provides a good picture of the variation in the data.

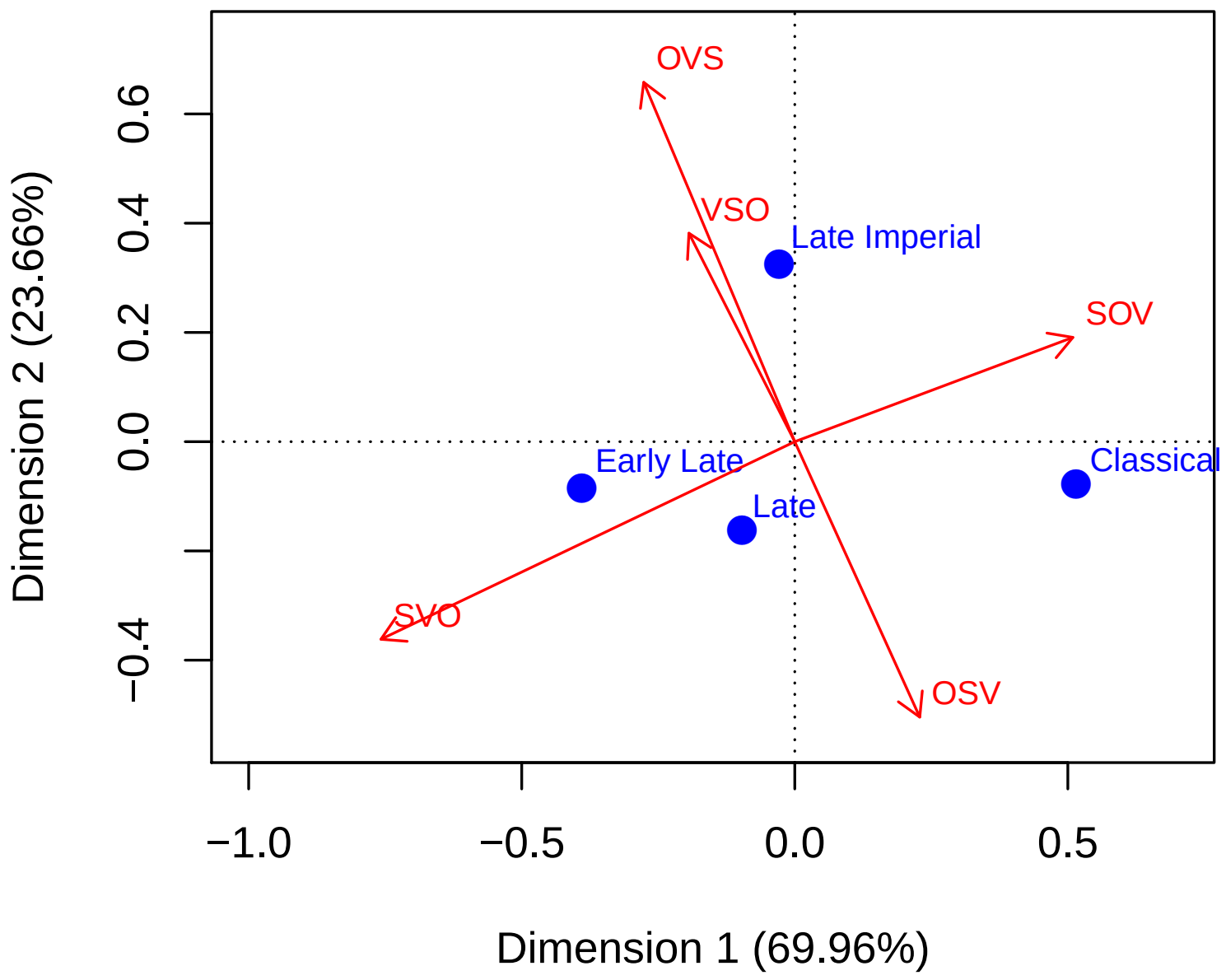

Figure 6.4: Correspondence Analysis I: Latin Word Order Patterns in Infinitival Clauses

These findings require some reflection, since correspondence analysis is a statistically robust technique (see section 5.3) that allows us to make some claims about the findings. According to the data, we can identify the following pattern of change:

(83) Classical Latin **SOV** → Late Latin **SVO**

This pattern also agrees with a traditional direction of word order change. Earlier in section 3.3.1, however, it was shown that the change from Latin to Romance languages undergoes an intermediate stage: SOV >TVX > SVO (35). If, in fact, infinitival clauses do not differ from the regular patterns in Latin, they should be able to reflect this intermediate stage as well. Table 6.5 illustrates all possible patterns with verb, object and any other constituent ($X$), except subject.

Table 6.5: Word Order Patterns in Latin Infinitival Clauses: Verb, Object and X (Other Constituents)

| Period | OVX | OXV | XOV | XOVX | VOX | VXO | XVO | XVOX |
|---|---|---|---|---|---|---|---|---|
| Classical Latin | 8 | 31 | 35 | 1 | 0 | 5 | 13 | 2 |
| Late Imperial | 3 | 17 | 23 | 0 | 6 | 3 | 8 | 0 |
| Early Late | 23 | 22 | 16 | 0 | 23 | 21 | 15 | 0 |
| Late | 5 | 16 | 8 | 0 | 1 | 5 | 27 | 0 |
| Sum | 39 | 86 | 82 | 1 | 30 | 34 | 63 | 2 |

Two observations can be made based on these raw frequencies: i) XOVX and XVOX are rare patterns in the present data and ii) XOV gradually decreases. Another notable observation is a great variation between OV and VO, even in Classical Latin, as illustrated in Table 6.6. Table 6.6 presents the distribution of two patterns OV and VO.[5] Although the OV order slowly decreases, it still remains very frequent.

Table 6.6: Word Order Patterns in Latin Infinitival Clauses: Verb and Object

| Period | OV/% | VO/% | Sum |
|---|---|---|---|
| Classical Latin | 98/77 | 30/23 | 128 |
| Late Imperial Latin | 55/66 | 29/34 | 84 |
| Early Late Latin | 66/35 | 124/65 | 190 |
| Late Latin | 60/63 | 35/37 | 95 |
| Sum | 279/56 | 218/44 | 497 |

The statistical relationship between these patterns, namely OV, OVX, OXV, VO, VOX, VXO, XOV and XVO, and chronological periods is presented in Figure 6.5:[6]

[5]These two patterns do not reflect the generic OV/VO order, they are actual verb+object and object+verb strings without any intervening material or subjects.

[6]The left topmost labels are XOVX and XVOX

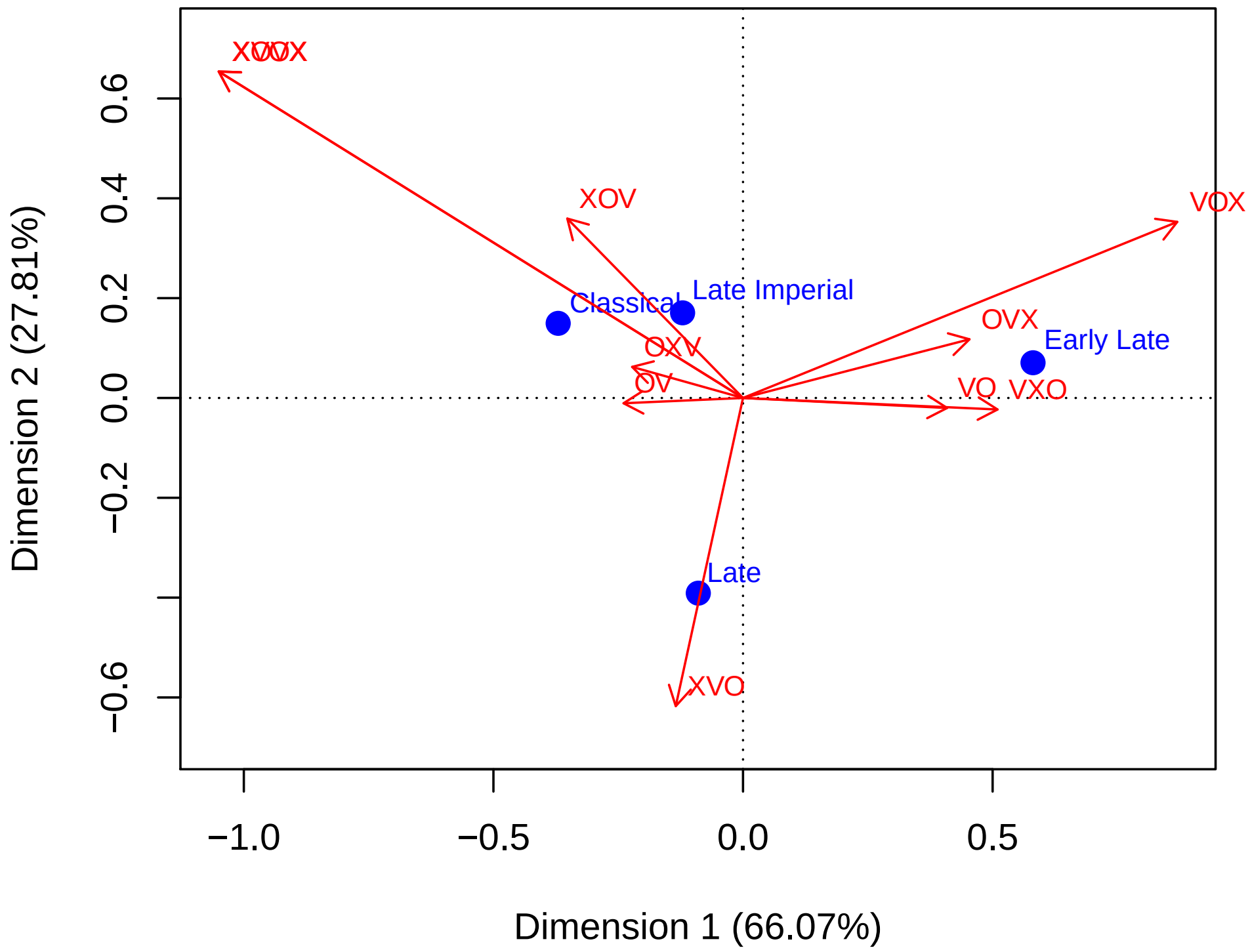

Figure 6.5: Correspondence Analysis II: Latin Word Order Patterns in Infinitival Clauses

First, both Classical and Late Imperial Latin are in the same region (top left), and they have a strong association with XOV, OXV and OV. Second, Early Late Latin and Late Latin are in opposite regions as compared to Figure 6.4. This suggests that they show a distinct relationship with respect to word order patterns. While Early Late Latin is strongly associated with OVX, VOX, VXO and VO, Late Latin has only one cluster with XVO order. Thus, the following path can be established:

(84) a. Classical/Late Imperial **XOV** → Early Late **OVX/VOX/VXO** → Late Latin **XVO**

b. Classical Latin **OV** → Early Late Latin **VO**

Indeed, Early and Late Latin display their association with the intermediate forms OVX

and XVO. That is, they reflect the change from SOV to SVO and from OV to VO via an intermediate stage. In this respect, infinitival clauses do not differ from the regular patterns in Latin. In fact, this finding provides strong support to the hypothesis of this study: the analysis of word order change in infinitival clauses can be applied to the word order phenomenon in general, reflecting language change.

### 6.1.2 Verb-Initial Word Order in Latin (VO)

Several studies have shown that the verb-initial position in finite clauses is considered a marked position. These constructions can carry emphasis and surprise, and can take part in contrastive structures (see section 2.2). In the present data, verb-initial constructions[7] occur in the following contexts: i) contrastive verb (85a), ii) contrastive nominal objects (85b) and ii) non-contrastive constructions (85c):

(85) a. Contrastive Verb

nemo dignus inventus est **aperire** librum nec **videre**
none-nom.sbj worthy find-pp be-3p.sg open-inf book-acc.obj not see-inf
eum
it-acc.obj

'no man was found worthy to open the book, nor to see it' (Jerome, 5.4)

b. Contrastive NP

propter acritudinem non patitur penetrare **cariem** neque
because bitterness-acc not allow-3p.sg penetrate-inf decay-acc.obj no
**eas bestiolas**, quae sunt nocentes
these-acc creatures-acc.obj which be-3p.pl injurious

'Because of its bitterness it prevents the entrance of decay and of those small creatures which are injurious' (Vitruvis, Book2 Chapter 9)

c. No Explicit Contrast

si haec afferre beatitudinem potest
if this-nom convey-inf beauty-acc.obj can-3p.sg

'if this can carry off happiness' (Boethius, 2.4)

[7]Only the following constructions are considered: VO, VOX, VXO.

In addition, it has been argued that this initial position is restricted to certain verbs in Classical Latin, e.g., motion verbs, ditransitives, perception verbs and causatives (Linde, 1923); this restriction, however, is weakened in later periods, where more verb types are seen in the initial position (Salvi, 2005). Based on this statement we would expect to observe only a small number of verb types in Classical Latin as compared to, for example, Late Latin. This hypothesis, however, does not tell us anything about the frequency of these verbs. Assuming that language structure is determined by usage and that frequency affects mental representation (see section 4.4), the weakening of this restriction could be expressed in terms of the entrenchment of the verb-initial position by a high frequency of certain verbs. Before testing this hypothesis, two points need to be addressed. First, it is traditionally argued that heavy constituents trigger VO order, e.g. Bauer (1995). Thus, restricting the dataset to light constituents would allow for a clearer pattern of the distribution. Second, Devine and Stephens (2006) claim that VO is an innovation in Classical Latin that occurs with abstract, non-referential or tail nouns, which are not used as discourse referents or used to refresh the hearer's memory about certain old referents (see also section 3.3.1). Such nouns are more likely to be pragmatically unmarked. The present data are coded for heaviness and information structure; therefore, we can examine these two assumptions. Let us look at heavy and light constituents. Recall that the data are coded for two different types of heaviness, namely heaviness by word count and heaviness by syntactic length. Figure 6.6a illustrates the effect of lengthy constituents by word count, where the x-axis marks the number of words. Figure 6.6b presents the effect of syntactic heaviness, where the x-axis marks light constituents (1) and heavy constituents (2). In this case, heaviness is determined by the presence of NP postmodifiers and coordinated NPs. Figure 6.6a shows that most of the postverbal nouns are one or two words long. Furthermore, most nouns are syntactically light, that is, these nouns are not coordinated or postmodified (see Figure 6.6b). Similarly, the distribution of syntactic length by chronological period reveals that postverbal nouns are mostly light (see Figure 6.6c). The observed pattern suggests that heavy constituents do not trigger VO order in verb-initial infinitival clauses.

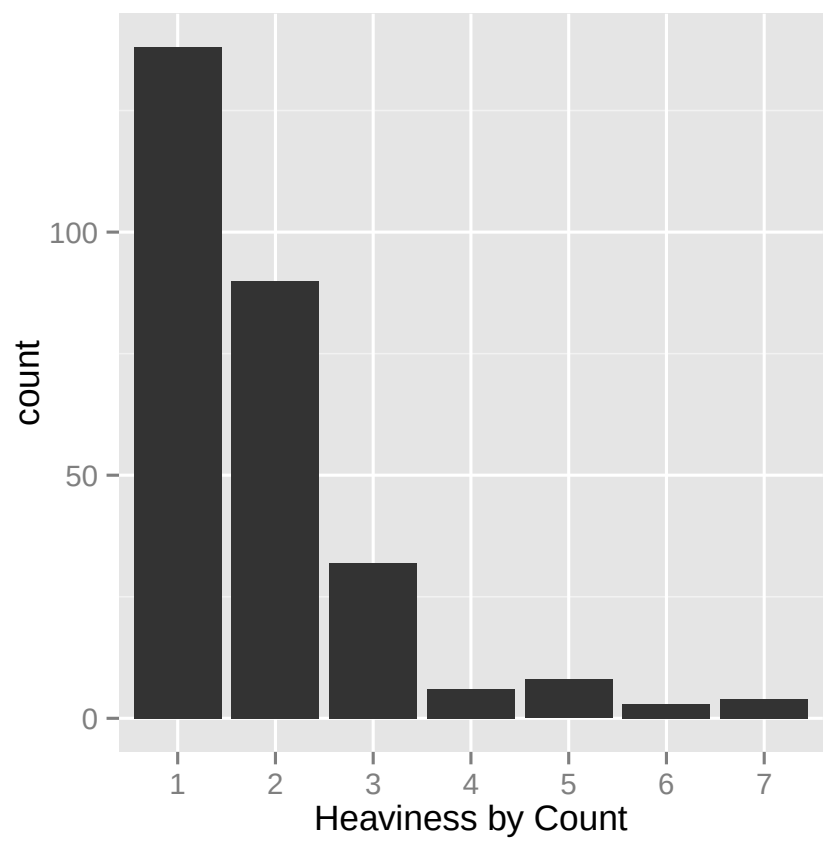

(a) Length by Word Count

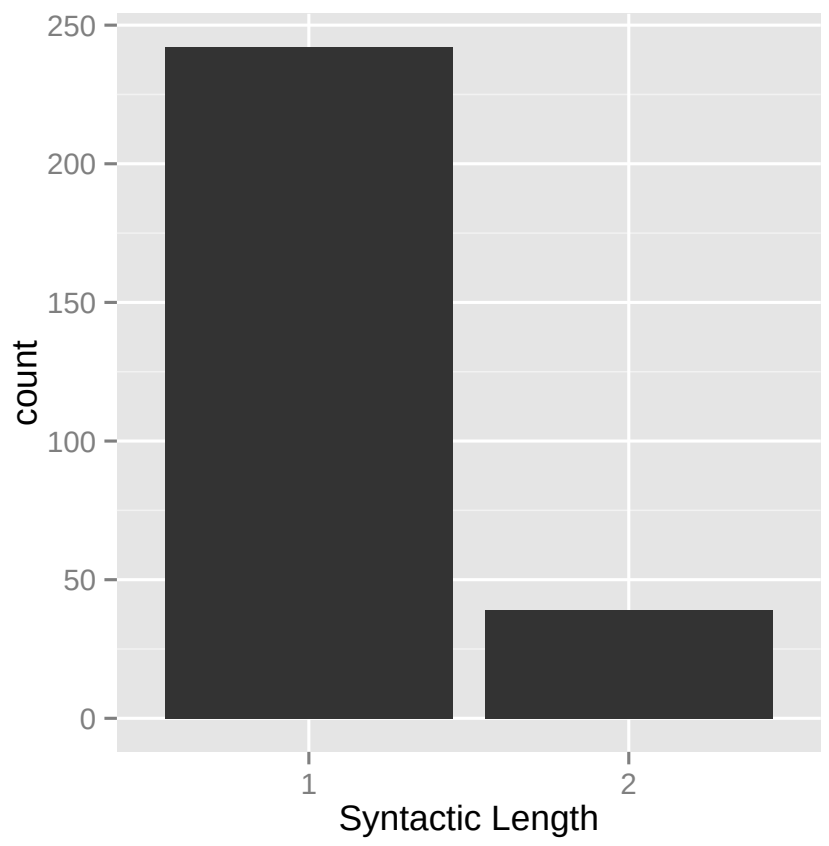

(b) Syntactic Heaviness

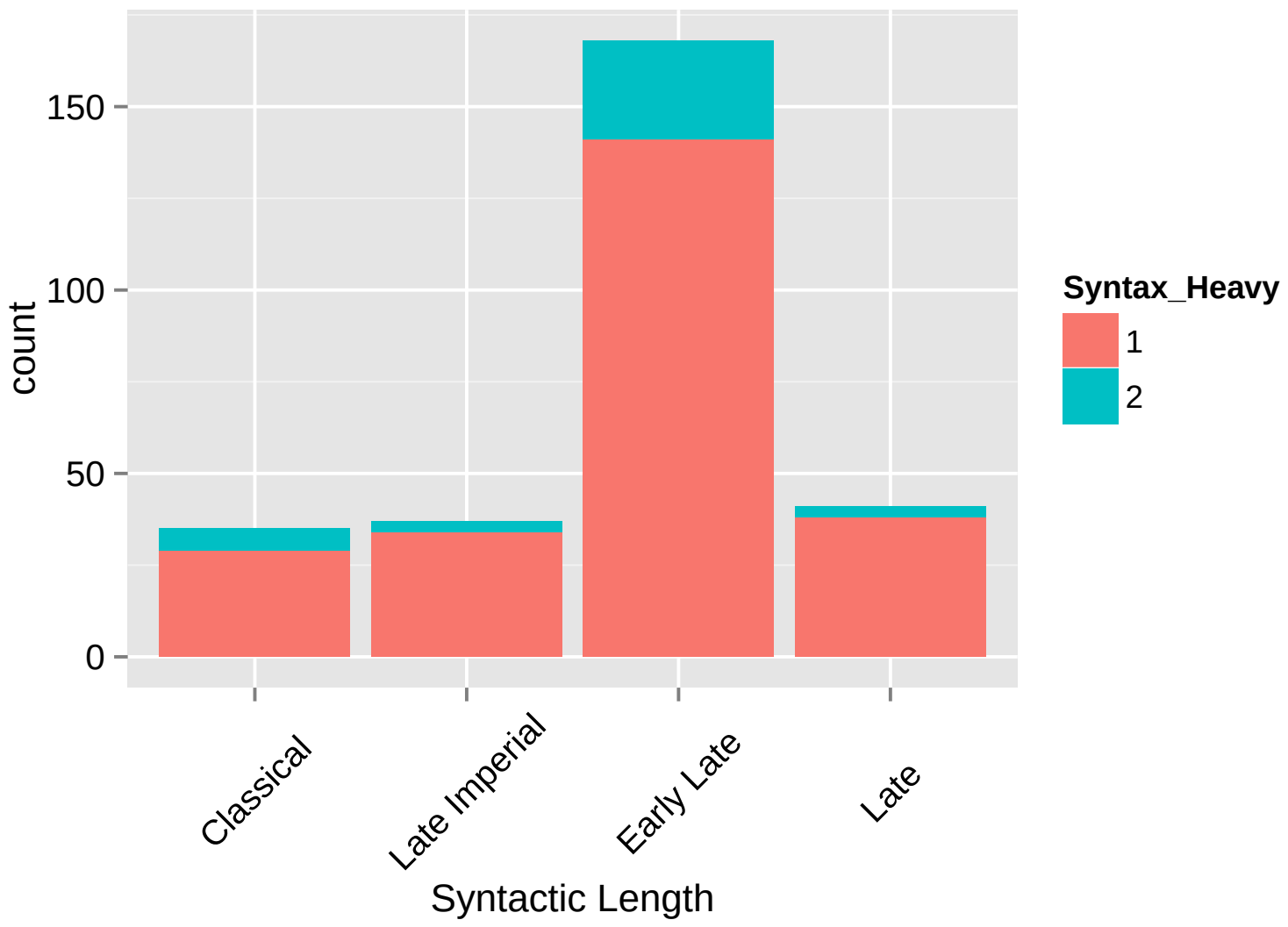

(c) Syntactic Heaviness by Period

Figure 6.6: Verb-Initial Order and NP Heaviness in Latin Infinitival Clauses (VO)

Let us now examine the distribution of information structure in the context of all postverbal object nouns (heavy and light) and compare the results with light postverbal nouns. Recall that information structure is classified on two levels: i) information status and ii) - informational relevance. The first level introduces the following types: i) *old* - tokens that can be evoked from previous context, ii) *accessible* - tokens that can be inferred from common knowledge and iii) *new* - tokens that are not recoverable from the context (see also section 1.3.2). The second level distinguishes between contrastive focus and new information focus (non-contrastive). Table 6.7 illustrates the results for all postverbal nouns, and Table 6.8 presents the information structure for light postverbal nouns.

Table 6.7: Information Structure of Postverbal NP (VO, VOX and VXO) in Latin Verb-Initial Infinitival Clauses

| Period | *New*/% | *Accessible*/% | *Old*/% | *NP Contrast*/% | *Inf. Focus*/% |
|---|---|---|---|---|---|
| Classical Latin | 15/43 | 1/3 | 19/54 | 2/6 | 33/94 |
| Late Imperial Latin | 11/30 | 7/19 | 19/51 | 4/11 | 33/89 |
| Early Late Latin | 71/42 | 44/26 | 53/32 | 21/13 | 147/87 |
| Late Latin | 26/63 | 4/10 | 11/27 | 7/17 | 34/83 |
| Sum | 123 | 56 | 102 | 34 | 247 |

Table 6.8: Information Structure of Light Postverbal NP (VO, VOX and VXO) (NP = One Word) in Latin Verb-Initial Infinitival Clauses

| Period | *New*/% | *Accessible*/% | *Old*/% | *NP Contrast*/% | *Inf. Focus*/% |
|---|---|---|---|---|---|
| Classical Latin | 3/30 | 1/10 | 6/60 | 1/10 | 9/90 |
| Late Imperial Latin | 8/32 | 5/20 | 12/48 | 4/16 | 21/84 |
| Early Late Latin | 36/52 | 11/16 | 22/32 | 10/15 | 59/85 |
| Late Latin | 16/64 | 3/12 | 6/24 | 6/24 | 19/76 |
| Sum | 63 | 20 | 46 | 21 | 108 |

There is a wide range of pragmatic features on postverbal nouns. In fact, the ratio between old and new information remains almost the same with heavy and light nouns. Another observation is that old information has a higher rate in Classical and Late Imperial Latin, while new information becomes more frequent in Late Latin. Although there are more non-contrastive nouns (new information focus), there is a slight increase in contrastive postverbal nouns. Recall that the aforementioned innovation in Classical Latin, proposed by Devine and Stephens (2006) (see section 3.3.1), affects only non-referential, abstract or tail nouns. Figure 6.7 shows that specific and non-specific nouns are more frequent in the postverbal position than abstract nouns.[8] However, there is an increase in abstract nouns in the postverbal position, as illustrated in table 6.9.

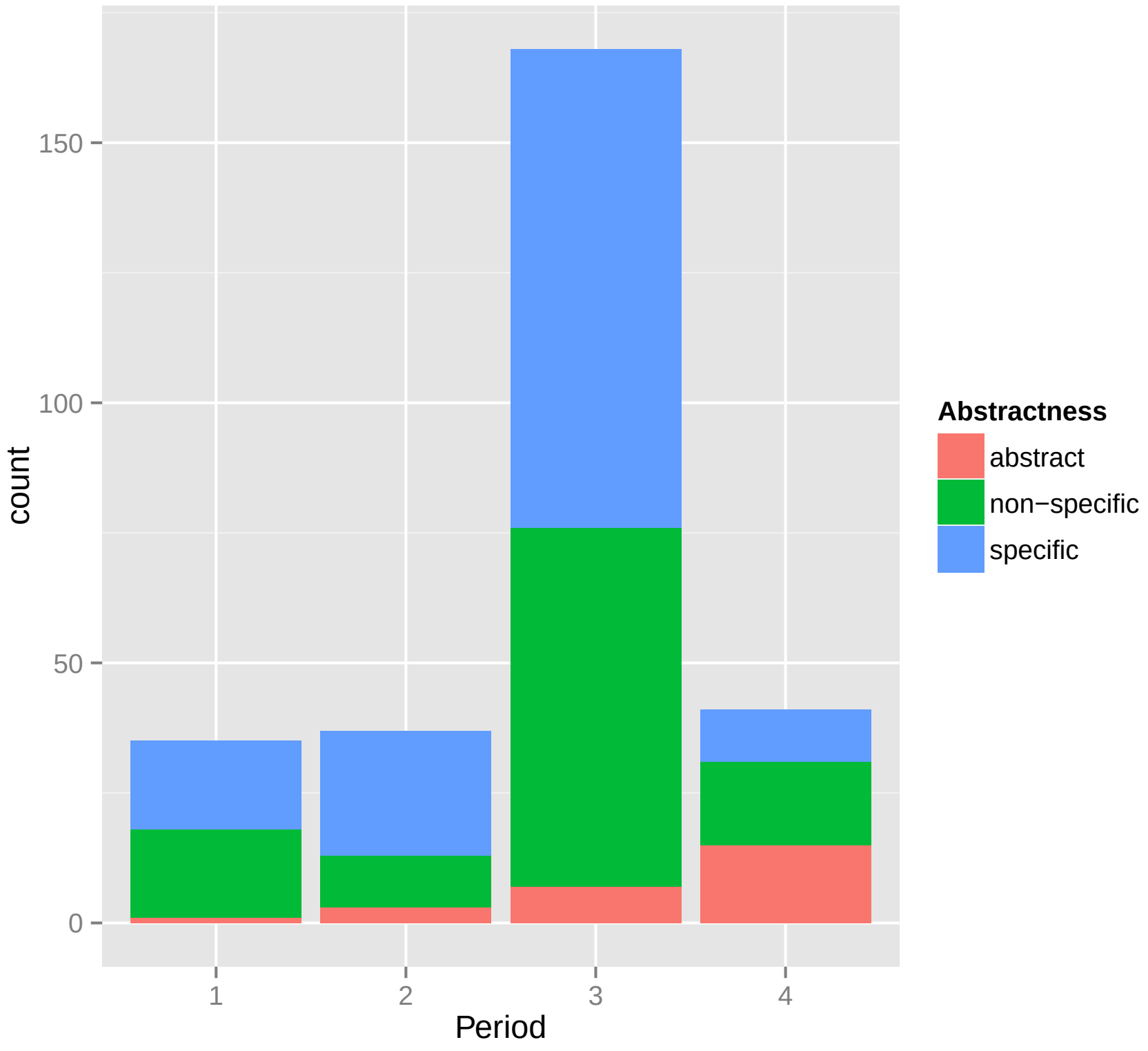

Figure 6.7: Specific/Non-Specific Postverbal NPs (VO, VOX and VXO) in Latin Verb-Initial Infinitival Clauses

[8] Abstract, specific and non-specific nouns are defined through context.

Table 6.9: Distribution of Specific/Non-Specific/Abstract Postverbal NPs (VO, VOX and VXO) in Latin Verb-Initial Infinitival Clauses

| Period | Abstract NP | Non-specific NP | Specific NP |
|---|---|---|---|
| Classical Latin | 1 | 17 | 17 |
| Late Imperial Latin | 3 | 10 | 24 |
| Early Late Latin | 7 | 69 | 92 |
| Late Latin | 15 | 16 | 10 |

Furthermore, Figure 6.8 demonstrates that non-animate postverbal nouns are more common in postverbal position than animate nouns.

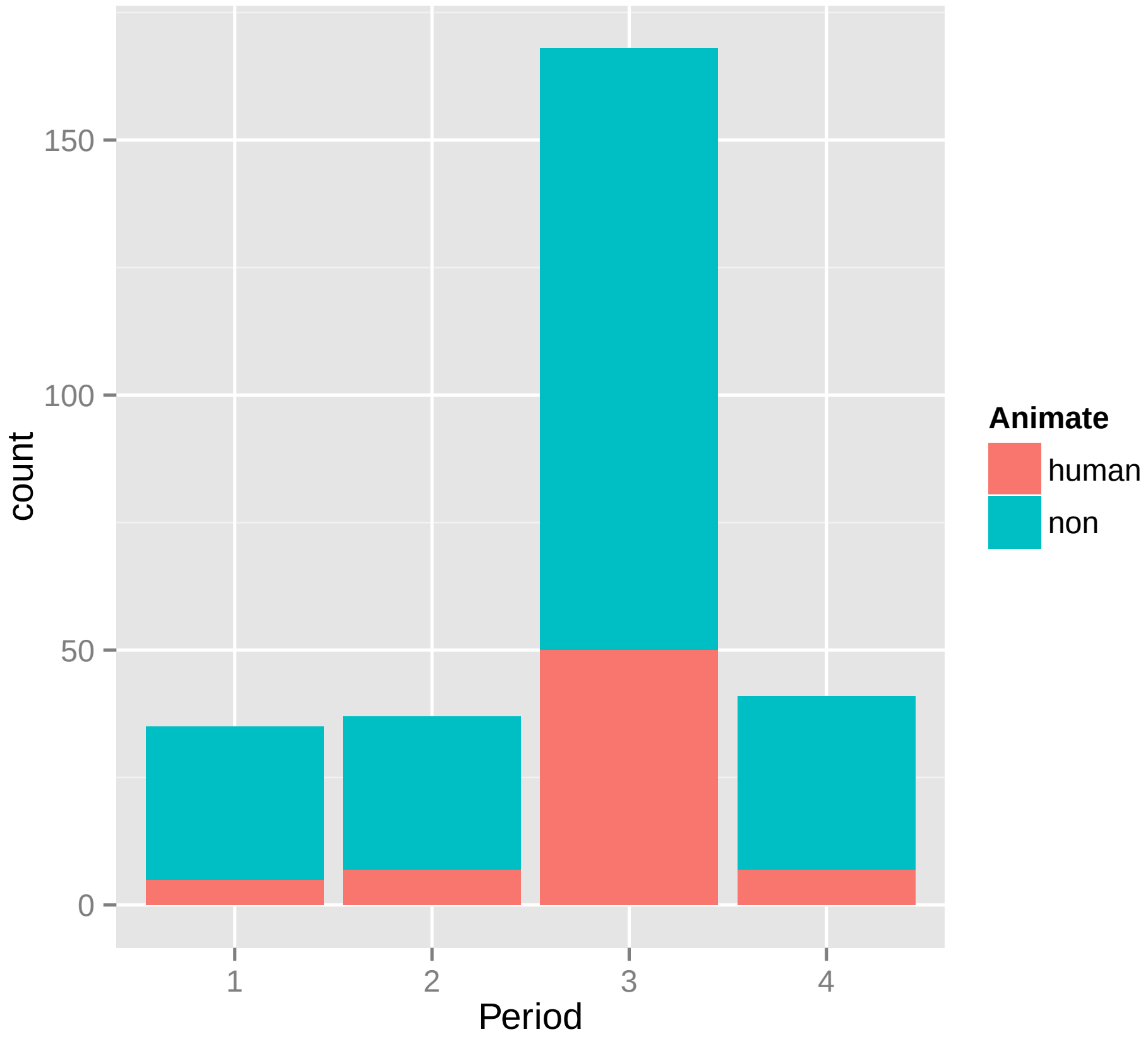

Figure 6.8: Animate/Non-Animate Postverbal NPs (VO, VOX and VXO) in Latin Verb-Initial Infinitival Clauses

Figure 6.9 and table 6.10 report the pragmatic values for each type of noun, namely abstract, non-specific and specific, across chronological period.[9] Two tendencies can be observed from Figure 6.9: i) an increase of *new information* with abstract and non-specific nouns and ii) an increase of *old information* with specific nouns.

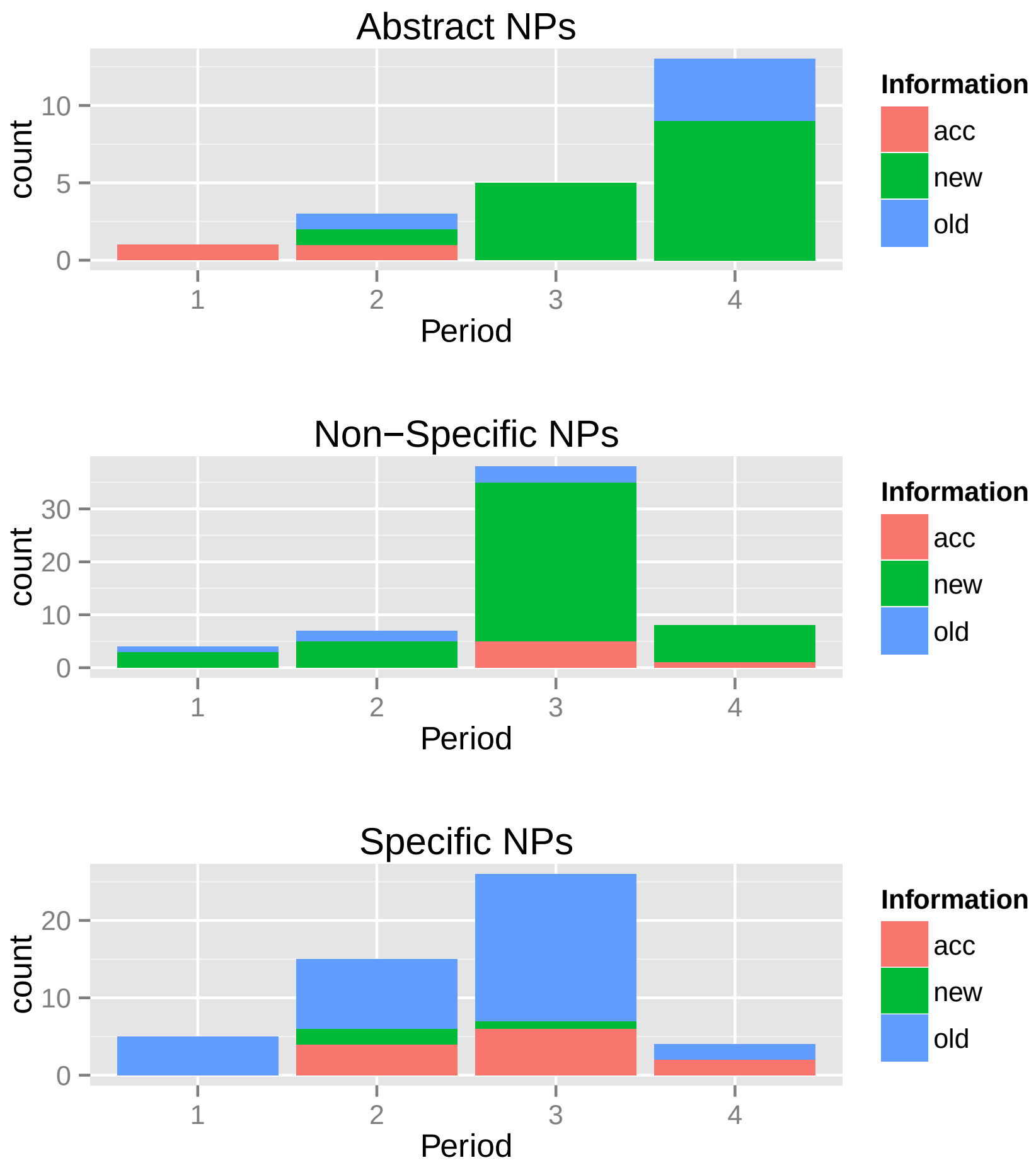

Figure 6.9: Pragmatic Values of Postverbal Light NPs (VO, VOX and VXO) in Latin Verb-Initial Infinitival Clauses

[9]This table reports only raw frequencies, as some frequencies are very small and the percentages will not be appropriate here.

Table 6.10: Pragmatic Values of Postverbal Light NPs (VO, VOX and VXO) in Latin Verb-Initial Infinitival Clauses

| Period | Abstract | | | Non-Specific | | | Specific | | |
|---|---|---|---|---|---|---|---|---|---|
| | acc | new | old | acc | new | old | acc | new | old |
| Classical Latin | 1 | 0 | 0 | 0 | 3 | 1 | 0 | 0 | 5 |
| Late Imperial Latin | 1 | 1 | 1 | 0 | 5 | 2 | 4 | 2 | 9 |
| Early Late Latin | 0 | 5 | 0 | 5 | 30 | 3 | 6 | 1 | 19 |
| Late Latin | 0 | 9 | 4 | 1 | 7 | 0 | 2 | 0 | 2 |
| Sum | 2 | 15 | 5 | 6 | 45 | 6 | 12 | 3 | 35 |

Finally, by restricting the dataset to non-specific and abstract nouns we can evaluate and compare their pragmatic values with the data from Devine and Stephens (2006). In addition, we can restrict the dataset to non-animate nouns, as Devine and Stephens (2006) show that animate nouns are more flexible with respect to their position. The results are reported in Table 6.11.[10]

Table 6.11: Information Structure of Non-Specific Light Nouns (VO, VOX and VXO) in Latin Verb-Initial Infinitival Clauses

| Period | *New* | *Accessible* | *Old* | *NP Contrast* | *Information Focus* |
|---|---|---|---|---|---|
| Classical Latin | 3 | 1 | 1 | 1 | 4 |
| Late Imperial Latin | 4 | 1 | 3 | 2 | 6 |
| Early Late Latin | 28 | 5 | 1 | 5 | 29 |
| Late Latin | 14 | 0 | 4 | 6 | 12 |
| Sum | 49 | 7 | 9 | 14 | 51 |

The rate of new information is higher than that for old information, and there is also an increase of new information over time for postverbal nouns. In addition, there is some evidence of nouns with contrastive focus in Classical and Late Imperial Latin. Finally, contrastive focus and information focus as well as new information increase their frequencies over time. Recall that in finite clauses in Classical Latin only pragmatically unmarked tail nouns and non-referential abstract nouns could appear postverbally. The results from infinitival clauses show that there is little evidence of pragmatically marked postverbal word order with *contrastive focus*. However, the data are not sufficient to make any assumptions. Consider the following two examples: the example from Devine and Stephens (2006) (42), repeated in (86a), and an example from the present study (86b). In both examples we have non-referential non-specific postverbal nouns, namely *legatos* 'envoys' and *venena* 'poison'. However, in (86b) there is also another non-referential noun *sicas* 'daggers', which is in the preverbal position. In addition, the example in (86b) has focus particles *ne ... sed* 'not only but'. This shows that word order variation and change are very complex phenomena that cannot be explained by pure syntactic analyses such as that of Devine and Stephens, which proposes a VO-leakage account, an innovation allowing an object to remain stranded postverbally in the VP rather than having to move to a preverbal position (Devine and Stephens, 2006) (see section 3.3.1).

[10]This table reports only raw frequencies, as some frequencies are very small and the percentages will not be appropriate here.

(86) a. (...) ut ad regem mitterent **legatos**
(...) that to king send-3p.pl envoys-acc.obj
'(...) that they send envoys to the king' (*Livy*)

b. que ne saltare et cantare sed etiam **sicas** vibrare et
who not jump-inf and sing-inf but also daggers-acc.obj wave-inf and
spargere **venena** didicerunt
scatter-inf poison-acc.obj learn-3p.pl.past
'who learned not only to dance and sing but also to wave daggers and to scatter poison' (Cicero)

Let us return to the hypothesis of verb restriction weakening. By using the reduced dataset, as discussed earlier, the distribution of verbs can be assessed in the reduced light contexts for a number of verbs in the initial position, their frequencies and their types, e.g., perception, causative.[11] Recall that verb-initial order is common with motion verbs;

[11]In the next section, we will see that the *heaviness* factor becomes relevant in Late Latin. Therefore, to avoid an influence from heavy nouns in Late Latin, they are excluded in this analysis.

auxiliary verbs; perception verbs, e.g. *videre* 'to see' and causative and ditransitive verbs, e.g. *dare* 'to give' (Linde, 1923). Table 6.12 reports the results from the present corpus.[12] Each verb type is presented with lemma frequency information, and perception, causative, motion and ditransitive verbs are in bold font.

Table 6.12: Initial Verbs in Latin Infinitival Clauses with Light NPs (VO, VOX and VXO)

| Period | Verb Types |
|---|---|
| Classical Latin | spargere (290) parare (816) exsugere (3) |
| Late Imperial | coniungere (202) effundere (254) operire (92) **accipere** (1899) extrahere (69) numerare (241) ornare (223) **perspicere** (177) **relinquere** (1486) **impertire** (28) exprobare (60) |
| Early Late | adhibere (258) prodere (114) **referre** (1257) **facere** (7733) figere (214) imponere (501) accipere (1899) conducere (157) diffamare (3) diligere (245) **dimittere** (386) eicere (192) **ferre** (3248) **habere** (5596) haurire (147) laudare (685) **mittere** (1943) parare (816) percutere (172) praedicare (138) **reddere** (1187) respondere (689) sanare (86) separare (63) sustinere (408) temptare (104) thesaurizare (3) **videre** (7001) manducare (3) **dare** (5625) **subire** (356) vocare (1152) |
| Late Latin | concedere (565) confluere (31) definire (42) **facere** (7733) **movere** (894) necare (140) occulere (39) praestare (537) replere (50) retinere (564) scire (1709) **scribere** (830) praeterire (266) soluere (542) spernere (150) |

As can be seen from Table 6.12, the present data show no evidence of motion, causative and perception verbs in Classical Latin for this specific context, namely light nouns with new information. In addition, we can see that the verbs are not very frequent.[13] In contrast, in

[12]Glosses are provided in the Appendix A

[13]Recall that frequency is based on lemma frequency from the OperaLatina Lexicon. It is not an actual verb frequency from the present data.

Late Imperial Latin, more verbs appear in this context, and there is evidence of perception, motion and ditransitive verbs, namely *perspicere* 'to observe', *impertire* 'to bestow', *accipere* 'to receive' and *relinquere* 'to leave'. In Early Late Latin, there is a considerable increase in the number of initial verbs with new information, including very frequent verbs such as *habere* 'to have', *mittere* 'to send' and *videre* 'to see'. Many of these verbs allow for an alternation with preverbal and postverbal NPs, e.g. *accipere librum* 'to take a book' and *panes accipere* 'to take bread'. In Late Latin, the list of verbs is smaller; however, most verbs exhibit only postverbal NPs, except the following verbs: *scire* 'to know', *praestare* 'to keep', *soluere* 'to free'. These three verbs also show one occurrence with a preverbal NP. It is clear from Table 6.12 that there is an increase in the number of initial verb types over time. The comparison with final verbs shows that there is also a constant decrease in the number of final verb types across period, as illustrated in Table 6.13.[14]

Table 6.13: Type Frequency of Initial Verbs (VO, VOX and VXO) and Final Verbs (OV, OXV and XOV) in Latin Light Infinitival Clauses

| Period | Verb-Initial/(%) | Verb-Final/(%) | Sum |
|---|---|---|---|
| Classical Latin | 3/(8) | 35/(92) | 38 |
| Late Imperial Latin | 11/(32) | 23/(68) | 34 |
| Early Late Latin | 32/(68) | 15/(32) | 47 |
| Late Latin | 15/(60) | 10/(40) | 25 |
| Sum | 61 | 83 | 144 |

Let us return to the earlier hypothesis with respect to verb restriction. If this hypothesis were true, we would expect only a restricted number of verbs in the initial position in Classical Latin and a subsequent weakening of this condition in later periods, with a considerable expansion of other verbs in the initial position. In the present data, only three verbs occur with postverbal nouns in Classical Latin. None of these verbs are motion or ditransitive verbs. In Late Imperial, there are more verbs in the initial position, including motion and ditransitive verbs. In Early Late Latin, we find an extensive list of verbs in the initial position. This fact suggests that Early Late Latin is the period of expansion. Recall that Salvi (2005) defines a three-phase passage from Latin to Romance word order (37), repeated here:

(87) **Latin** Left periphery | (Focus) (V) [SOXV] | Right periphery

[14]Frequency in Table 6.13 represents a type frequency, namely the unique verb type, in contrast to token frequency, which is the count of all verbs.

**Phase 1** Left periphery | (Focus) **V** [SOX] | Right periphery

**Phase 2** Left periphery | (**Theme**/Focus) **V** [SOX] | Right periphery

The passage from Latin (First Stage) to Phase 1 (Second Stage) is characterized by an expansion of all verbs in the initial position. As a result, this expansion yields an unmarked verb-initial order, where verb preposing becomes a norm. As Salvi (2005) states, in this new phase fronted focused constituents and verbs are not in a complementary distribution, as is the case in Classical Latin. Let us examine our data. First, the present findings suggest that Early Late Latin can be described as *Phase 1* (Second Stage) of the change: there is an extensive list of initial verbs in pragmatically unmarked context, namely new information. Second, there is evidence of cases with focused constituents. Consider the examples in (88):

(88) a. non veni **pacem** mittere sed gladium
not come-1p.sg peace-acc.obj send-inf but sword-acc.obj
'I came not to send peace, but the sword' (Jerome M10.34)

b. quaerebant principes sacerdotum et scribae mittere in
seek-3p.pl.past chief-nom.sbj priests-gen and scribes-nom.sbj throw-inf on
illum **manus** illa hora
him-acc hands-acc.obj that-abl hour-abl
'the chief priests and the scribes sought to lay hands on him the same hour' (Jerome L20.19)

In (88a) we have a ditransitive verb *mittere* 'to send' with a preverbal noun bearing a contrastive focus *pacem* 'peace', whereas in (88b) the same verb occurs with a non-contrastive (information focus) noun *manus* 'hands' in the postverbal position. On the other hand, such evidence is not found in earlier stages. Keeping in mind the small size of this corpus, these findings confirm Salvi's hypothesis with respect to word order change: First, there is a small restricted group of verbs in the initial position. This group is further expanded to other verbs, including very frequent verbs. As a result, this preposing becomes a norm. The second stage concerns noun fronting. In earlier stages this fronting and verb preposing were in complementary distribution. However, from Early Late Latin there is slight evidence that both a fronted noun and a preposed verb can co-occur. Therefore, the present data show that Early Late Latin corresponds to Phase 1 (Second Stage) of change in Salvi's hypothesis.

Finally, it has been shown that verb-initial structures increase their frequencies not only in main clauses but also in subordinate clauses, albeit at a slower rate (Salvi, 2004) (see also Table 3.1). Given that AcI constructions are traditionally considered full sentences, the rate of verb-initial order in the present data can be evaluated in AcI (full sentence) and other infinitives (reduced structure). Figure 6.10 illustrates the distribution of verb-initial order in AcI and other infinitives, and Table 6.14 reports their frequencies.

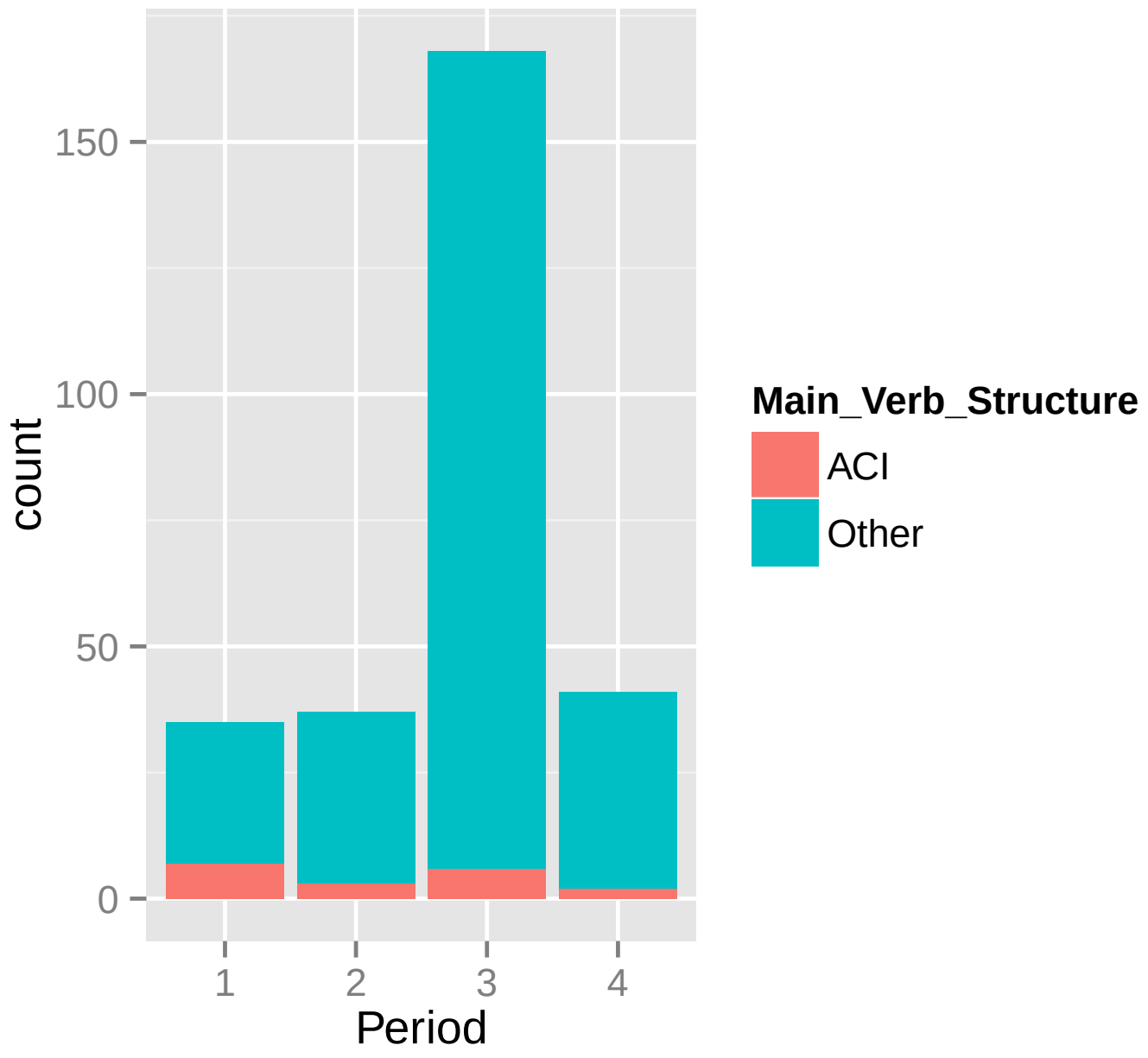

Figure 6.10: Verb-Initial Structure (VO, VOX and VXO) in AcI (full) and Other Infinitives (Reduced) in Latin

Table 6.14: Verb-Initial Structure (VO, VOX and VXO) in AcI (full) and Other Infinitives (Reduced) in Latin

| Period | ACI/% | Other/% |
|---|---|---|
| Classical Latin | 7/20 | 28/80 |
| Late Imperial Latin | 3/8 | 33/92 |
| Early Late Latin | 6/4 | 161/96 |
| Late Latin | 2/5 | 37/95 |
| Sum | 18 | 259 |

From Figure 6.10, it is clear that the relative frequency of verb-initial order is higher in reduced clauses than in AcI.[15] In addition, there is a drastic increase in the frequency for reduced infinitives in Early Late Latin. At this point, the data do not show a robust connection between type of clause and verb-initial order with respect to chronological period.

### 6.1.3 Verb-Final Word Order in Latin

It has been traditionally claimed that verb-final order is the most unmarked position in Latin, which was inherited from Proto-Indo-European. Earlier, in section 2.2, we also saw that not all verb-final orders are considered unmarked, as the initial position can host focalized and topicalized constituents. Table 6.15 illustrates the distribution of information structure for all preverbal nouns in the present data.[16]

[15]Classical Latin - 1, Late Imperial Latin - 2, Early Late Latin - 3 and Late Latin - 4.
[16]Verb-final structures in this analysis are OV, OXV and XOV.

Table 6.15: Information Structure of Preverbal NP (OV, OXV and XOV) in Latin Infinitival Clauses

| Period | *New*/% | *Accessible*/% | *Old*/% | *NP Contrast*/% | *Inf. Focus*/% |
|---|---|---|---|---|---|
| Classic Latin | 89/54 | 33/20 | 42/26 | 32/20 | 132/80 |
| Late Imperial Latin | 41/43 | 22/23 | 32/34 | 15/16 | 79/84 |
| Early Late Latin | 45/44 | 30/29 | 27/27 | 31/31 | 69/69 |
| Late Latin | 27/32 | 27/33 | 29/35 | 14/17 | 69/83 |
| Sum | 202 | 112 | 130 | 92 | 349 |

Like postverbal nouns (Table 6.7), preverbal NPs can take on a wide range of information structure features. There is also a decrease in *new* information, as compared to *old* information, which shows a slight increase over time. Interestingly, the ratio between *new* and *old* information decreases from 2:1 in Classical Latin to 0.9:1 in Late Latin. Recall that with postverbal nouns *old* information is more frequent until Late Latin, where *new* information becomes dominant (see Table 6.7). That is, the distinction between information status features, namely *old* and *new*, for preverbal nouns lessens, whereas this distinction is heightened for postverbal nouns in Late Latin. What about the effect of heaviness on preverbal NPs and information structure? Figure 6.11 demonstrates two types of heaviness, namely word length and syntactic length. Weight distribution is almost identical to the one found for postverbal nouns in Figure 6.6, that is, most of preverbal nouns are one or two words long and they are syntactically light. However, the chronological distribution of syntactic length shows that heavy preverbal NPs have a higher frequency in Classical Latin (see Figure 6.11c), whereas heavy postverbal NPs have a higher frequency in Early Late Latin (see Figure 6.6c). An example of a syntactically heavy preverbal NP is illustrated in (89):

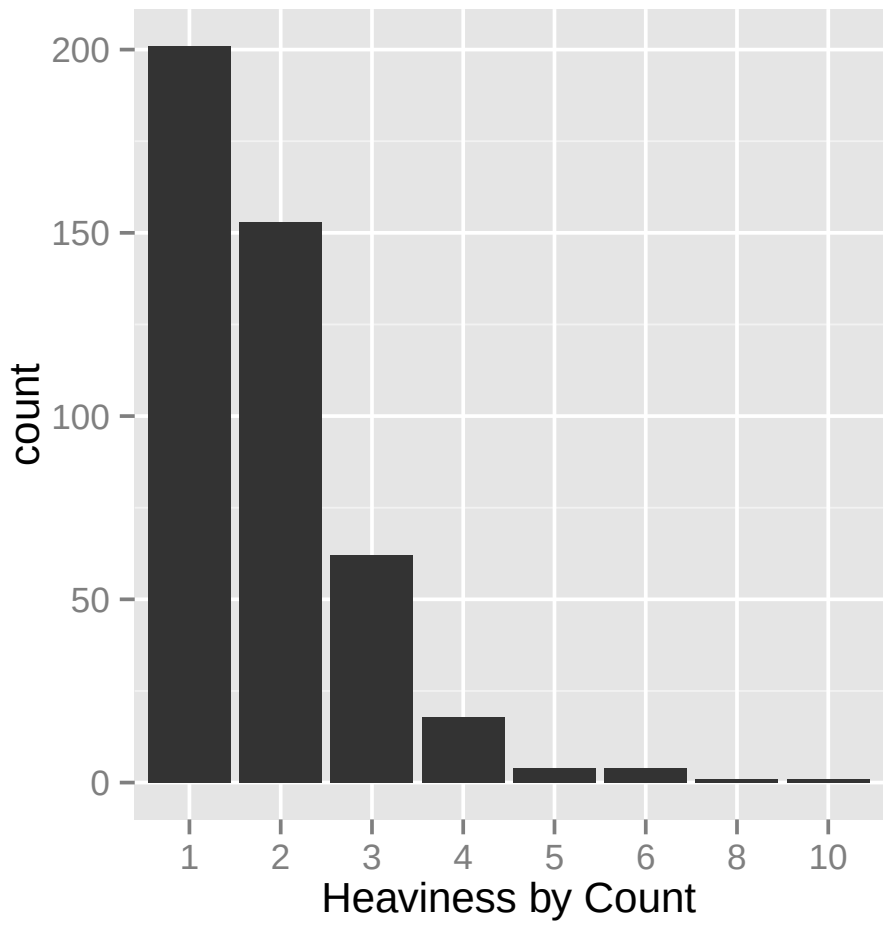

(a) Length by Word Count

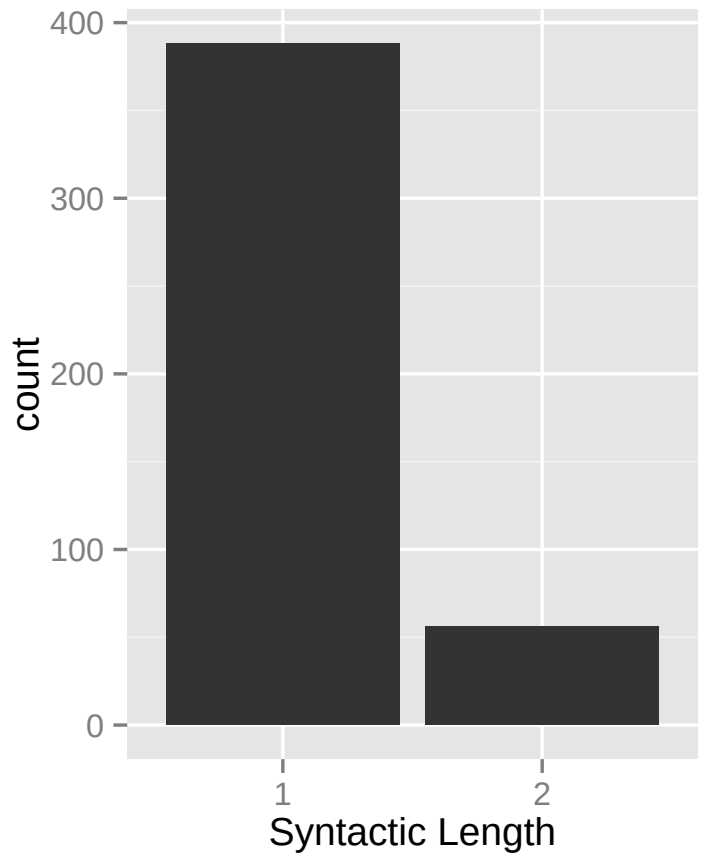

(b) Syntactic Heaviness

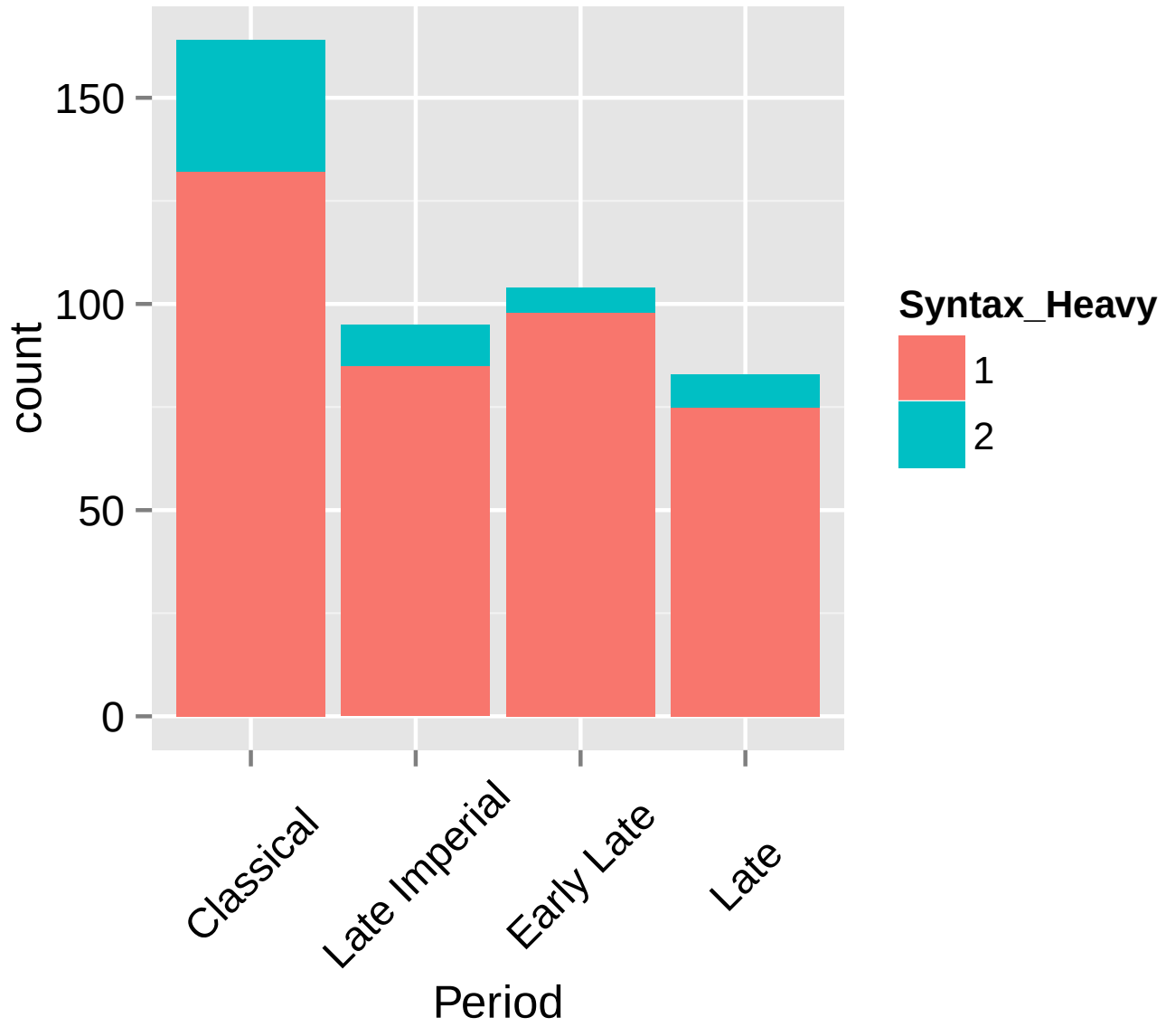

(c) Syntactic Heaviness by Periods

Figure 6.11: Verb-Final Order and NP Heaviness in Latin Infinitival Clauses (OV, OXV and XOV)

(89) **opes** **suas**, **imparatam** **rem** **publicam**, **magna**
work-acc.obj his-acc unprepared-acc thing-acc.obj public-acc great-acc
**praemia** **coniurationis** docere
prizes-acc.obj conspiracy-gen point-inf

‘He pointed out his own resources, the unprepared condition of the state, the great prizes of conspiracy’ (Sallust, 17.1)

Let us now examine the information structure of light preverbal nouns so that we can compare the results with light postverbal nouns. Table 6.16 reports the results. The decrease in *new* information is still noticeable.

Table 6.16: Information Structure of Light Preverbal NP (OV, OXV and XOV) (NP = One Word) in Latin Verb-Final Infinitival Clauses

| Period | *New*/% | *Accessible*/% | *Old*/% | *NP Contrast*/% | *Inf. Focus*/% |
|---|---|---|---|---|---|
| Classical Latin | 47/60 | 15/19 | 17/22 | 11/14 | 68/86 |
| Late Imperial Latin | 23/51 | 13/29 | 9/20 | 5/11 | 48/89 |
| Early Late Latin | 25/49 | 11/22 | 15/30 | 11/22 | 39/78 |
| Late Latin | 10/42 | 5/21 | 9/38 | 4/17 | 20/83 |
| Sum | 105 | 44 | 50 | 31 | 167 |

Finally, it has been argued that verb-final order occurs more frequently in subordinate clauses than in main clauses (Linde, 1923) (see also Table 2.1). Figure 6.12 demonstrates the distribution of verb-final order in AcI (full clauses) and other infinitives (reduced clauses), and Table 6.17 reports their frequencies.

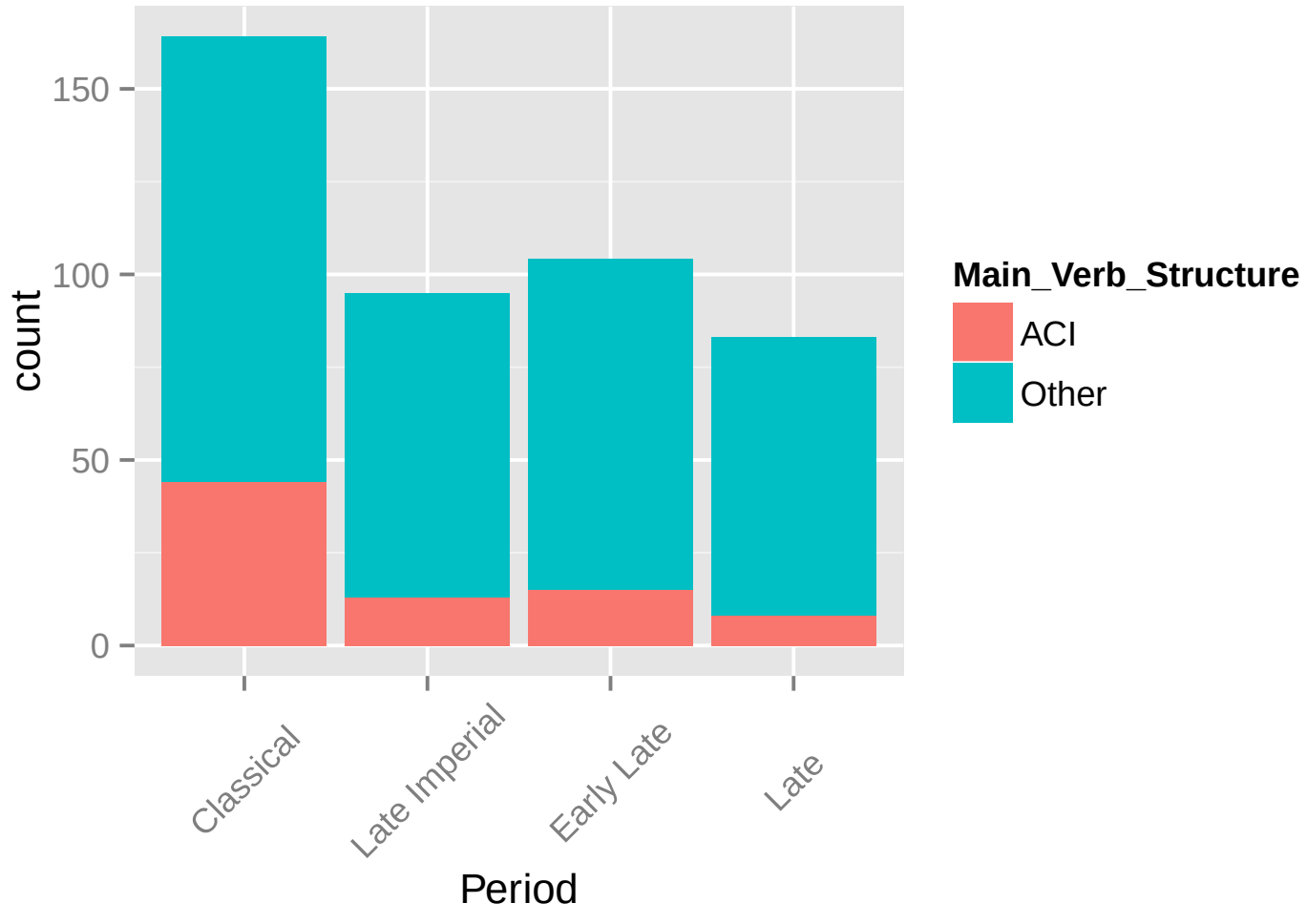

Figure 6.12: Verb-Final Structure in AcI and Other Infinitives in Latin (OV, OXV and XOV)

Table 6.17: Verb-Final Structure in AcI and Other Infinitives in Latin (OV, OXV and XOV)

| Period | ACI/% | Other/% |
|---|---|---|
| Classical Latin | 45/27 | 119/73 |
| Late Imperial Latin | 13/14 | 82/86 |
| Early Late Latin | 15/15 | 87/85 |
| Late Latin | 8/10 | 75/90 |
| Sum | 81 | 363 |

From table 6.17, two observations can be made. First, the rate of verb-final order is higher in reduced clauses than in AcI. Second, there is a constant increase of verb-final order in reduced clauses from 73% in Classical Latin to 90% in Late Latin. In parallel, there is a constant decrease of AcI from 27% in Classical Latin to 10% in Late Latin. These findings suggest a possible correlation between word order and change in type of clause. In the next section, I will look in greater detail at clause types and infinitival positions.

### 6.1.4 Infinitival Clauses: Type and Position

In section 3.3.2, we saw that the verb form plays a role in word order change (Zaring, 2010). Furthermore, in sections 1.3.3.1 and 5.2.2.4 we learned that infinitival clauses are not homogeneous syntactic structures (Cinque, 2004; Iovino, 2010). The results from the previous sections also demonstrate that there are some changes in clause type, namely decrease in AcI and increase in reduced infinitives (see Figure 6.10 and Figure 6.12); however, the data frequency suggests that these changes are not related to the change in word order. Let us examine infinitival clauses in greater detail with respect to clause type, namely the type of main verb. Recall that infinitival clauses are grouped into the following classes: i) Accusativus cum Infinitivo, ii) Raising structure, iii) Control structure, iv) Simple infinitive, v) Restructuring structure and vi) Prepositional infinitive (see section 5.2.2.4). Table 6.18 summarizes the raw frequencies for these five classes.

Table 6.18: Infinitival Clause Types in Latin

| Period | AcI | Control | Raising | Restructuring | Simple |
|---|---|---|---|---|---|
| Classical Latin | 87 | 19 | 7 | 111 | 40 |
| Late Imperial Latin | 45 | 13 | 6 | 77 | 37 |
| Early Late Latin | 62 | 18 | 1 | 174 | 100 |
| Late Latin | 33 | 4 | 11 | 82 | 49 |
| Sum | 227 | 54 | 25 | 444 | 226 |

It appears that the most frequent clause type is restructuring verbs. In addition, there is also a noticeable decrease in AcI constructions. Although it is argued that prepositional infinitives emerge from the 2nd century AD (section 1.3.3.1), the present data exhibit no evidence of prepositional infinitives. Let us look at the OV/VO distribution in these five classes, as illustrated in Figure 6.13. In this plot, the y-axis represents categorical variables, namely verb types, and the x-axis shows chronological periods. The dots correspond to the number of tokens; the red color indicates OV order and the blue color indicates VO order.

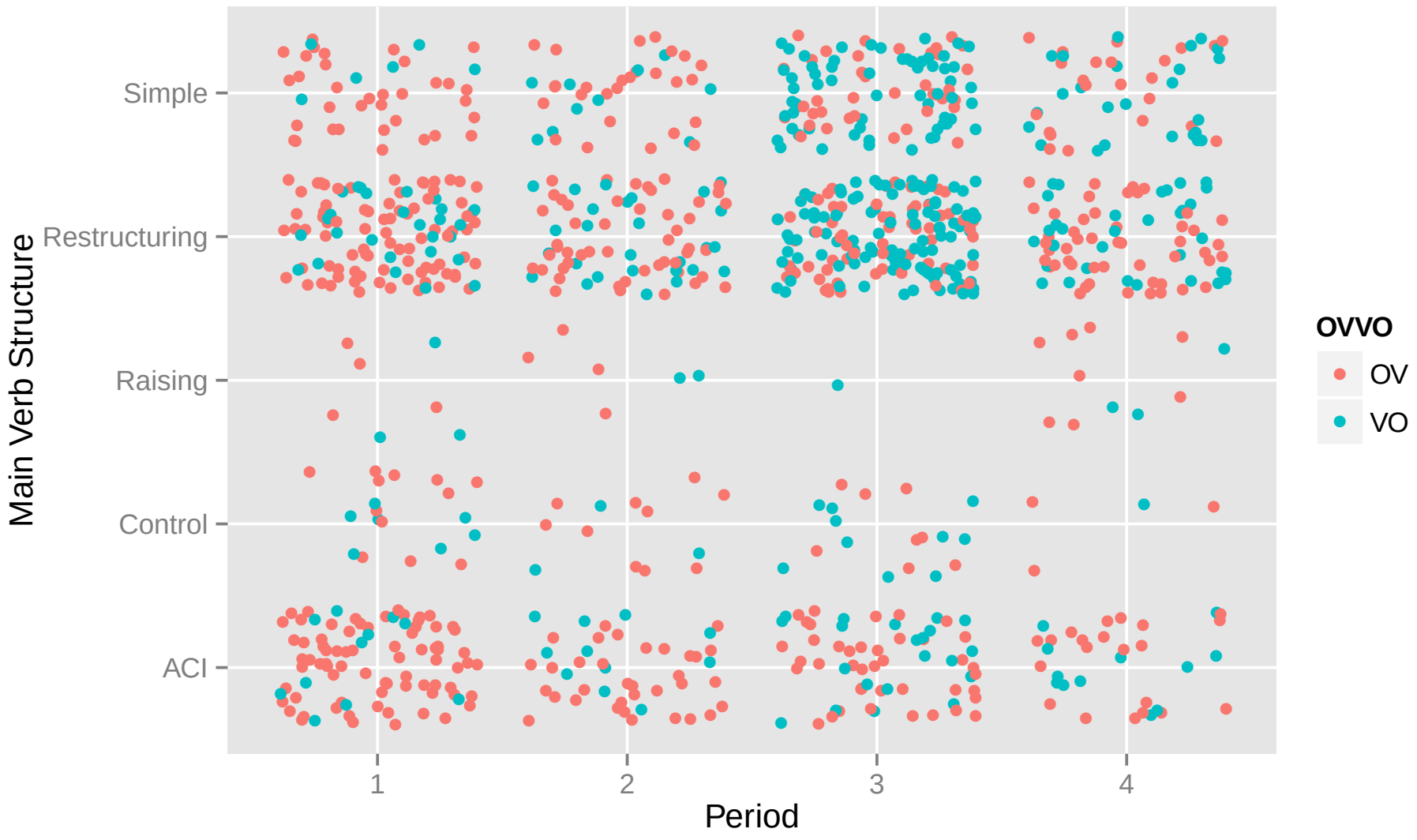

Figure 6.13: Clause Structures and OV/VO Distribution across Periods in Latin Infinitival Clauses

The plot in Figure 6.13 allows us to detect a very sparse distribution of two verbal categories, namely *Raising* and *Control*. Instead of discarding these two types, I conflate these two groups with the *Simple* group (impersonal constructions), as they traditionally do not differ in their syntactic structures, as compared to restructuring and AcI verbs. Table 6.19 demonstrates the frequencies of OV and VO distribution for these verbs.

Table 6.19: OV/VO Order and Clause Type in Latin Infinitival Clauses

| | AcI | | | Other | | | Restructuring | | |
|---|---|---|---|---|---|---|---|---|---|
| Period | OV | VO | Sum | OV | VO | Sum | OV | VO | Sum |
| 1 | 76/(87) | 11/(13) | 86 | 50/(76) | 16/(24) | 66 | 85/(76) | 27/(24) | 112 |
| 2 | 34/(76) | 11/(24) | 45 | 41/(73) | 15/(27) | 56 | 49/(64) | 28/(36) | 77 |
| 3 | 41/(66) | 21/(34) | 62 | 46/(39) | 73/(61) | 119 | 65/(37) | 109/(63) | 174 |
| 4 | 21/(64) | 12/(36) | 33 | 32/(50) | 32/(50) | 64 | 49/(60) | 33/(40) | 82 |
| Sum | 172 | 55 | 226 | 169 | 136 | 305 | 248 | 197 | 445 |

There is a relatively slow decline of OV and a slow increase of VO in all types of infinitival clauses. Let us examine the VO distribution in infinitival clauses governed by these verb types, as shown in Figure 6.14.

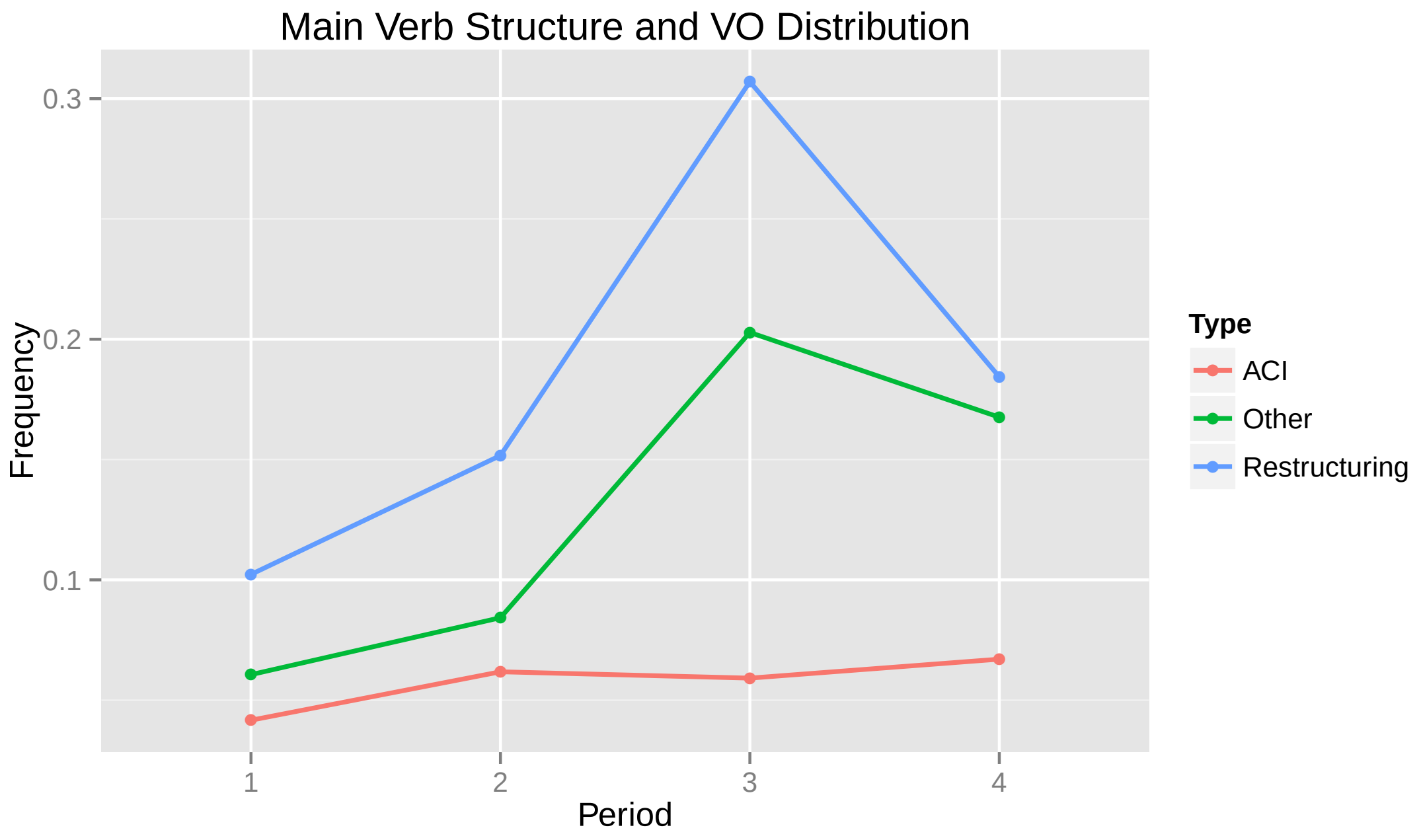

Figure 6.14: Frequency of VO Order in Infinitives with Respect to Main Verb in Latin Infinitival Clauses

Figure 6.14 shows that the distribution of VO is not homogenous across all categories. We can see that the *Restructuring* group maintains the highest rate of VO order, whereas *AcI* displays the lowest rate of VO. In addition, both the *Restructuring* and *Other* groups exhibit a constant increase in VO, with a drop in VO frequencies during the Late Latin period. Let us examine the *Restructuring* group in greater detail. Recall that this group consists of a special category of main verbs, namely modal, aspectual, perceptual and causative verbs. Some studies show that these verbs form a monoclausal structure with their infinitives, compared to a biclausal structure between a main verb and its infinitive (Rizzi, 1976; Cinque, 1998). It has been traditionally assumed that there exists a 'tight connection' between Vmain (main verb) and Vinf (infinitive) in this group. In this view, one would expect a tight cluster between a main verb and an infinitive, with an object following the infinitive (Rizzi, 1976). To determine whether such a tight relation exists in Latin restructuring verbs, the postposed restructuring infinitival clauses are extracted from the present data. Table 6.20 reports the results for the following two combinations of restructuring verbs: i) Main verb + Infinitive +

Nominal direct object and ii) Main verb + Nominal direct object + Infinitive. These results reveal that both combinations occur frequently in all four periods of Latin, with a very small decrease of Vmain+Object+Vinf over time. These findings suggest that VO order is not induced by this type of structure.

Table 6.20: Restructuring Verbs with Postposed Infinitives in Latin

| Period | Vmain+Vinf+Object (%) | Vmain+Object+Vinf (%) | Sum |
|---|---|---|---|
| Classical Period | 21/(43) | 28/(57) | 49 |
| Late Imperial | 8/(22) | 29/(78) | 37 |
| Early Late | 103/(69) | 47/(31) | 150 |
| Late | 20/(56) | 16/(44) | 36 |
| Sum | 120 | 152 | 272 |

Finally, let us examine the statistical relation between VO order and verb type across time. This probabilistic relationship can be determined through the outcome of the logistic regression for each group.[17] Figure 6.15 presents the results of the logistic transformation and Table 6.21 summarizes the estimated parameters for slopes and intercepts.

[17] R function for each verbal category: lrm(OVVO~Period,data)

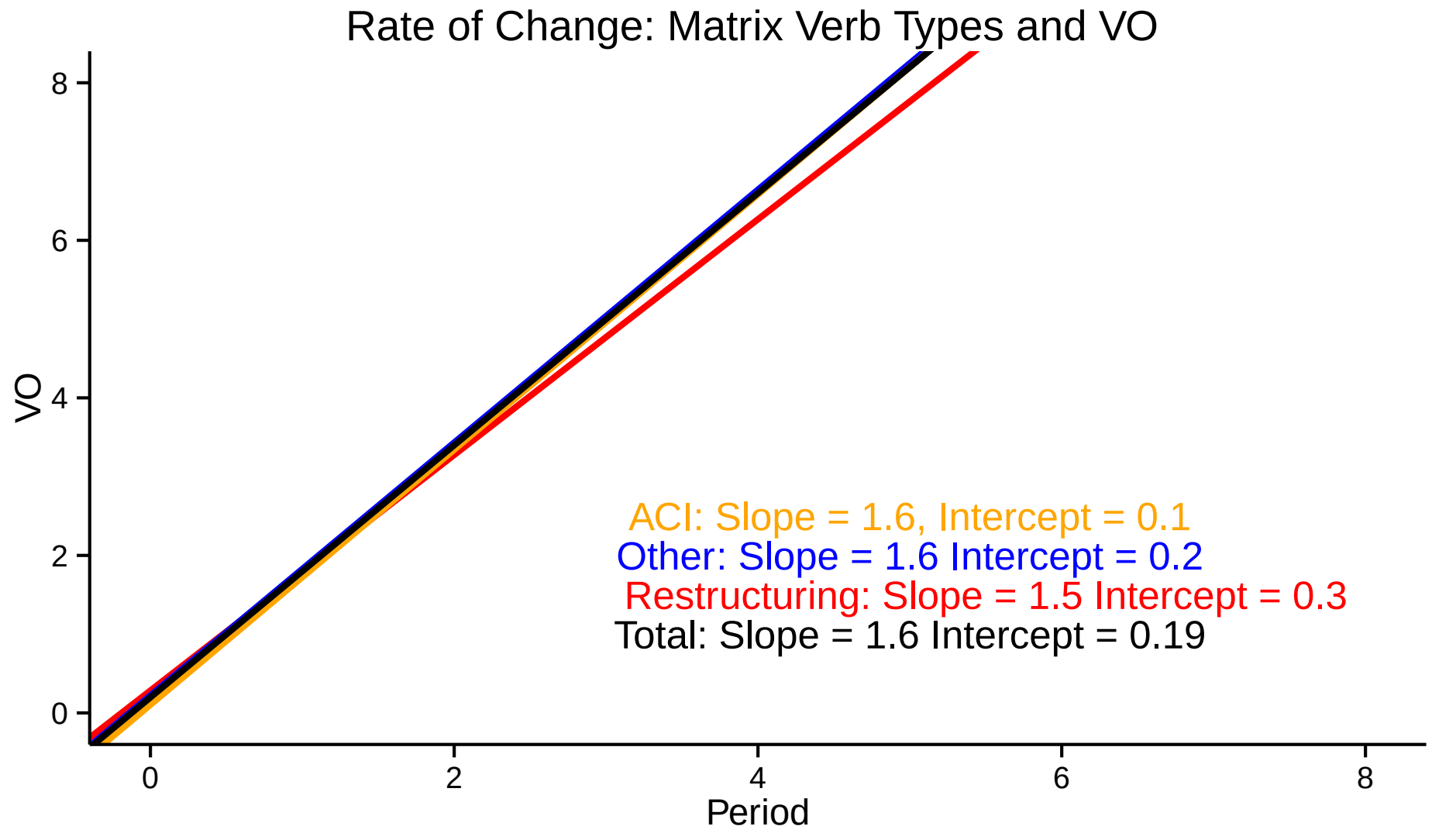

Figure 6.15: VO Rate in Infinitival Clauses across Different Types of Matrix Verbs in Latin Infinitival Clauses

Table 6.21: Slope and Intercept Parameters of the Frequency of VO order in Latin Infinitival Clauses: Verb Types

| Verb Types | Slope Parameter | Intercept Parameter |
|---|---|---|
| AcI | 1.62 | 0.104 |
| Restructuring | 1.5 | 0.284 |
| Other | 1.6 | 0.223 |

Figure 6.15 shows that the rate of change for all types of structures is almost the same - around 1.5 and 1.6. In fact, the *chi-square* test demonstrates that these differences are not significant ($p - value = 0.9579$). What can we say about these findings? This result shows that the frequency of VO order increases at a very similar rate in *AcI*, *Restructuring* and *Other* verbs. This fact strongly suggest that the process of word order change is similar in these three contexts.

The data also reveal a variation in the OV/VO distribution with respect to the position of main verbs[18] as shown in Figure 6.16 and table 6.22. While independent infinitives become less and less frequent, the VO form seems to continue its diffusion. For example, independent infinitives have almost disappeared in our data by the Late Latin period (6th century); VO order in Early Late Latin, however, demonstrates a higher frequency rate (45%) in comparison with Classical Latin (9%). In a similar fashion, the frequency rate of VO slightly increases in preposed infinitives from 6% in Classical Latin to 27% in Late Latin. Postposed infinitives, however, have by far the highest rate of VO form. In addition, postposed infinitives show a constant increase of VO.

[18]See section 5.2.2.2.

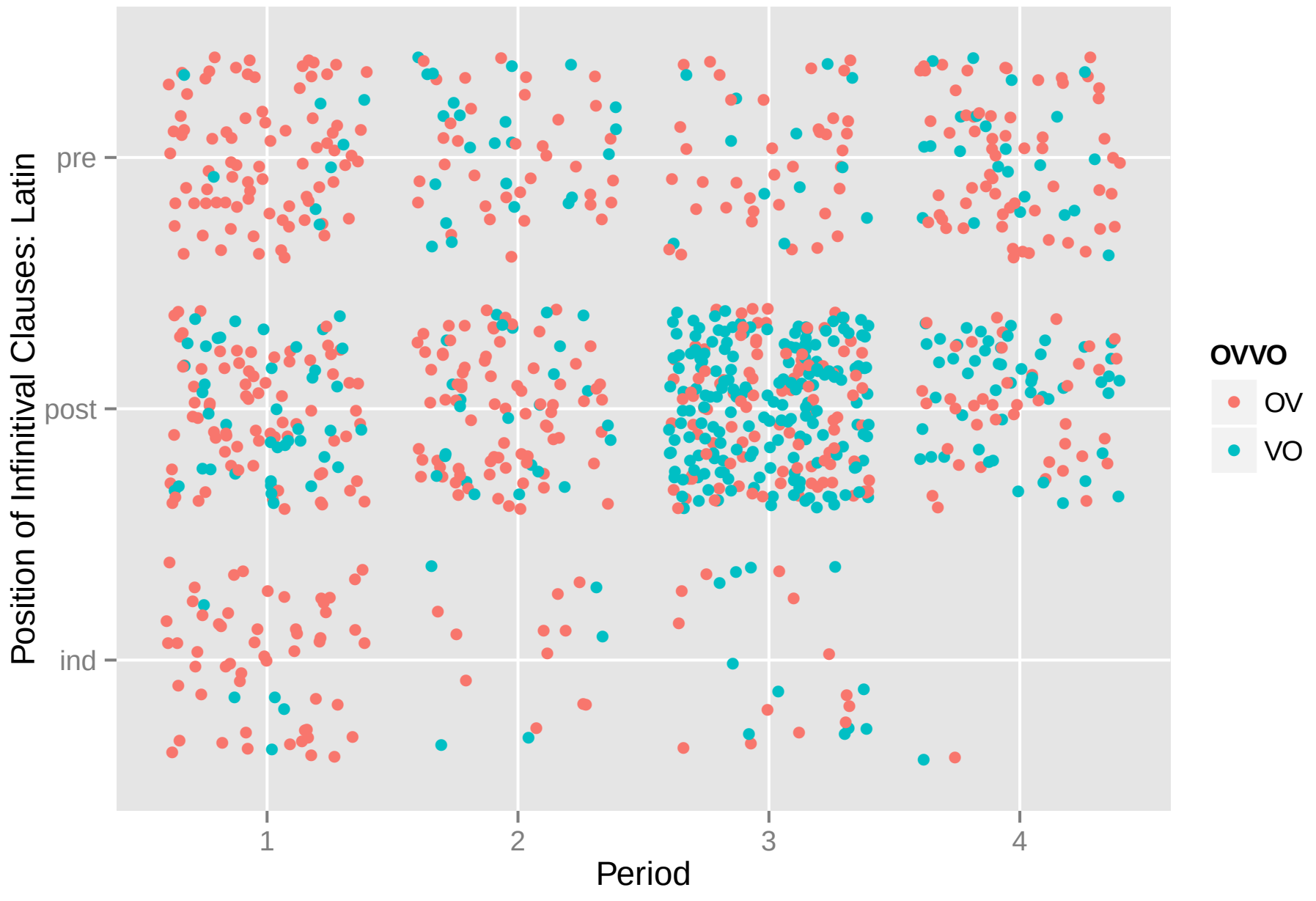

Figure 6.16: Position of Infinitival Clauses across Periods in Latin Infinitival Clauses

Table 6.22: OV/VO Order and Position of Infinitival Clauses in Latin

| | Preposed | | | Independent | | | Postposed | | |
|---|---|---|---|---|---|---|---|---|---|
| P. | OV/(%) | VO/(%) | Sum | OV/(%) | VO/(%) | Sum | OV/(%) | VO/(%) | Sum |
| 1 | 81/(91) | 8/(9) | 89 | 54/(92) | 5/(9) | 59 | 75/(65) | 41/(35) | 116 |
| 2 | 35/(60) | 23/(40) | 58 | 11/(69) | 5/(31) | 16 | 79/(76) | 25/(24) | 104 |
| 3 | 36/(74) | 13/(26) | 49 | 13/(54) | 11/(45) | 24 | 103/(37) | 179/(63) | 282 |
| 4 | 62/(73) | 23/(27) | 85 | 1 | 1 | 2 | 39/(42) | 53/(58) | 92 |
| Sum | 214 | 67 | 281 | 79 | 22 | 101 | 296 | 298 | 594 |

Let us plot these three positions separately in the logistic regression and estimate their predicted rate of slope and intercept. The rate of VO change for preposed, postposed and independent infinitives is illustrated in Figure 6.17 and the summary of their values is presented in Table 6.23.

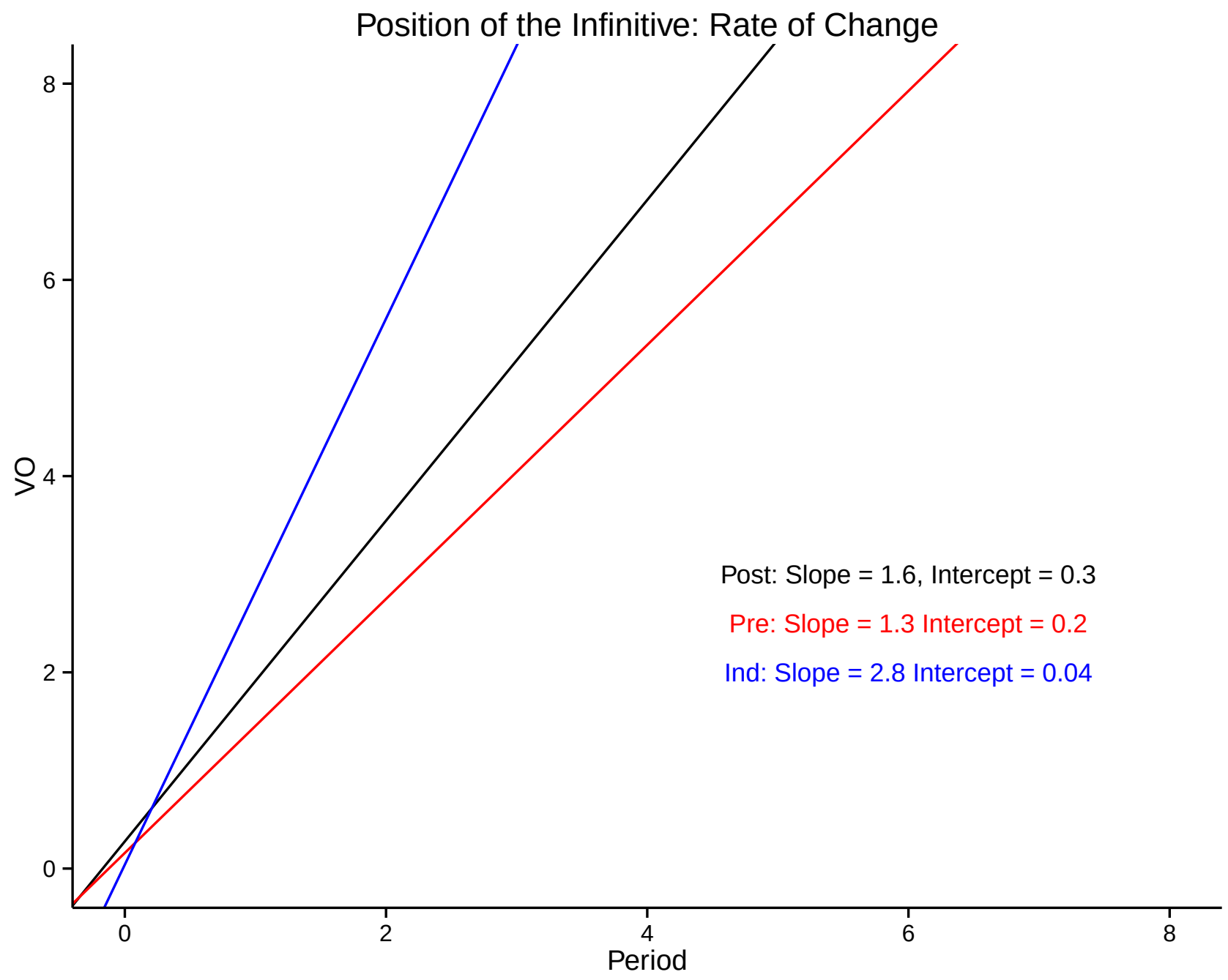

Figure 6.17: VO Rate across Different Types of Infinitival Position: Latin

Table 6.23: Slope and Intercept Parameters of the Frequency of VO order in Latin: Position of Infinitives

| Verb Types | Slope Parameter | Intercept Parameter |
|---|---|---|
| Preposed | 1.29 | 0.159 |
| Postposed | 1.63 | 0.27 |
| Independent | 2.78 | 0.0394 |

We can see from Figure 6.23 that there is a difference in the slope and intercept values, for example, the highest slope of 2.8 with independent infinitives and the lowest slope of 1.3 with preposed infinitives. However, these differences are not found to be significant ($p-value = 0.8623$). That is, the frequency of VO order is increasing at a similar rate in all these contexts. In fact, the small differences that are observed in the rate in each position are more likely to reflect some contextual influences on word order patterns. For instance, it has been traditionally claimed that independent infinitives are full-fledged sentences (see section 1.3.3.1), as they allow for a subject, tense and mood. As a full sentence, these structures are more likely to allow for a greater range of stylistic and pragmatic variations, which would affect word order patterns.

So far, the examination of the change in position of infinitival clauses and the change in frequency of verb types is independent of the change from OV to VO. That is, while certain contexts, such as restructuring verbs and postposed infinitives, may have influence on VO order, the changes that these structures undergo are independent of the word order change, for example, a decrease in preposed infinitives or a decrease in AcI structures. In order to evaluate this hypothesis statistically, we need to calculate a rate of change for each context, namely the position of infinitives, verb types and OV/VO change. The obtained values of slopes and intercepts for each construction are plotted in Figure 6.18 and the summary of slope and intercept parameters is presented in Table 6.24. In the plot, the red line represents word order change with a slope of 1.6, the blue line displays the change in verb types with a slope of 0.56 and the black line shows the rate of change in the position of infinitival clauses with a slope of 0.2. The significance test shows no significance across three changes ($p - value = 0.3409$). This result provides clear evidence that these three changes are not related, at least in the present data. While keeping in mind the small size of the corpus, these results suggest that there is no correlation between position of infinitives, the change from OV to VO and the decline of AcI.

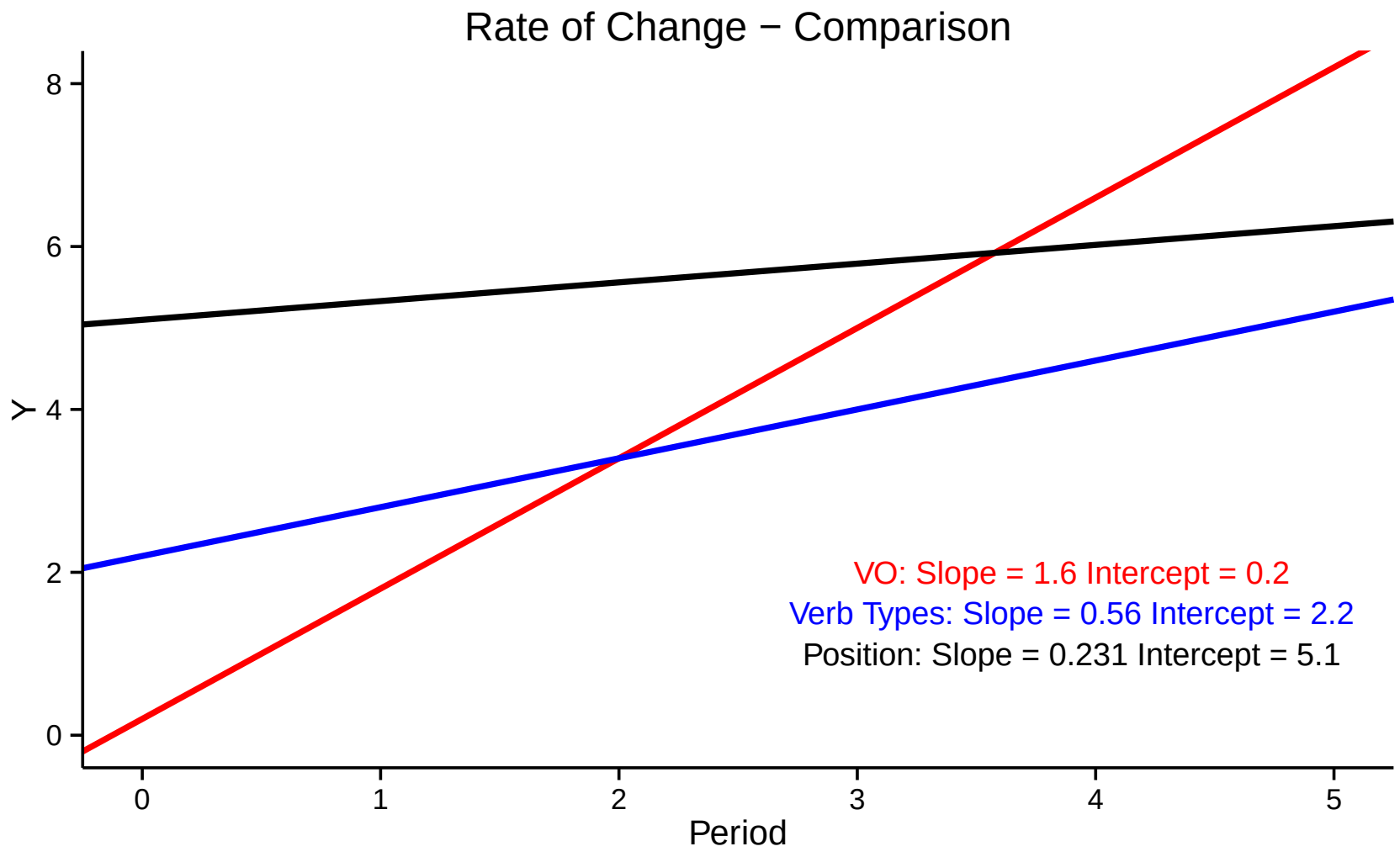

Figure 6.18: Rate of Change in Three Contexts: VO, Position and Verb Type in Latin Infinitival Clauses

Table 6.24: Slope and Intercept Parameters: OV/VO Change, Position of Infinitives and Verb Type in Latin Infinitival Clauses

| Verb Types | Slope Parameter | Intercept Parameter |
|---|---|---|
| OV/VO | 1.6 | 0.19 |
| Position | 0.23 | 5.13 |
| Verb Type | 0.568 | 2.21 |

## 6.2 Multi-Factorial Analysis

In this section word order change will be analyzed by means of a multivariate model. Earlier, in section 6.1.1 I used a univariate model, a model with only one independent factor, namely chronological period. The multivariate model allows for examination of the interplay between independent factors, namely syntactic, pragmatic, semantic and sociolinguistic, and word order alternation. However, unequally distributed data across factor groups and multiple interactions between factors are common issues in sociolinguistic analysis. Combining various statistical methods is beneficial in such cases, as it provides more confidence in the results. Let us first look at the overall importance of the variables. That is, the results will tell us which variables would be better predictors for OV/VO variation if we combine all the variables together. This analysis can be performed with *Random Forest* using the R package *partykit* (see section 5.3). Random Forest is a statistically robust technique that makes it possible to visualize the combined effect of factor groups and see which factors are statistically more important (see section 5.3). Here, the Latin model includes all linguistic factors, and they are plotted according to their importance, as shown in Figure 6.19.[19] All independent factors placed to the right of the dashed vertical line are considered significant.

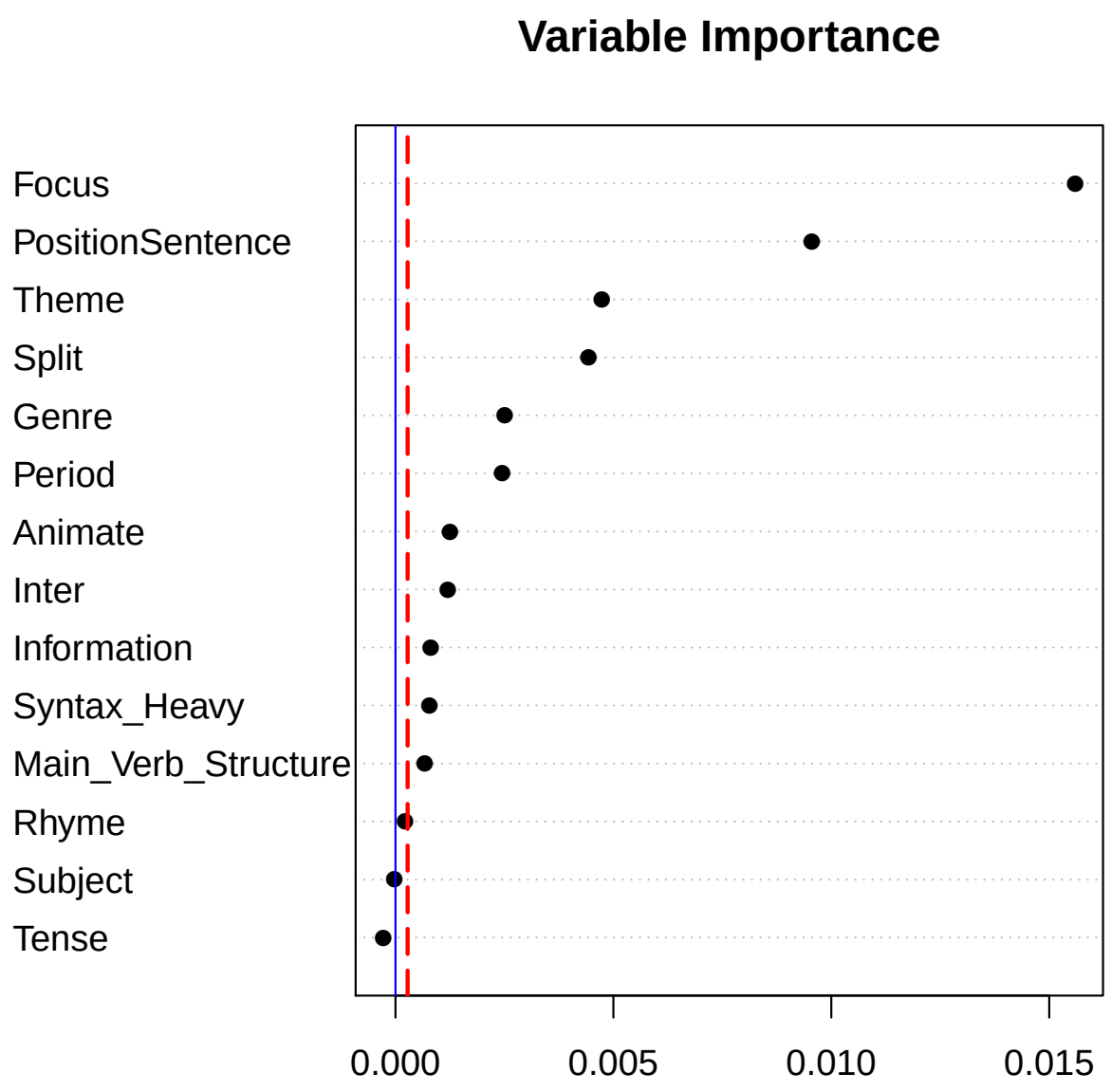

Figure 6.19: Random Forest Analysis - Latin Infinitival Clauses

[19] I attempted to include frequency and both types of heaviness in the model, but due to the number of factor level, small dataset and memory allocation for tree training and pruning, it proved to be computationally demanding.

The model shows that *Focus* is by far the most important factor group in Latin.[20] From this model, it is possible to identify the following significant factors, presented in order of their importance: *Position*, *Theme*, *Split*, *Genre*, *Period*, *Animacy*, *Intervening constituents*, *Information Status*, *Heaviness* and *Verb Type*. In the previous section, we observed that there is a difference in VO frequency among postposed, preposed and independent verbs (Table 6.16). In addition, the VO frequency is different based on verb types (Table 6.19). Here, we have statistical evidence that these two syntactic factors indeed influence OV/VO alternation. Furthermore, information structure analysis in section 6.1.3 and section 6.1.2 has demonstrated that there is a vast range of pragmatic values on preverbal and postverbal nouns. This observation is confirmed here by a robust statistical analysis: information status (new, old and accessible information) and information relevance (contrastive and information focus) are statistically significant for word order alternation. *Rhyme*, *Subject* and *Tense* do not show any significance for word order alternation in this model. It is traditionally argued that the presence of subject is more likely to trigger VO order (Bauer, 1995). Non-significance of *Subject*, namely presence or absence of Subject, contradicts this statement. Recall also the findings from section 6.1.1 that do not reveal any noticeable frequency effect on word order alternation with different types of subjects (see Table 6.4). From Figure 6.19 we can also notice many factors that have not been analyzed yet, e.g. animacy, genre, theme, rhyme. That is, these results demonstrate that word order change is, indeed, a very complex phenomenon that is interwoven with multiple factors, such as pragmatic, syntactic and semantic factors. This analysis, however, does not provide a detailed view on how each factor influences OV/VO alternation. For example, we know that *Focus* is relevant; however, there is no indication which type of focus influences VO order, e.g. contrastive focus or information focus. In order to examine each factor, conditional tree analysis is applied (see section 5.3). Such a conditional tree looks at the factor ranking inside each factor group. Each factor group is represented by an oval (node). The higher the node is on the tree, the stronger is its association with OV/VO alternation. The branches from each node represent the split between its values. Finally, on the bottom, we have the proportions of OV and VO orders. If we plot all the factors on one tree, such a tree will make the plot very complex for interpretation. Instead, I analyze factors by their groups, namely sociolinguistic, pragmatic and syntactic factors.

[20]Initially, the data are coded for contrastive focus and information focus as well as focus on the verbs. Given that the number of such verbs is very small, these tokens are recoded as information focus.

I begin with sociolinguistic factors. These factors include *Genre*, *Theme* and *Rhyme*. Figure 6.20 illustrates the output of the model with the aforementioned factors. *Theme* is the most important predictor for OV/VO order. This predictor is differentiated into two groups: i) *history* and *literature* and ii) *religion*. This fact tells us that there is a difference in the use of OV and VO orders in religious texts as compared to historical and literary texts. *Religion* further intersects with *Genre*, and the highest VO rate is shown in *narrative* genre as compared with *letters*. The second branch (*history* and *literary*) is further differentiated by period. The division identifies Early Late Latin as a boundary line. Both Classical and Imperial Latin exhibit a slightly higher VO rate in *letters* and *speech*, as compared to *treatise*. Interestingly, Late Latin demonstrates a higher VO rate in metric prose (50%) as compared to non-metric prose (20%).

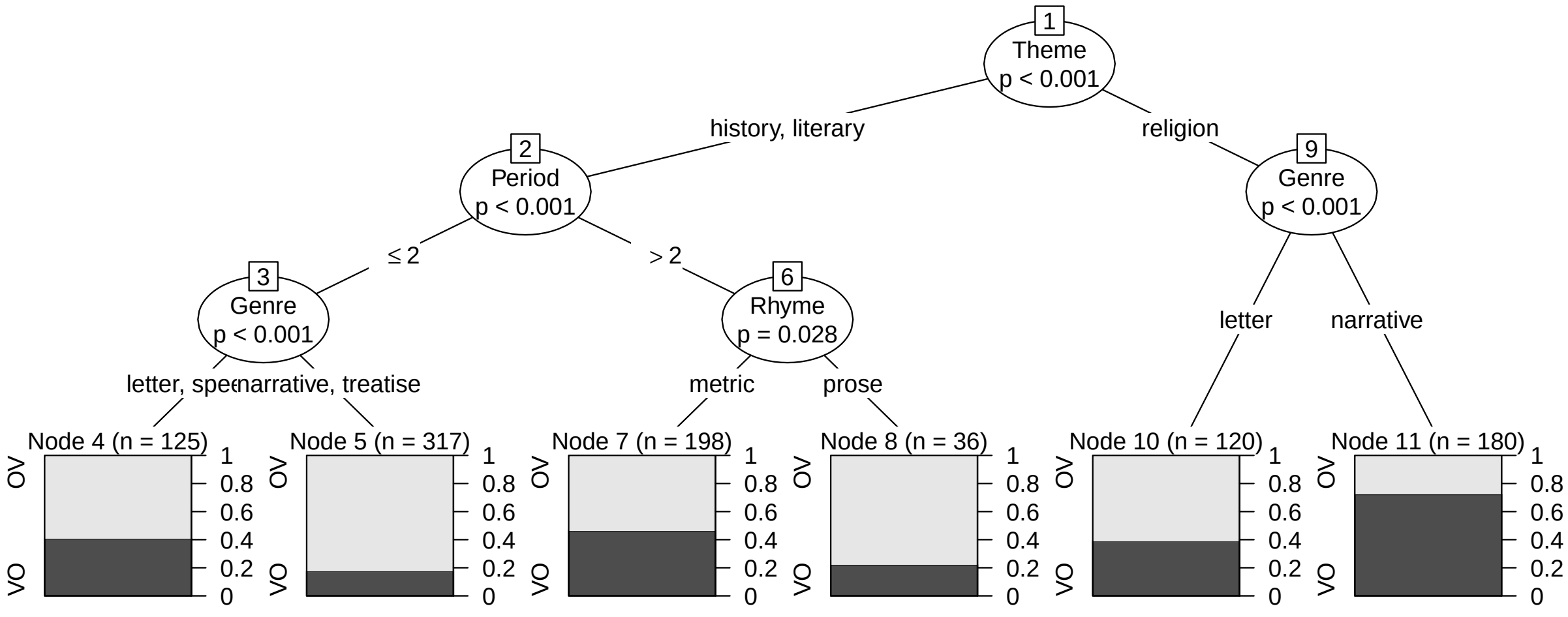

Figure 6.20: Sociolinguistic Factors - Latin Infinitival Clauses

Let us examine in more detail the *Theme* factor. The detailed examination of the distribution of theme by period is presented in Table 6.25.

Table 6.25: Sociolinguistic Factor - Theme: Latin Infinitival Clauses

| Period | history | literary | religion |
|---|---|---|---|
| Classic Latin | 216 | 48 | 0 |
| Late Imperial Latin | 0 | 178 | 0 |
| Early Late Latin | 80 | 0 | 275 |
| Late Latin | 34 | 120 | 25 |

*Religious* texts are available only in Early Late and Late Latin, which could explain the high rate of VO order. However, in the tree (see Figure 6.20) religious letters show only 40% of VO, whereas religious narratives display 70% of VO. Table 6.26 illustrates the distribution of OV/VO in letters and narratives. In the 4th century narratives and letters are specimens of the same author - Jerome. The narrative genre is presented by Jerome's translation of Vulgar Bible,[21] and the letter genre is Jerome's personal communication on various religious subjects. Thus, the data provide evidence of word order alternation influenced by stylistic factors from the same author.

Table 6.26: Sociolinguistic Factor - Genre (Letters and Narrative): Latin Infinitival Clauses

| | Letters | | Narrative | |
|---|---|---|---|---|
| Period | OV/% | VO/% | OV/% | VO/% |
| Early Late Latin (Jerome) | 32/55 | 26/45 | 50/28 | 130/72 |
| Late Latin | 17/68 | 8/32 | 26/76 | 8/24 |

Syntactic factors are shown in Figure 6.21. *Position of Infinitives* is identified as the most important factor. The tree branches into i) *independent* and *preposed* infinitives and ii) *postposed* infinitives. The latter is differentiated by period, and again Early Late Latin is marked as a border line. From Early Late Latin word alternation is further conditioned by *main verb structure*, showing a very high rate for VO order with *restructuring* and *other* verbs in comparison with *AcI*. In contrast, *independent* and *preposed* infinitives show only a slight increase in VO order after Classical Latin.[22] Although the other syntactic factors (subject, tense, split, intervening) were included in the model, their ranking is not found to be significant in this model.

[21] The high VO rate is most likely to be a result of translation influence.

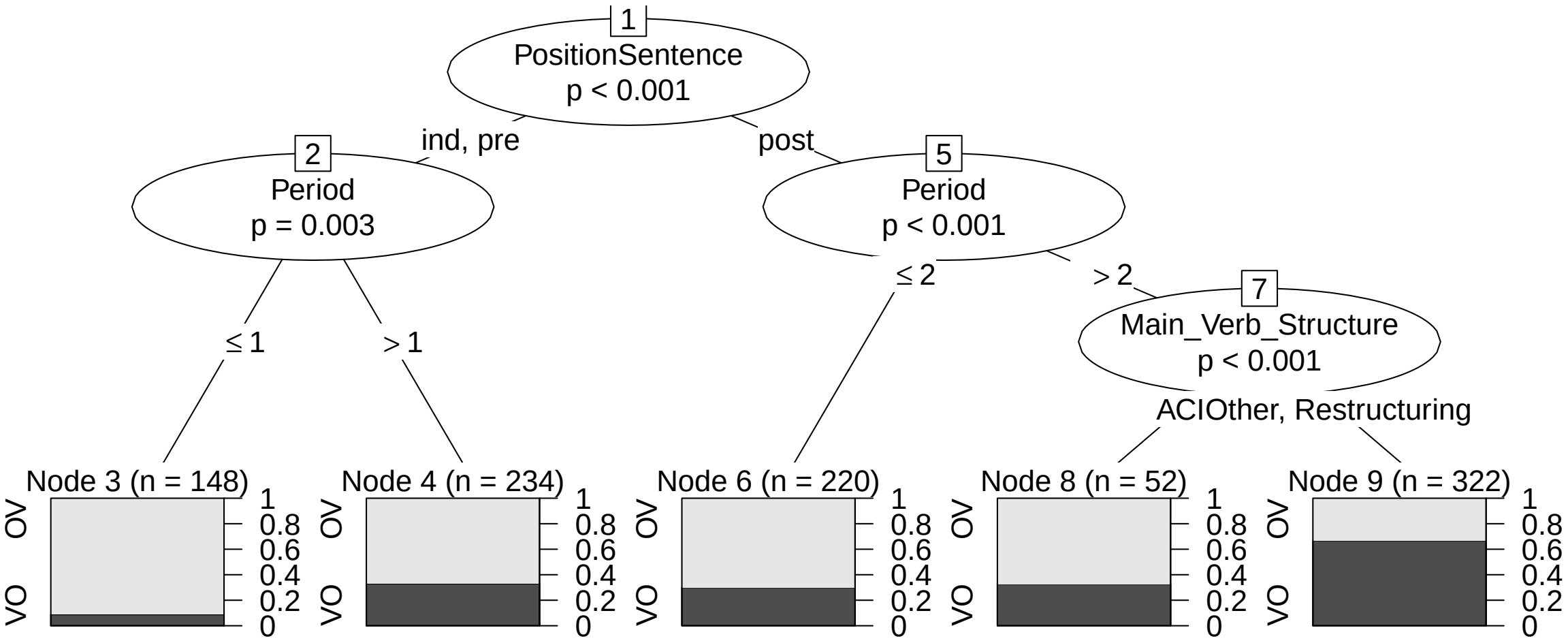

Figure 6.21: Syntactic Factors - Latin Infinitival Clauses

Pragmatic factors are shown in Figure 6.22. Pragmatic factors are important in identifying a basic word order in discourse-configurational languages. Recall from section 1.3.2 that basic word order is defined as an unmarked prosodic pattern of the language and that the unmarked pattern is determined by information structure categories, such as new information and information focus (Hinterhölzl, 2010). It is also argued that the influence of pragmatic factors on syntax can be overshadowed by heavy constituents (Taylor and Pintzuk, 2012b). Therefore, the inclusion of the factor *Heaviness* is necessary to control for its influence. From Figure 6.22 it is clear that the best predictor is *Period*. That is, the influence of information structure is not homogenous across periods. Let us examine differences between word order and information structure. In Classical and Imperial Latin *information status* is a greater predictor for word order. We see that *old* information has a slightly higher rate for VO order than *new* and *accessible*. This is an interesting fact, as

[22]Classical Latin -1, Late Imperial - 2.

*new information* often serves to identify a non-marked word order (see section 1.3.1). In contrast, a better predictor for Late Latin is *information relevance*: i) contrastive focus and ii) information focus. Information focus further shows another chronological split, namely Early Late and Late Latin. In Late Latin, *new* information has a much higher VO rate than *old* and *accessible* information. Once more, this fact can be used to identify a basic word order. In contrast to Classical and Late Imperial Latin (Node 2 in the tree), where *old* information has a higher VO rate, this time *new* information plays a role for postverbal objects. Given the assumption that new information is a necessary condition in the non-marked word order, the findings suggest that in Classical and Imperial Latin OV is a non-marked word order, whereas in Late Latin, the basic word order is VO. On the other hand, in Early Late Latin we can observe the effect of heaviness on VO order. That is, syntactically heavy constituents show a very high rate for VO order.

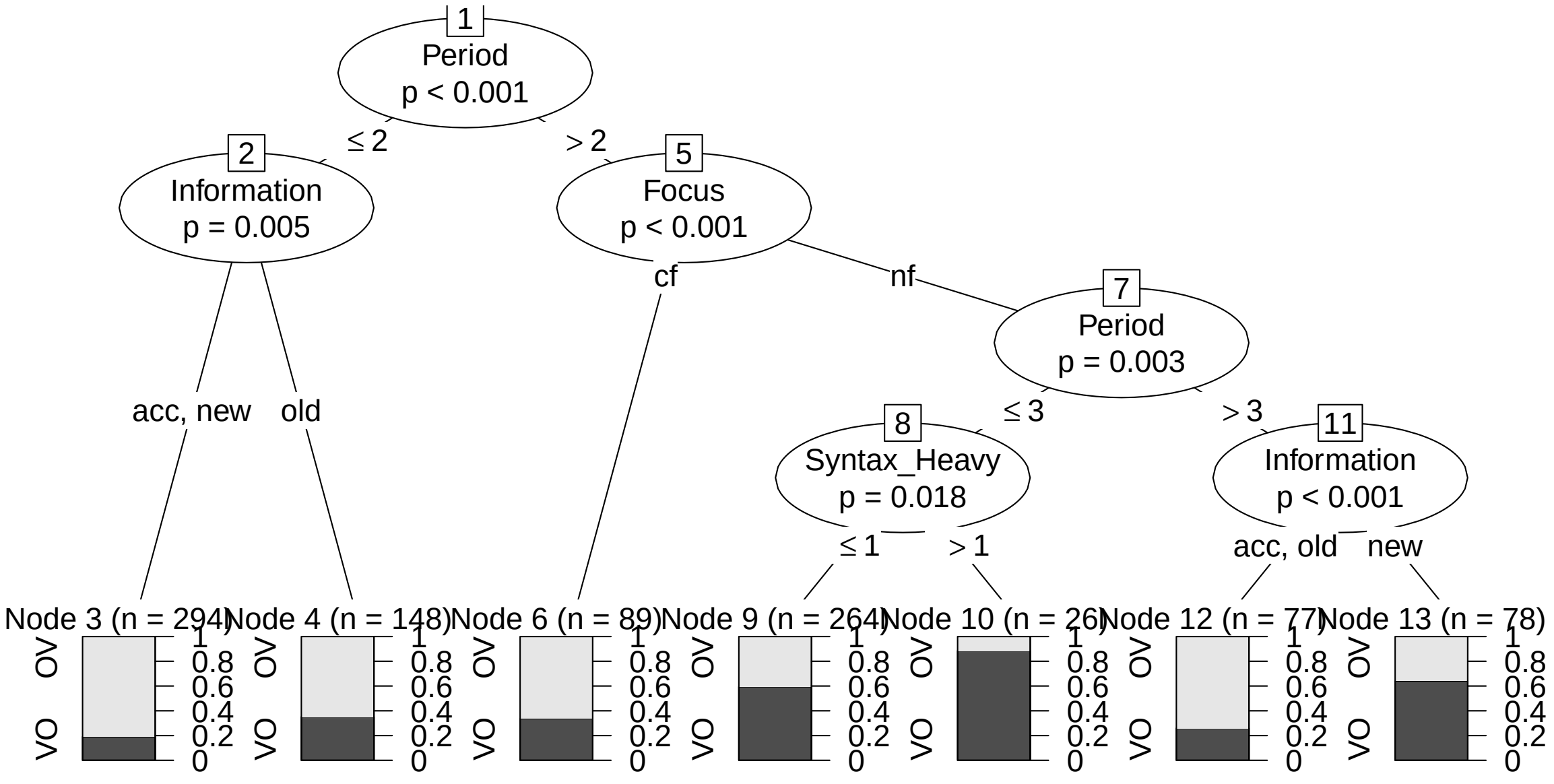

Figure 6.22: Pragmatic Factors - Latin Infinitival Clauses

As mentioned earlier, lengthy constituents are a common cross-linguistic factor that triggers VO order; thus, we can obtain a better picture by eliminating heavy constituents from the analysis, illustrated in Figure 6.23, where only light constituents are considered (words≤3).

The rate of light postverbal NPs still remains high, supporting the earlier hypothesis that VO is a basic word order in Late Latin. VO order occurs about 60% of the time in Early Late and Late Latin with nouns bearing new information, whereas in Classical and Late Imperial Latin, only 20% of postverbal nouns carry new information.

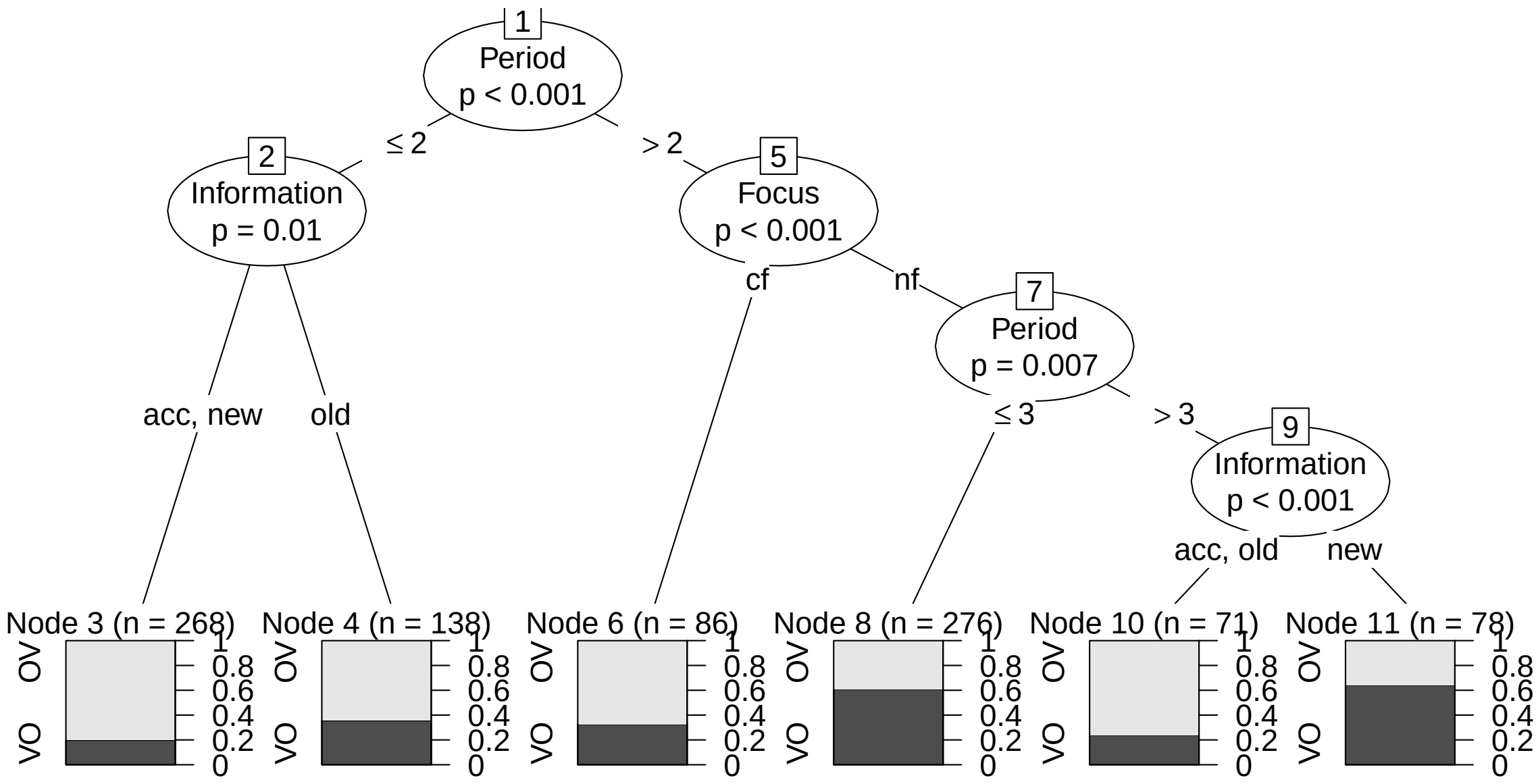

Figure 6.23: Pragmatic Factors with Light Constituents - Latin Infinitival Clauses

So far, I have presented the relation of variables to the predicted value, namely OV and VO, by their relative importance and by their ranking inside each factor. While ranking is a useful tool for understanding the data, this technique is used only for data exploration, as it is subject to data variability and size: Each node is associated with the smallest p value, and the larger the number of observations, the more likely the p value is to be significant. Thus, the next step is to apply a more robust statistical technique. While there are various methods available for multi-factorial analysis in the sociolinguistic field, e.g., *GoldVarb* and *Rbrul*, the use of logistic regression and Bayesian interpretation of probabilities (see section 5.3) not only improve analytical methods, but also make variationist research accessible to other quantitative fields and sub-disciplines of linguistics, e.g. corpus linguistics, cognitive linguistics and computational linguistics. Furthermore, recent advances in statistics and statistical packages offer improved methods for evaluating statistical significance in sociolinguistic studies, namely the combination of fixed and mixed logistic regression models (see section 5.3). In the fixed-effect model we treat each observation as independent of the others. This assumption is not always suitable, especially when we deal with individuals and words. In diachronic studies, tokens are usually represented by various authors or texts, and each author contributes more than one token, that is, tokens are grouped by individual. Similarly, one token does not always represent a unique word, rather they are grouped into word type clusters. A frequent word type may contribute many tokens, whereas an infrequent word type may have only one single token. In order to control for this by-word and by-speaker imbalance, the mixed model is used. In recent years, many sociolinguistic, phonological and morpho-syntactic studies have advocated for the use of this mixed model (see Johnson (forthcoming)); this model, however, has not been evaluated in diachronic syntactic studies. Therefore, this study implements both models, examining the effect of the pragmatic and syntactic factors identified in earlier models. Recall that the predicted variable is binary, namely OV and VO. For these models, the outcome of interest is the probability of VO. Figure 6.19 provides us with the most important factor groups for the predicted variable OV/VO: 1) *Focus* (information relevance), 2) *Position* (position of infinitival clauses), 3) *Theme*, 4) *Split* (split of NPs), 5) Genre 6) *Period.* While *Verb Type* and *Information status* seem to be less important than the first six factors, they will also be included in the fixed model to determine their influence on OV/VO. In contrast, *Theme* and *Genre* are not included in this model, since the examination of sociolinguistic factors in Figure 6.20 has shown that religious letters and narratives occur only in Late Latin. Finally, the frequency effect (log) of infinitival verbs is also included in the present models. The complete model is illustrated in (90) (see section 5.3.5 for more detail on this model).

(90)
```
OVVO[i] ~ dbern(mu[i])
mu[i]<-1/(1+exp(-(b0 + b1*Period[i] + b2*InformationStatus[i]
+ b3*InformationRelevance[i] + b4*VerbType[i]
+b5*PositionSentence[i] + b6*Split[i] + b*7Frequency[i])))
```

The following reference values have been chosen for the independent factors in (90): *Information Status* (new), *Information Relevance* (non-contrast), *Verb Type* (AcI), *Position* (postposed) and Split (no split). Thus, the application value VO is tested against those values. For example, in case of positive significant results (HDI is greater than zero) for *information relevance* parameter, contrastive focus will favor VO order; if the results are below zero (HDI is less than zero), contrastive focus will favor OV order. If the HDI values include zero, the influence is not significant. In addition, the model includes log(frequency) to examine any potential influence. The obtained posterior distribution for each parameter (factor) will show us whether there is robust statistical significance to predict a VO diffusion in Latin. Figure 6.24 shows the posterior distribution for each parameter from the model in (90).

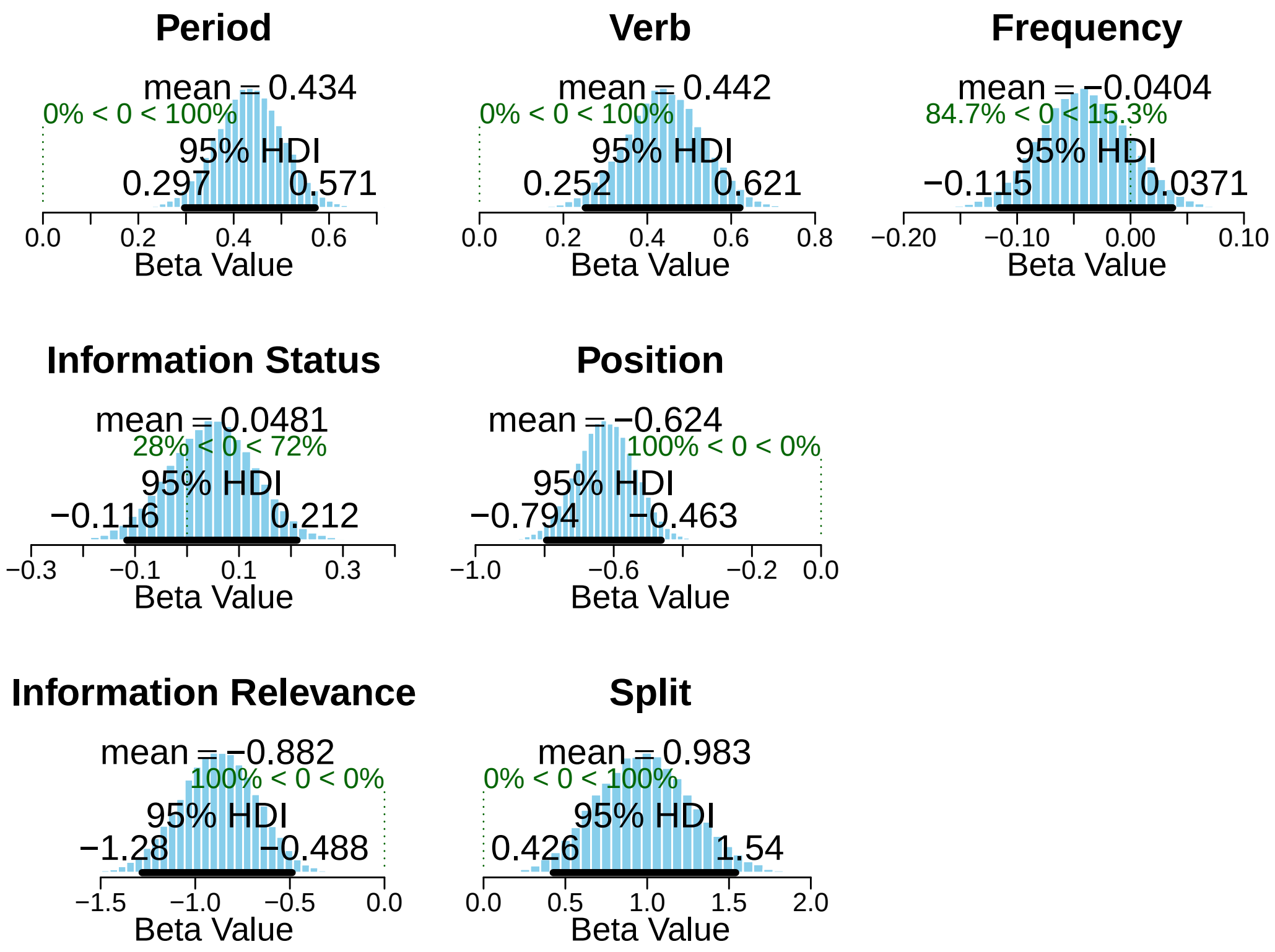

Figure 6.24: Posterior Distribution: Fixed Effect Model for Latin Infinitival Clauses

Let us look first at the factors that are not credible, that is, for which the zero value is a part of the 95% HDI interval (see section 5.3.5). The non-credible factors are 1) *Information status* and 2) *Frequency*. In contrast, the credible factors are 1) *Period*, 2) *Information relevance*, 3) *Verb Type*, 4) *Position* and 5) *Split*. The first factor, *Period*, provides statistical evidence that the probability of VO order increases over time.[23] The second factor, *Information relevance*, has negative values, that is, contrastive focus is more likely to predict OV order. The third factor, *Verb type*, demonstrates that VO order is more likely with non AcI verbs, namely *restructuring* and *other* verbs. The fourth factor, *Position*, shows that OV (95% HDI is less than zero) is more likely when infinitives are preposed or independent. Finally, the last factor, *Split*, provides evidence that split NPs are more likely to have VO order. For example, if the noun and its adjective are split, the order is more likely to be XVO.

Let us turn now to a mixed model. As Johnson (forthcoming) points out, non-inclusion of individual speakers may cause a Type I error, where a chance effect may be interpreted as a significant difference in the population. On the other hand, a mixed effect model may produce a Type II error, where a real difference in the population is not recognized as significant. Thus, the comparison of two models will reinforce the significance of the variables if they are identified as significant in both models. Figure 6.25 reports the results from the mixed model with the same factors as in previous fixed model. This time, the model also includes authors and words as random effects, as illustrated in (91).[24]

(91)
```
for (i in 1:N){
OVVO[i] ~ dbern(mu[i])
mu[i]<-1/(1+exp(-(b0 +  b1*Period[i] + b2*Focus[i] +
b3*Information[i]  + b4*Position[i]+
b5*Verb[i]+b6*Split[i]
```

[23] Period is a continuous variable, that is, the greater the period is, the more likely it is to have a VO order (95% HDI is greater than zero).

[24] Since *frequency* is not significant in a fixed model, it is excluded from the mixed model, given the complexity involved in mixed model processing.

```
 + u[Author[i]]+y[Word[i]]))))}
for (j in 1:M) {
u[j] ~ dnorm(0,tau)}
for (j in 1:W) {
y[j] ~ dnorm(0,tau)}
```

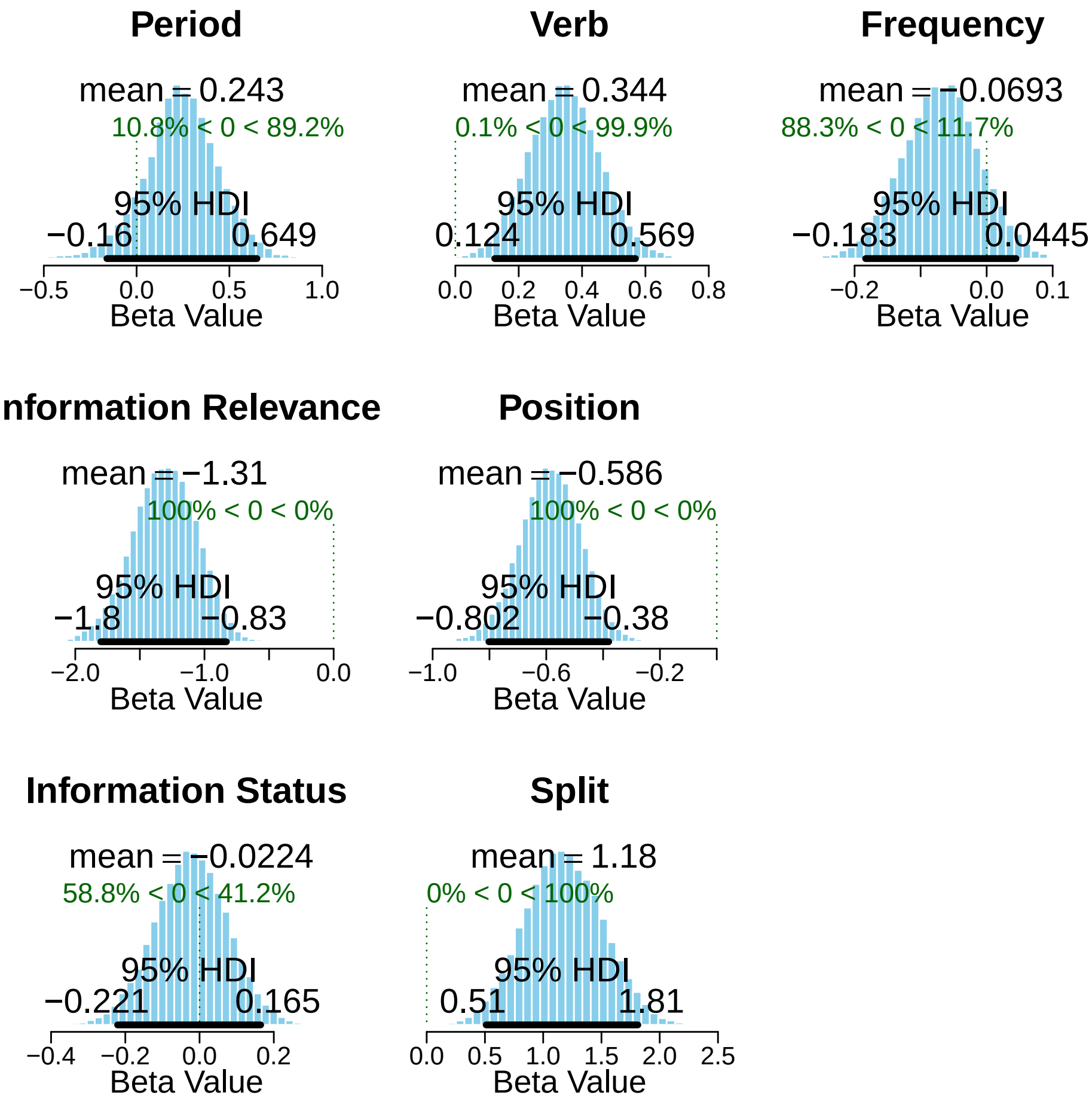

Figure 6.25: Posterior Distribution: Mixed Effect Model (Authors and Words) for Latin Infinitival Clauses

Presented in Figure 6.25, the results from this mixed model show that *Information relevance*, *Verb type*, *Position* and *Split* are significant parameters. VO order is more likely with non-contrastive focus, non AcI verb type, postposed infinitive and a split noun. The comparison between fixed and mixed models allows for a stronger evaluation of independent factors, because fixed models are prone to Type I errors, where a chance effect can be mistaken for a real significance. In contrast, mixed models are subject to Type II errors, where a small population difference may not be recognized as significant (see Johnson (forthcoming) for more details on Error Types). Thus, if the factor is found significant in both models, this provides a stronger evidence for its statistical importance. Thus, there is a strong confirmation for the following claims in the data:

1. Postverbal NPs are more likely to bear non-contrastive information focus
2. Postposed infinitives are a favorable condition for VO order diffusion
3. Non AcI verbs (*restructuring* and *other*) are a favorable context for VO order
4. Type frequency of infinitival verb is not significant for OV/VO variation in Latin

The hypothesis with respect to the role of old/new information is not shown to be statistically significant. However if we look back at Figure 6.23, it is clear from the inference tree that the tendency for new information to be on postverbal NPs develops in Late Latin (6th century). In fact, in Classical Latin VO displays a slight preference for old information. In contrast, focus seems to maintain the same distribution in our data across time, namely contrastive focus - preverbal and non-contrastive focus - postverbal NPs. Interestingly, the *Split* factor remains a very strong predictor for VO order. Let us examine this factor in greater detail. Figure 6.26 illustrates that in Classical and Early Late Latin, XVO and OVX are equally possible with split nouns, as illustrated in (92). However, in Late Latin 80% of split nouns are XVO. Finally, the hypothesis with respect to frequency is not shown to be statistically robust.

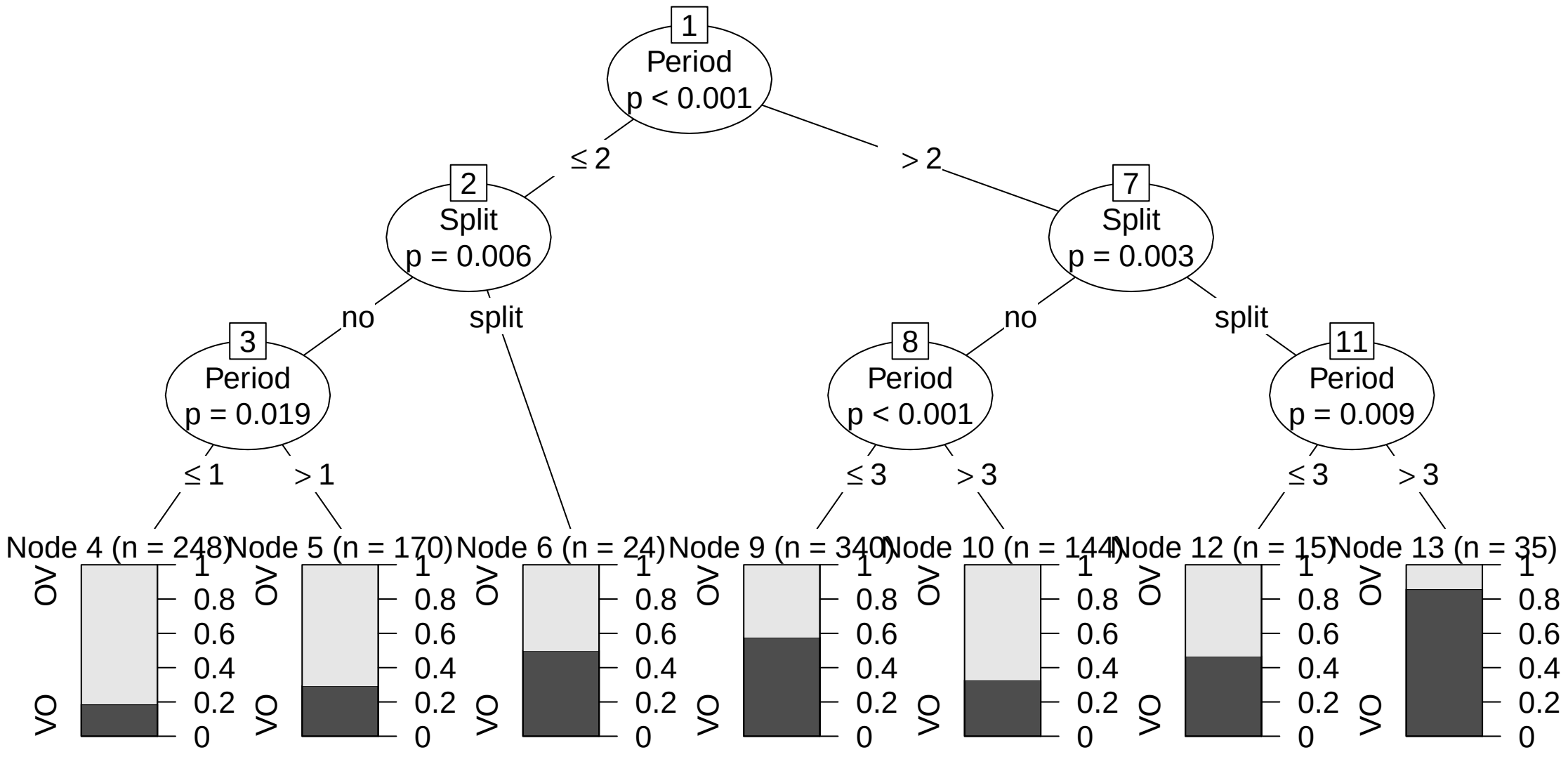

Figure 6.26: Split NPs in Latin Infinitival Clauses

(92) a. XVO

colloquendi dare facultatem
discuss-gen. give-inf opportunity-acc.obj

'give an opportunity to talk (Caesar, *De Bello Gallico* 2:26)

b. OVX

latebras captare secretas
hiding-places capture-inf secret-acc

'he seeks out secret lurking places' (Ammianus, *Book16* 7.7)

## 6.3 Summary

In this chapter, I have examined word order in Latin. First, I have presented an exploratory analysis, showing the constant increase of VO over time. The results show a traditional direction of word order change from SOV to SVO via a transitional stage XVO/OVX observed in Late Latin. Furthermore, the examination of verb-initial order allowed us to evaluate two models of word order change in Latin: i) Salvi (2005) and ii) Devine and Stephens (2006). According to Salvi's model, the passage from Latin to Romance languages comprises two steps: i) Restricted class of verb in the initial position and an expansion of other verbs into the initial position at later periods and ii) co-occurrence of focus and preposed verbs. First, the present data demonstrate a small group of infrequent verbs in Classical Latin, followed by a slightly larger group that includes ditransitive and perception verbs in Late Imperial Latin. Early Latin provides evidence for a large group of very frequent verbs in the initial position, suggesting that this period corresponds to Phase I in Salvi's model. Second, there is some evidence of preverbal contrastive constituents with these verbs, suggesting that Early Late Latin is, indeed, Phase I of the change. In contrast, the present data do not confirm the hypothesis for VO-leakage (Devine and Stephens, 2006), an innovation which allows for an object to remain in the postverbal position instead of moving to a preverbal position. In this approach, one of the criteria for VO leakage is non-referentiality, abstractness and pragmatical neutrality of postverbal nouns. The data demonstrates a range of pragmatical values on postverbal nouns, including contrastive focus.

Subsequently, I have shown that Latin infinitival clauses are not homogenous with respect to VO diffusion. Restructuring verbs and postposed infinitives are the most favorable syntactic contexts for the VO order. While both *restructuring* and *other* verb types are significant for VO order, the rate of VO with restructuring verbs is higher. In addition, the evaluation of infinitival position and verb types reveals that there are some changes that are not related, at least directly, to OV/VO change, namely the decrease of AcI and decline in preposed infinitives. Next, I have performed a multi-variate analysis and identified the strongest predictors for VO order. Among these predictors are *Information relevance* (contrastive focus and information focus) and *Position* (preposed, postposed and independent infinitives). The comparison between fixed and mixed models has further confirmed these findings by identifying postposed infinitives, restructuring verbs and non-focused NPs as the best predictors for VO order in Latin. Finally, I have demonstrated that from Early Late Latin tokens with information focus and new information are more likely to exhibit VO order. In contrast, in Classical Latin, new information predicts OV order. That is, there is a change in information structure between Late Imperial and Early Latin. This change also suggests that the period of Early Late Latin is a turning point for word order change.

# Chapter 7

# Quantitative Analysis: Old French Word Order

*[O]ne generation is more likely to differ from its predecessor in the frequency with which its speakers use certain forms than in whether those forms are possible at all*

*(Kroch, 1989b, 348)*

This chapter provides an analysis of word order in infinitival clauses in Old French. In the first section, I will present descriptive characteristics of word order patterns and word order alternation. Particularly, I will examine the effect of information structure on preverbal and postverbal nouns. In the second section, I will present a multi-factorial analysis and evaluate the effect of sociolinguistic and linguistic factors on word order alternation and change.

## 7.1 Descriptive Analysis

### 7.1.1 Word Order Patterns

Table 7.1 and Figure 7.1 show the distribution of OV/VO order in Old French. A few trends in the data are immediately obvious. Compared with the Latin data, where VO order initially was 20%, the data in Old French exhibits a higher initial rate (46%). Interestingly, the same rate (around 50%) continues until the 12th century. Only from the 13th century can we observe a constant increase, running from 66% in the 13th century to 73% in the 14th century.

Table 7.1: OV/VO Distribution in Infinitival Clauses: Old French

| Period (century) | OV | % | VO | % | Total |
|---|---|---|---|---|---|
| 10th | 13 | 54 | 11 | 46 | 24 |
| 11th | 66 | 46 | 79 | 54 | 145 |
| 12th | 246 | 56 | 193 | 44 | 439 |
| 13th | 57 | 34 | 83 | 66 | 166 |
| 14th | 200 | 27 | 538 | 73 | 738 |
| Total | 582 | 38% | 930 | 62% | 1512 |

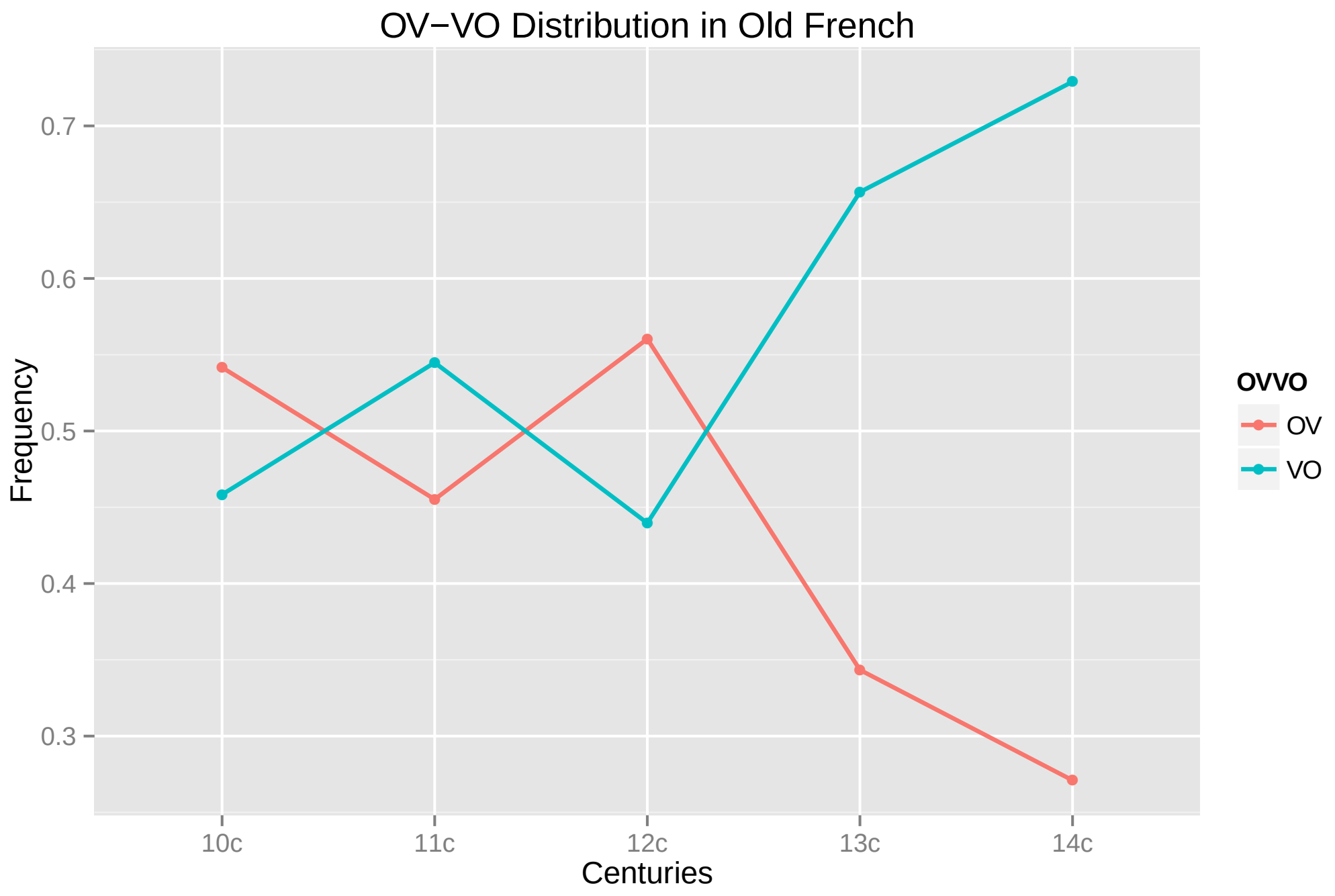

Figure 7.1: Frequency of Word Order in Infinitival Clauses: Old French

Figure 7.1 suggests two stages of distribution in our data: i) a plateau effect between the 10th and the 12th centuries and ii) a steep increase in VO after the 12th century. Figure 7.2 shows the rate of change for VO order, using the logistic transform (see section 5.3). The x-axis represents periods as a continuous scale, and the red line is the logistic transform with a slope equal to 1.67 and an intercept equal to 0.002. There is a statistically significant rise of VO order in Old French, as the slope is not equal to zero.

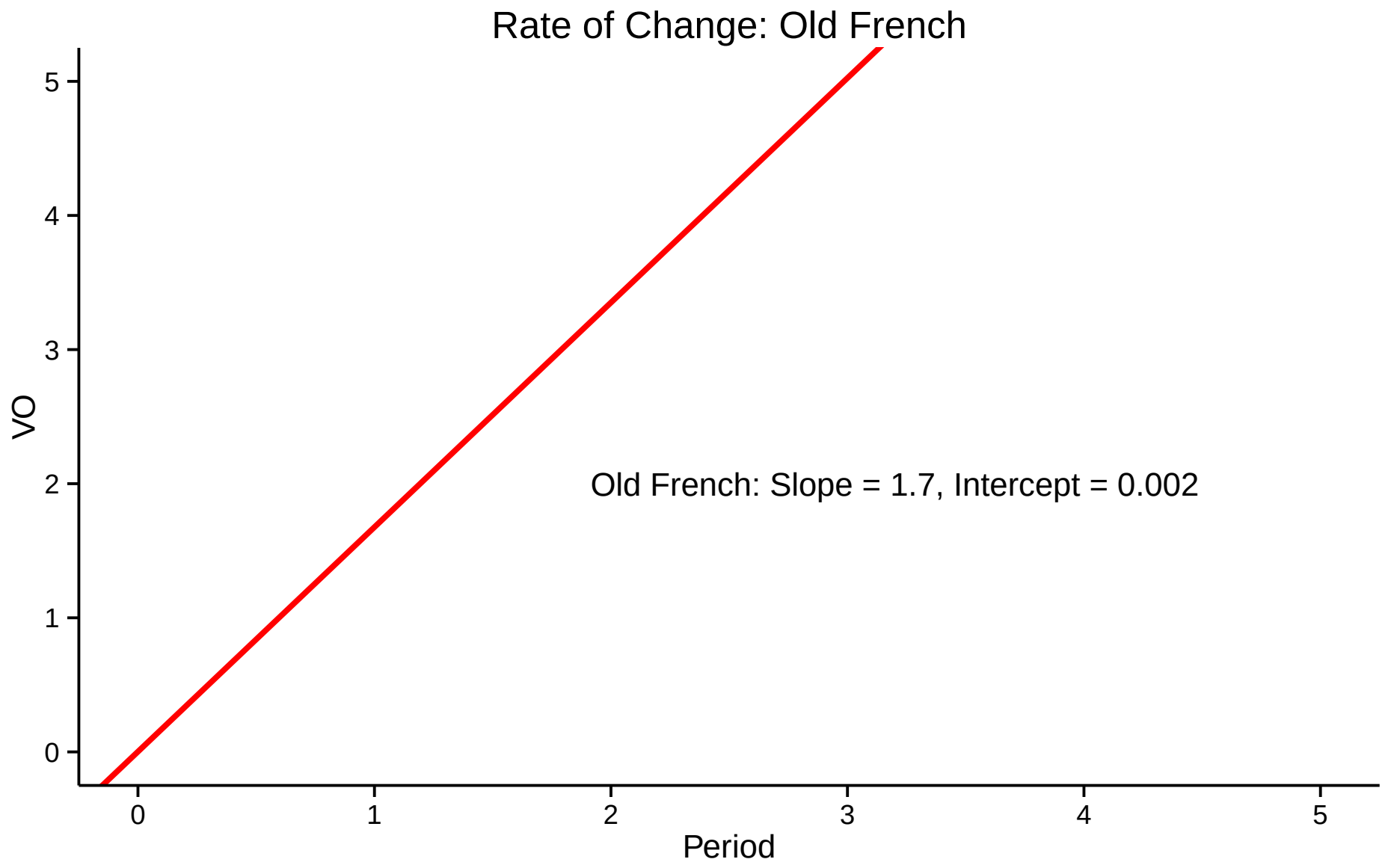

Figure 7.2: Rise of VO in Infinitival Clauses in Old French (Logit Transform)

Let us examine word order patterns in Old French. The present data exhibit the following word order patterns:

(93) a. XOV

| | | | | | |
|---|---|---|---|---|---|
| Et | a | sos | sancz | honor | porter |
| and | to | his | saints | honor-acc.obj | carry-inf |

‘and to honor his saints’ (Saint Léger)

b. XVO

| | | | | | |
|---|---|---|---|---|---|
| et | del | chevalier | demander | le | non |
| and | of | knight | ask-inf | the | name-acc.obj |

‘and to ask the name of this knight’ (Yvain, 55.1910)

c. VXO

Ge vuel avoir a toi afere
I want have-inf to you matter-acc.obj

'I have a matter to discuss with you' (Tristan, 3912)

d. OXV

Lors li vont son cheval fors treire
toward him go-3p.pl his horse-acc.obj forward bring

'they bring him his horse' (Yvain, 127.4376)

e. VOX

Pur tenir l' ordre e la meisun.
for have the order in the house

'in order to keep order in house' (MARIEF$_L$AIS,.3913)

Table 7.2 illustrates the raw frequencies for all possible patterns with verb, object and any other constituent (X). We can see that OXV and XOV are the least frequent orders in the data. There is also a notable increase in VOX order. Furthermore, OVX and XVO are almost equal in their raw frequencies except during the 13th century, where the present corpus has no data for XVO. Finally, OV order is dominant until the 13th century, where VO order is prevalent.

Table 7.2: Word Order Patterns in Infinitival Clauses in Old French: Verb, Object and X (Other Constituents)

| Period | OV | OVX | OXV | XOV | VO | VOX | VXO | XVO |
|---|---|---|---|---|---|---|---|---|
| 11th | 40 | 1 | 1 | 1 | 32 | 1 | 0 | 0 |
| 12th | 251 | 14 | 7 | 10 | 207 | 16 | 13 | 14 |
| 13th | 36 | 21 | 1 | 0 | 74 | 23 | 11 | 0 |
| 14th | 171 | 19 | 7 | 3 | 392 | 106 | 23 | 17 |
| Sum | 498 | 55 | 16 | 14 | 705 | 146 | 47 | 31 |

From Table 7.2 it is not clear how all these patterns are related with respect to chronological periods. The association between word order patterns and periods can be analyzed using the Correspondence Analysis method, which statistically examines proximity between two factors, e.g. word order patterns and chronological periods, in a two dimensional scale (see section 5.3). In Figure 7.3 we can see that the 11th and the 12th centuries are grouped together. Furthermore, they are strongly associated with OV and XOV. The 13th century region shows an association with OVX and VXO. The XVO order lies between regions, that is, it is not associated with a specific century. Finally, the 14th century is associated with VOX and VO. This plot represents 95% of the variation in the data.

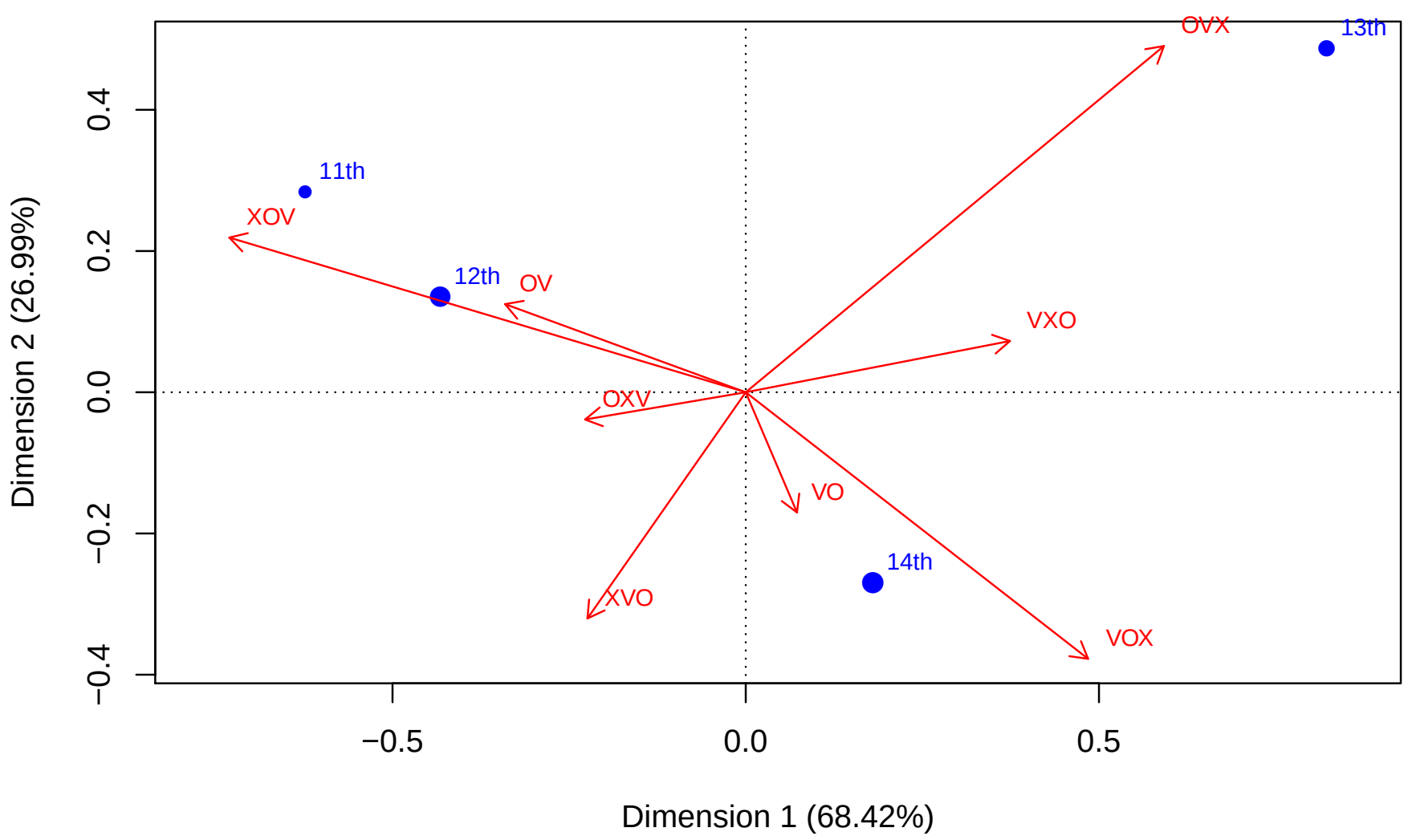

Figure 7.3: Correspondence Analysis: Old French Word Order Patterns in Infinitival Clauses

Let us reflect on the obtained association patterns. According to the data, we can identify the following patterns of change:

(94) 11-12th **(X)O(X)V** → 13th **OVX/VXO** → 14th **VO(X)**

This pattern mirrors a traditional pattern of change from OV to VO via an OVX/XVO stage in main clauses. However, in order to evaluate this statement with respect to these results, it is necessary to examine the information structure of preverbal and postverbal nouns. Recall from section 2.3 that Old French is traditionally described as a verb-second (V2) language (Thurneysen, 1892; Skarup, 1975) and that the first position in a sentence is often reserved for a topicalized or focused constituent. Let us evaluate the effect of information structure on the OV/VO alternation. Recall that the present corpus is codified for two levels of information structure (section 1.3.2): i) information status and ii) informational relevance. The first level describes the cognitive status of nouns: i) *old* - tokens that can be evoked from previous context, ii) *accessible* - token that can be inferred from common knowledge and iii) *new* - tokens that are not recoverable from the context. The second level examines informational salience of nouns and distinguishes between contrastive focus and information focus (non-contrastive). Thus, if the present corpus reflects a V2 language, we would expect to find preverbal object nouns with marked pragmatic values. Table 7.3 illustrates the distribution of information structures of preverbal and postverbal nouns.[1] In the 11th century, there is no evidence of contrastive nouns in the preverbal position, and there are only four contrastive nouns in the postverbal position. Furthermore, old information has a higher frequency preverbally than postverbally. In contrast, there are slightly more cases of new information in the postverbal position. In the 12th century, there are more contrastive nouns and old information in the preverbal position, suggesting that preverbal position is, indeed, often reserved for pragmatically marked constituents. In contrast, in the 13th century, there is an increase of *new* and *old* information in the postverbal position. Similarly, in the 14th century we can observe more *old information* and *contrastive focus* on postverbal nouns than on preverbal nouns.

[1]The relative frequencies are calculated separately per each category, namely *new*, *old*, *acc*, *contrastive focus* and *new information focus*.

Table 7.3: Information Structure and Word Order in Infinitival Clauses in Old French

(a) Information Status

| | Accessible | | New | | Old | |
|---|---|---|---|---|---|---|
| Period | OV/% | VO/% | OV/% | VO/% | OV/% | VO/% |
| 11th | 6/75 | 2/25 | 9/41 | 13/59 | 28/61 | 18/40 |
| 12th | 52/47 | 58/53 | 103/51 | 100/49 | 127/58 | 92/42 |
| 13th | 10/23 | 33/77 | 24/39 | 38/61 | 23/38 | 38/62 |
| 14th | 39/20 | 160/80 | 53/21 | 197/79 | 108/37 | 181/63 |
| Sum | 107 | 253 | 189 | 348 | 286 | 329 |

(b) Information Relevance

| | Contrast | | Inf. Focus | |
|---|---|---|---|---|
| Period | OV/% | VO/% | OV/% | VO/% |
| 11th | 0 | 4 | 42/60 | 28/40 |
| 12th | 39/70 | 23/37 | 242/52 | 226/48 |
| 13th | 10/63 | 6/37 | 47/31 | 103/69 |
| 14th | 34/43 | 46/57 | 166/25 | 491/75 |
| Sum | 83 | 79 | 497 | 848 |

As mentioned earlier in section 2.3, lengthy constituents often trigger VO order. Let us examine whether this factor triggers VO order and interposes its influence between pragmatic factors and word order. The present data are coded for heaviness, therefore we can test its influence. Recall that the data are coded for two different types of heaviness, namely heaviness by word count and heaviness by syntactic length. Figure 7.4a illustrates the effect of lengthy constituents by word count, where the x-axis marks the number of words. Figure 7.4b presents the effect of syntactic heaviness, where the x-axis marks light constituents (1) and heavy constituents (2). In this case, heaviness is determined by the presence of NP postmodifiers and coordinated NPs. Figure 7.4a shows that lengthy constituents with more than two words trigger VO order. Similarly, in Figure 7.4b we see that syntactically long

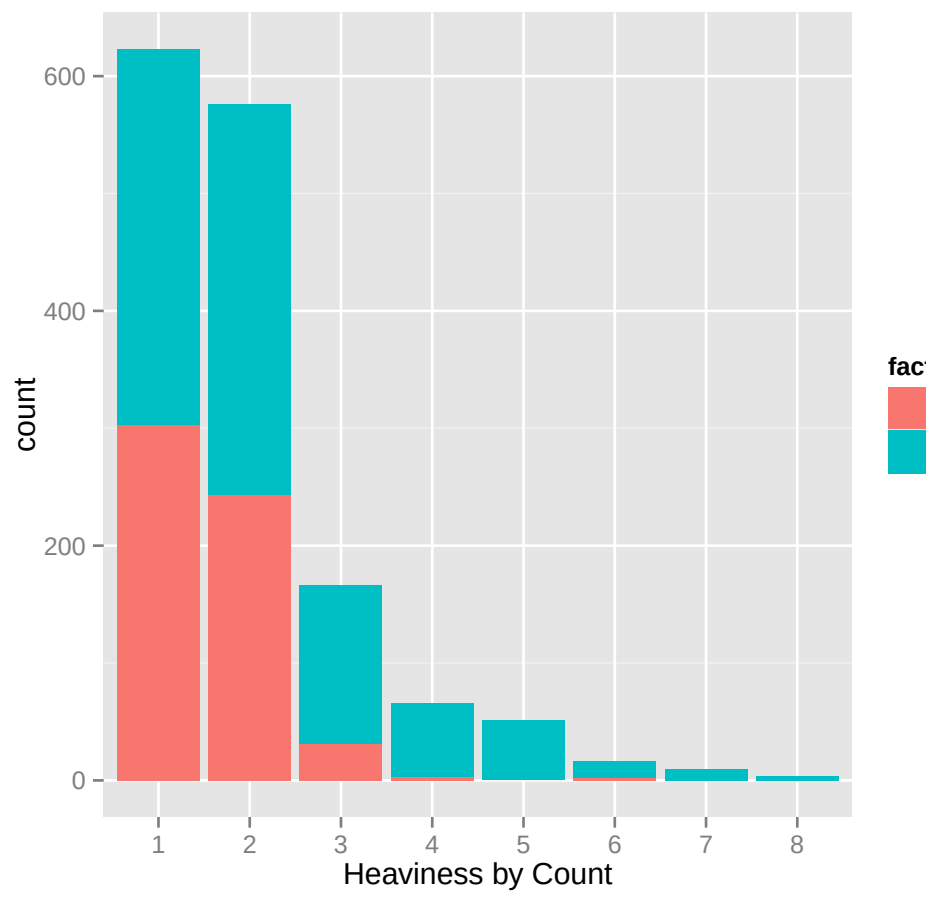

(a) Length by Word Count

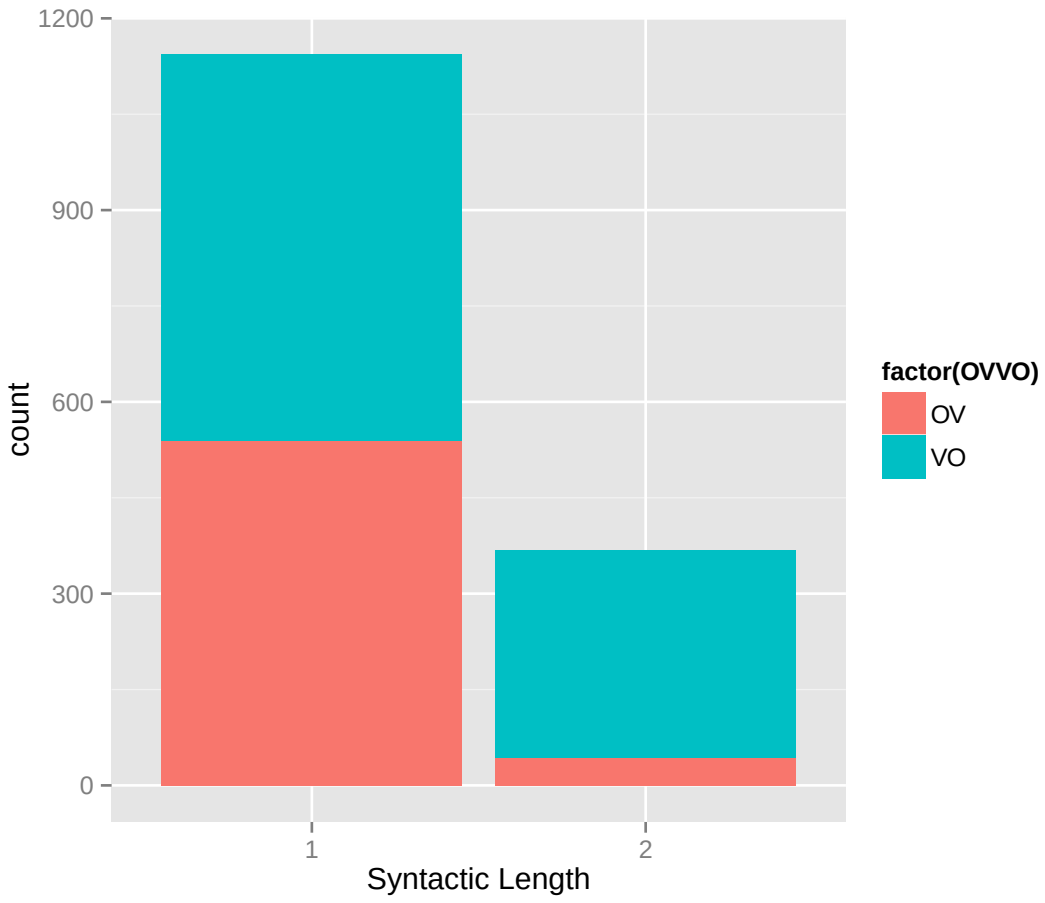

(b) Syntactic Heaviness

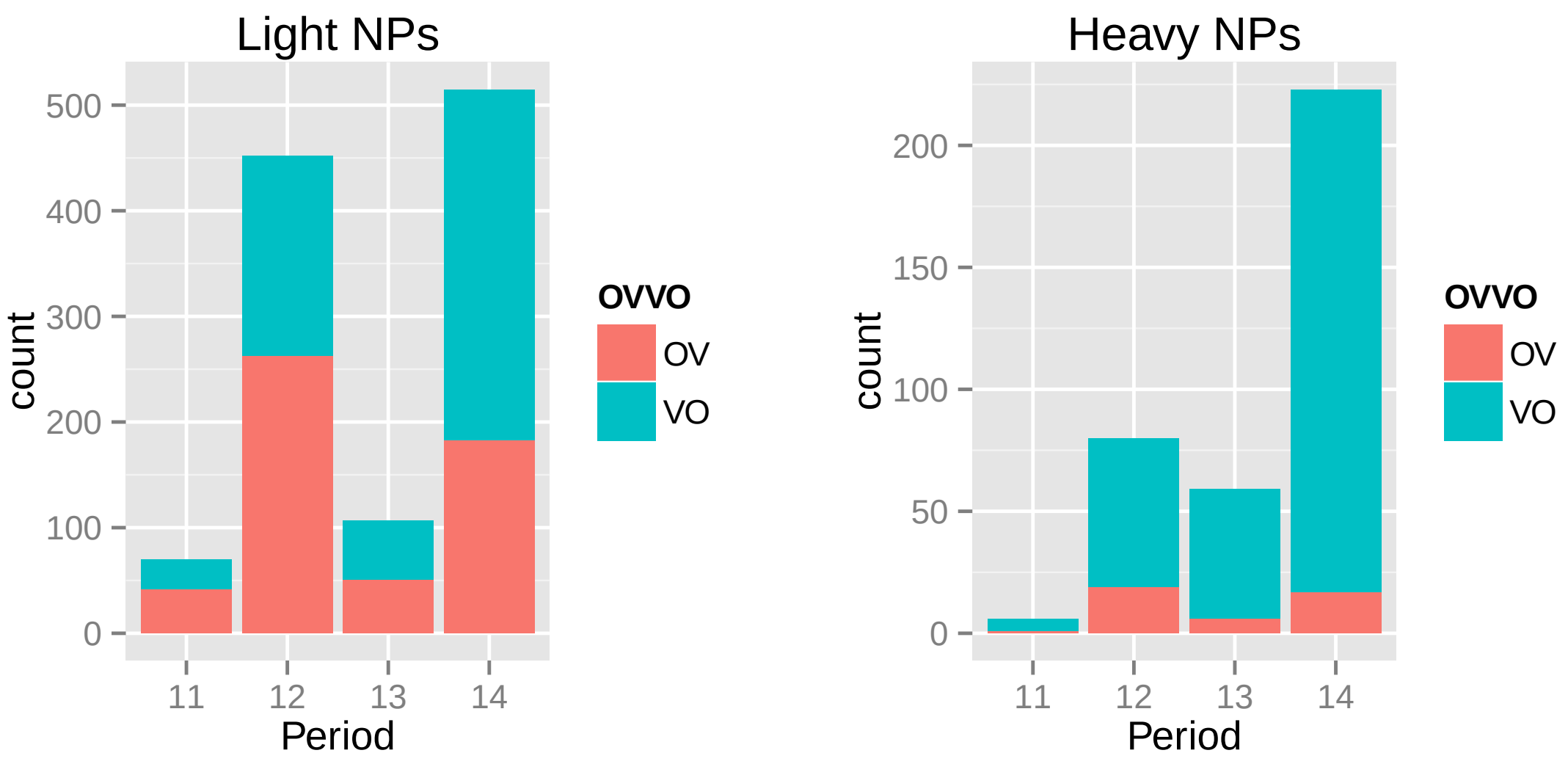

(c) Syntactic Heaviness by Periods

Figure 7.4: Word Order and NP Heaviness in Infinitival Clauses in Old French

NPs also trigger VO order. Across chronological periods, heavy constituents are also predominantly VO, as illustrated in Figure 7.4c. Therefore, it is necessary to reduce the dataset to constituents with light weight for better evaluation of information structure (see table 7.4). However, even without the effect of heavy constituents, the tendency remains similar: preverbal nouns with *old information* are more frequent than nouns with *new information* in the 11th-12th centuries, with a ratio of 2:1. In contrast, in the 13th century the ratio between *old* and *new* is 1:1, followed by a ratio 1:1.5 in the 14th century. Furthermore, the frequency of *contrastive information* for preverbal nouns is higher than it is for postverbal nouns, except during the 11th century, for which there are no contrastive preverbal data attested.

Table 7.4: Information Structure and Word Order in Infinitival Clauses in Old French (Syntactically Light NPs)

(a) Information Status

| | Accessible | | New | | Old | |
|---|---|---|---|---|---|---|
| Period | OV/% | VO/% | OV/% | VO/% | OV/% | VO/% |
| 11th | 6/86 | 1/14 | 9/43 | 12/57 | 27/64 | 15/36 |
| 12th | 47/51 | 46/49 | 94/54 | 79/46 | 122/66 | 64/34 |
| 13th | 8/38 | 13/62 | 22/51 | 21/49 | 21/49 | 22/51 |
| 14th | 37/30 | 86/70 | 48/28 | 125/72 | 98/45 | 121/55 |
| Sum | 98 | 146 | 173 | 237 | 268 | 222 |

(b) Information Relevance

| | Contrast | | Inf Focus | |
|---|---|---|---|---|
| Period | OV/% | VO/% | OV/% | VO/% |
| 11th | 0 | 4 | 41/64 | 23/36 |
| 12th | 30/68 | 14/32 | 232/57 | 174/43 |
| 13th | 10/83 | 2/17 | 41/43 | 54/57 |
| 14th | 31/57 | 23/43 | 152/33 | 308/67 |
| Sum | 71 | 43 | 466 | 559 |

Thus, we see a wide range of pragmatic values for both preverbal and postverbal nouns, with a tendency for *old information* to be associated with preverbal nouns in the 11th-12th centuries. The findings also reveal that preverbal objects often carry *new information*. For example, the ratio of *new information* for OV/VO in the 11-12th centuries is 1.14:1 (OV - 103 and VO -80). Similarly, *new information focus* occurs in the preverbal position with a ratio 1.4:1 (OV-273 and VO - 197). These facts suggest that the preverbal position can be pragmatically unmarked. Section 2.3 showed that the range of pragmatic features of preverbal nouns in Old French decreases over time (Marchello-Nizia, 1995; Zaring, 2010; Labelle and Hirschbühler, 2012). For example, in the 12th century, preverbal objects can be marked or unmarked and refer to *old* and *new* information. In fact, Marchello-Nizia (1995) notes that there are more *rhematic* preverbal nouns (new information) than *thematic*. By the 13th century the information role is restricted: the preverbal object only carries marked pragmatic features, such as marked theme and marked rheme. The present codification allows for such an evaluation. In addition, the present corpus makes it possible to examine postverbal nouns in contrast to previous studies that have investigated only the information structure of preverbal nouns. Figure 7.5 illustrates the information status of preverbal nouns across time, where the x-axis represents chronological periods and the y-axis displays three types of information status: new, old, accessible. Information relevance is illustrated by colors: contrastive focus - red and and information focus - blue.

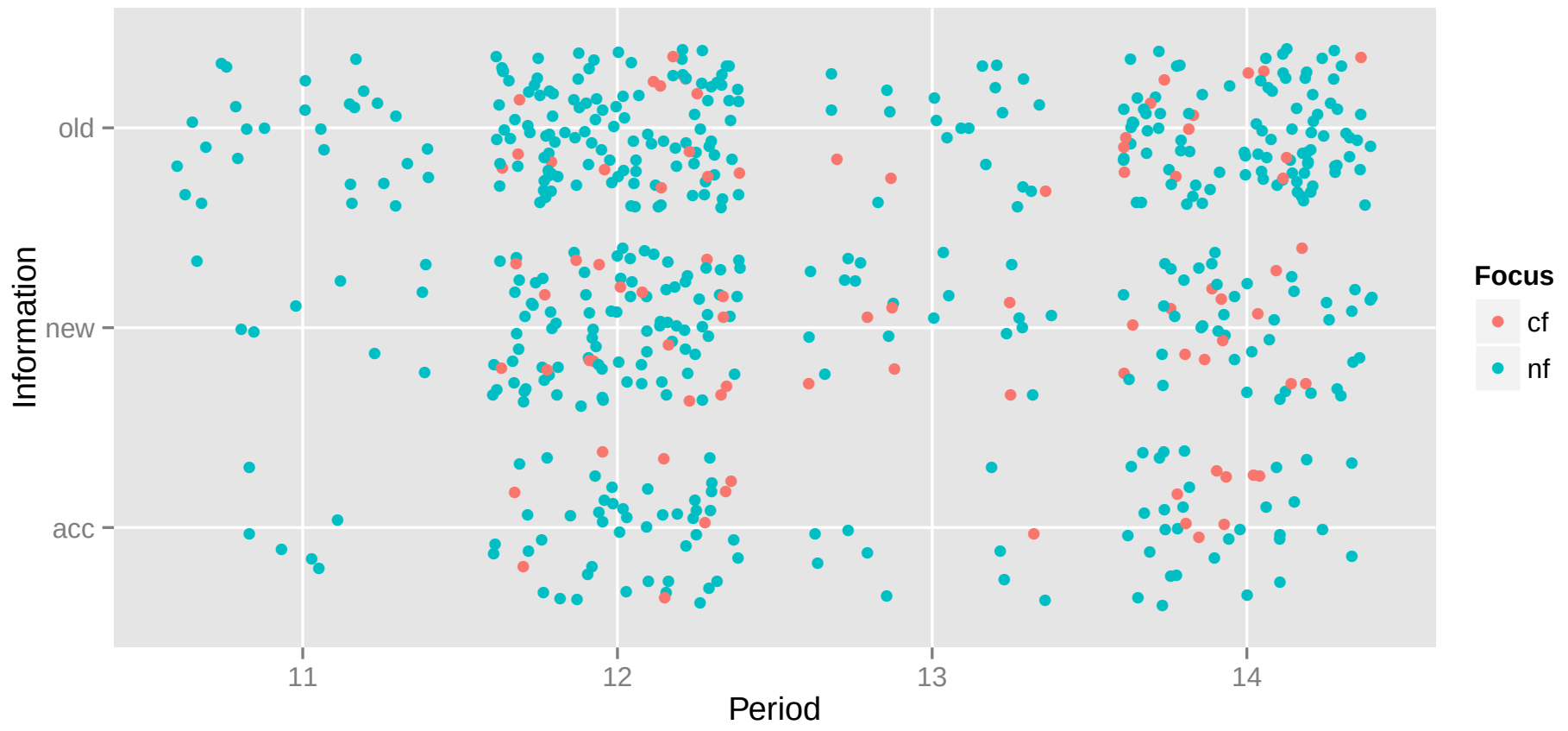

Figure 7.5: Preverbal NP and Information Structure in Infinitival Clauses: Old French

The range of information structure, namely new, old and accessible, remains constant across four chronological periods. It is also noticeable that in the 11th and 13th centuries, the data are more sparse with respect to preverbal nouns. In contrast, there is no evidence of contrastive focus in the 11th century. However, from the 12th century onward contrastive focus is found with both old and new information, suggesting that the range of information structure remains the same. Thus, the data show that preverbal nouns can express i) contrastive focus with new and old information and ii) information focus with new and old information. Labelle and Hirschbühler (2012) also look at information status from Early Old French to the early 14th century (see section 2.3). Their findings show that preverbal objects tend to display *Focus* (new information) until 1205 and *Topic* (old and accessible information) after 1225. Their Figure 2.1 is repeated here:

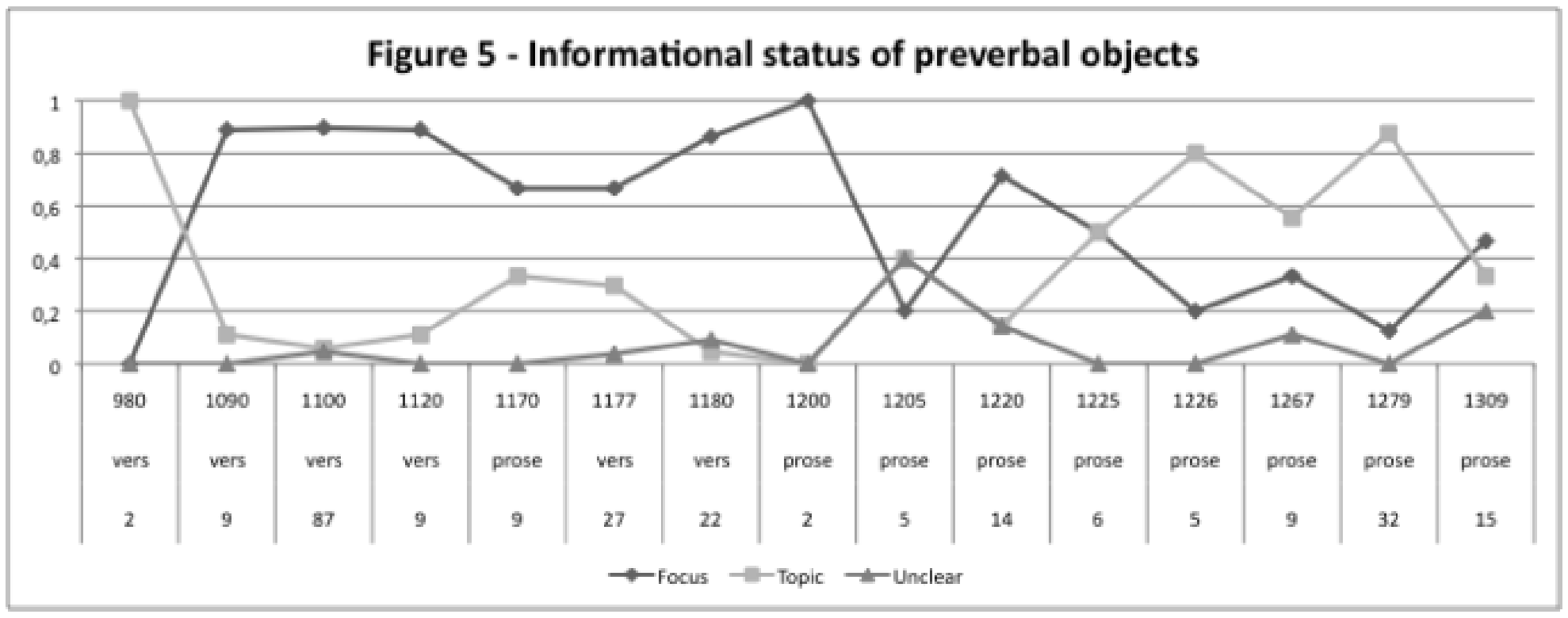

Figure 7.6: Information Status of Preverbal Nominal Objects in Old French: Finite Clauses (Labelle and Hirschbühler, 2012, 20)

Let us compare our infinitival clauses with that of Labelle and Hirschbühler (2012) for finite clauses. The present data are plotted in Figure 7.7, where the x-axis represents four chronological periods and the y-axis shows the relative frequencies of new, old and accessible types. In contrast to Figure 7.6, there is an increase in new information until the 13th century, followed by a decline in the 14th century. On the other hand, there is a decline in old information, followed by an increase in the 14th century. The frequency rate of accessible information remains almost constant (∼20%).

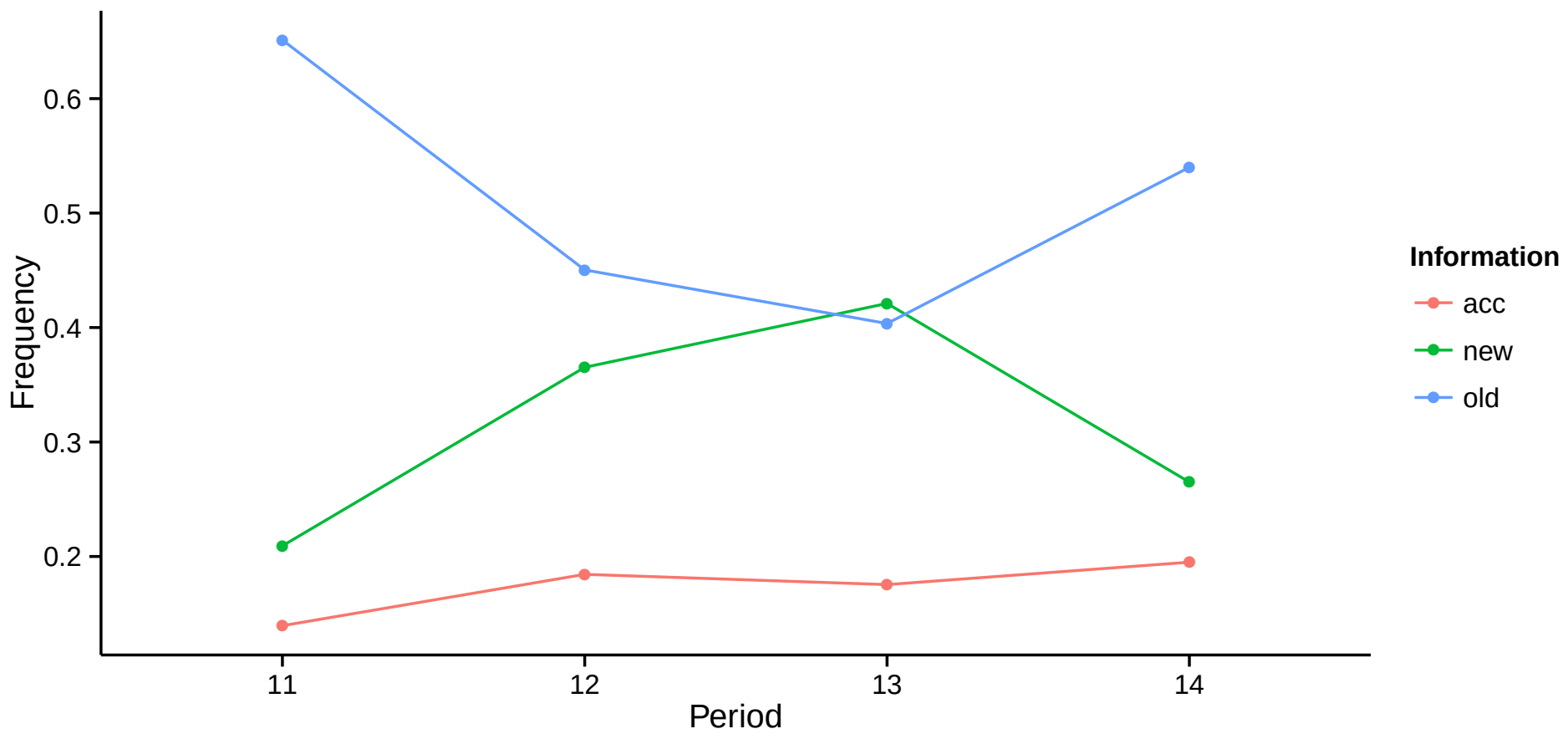

Figure 7.7: Information Status of Preverbal Nominal Objects in Infinitival Clauses in Old French: Period

A more detailed analysis is presented in Figure 7.8, where the x-axis represents the individual dates of manuscripts. It appears that *old* information is dominant until the mid 11th century where both *old* and *new information* co-exist at about 50%. There is also an increase in accessible information at the same time, followed by a stable rate (∼20%), as illustrated in Figure 7.8a. In Figure 7.8b *old* and *accessible* information are merged.[2] While the relative frequency for *Old* information after the merger has changed, the tendency remains the same: the present data show a stability in information structure from the mid 11th century until the 14th century.

[2]Recall that Labelle and Hirschbühler (2012) assign *Topic* to preverbal nouns with *old* and *accessible* information.

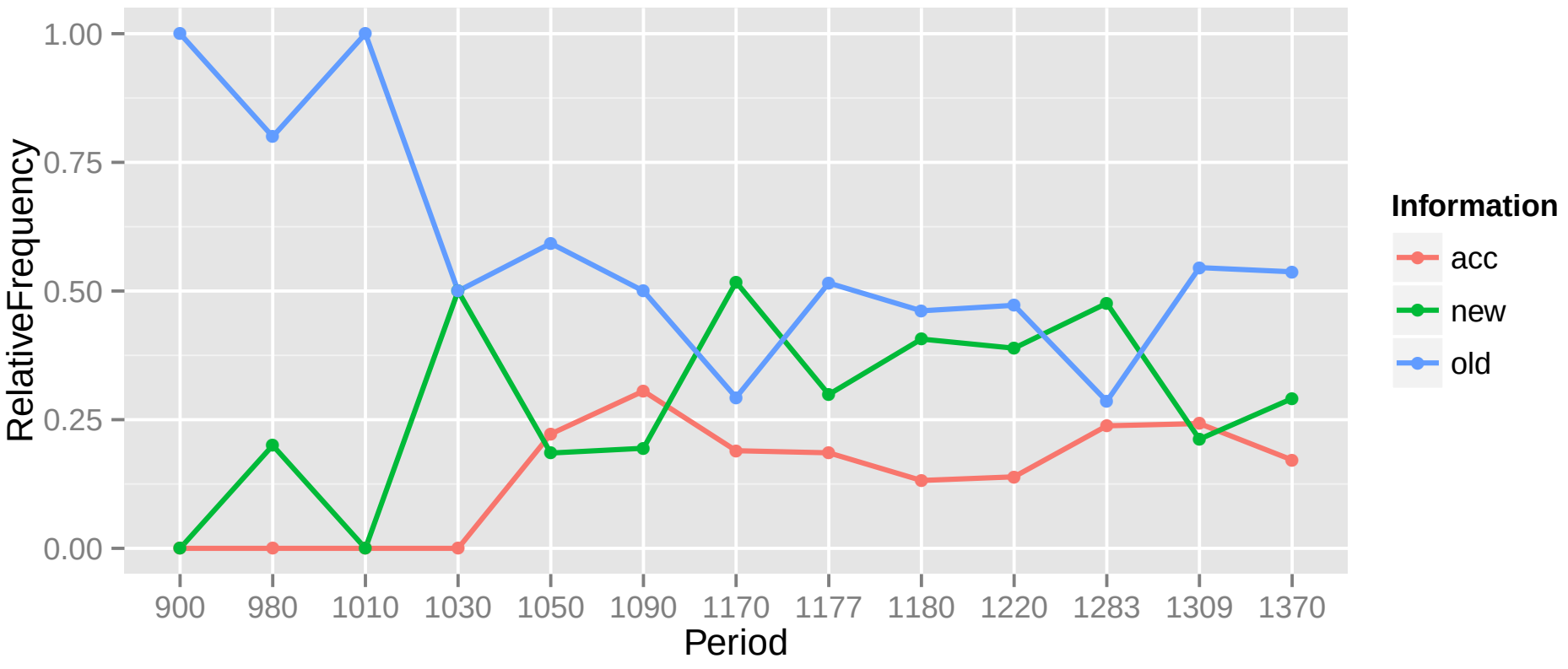

(a) Old, New and Accessible Information

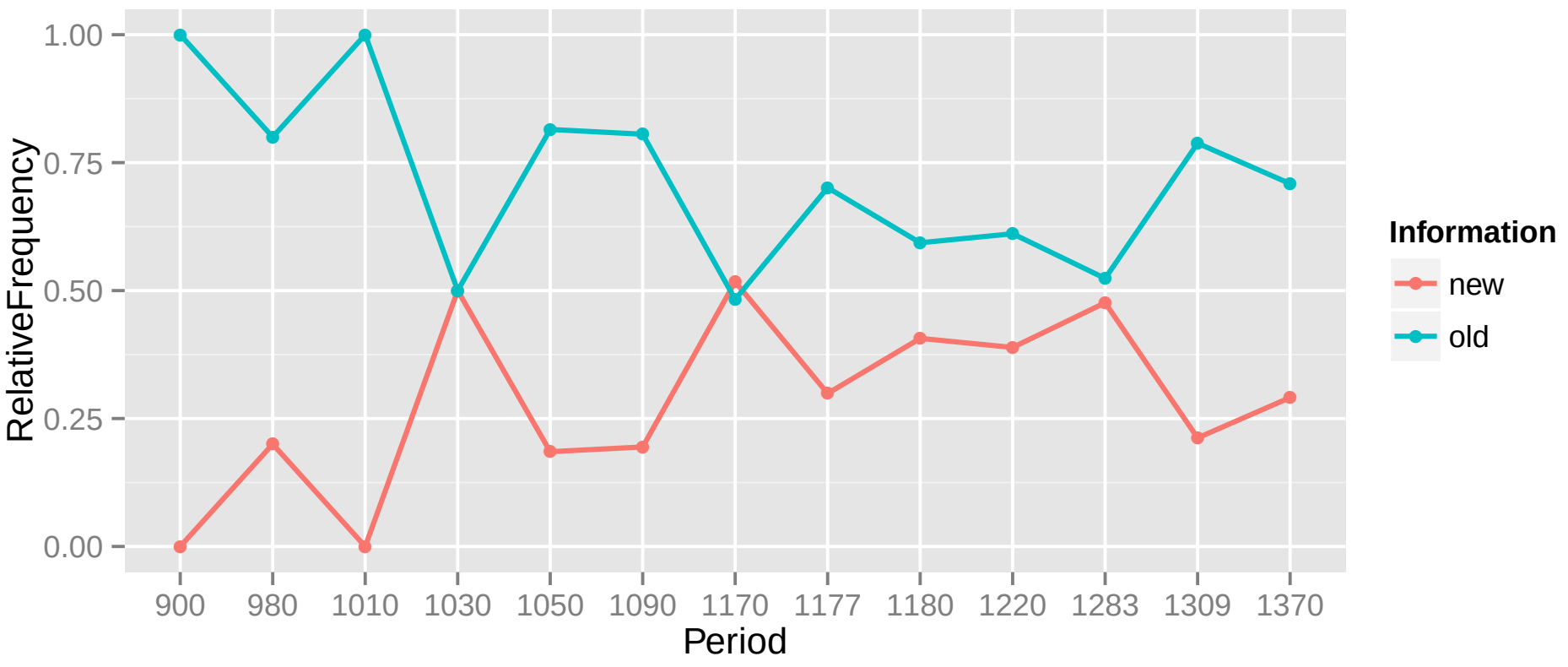

(b) Old and New Information

Figure 7.8: Information Status of Preverbal Nominal Objects in Infinitival Clauses in Old French: Date of Manuscript

Let us turn now to the postverbal nouns. Figure 7.9 illustrates the information status of postverbal nouns across time, where the x-axis represents chronological periods and the y-axis displays three types of information status: *new*, *old*, *accessible*. Information relevance is illustrated by colors: *contrastive focus* - red and and *new information focus* - blue. Similar to the preverbal nouns (see Figure 7.5), all three categories of information status are present in postverbal nouns. Furthermore, there is a similarity with respect to data sparsity in the 11th century. In contrast, there is a difference in information relevance. *Contrastive focus* is attested with *old information* in the 11th century. In the 12th and 14th centuries both *contrastive* and *new information* focus are present with all three types of information status, whereas in the 13th century there is no attested evidence for contrast with *old information*, as compared to the preverbal nouns.

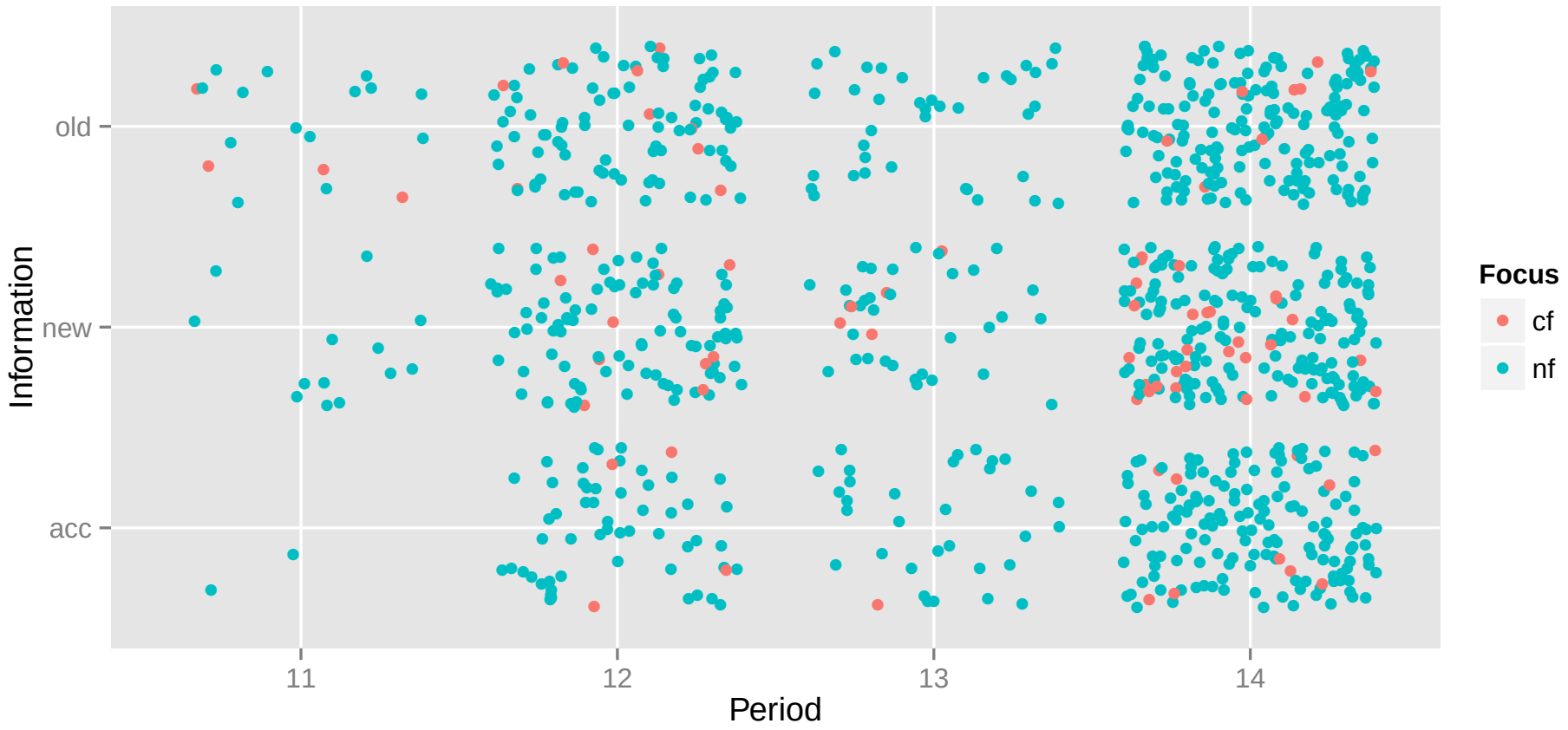

Figure 7.9: Postverbal NP and Information Structure in Infinitival Clauses in Old French

Figure 7.10 illustrates the distribution of information status across four periods. The data show a stability of *new* and *old* information from the 12th century onward at the same frequency rate (∼40%). There is also a small decrease in *old information* from ∼55% in the 11th century to 40% in the 12th century, and there is an increase in *accessible information* from ∼5% to 30% by the 13th century. These facts imply that pragmatic factors have some effect on postverbal nouns in Early Old French; this effect, however, has diminished over time.

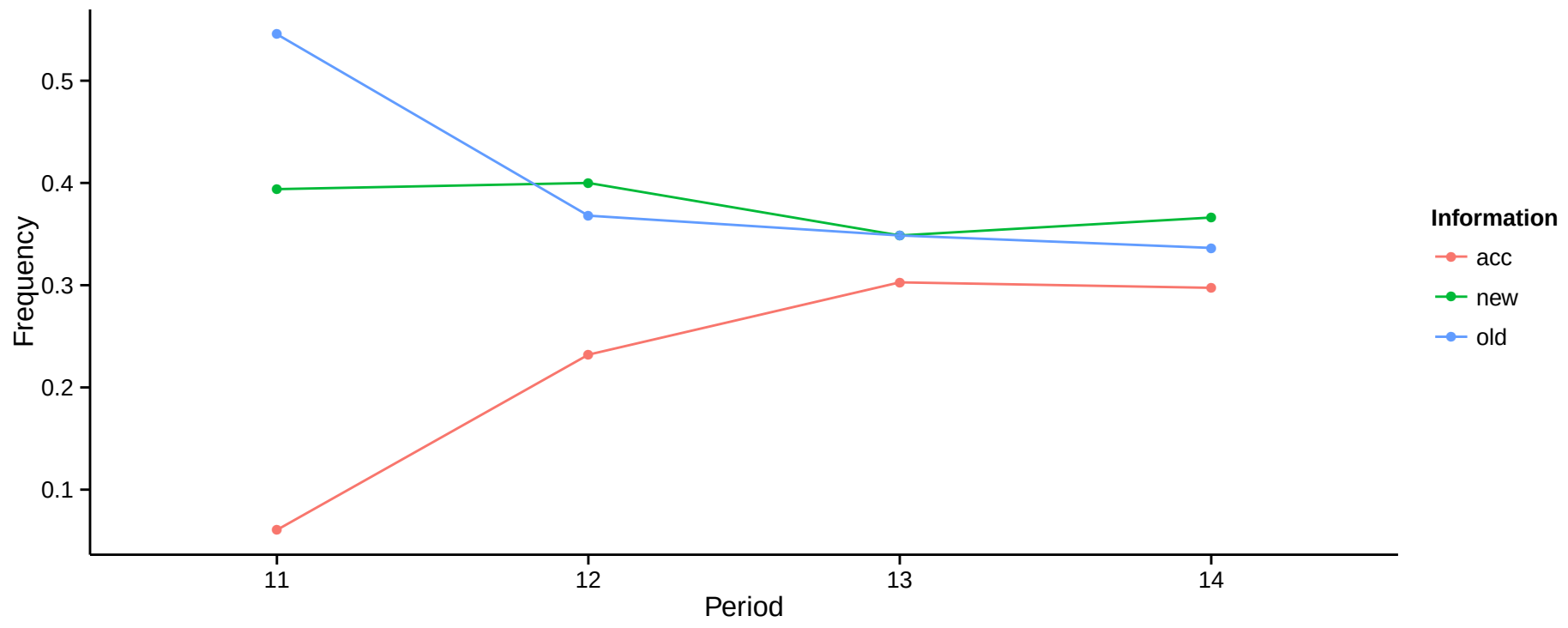

Figure 7.10: Postverbal NP and Information Stratus in Infinitival Clauses in Old French: Period

Figure 7.11 shows a more detailed distribution of information structure. Similar to Figure 7.9, there is a decrease in old information and an increase in new information by the mid 11th century. The remaining data show a stable variation between old, new and accessible information, with one small drop of new information in the year 1283.

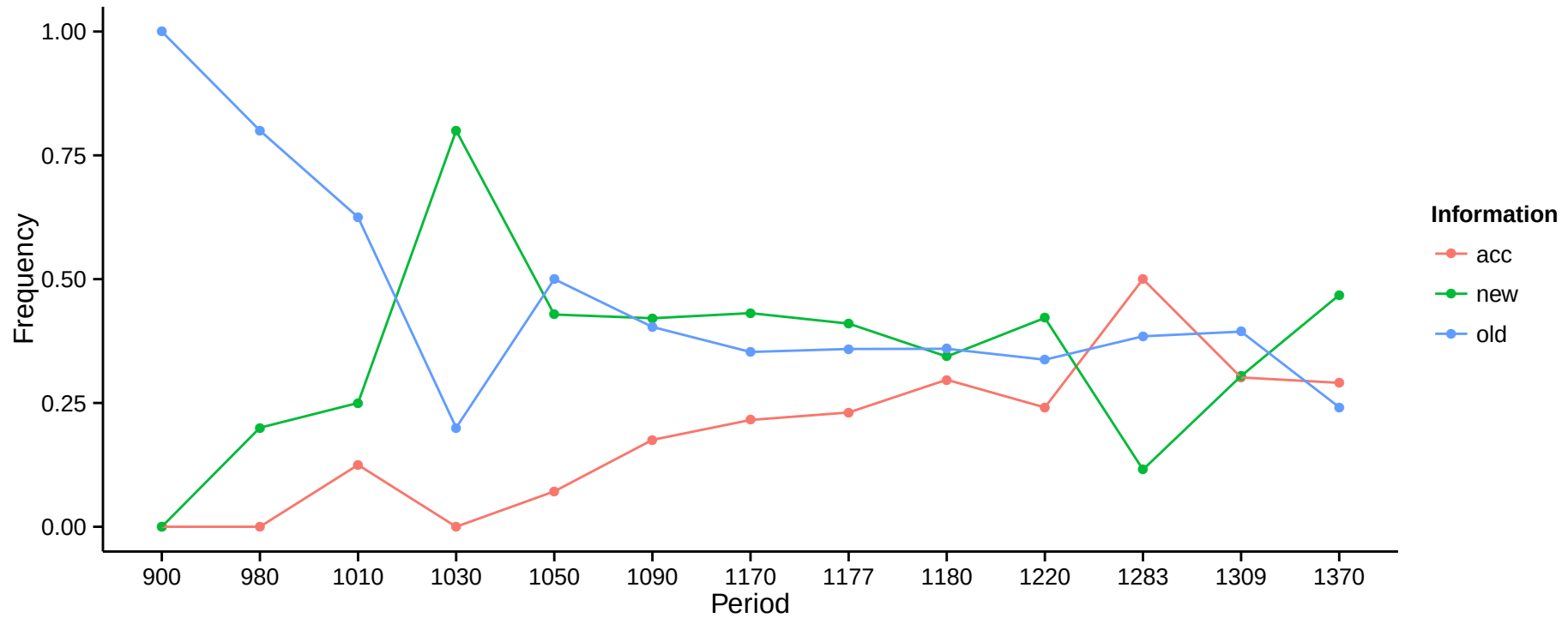

Figure 7.11: Postverbal NP and Information Status in Infinitival Clauses in Old French: Date of Manuscript

Thus far, the data demonstrate that the range of information structure remains closely similar with postverbal and preverbal nouns. In addition, the period from the 12th until the 13th century shows a relative stability in the variation. However, there are two notable differences: i) Preverbal nouns show a greater difference in the 11th century, namely 65% of old and 20% of new information, whereas postverbal nouns show a very small difference, namely 55% of old and 40% of new information; ii) There is a decline in new information and an increase in old information for preverbal nouns, while postverbal nouns show a stable variation. These facts suggest that by the 12th century the postverbal position gains some stability on the level of information structure. On the other hand, the preverbal position is unstable, showing a decline in new information in the 14th century and an increase in old information in the 14th century. Apparently, a pure pragmatic hypothesis is not sufficient to explain word order change in Old French. In the next section, I will look at syntactic factors, namely position of infinitives and type of matrix verbs.

### 7.1.2 Infinitival Clauses: Type and Position

Clause type has frequently been shown to affect word order (section 2.3). From Early Old French OV order is found more frequently in subordinate clauses than in main clauses (see Table 2.4). Furthermore, verb form exhibits an influence on word order. For example, OV order appears very frequently with infinitives as compared to finite verbs (Zaring, 2010). In this section, two more syntactic categories are addressed with respect to word order: i) Position of infinitives and ii) Type of infinitives. It has been shown that there is no evidence of an independent AcI construction in Old French in contrast to Latin (Bauer, 1999). Indeed, the attested infinitival positions in the corpus are the following: i) preposed infinitive, ii) postposed and iii) prepositional infinitive.[3] Figure 7.12 illustrates the distribution of infinitival clauses with respect to OV/VO order[4] and Table 7.5 illustrates their raw frequencies. It is evident that preposed infinitives have nearly disappeared in Old French. These infinitives almost entirely exhibit OV order, which suggests their archaic character. On the other hand, from the 12th century there is a development of prepositional infinitives. First, the ratio of OV/VO in these infinitives is 2:1 in the 12th century, followed by 1:1 in the 13th century and finally 1:2 in the 14th century. In contrast, the postposed infinitives display the highest frequency rate of VO from the 12th century onward. These findings suggest that postposition is a norm in Old French, preposed infinitives are uncommon and there is an on-going development of prepositional clauses.

[3]Prepositional clauses are not coded by their position with respect to the main verb of the sentence, since infinitives are governed by the preposition and are always postposed with respect to their preposition. However, all types of prepositions are included here, namely aspectual verbs (prepositional clause introduced by à, de) and other prepositional clauses.

[4]Prepositional - *prep*, preposed - *pre* and postposed - *post*.

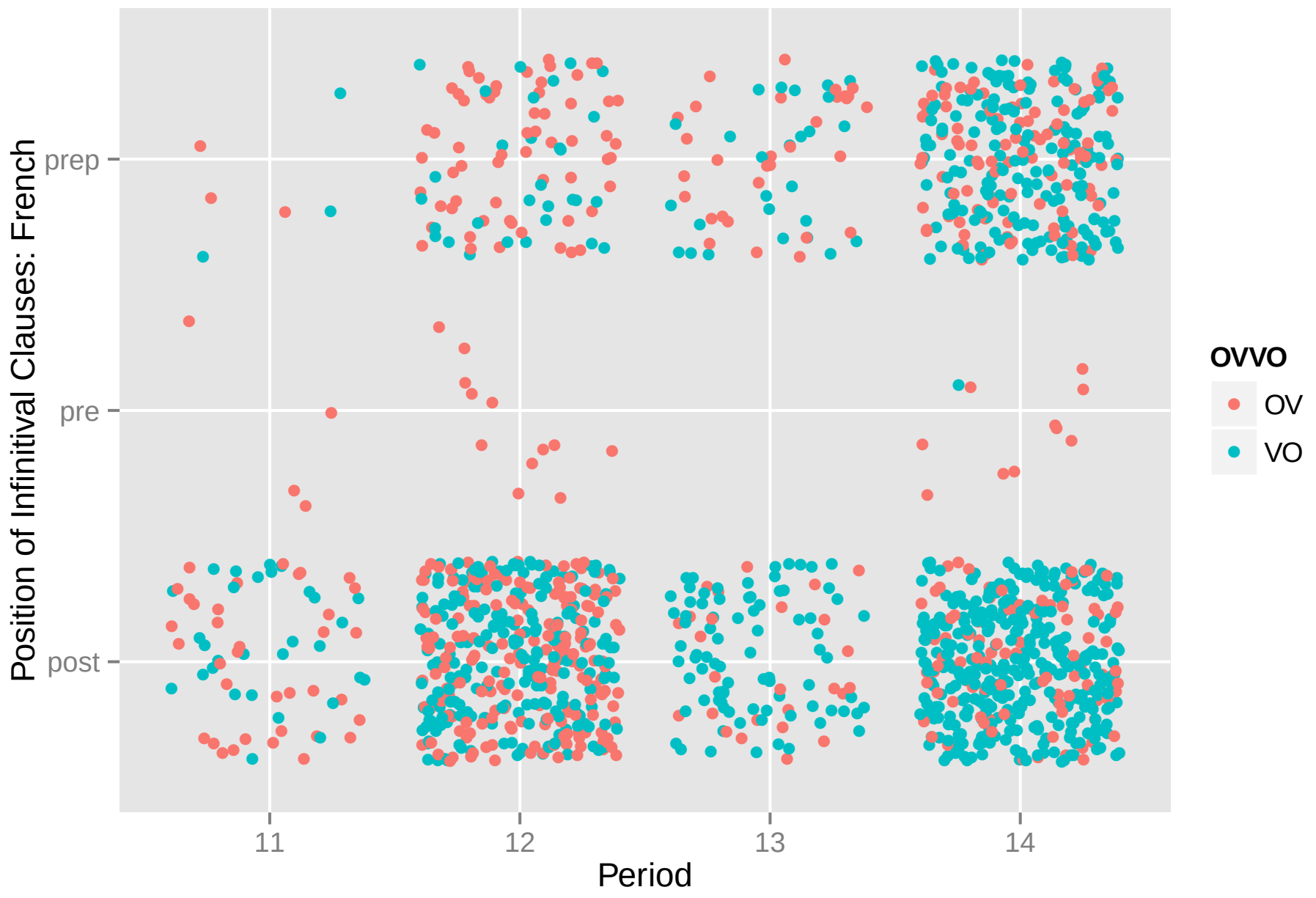

Figure 7.12: Position of Infinitival Clauses with Respect to Main Verbs across Periods in Infinitival Clauses in Old French

Table 7.5: OV/VO Order and Position in Infinitival Clauses with Respect to Main Verbs in Old French

| | Postposed | | Preposed | | Prepositional | |
|---|---|---|---|---|---|---|
| Period | OV | VO | OV | VO | OV | VO |
| 11th | 36 | 30 | 4 | 0 | 3 | 3 |
| 12th | 208 | 220 | 12 | 0 | 62 | 30 |
| 13th | 26 | 83 | 0 | 0 | 31 | 26 |
| 14th | 92 | 352 | 10 | 1 | 98 | 185 |
| Sum | 362 | 685 | 26 | 1 | 194 | 244 |

The rate of VO change for prepositional and postposed infinitives is illustrated in Figure

7.13.[5] There is no significant difference in their values, suggesting that the frequency of VO order increases at the same rate in postposed infinitives and prepositional infinitives.

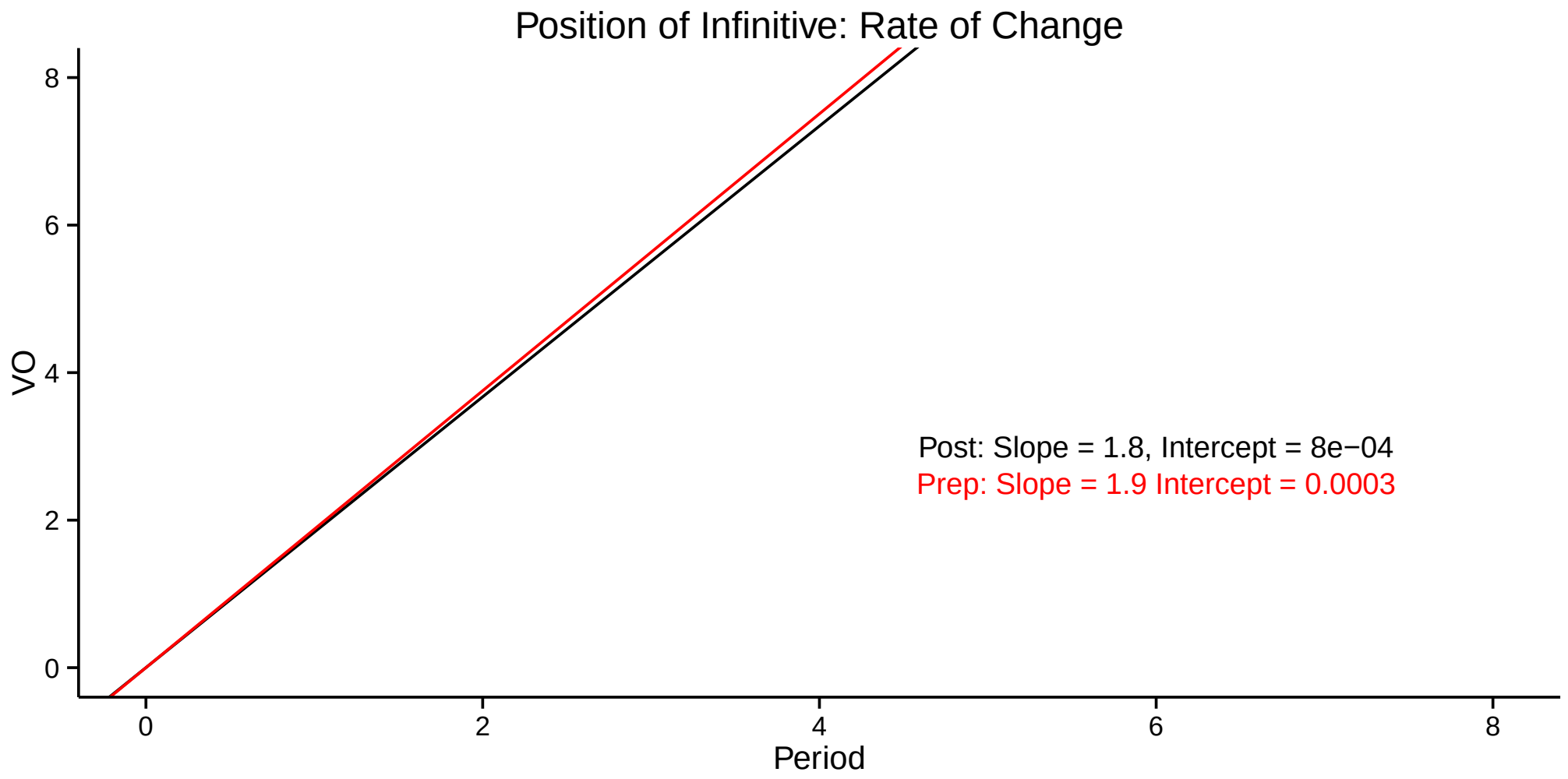

Figure 7.13: VO Rate across Different Types of Infinitival Position in Old French

The second syntactic category that is addressed in this section is the type of infinitival clause. Recall that infinitival clauses are classified into the following classes: i) Accusativus cum Infinitivo, ii) Raising structure, iii) Control structure, iv) Simple infinitive, v) Restructuring structure and vi) Prepositional infinitive[6] (see section 5.2.2.4). Table 7.6 summarizes the raw frequencies for the four classes, as there is no evidence of AcI construction in Old French. Prepositional and Restructuring verbs are the two most common infinitival types in Old French.

[5]It was not possible to include the *preposed* group, as it contains missing values.

[6]In this factor, prepositional infinitives are treated as a syntactic structure, and the codification distinguishes between prepositional verbs, classified as aspectual (see section 5.2.2.4) and other types of prepositions.

Table 7.6: Infinitival Clause Types in Old French

| Period | Control | Prepositional | Restructuring | Simple |
|---|---|---|---|---|
| 11th | 2 | 6 | 61 | 7 |
| 12th | 13 | 94 | 410 | 15 |
| 13th | 0 | 57 | 109 | 0 |
| 14th | 0 | 286 | 433 | 19 |
| Sum | 15 | 443 | 1013 | 41 |

Figure 7.14 illustrates the distribution of these verbs with respect to OV/VO order, and their raw frequencies are reported in Table 7.7. Similar to the Latin data, there are two verbal categories, with a sparse distribution, namely *Simple* and *Control.* (see Figure 7.14). These two groups are merged together to form the *Simple* group. Figure 7.14 also demonstrates that the *Prepositional* group starts with only a few tokens in the 11th century and continues increasing over time. This fact suggests that this group is an innovation. Recall that the data do not show any prepositional infinitival verbs in Latin.

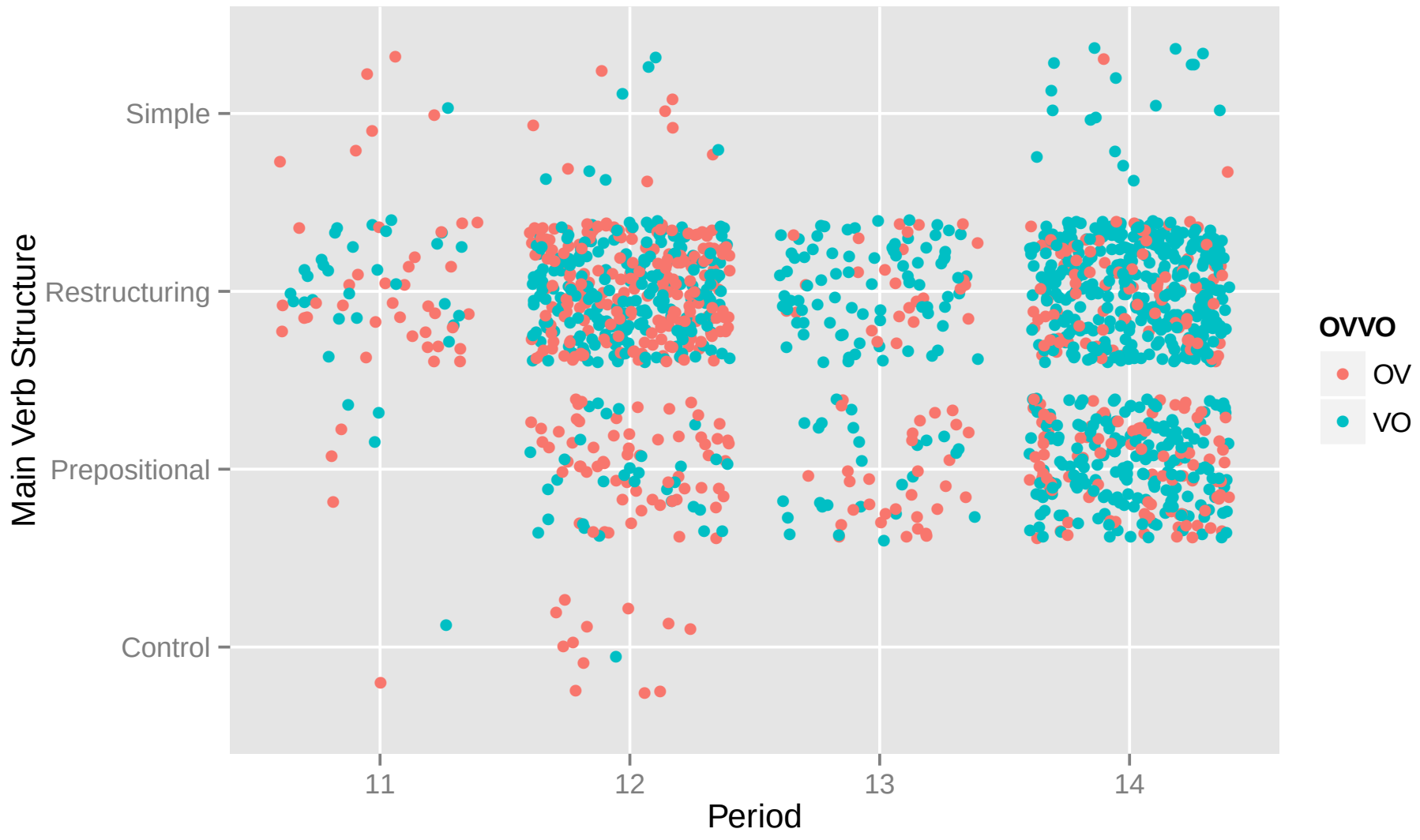

Figure 7.14: Main Verb Structures across Periods in Infinitival Clauses Old French

Table 7.7: OV/VO Order and Main Verb Structures in Infinitival Clauses in Old French

| | Control | | Prepositional | | Restructuring | | Simple | |
|---|---|---|---|---|---|---|---|---|
| Period | OV | VO | OV | VO | OV | VO | OV | VO |
| 11th | 1 | 1 | 3 | 3 | 33 | 28 | 6 | 1 |
| 12th | 12 | 1 | 63 | 31 | 199 | 211 | 8 | 7 |
| 13th | 0 | 0 | 31 | 26 | 26 | 83 | 0 | 0 |
| 14th | 0 | 0 | 98 | 188 | 100 | 333 | 2 | 17 |
| Sum | 13 | 2 | 195 | 248 | 358 | 655 | 16 | 25 |

These three infinitival groups differ in their VO distribution, as shown in Figure 7.15. Interestingly, the *Restructuring* group maintains the highest frequency of VO order, similar to the Latin data. The lowest, flat rate is observed with the *Other* group, whereas the *Prepositional* groups exhibit a constant increase in VO.

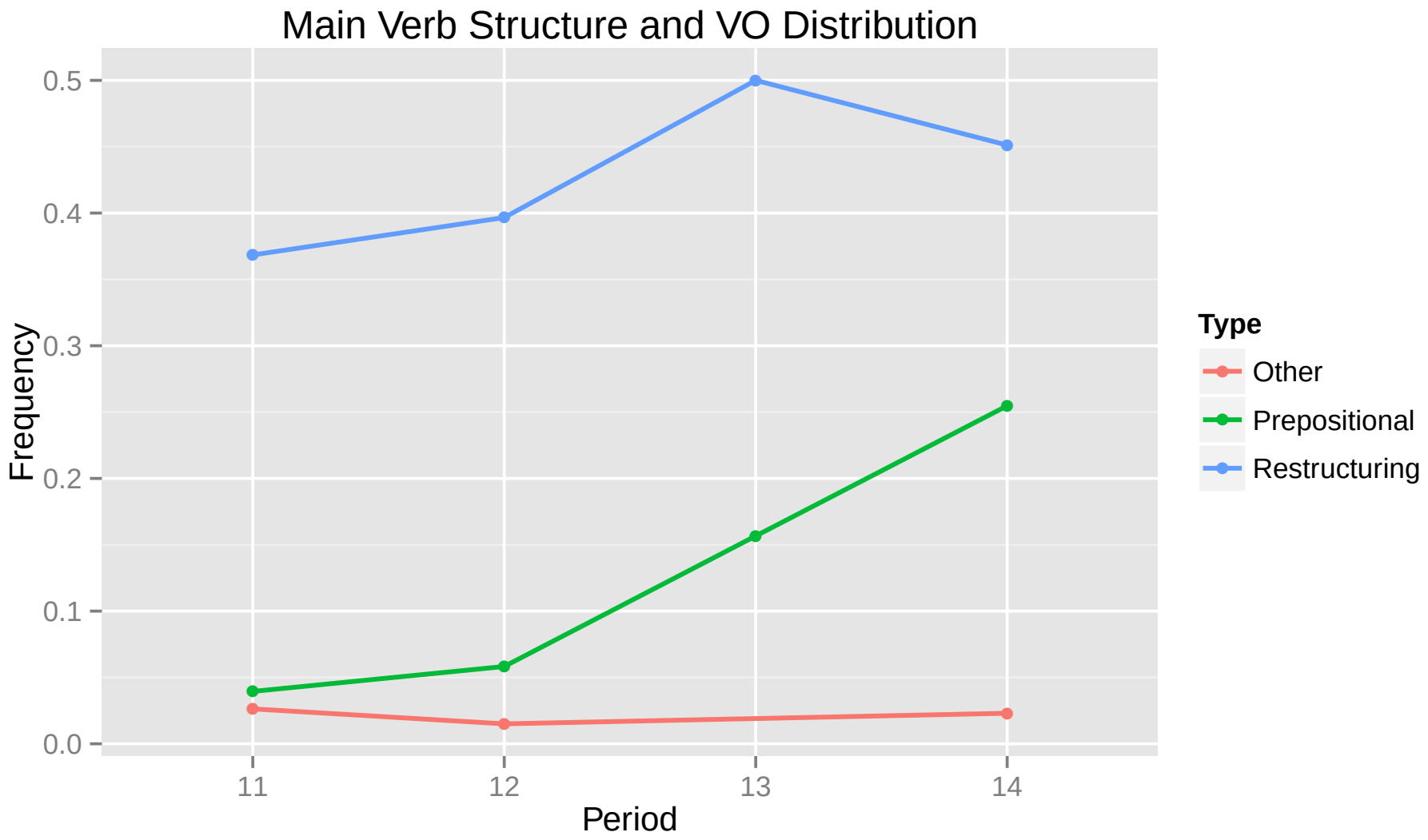

Figure 7.15: Frequency of VO Order in Infinitival Clauses with Respect to Main Verb in Old French

Let us examine restructuring verbs in greater detail. In order to determine whether there exists a tight cluster between a main verb and an infinitival verb, the subset of postposed restructuring verbs was extracted. Table 7.8 reports the results. While there is a constant decrease in OV order, the occurrence of a direct object between the main verb and its infinitive is still very frequent, especially in the 11th and 12th centuries. These findings suggest that the high frequency of VO with restructuring verbs is not induced by the tight relation between the two verbs, namely Vmain and Vinfinitive.

Table 7.8: Restructuring Verbs with Postposed Infinitives in Old French

| Period | Vmain+Object+Vinf/(%) | Vmain+Vinf+Object/(%) | Sum |
|---|---|---|---|
| 11th | 30/(52) | 28/(48) | 58 |
| 12th | 190/(47) | 211/(53) | 401 |
| 13th | 26/(24) | 83/(76) | 109 |
| 14th | 90/(21) | 332/(79) | 422 |
| Sum | 336 | 654 | 990 |

So far, we have not examined the information structure distribution in prepositional and restructuring verbs. Since prepositional verbs show an ongoing development and a slow increase in VO, I will start with this type. Table 7.9 and Figure 7.16 illustrate the distribution of information structure with OV order. First, there is a similarity in the information structure for preverbal nouns in prepositional verbs and the overall picture of preverbal nouns in Figure 7.7, namely an increase in *new information* until the 13th century and a decrease in *old information* in the 13th century (see Figure 7.16a). This distribution also shows that *old information* category is predominant in the 11th, 12th and 14th centuries (see Figure 7.16b). In addition, a wide variance between *old* and *new information* suggests a sensitivity of preverbal word order for pragmatic values.

Table 7.9: Information Structure in Prepositional Verbs in Old French: Preverbal Nouns

| Period | Accessible | New | Old | Sum |
|---|---|---|---|---|
| 11th | 1 | 0 | 2 | 3 |
| 12th | 10 | 17 | 36 | 63 |
| 13th | 8 | 14 | 9 | 31 |
| 14th | 9 | 25 | 64 | 98 |
| Sum | 28 | 56 | 111 | 195 |

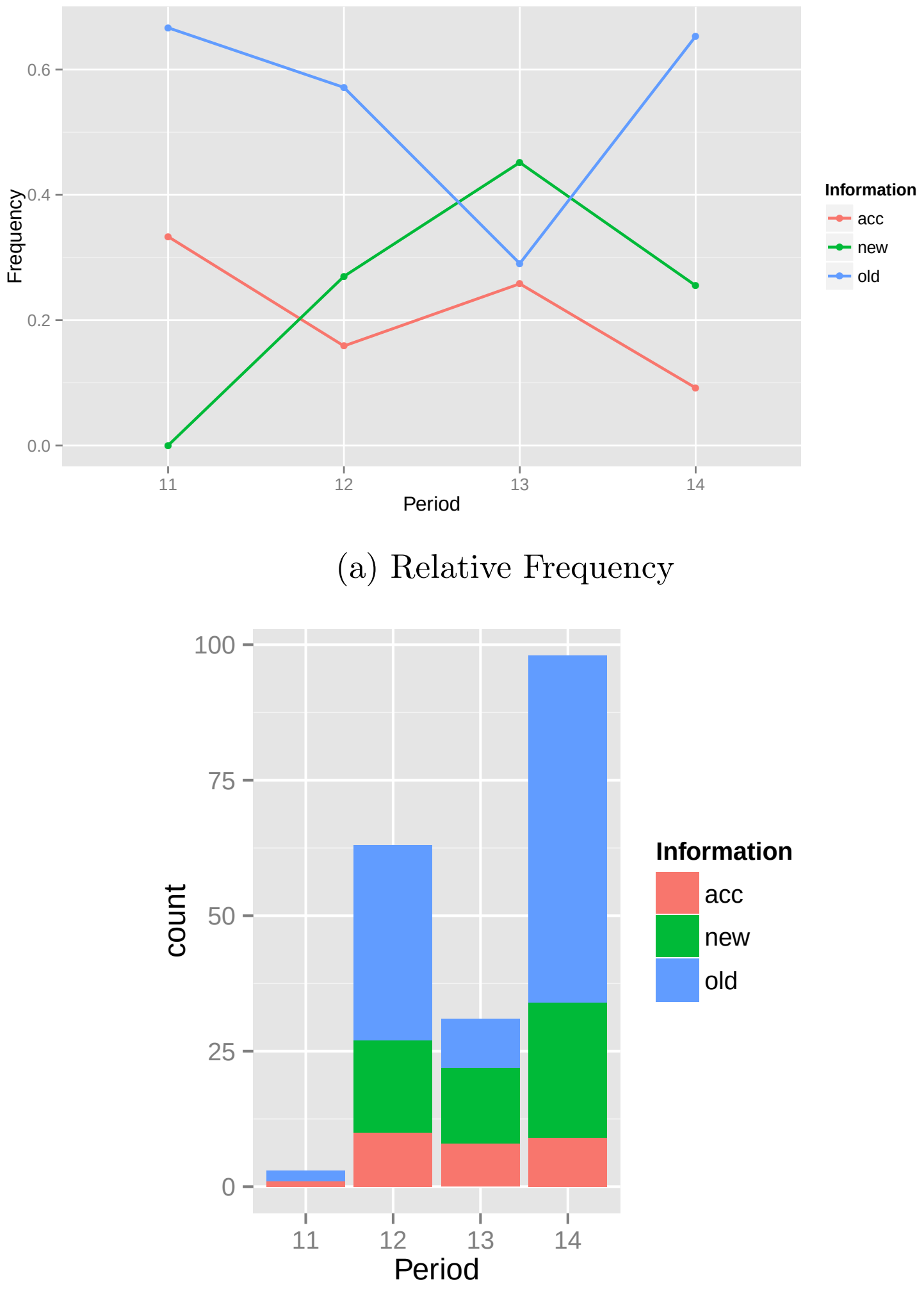

(a) Relative Frequency

(b) Frequency

Figure 7.16: Information Structure in Prepositional Verbs in Old French: Preverbal Nouns

Figure 7.17 and table 7.10 examine postverbal nouns and information structure in prepositional infinitives. In a similar fashion, we find a similarity between postverbal nouns in prepositional clauses (Figure 7.16) and overall distribution of postverbal nouns (Figure 7.11), namely a relative stability and small variance of information status categories in postverbal nouns in the 12th, 13th and 14th centuries. Finally, the small variance between information structure categories in this group suggests that postverbal word order weakens its sensitivity to pragmatic values. It should be noted that the 11th century is represented by 6 tokens only, which makes it hard to make any assumptions about information structure in this period.

Table 7.10: Information Structure in Prepositional Verbs in Old French: Postverbal Nouns

| Period | Accessible | New | Old | Sum |
|---|---|---|---|---|
| 11th | 0 | 0 | 3 | 3 |
| 12th | 10 | 7 | 14 | 31 |
| 13th | 5 | 9 | 12 | 26 |
| 14th | 45 | 74 | 69 | 188 |
| Sum | 60 | 90 | 98 | 248 |

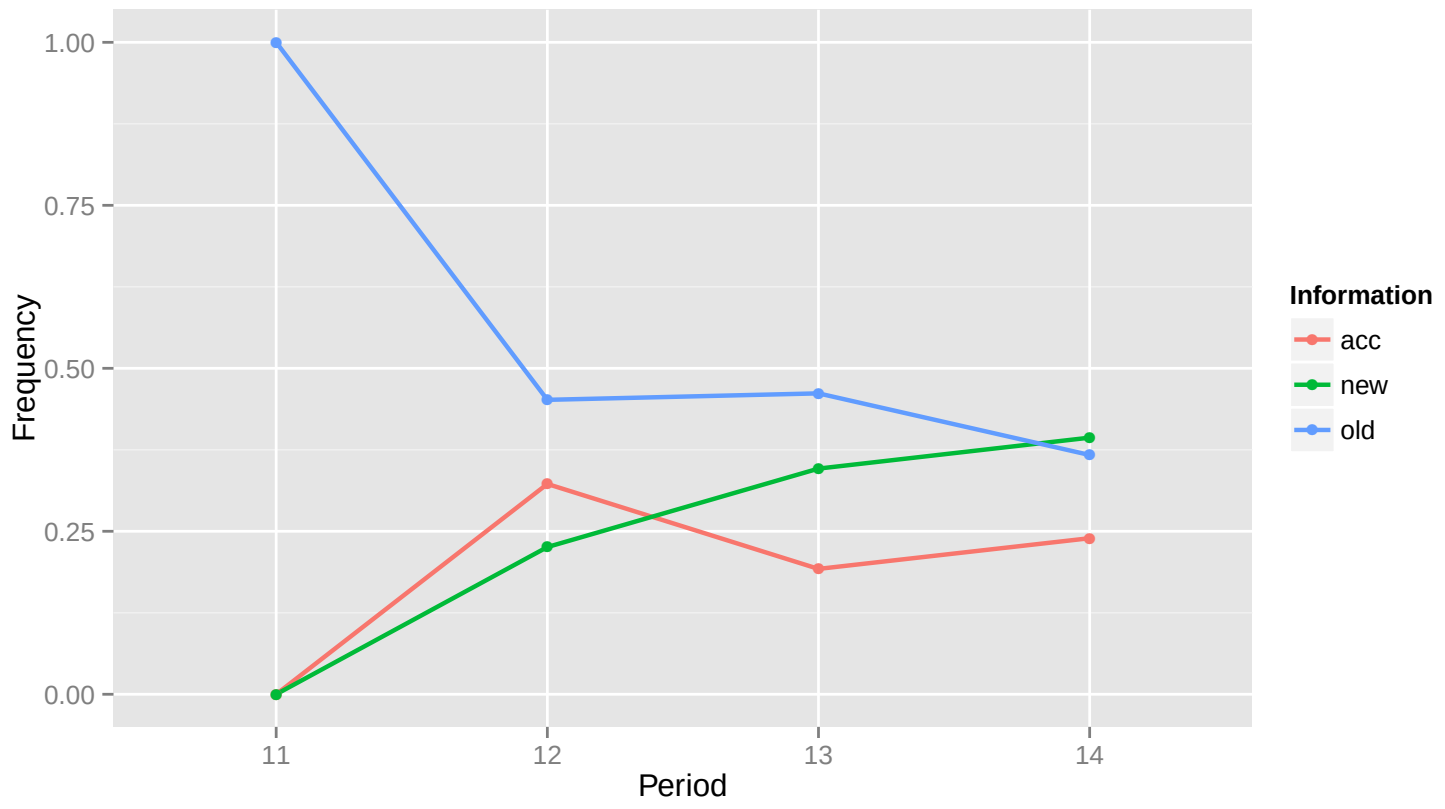

(a) Relative Frequency

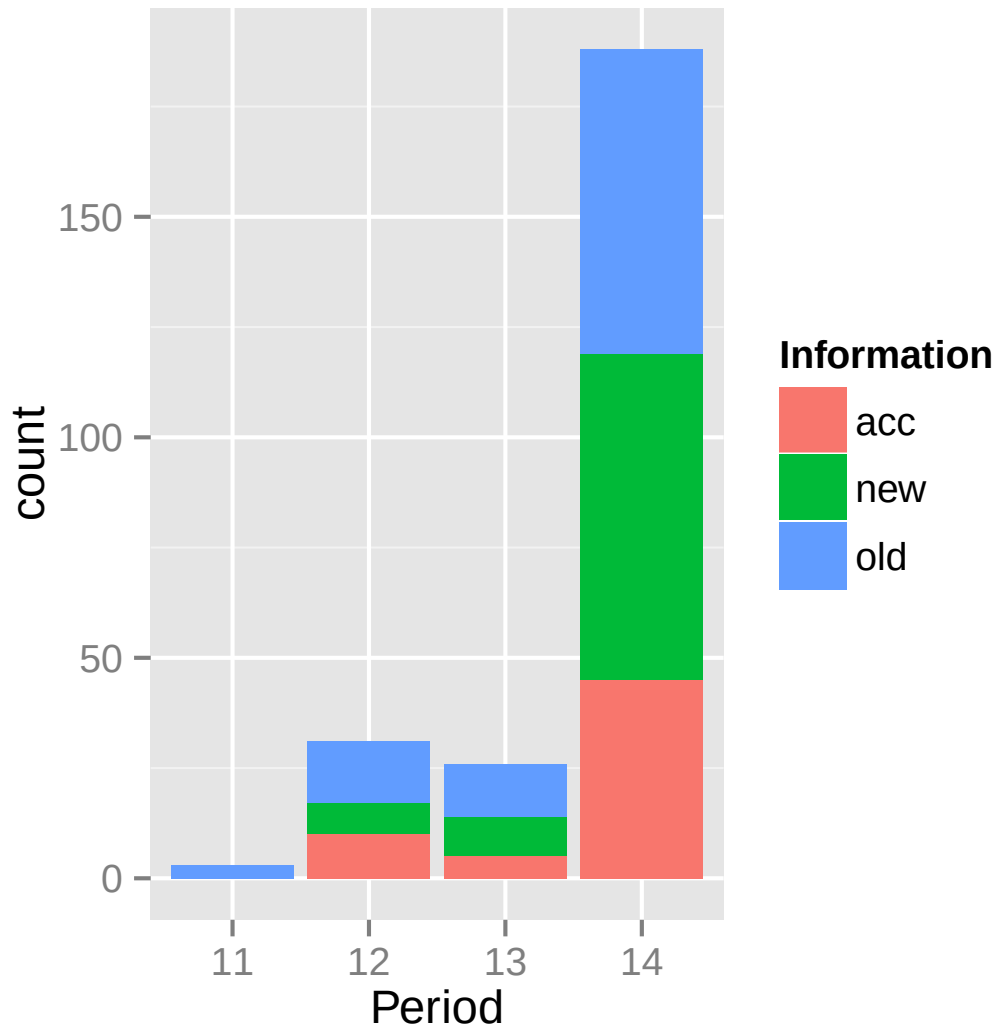

(b) Frequency

Figure 7.17: Information Structure in Prepositional Verbs in Old French: Postverbal Nouns

Table 7.11 and Figure 7.18 illustrate the distribution of information structure with OV order in restructuring verbs. Preverbal nouns demonstrate a dominance of *old information* and a wide variance of information structure categories, suggesting the sensitivity of preverbal nouns to pragmatic values.

Table 7.11: Information Structure in Restructuring Verbs in Old French: Preverbal Nouns

| Period | Accessible | New | Old | Sum |
|---|---|---|---|---|
| 11th | 4/(12) | 7/(21) | 22/(67) | 33 |
| 12th | 39/(20) | 75/(38) | 85/(42) | 199 |
| 13th | 2/(7) | 10/(39) | 14/(54) | 26 |
| 14th | 29/(29) | 28/(28) | 43/(43) | 100 |
| Sum | 74 | 120 | 164 | 358 |

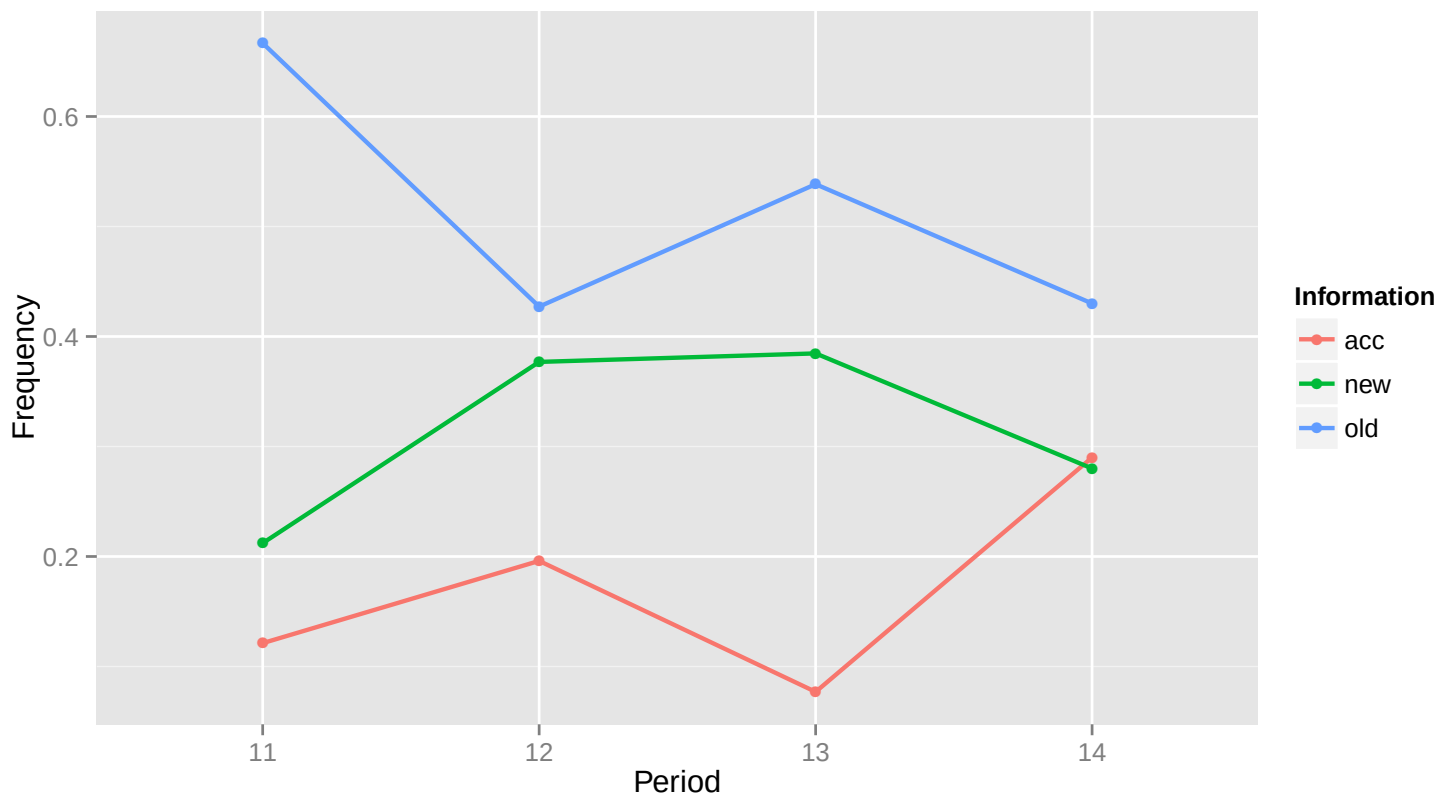

(a) Relative Frequency

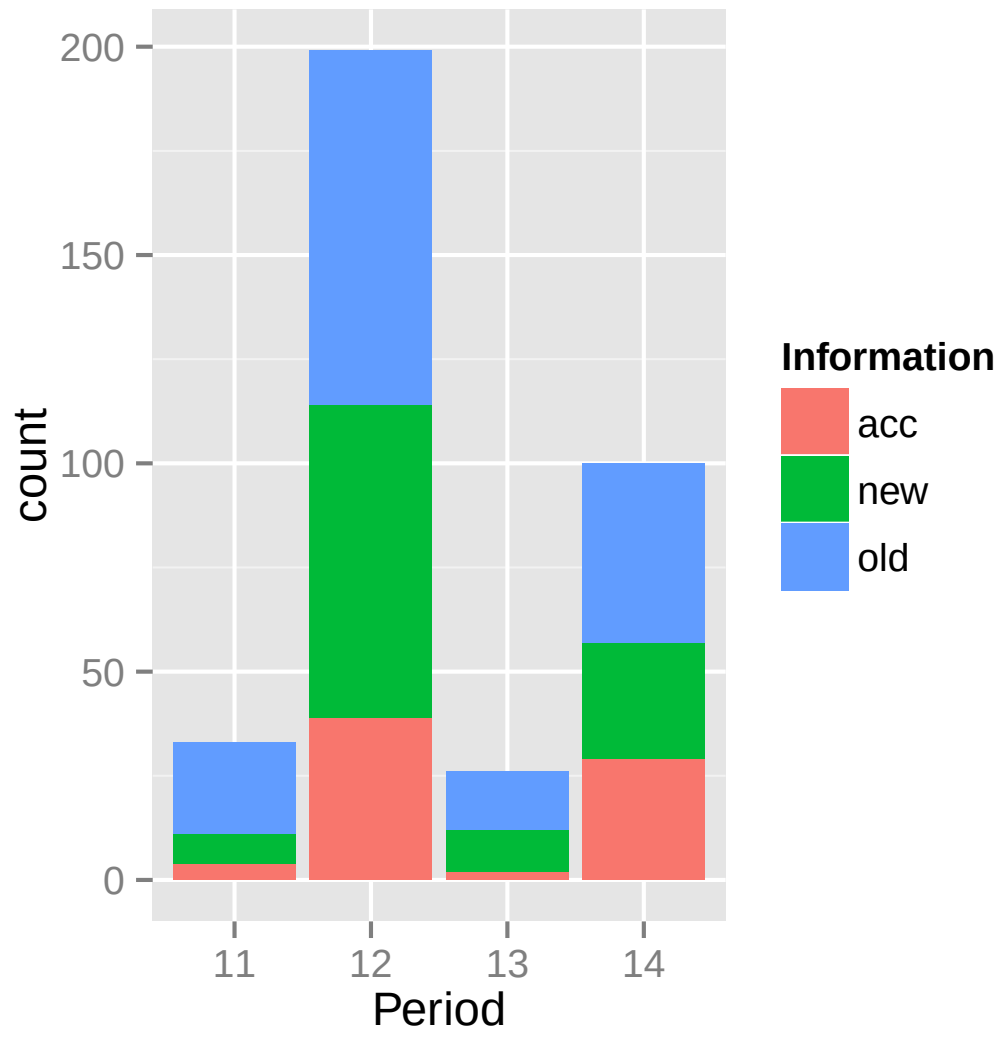

(b) Frequency

Figure 7.18: Information Structure in Restructuring Verbs in Old French: Preverbal Nouns

In contrast, in Figure 7.19 postverbal nouns show a slightly higher rate of *new information*. In addition, there is a relatively small variance between *old* and *new* information. This fact suggests that postverbal nouns are less sensitive to information structure categories than preverbal nouns. By comparing these two types of nouns, two similarities emerge: i) a large variance between information structure categories in preverbal nouns and the prevalence of *old information* and ii) a small variance between information structure categories in postverbal nouns.

Table 7.12: Information Structure in Restructuring Verbs in Old French: Postverbal Nouns

| Period | Accessible | New | Old | Sum |
|---|---|---|---|---|
| 11th | 2/(7) | 12/(43) | 14/(50) | 28 |
| 12th | 48/(23) | 89/(42) | 74/(35) | 211 |
| 13th | 28/(34) | 29/(35) | 26/(31) | 83 |
| 14th | 104/(31) | 122/(37) | 107/(32) | 333 |
| Sum | 182 | 252 | 221 | 655 |

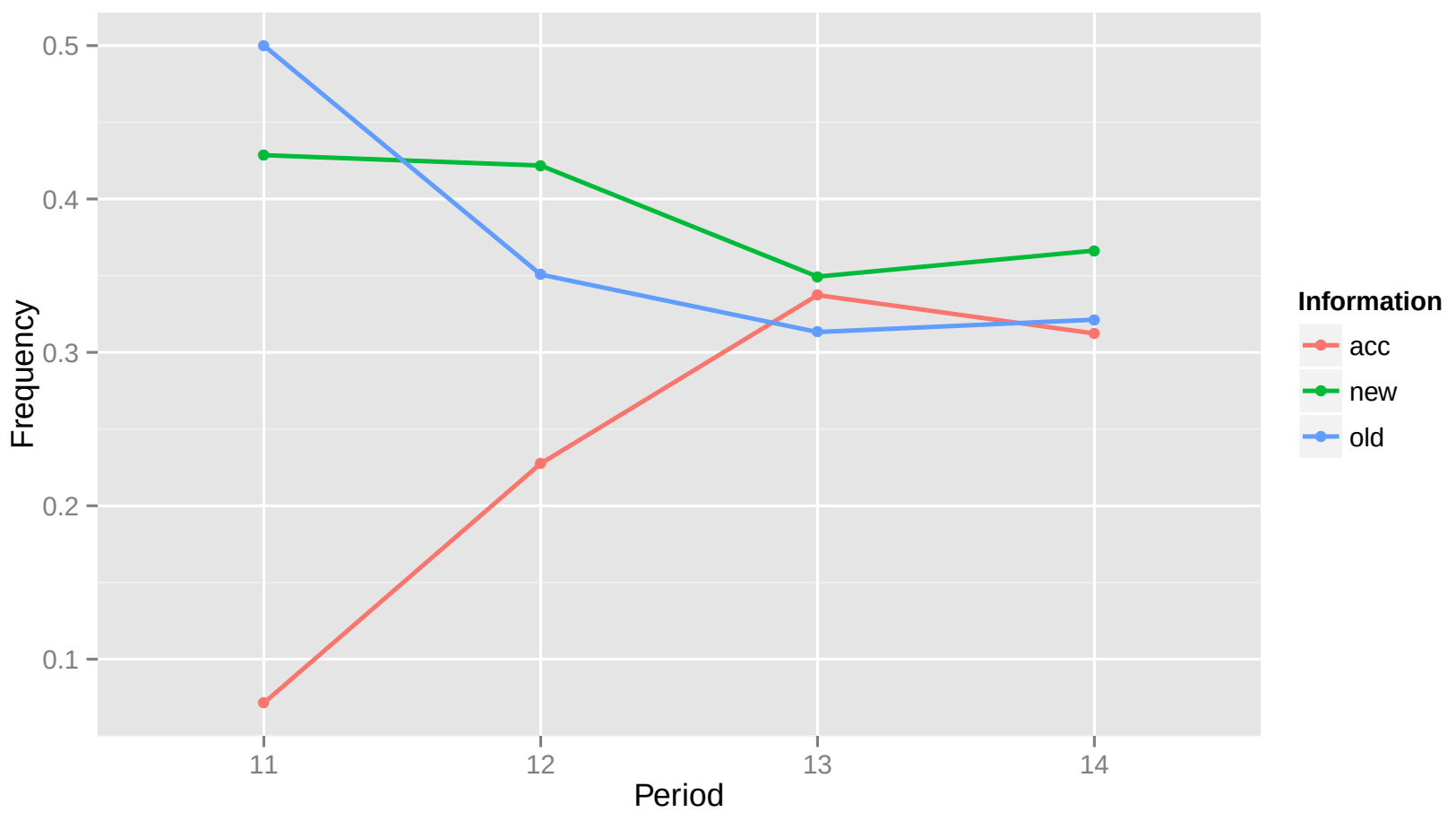

(a) Relative Frequency

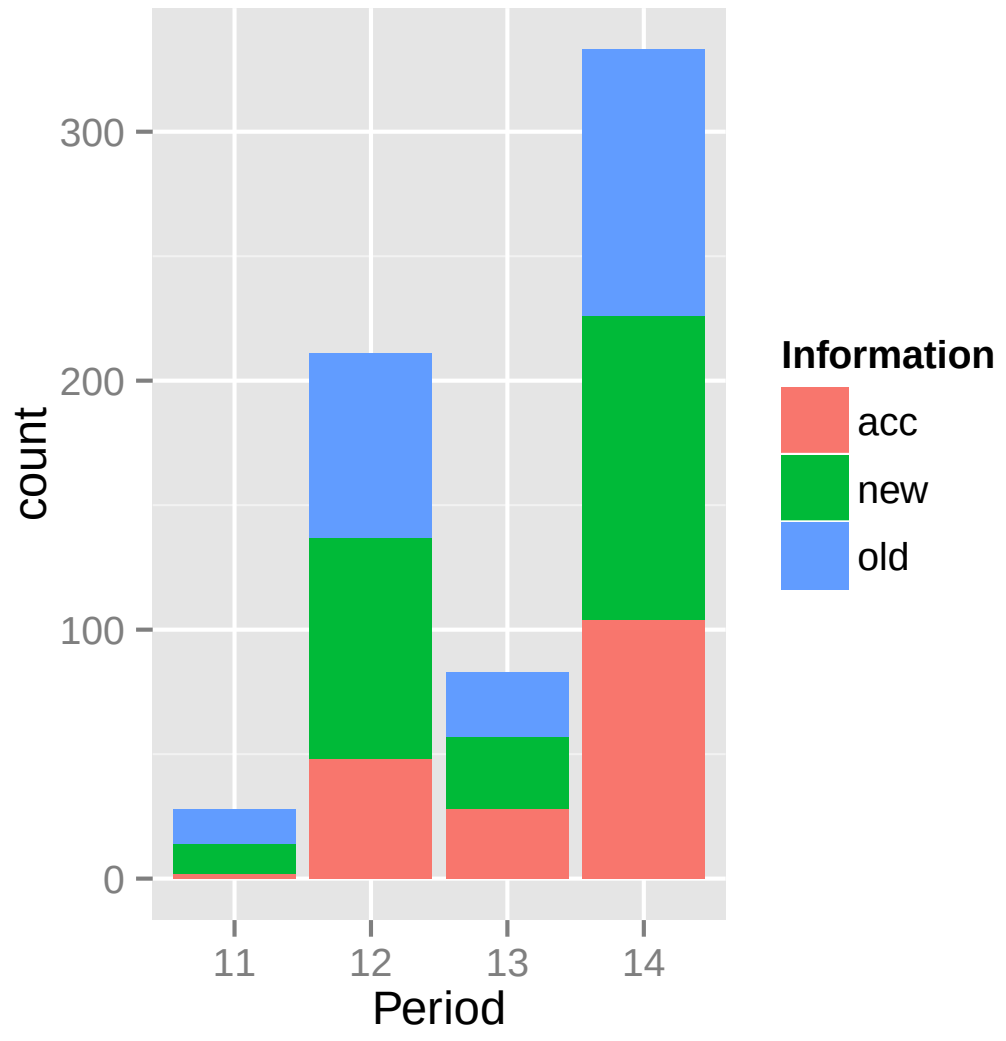

(b) Frequency

Figure 7.19: Information Structure in Restructuring Verbs in Old French: Postverbal Nouns

Finally, let us examine the rate of change for each category. The summary is presented in Figure 7.20. The highest slope value (1.8) and the smallest intercept value (0.0003) are observed with prepositional infinitives. Restructuring verbs show the intercept (0.3) with a slope of 1.5. A comparison between the two models, namely Latin and Old French, will be made in chapter 8; however, it is worth noticing that the *Restructuring* group is the only type that actually preserves the same intercept (0.3) and the same slope (1.5) in Latin and Old French. While there is a small deviation between prepositional and restructuring verbs in the Old French data, this difference is not significant ($p - value = 0.7342$), suggesting that the rate fluctuation reflects some contextual influences, as explained earlier in section 6.1.1. That is, the frequency of VO order increases at the same rate in prepositional and restructuring verbs, suggesting that the process of word order change is the same in these contexts.

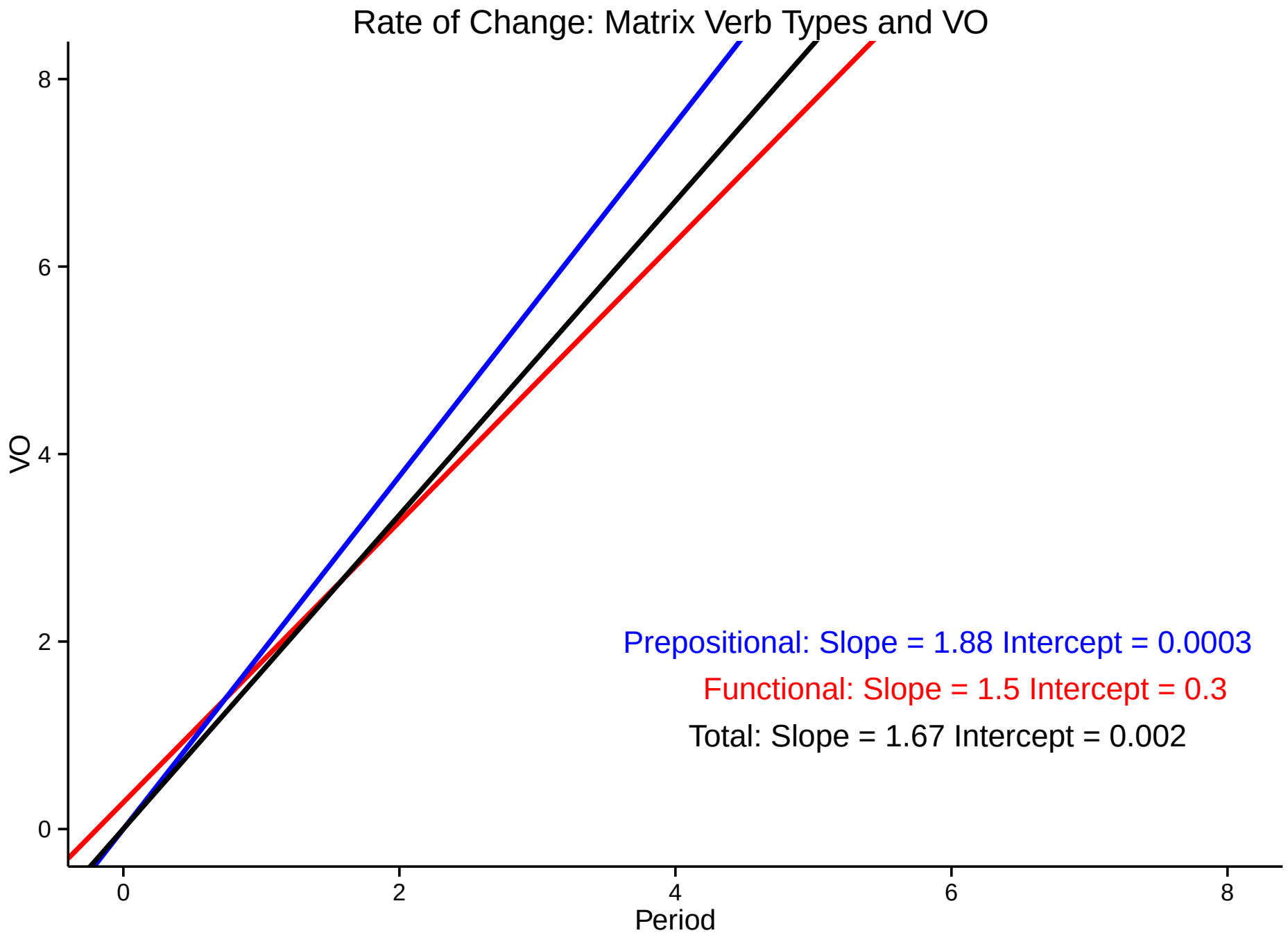

Figure 7.20: VO Rate in Infinitival Clauses across Different Types of Matrix Verbs: Old French

## 7.2 Multi-Factorial Analysis

This section examines the interplay of various factors and their significance on word order variation. First, let us look at the relative importance of all predictors.[7] This analysis is performed with *Random Forest* using the R package *partykit* (see section 5.3). Random Forest is a statistically robust technique that makes it possible to visualize the combined effect of factor groups and see which factors are statistically more important (see section

[7] *Frequency* is excluded due to allocation memory issues. This factor requires a lot of memory, and Random Forest analysis is very memory consuming in its own right.

5.3). The model in Figure 7.21 includes all linguistic factors, and they are plotted according to their importance. Note that *heaviness* is by far the most significant predictor for word order alternation. Indeed, it is traditionally argued that in Old French heavy constituents contribute to VO order. Consequently, the following significant factors are important, listed by order of their importance: *Information*, *Rhyme*, *Period*, *Position*, *Verb Type*, *Information Relevance*, *Animacy*. In the previous section, we observed that position of infinitive and verb types show differences in OV/VO distribution (see Table 7.5 and Figure 7.15). These observations are confirmed here by the statistical test. Furthermore, we see that information structure plays an important role in the variation, specifically information status, namely old and new information. On the other hand, *Split*, *Theme*, *Intervening Constituents* and *Genre* do not seem to play any significant role in this alternation. Since heaviness is the predominant factor, it is more likely that heaviness will hinder the relationship between word order and other factors. Thus, it is necessary to eliminate heavy constituents from our data. In the consequent analyses, the dataset is reduced to light nouns.

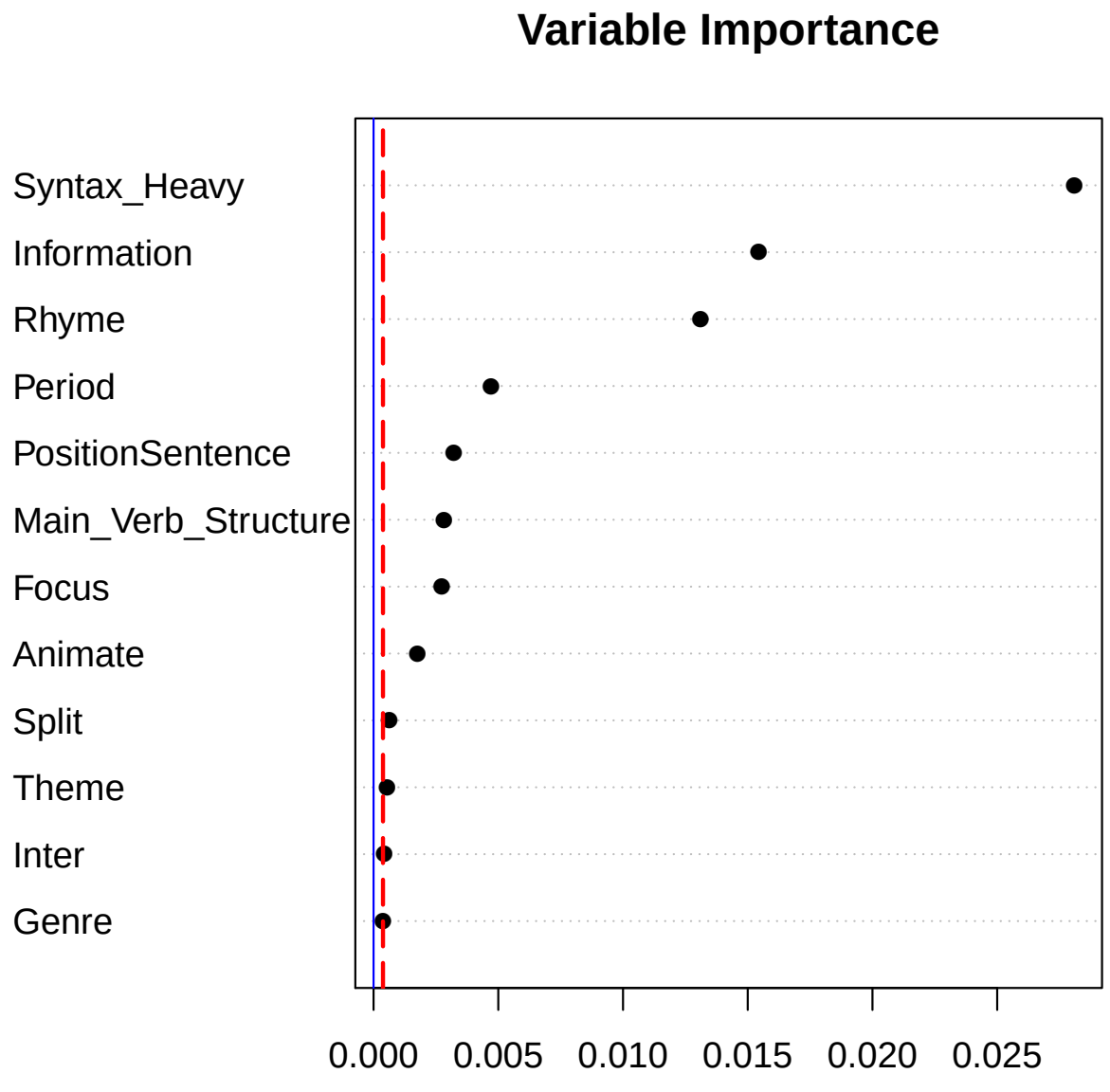

Figure 7.21: Random Forest Analysis - Old French

The Random forest analysis allows us to see how all the factors interact with word order alternation; however, it does not provide a detailed view on how each factor influences OV/VO alternation. For example, we know that *Information* is relevant; however, there is no indication of which type of information favors VO order, e.g. *new*, *old*, *accessible.* In order to examine each factor, the conditional tree analysis is applied (see section 5.3). Each factor group in the conditional tree is represented by an oval (node). The higher the node is on the tree, the stronger is its association with OV/VO alternation. The branches from each node represent the split between its values. Finally, on the bottom, we have the proportions of OV and VO orders. Plotting all the factors at once will make the plot very complex for interpretation. Instead, I analyze factors by their groups, namely sociolinguistic, pragmatic and syntactic factors. The sociolinguistic group is presented in Figure 7.22. This group includes *Genre*, *Rhyme* and *Theme.* It turns out that the main sociolinguistic predictor for word order alternation in Old French is *rhyme.* While rhyme is traditionally assumed to play a part in word order linearization, it has not been proven statistically to my knowledge until the present study. Metric verses further split by the *Theme* factor where a higher VO rate is shown by the *historical* theme. With respect to prose, it is differentiated by *genre*, showing the highest rate for VO with *narrative*, *speech* and *biography* in comparison to *treatise.* Notice that *theme* does not play any significant role.

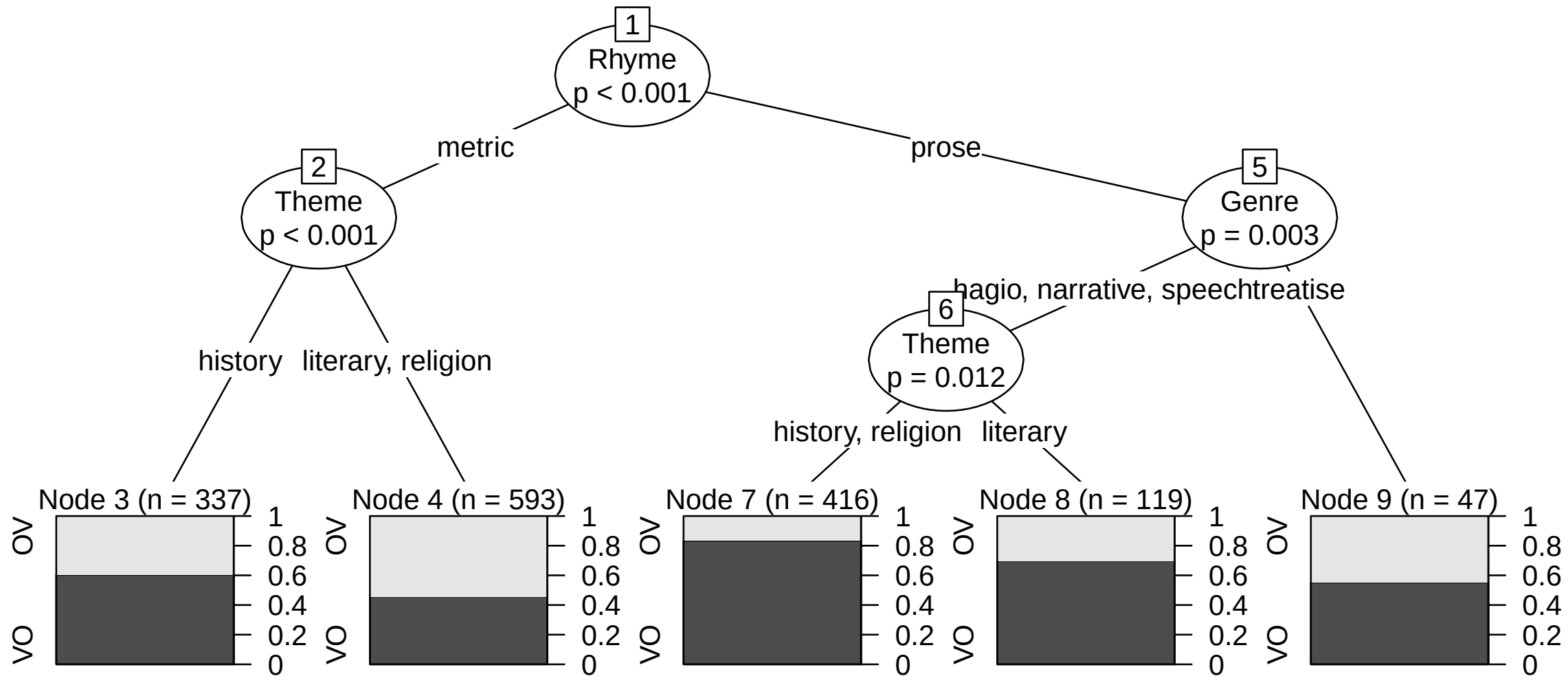

Figure 7.22: Sociolinguistic Factors - Old French

Syntactic factors are shown in Figure 7.23. This model depicts the ranking of syntactic factors. Once more, data are differentiated by *period*, with the 12th century as a boundary line. Before the 13th century, only *restructuring* verbs display a higher VO order than the remaining types. From the 13th century the picture seems to change. At this point, *other* verbs are joined with *restructuring* and show a very high VO rate. In contrast, *prepositional* verbs seem to diffuse VO order slowly; by the 14th century, they still have not reached the level of *restructuring* and *other* verbs. These results suggest that restructuring verbs are the most favorable contexts for VO form in Early Old French, as VO frequency is higher with these verbs.

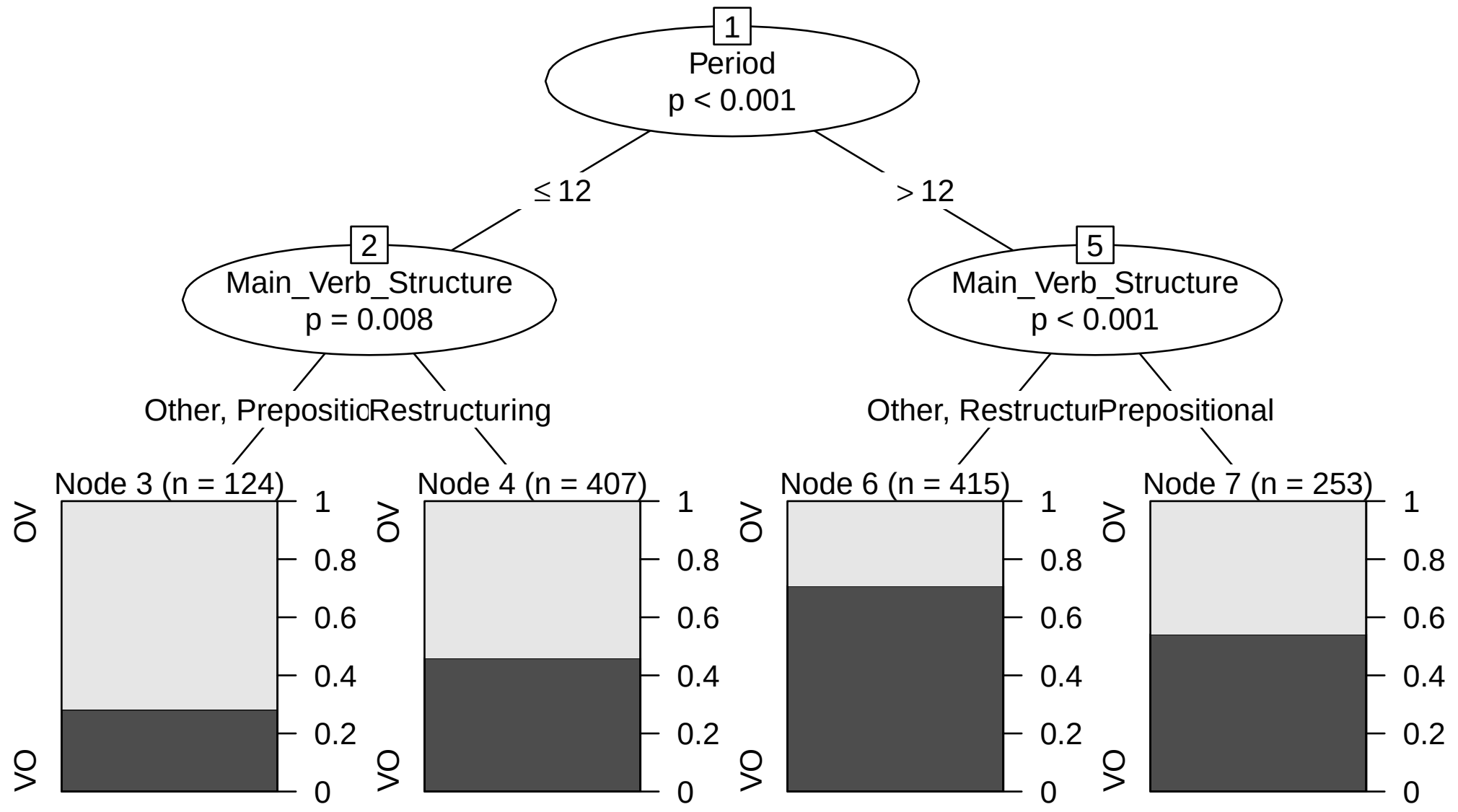

Figure 7.23: Syntactic Factors - Old French

Pragmatic factors are illustrated in Figure 7.24. Similar to syntactic factors, *period* is by far the most important predictor in Old French. The split follows the same boundary line as identified for syntactic and sociolinguistic factors. In the 12th century and earlier, *information* is a better predictor for the data. *Accessible* and *new* information show a higher VO rate than *old* information. In contrast, after the 12th century, the best predictor is *focus*. While the OV rate is higher with *contrastive* focus, *non-contrastive* focus is further differentiated by *information*. Surprisingly, *old* information shows almost the same rate of VO as *accessible* and *new* information in the 13th century. In fact, these results suggest that from the 13th century information status discontinues its role as a trigger for OV order. In contrast, focus is still an active player in OV order. These findings also support the results of Zaring's (2010) and Marchello-Nizia's (1995) studies, which demonstrate that the range of pragmatic functions on preverbal objects decreases by the 13th century. Another interesting detail is with respect to the 11th and 12th centuries. The question is whether the basic word order in Early Old French is VO or OV. Under the earlier assumption that new information is a key element for identifying unmarked word order, it seems that Old French is somewhere in between. The OV rate is high with new and *accessible* as well as *old* information. While there is clear indication that VO is dominant with *new* information in Late Latin (see Figure 6.23), Old French does not seem to fit the same VO structure. On the other hand, it is often argued that Old French is better described as a Verb Second (V2) language. We have seen earlier that this stage is considered a transitional TVX stage, where the initial position is often hosted by focalized and topicalized constituents. The results in section 7.1.1 showed that our data display a wide range of pragmatical values on preverbal nouns. Thus, this fact suggests that the data in the 11th-12th centuries are better described as a V2 period, based on the information structure distribution.

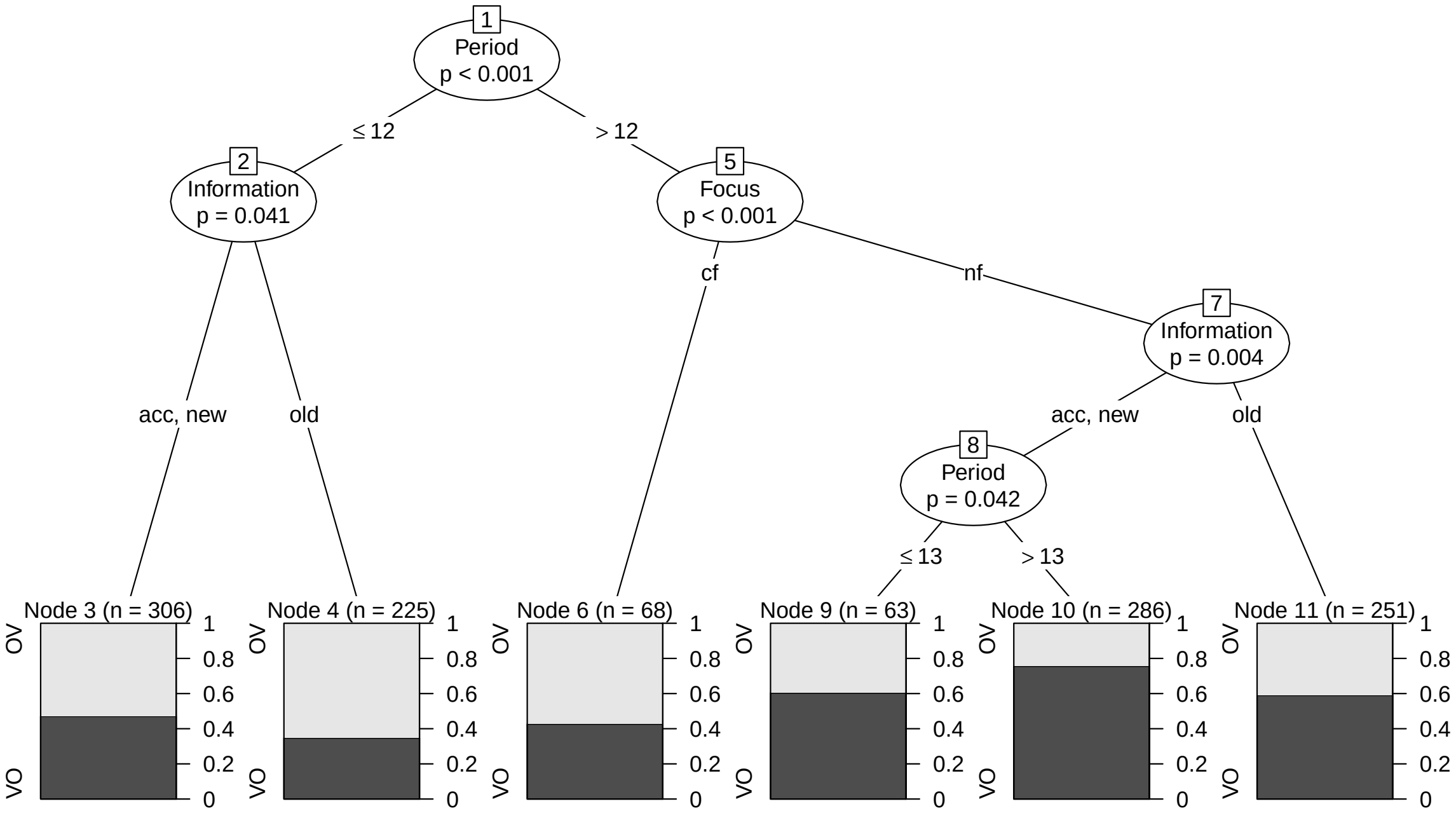

Figure 7.24: Pragmatic Factors with Light Constituents - Old French

Following the same method described in the previous chapter (see section 6.2) I apply fixed and mixed model to the variables. Based on the results from random and conditional trees, the following hypotheses will be tested: i) the effect of contrastive focus on preverbal NPs, ii) the effect of old information on preverbal NPs and iii) the influence of restructuring verbs on VO. Since heaviness is shown to be the strongest predictor for VO order in a random forest analysis (see Figure 7.21), only tokens that are less than three words will be considered for these models. The first model is illustrated in (95). The following reference values have been chosen as independent factors in (95): *Information Status* (old), *Information Relevance* (non-contrast), *Verb Type* (restructuring), *Position* (postposed).

(95)
```
OVVO[i] ~ dbern(mu[i])
mu[i]<-1/(1+exp(-(b0 + b1*Period[i] +
b2*InformationStatus[i] + b3*InformationRelevance[i] +
b4*VerbType[i] + b5*PositionSentence[i]
+b6*Frequency[i])))
```

Figure 7.25 demonstrates the posterior distribution. The non-credible factor is Main verb position,[8] with frequency as an additional factor. In contrast, the credible factors are 1) *Period*, 2) *Information status*, 3) *Position of infinitival clause*, 4) *Information relevance* and 5) *Frequency.* With each chronological increase, there is a statistically significant increase of VO. The second factor, *Information status*, provides statistical evidence that the probability of VO order increases when NPs carry new or accessible information. The third factor, *Position*, has negative values, that is, prepositional and preposed infinitives are more likely to predict OV order in comparison with postposed infinitives. The fourth factor, *Information relevance*, shows that VO is more likely with non-contrastive NPs. Finally, the positive values for the factor *Frequency* demonstrate that the more frequent the verb, the more likely it is for the word order to be VO.

[8]The factor is not credible when the zero value is a part of the HDI interval.

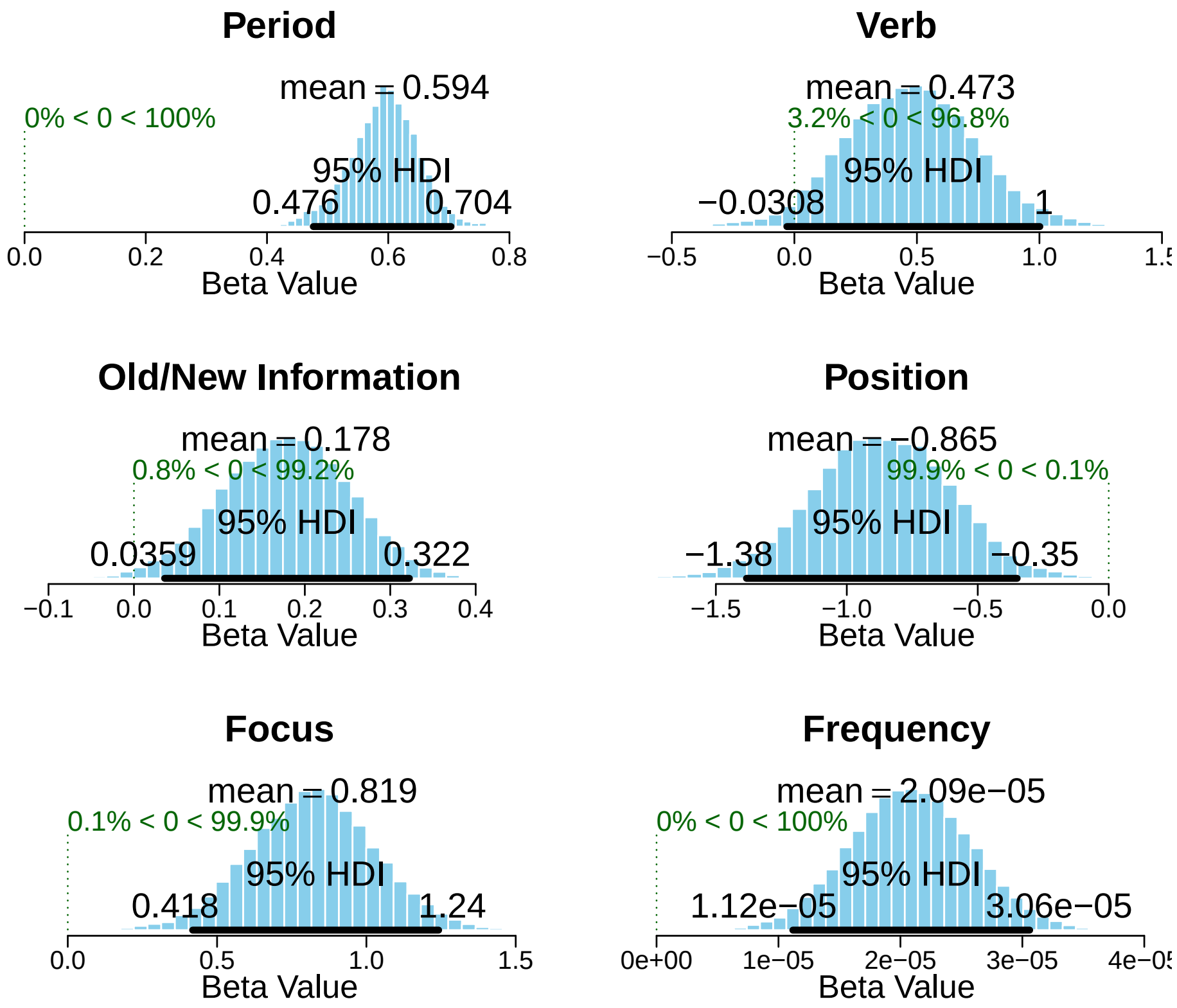

Figure 7.25: Posterior Distribution 1: Fixed Effect Old French Model

The model for mixed effect regression is presented in (96). This model includes authors and words, treated as random effects. This model also includes Rhyme and Animacy, as these factors might be clustered by individual authors.[9]

(96)
```
mu[i]<-1/(1+exp(-(b0 +  b1*Period[i] +
b2*InformationRelevance[i] + b3*InformationStatus[i] +
b4*PositionSentence[i]+ b5*Animacy[i]+b6*Rhyme
            + u[Author[i]]+y[Word[i]])))
```

Figure 7.26 reports the results from the mixed model. First, this model confirms the strongest predictors from the fixed model: 1) *Information status* (new/accessible), 2) *In-*

[9]Two factors are excluded from this model: i) Main verb structure - it was not significant in the fixed model and ii) Frequency.

*formation relevance* (non-contrastive) and 3) *Position* (postposed). In addition, the model shows that *Rhyme* is also significant, that is, non-metric prose is more likely to be VO. In contrast, *Animacy* is not a credible factor. However, from the posterior distribution (the central point), we can notice a negative tendency for non-human referents toward OV.[10]

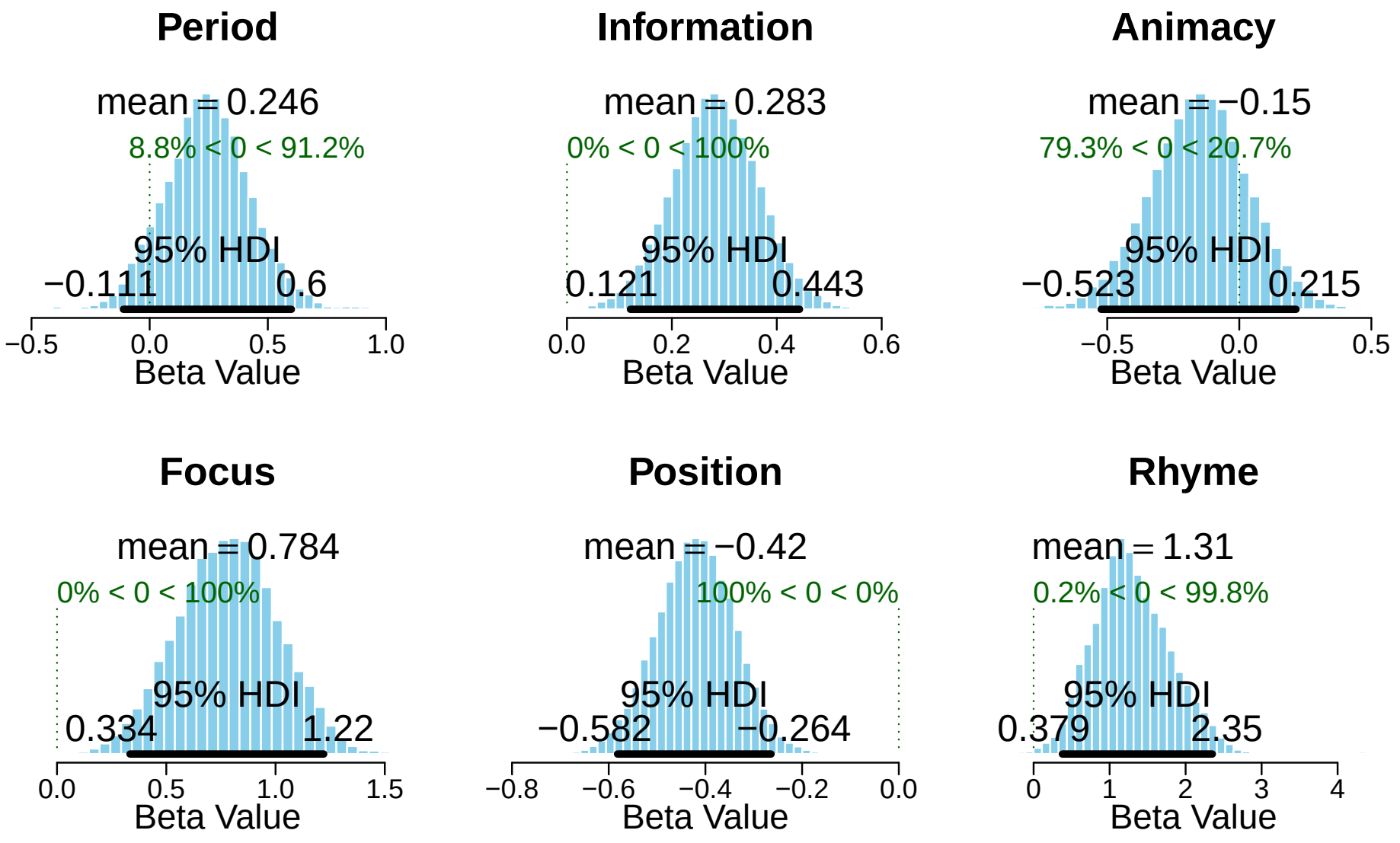

Figure 7.26: Posterior Distribution 2: Mixed Effect Old French Model

Thus, both models have confirmed the following hypotheses:

1. Postverbal NPs are more likely to carry non-contrastive new information focus

2. Postverbal NPs are more likely to carry new and accessible information

3. Postposed infinitives are a favorable condition for VO order diffusion

4. There is a frequency effect on VO distribution: The more frequent the verb is, the more likely it is to have VO order.

These facts show that pragmatic features remain stable across time, that is, contrastive focus and old information are features associated with a preverbal position and new/accessible and non-contrastive focus are associated with a postverbal position. My model, however, is not able to statistically confirm the effect of main verb types on word order distribution.

[10]Reference value is *human*.

## 7.3 Summary

In this chapter, I have examined word order distribution in Old French. I have shown that word order change in infinitival clauses mirrors a traditional pattern associated with Old French, namely the shift from OV to VO via OVX/XVO. Furthermore, I have identified the following chronological periodization of change: 11-12th **(X)O(X)V** → 13th **OVX/VXO** → 14th **VO(X)**. I have also noticed several parallel developments in the data: first, the emergence and subsequent increase of prepositional infinitives, which until the 14th century exhibit the dominance of OV order; second, the disappearance of preposed infinitival clauses, which almost exclusively show their preference for OV order and suggest their archaic use. Furthermore, I have shown that restructuring verbs are the most favorable context for VO order. The statistical models have further provided evidence that word order alternation is a result of information structure, that is, preverbal order is associated with old information and contrastive focus, whereas the postverbal position is predicted by non-contrastive focus and new or accessible information. These facts strongly suggest that Old French is not an OV language. However, given the stable OV/VO variation between the 10th and 12th centuries, these findings also suggest that Old French is not a fixed VO language until the 14th century. That is, Old French is at the intermediate stage, OVX/XVO.

# Chapter 8

# From Latin to Old French

*Change is not made up of independent adjustments,*
*but occurs as part of a system of interrelated changes.*
*(Tummers et al., 2005, 236)*

In the previous chapters Latin and Old French were analyzed as two separate units, reflecting their individual character in processes influencing word order change. This chapter will incorporate what was learned in chapter 6 (Latin) and chapter 7 (Old French) and compare results with previous analyses (see chapter 2). The combined model will be further constructed to examine word order alternation as a continuum from Latin to Old French. The obtained results will make it possible to detect which factors remain stable in their VO predictions and which change across time. The stable constraints will indicate a synchronic alternation, whereas the unstable factors will present evidence for diachronic change.

## 8.1 Factor Comparison

First, I will review variables that are often used to account for word order alternation and for changes from OV to VO order. Table 8.1 provides an overview of these variables for Latin, and Table 8.2 reviews variables for Old French. The central column names variables, and the left and right columns represent values. Note that the values can be described as OV and VO or verb-final and verb-initial.

| OV/Verb-Final | Factor Groups | VO/Verb-Initial |
|---|---|---|
| Subordinate | ← **Type of Clause** → | Main and Imperative |
| | (Bolkestein, 1989; Skopeteas, 2011) | |
| | **Length** → | Lengthy Constituents |
| | (Haida, 1928; Bauer, 1995) | |
| Conservative Classical | ← **Register** → | Semi-Literate |
| | (Adams, 1976) | |
| | **Subject** → | Explicit |
| | (Bauer, 1995) | |
| | **Verb Type** → | Motion, Causative, Mental |
| | (Dettweiler, 1905; Bauer, 1995) | |
| Topicalized Object | ← **Topicalization** | |
| | (Knoth, 2006) | |
| Focus, Emphasis | ← **Focalization** → | Non-Presupposed |
| | (Pinkster, 1990; Knoth, 2006) | |
| | (Halla-aho, 2008; Kiss, 1998) | |
| | **Abstractness** → | Abstract, Non-Referen. NP |
| | (Devine and Stephens, 2006) | |
| Non-Animate NP | ← **Animacy** | |
| | (Devine and Stephens, 2006) | |
| Unmarked | ← **Markedness** | |
| | (Brugmann and Delbrück, 1900) | |

Table 8.1: Overview of Proposed Factors: Latin

| **OV/Verb-Final** | **Factor Groups** | **VO/Verb-Initial** |
|---|---|---|
| Subordinate | ← **Type of Clause** → | Main and Imperative |
| | (Rickard, 1962; Marchello-Nizia, 1995) | |
| | **Length** → | Lengthy Constituents |
| | (Pearce, 1990) | |
| | **Subject** → | Explicit |
| | (Marchello-Nizia, 1995) | |
| Non-Finite | ← **Verb Form** → | Finite |
| | (Zaring, 2010) | |
| Topicalized Object | ← **Topicalization** | |
| | (Adams, 1987; Marchello-Nizia, 1995) | |
| Focus | ← **Focalization** | |
| | (Adams, 1987; Marchello-Nizia, 1995) | |
| Verses | ← **Metrics** | |
| | (Rainsford et al., 2012) | |
| | **Markedness** → | Unmarked |
| | (Bauer, 1999) | |

Table 8.2: Overview of Proposed Factors: Old French

A comparison of variables in Latin and Old French by previous authors (Table 8.1 and Table 8.2) reveals several stable tendencies: lengthy constituents, topicalization and focalization cross-linguistically trigger OV order. In addition, subordinate clauses, explicit subjects and literate register or verse seem to be favorable contexts for OV in both languages. There is also a shift in the direction of prediction by markedness, namely OV in Latin and VO in Old French. The factors that are not cross-listed in both languages are animacy, abstractness, verb type and verb forms. Thus, in Latin OV is often found with non-animate NPs, whereas abstract or non-referential NPs are more likely to occur with VO order. Additionally, verbal semantics in Latin has some influence on word order, that is, VO is observed with causatives and verbs of motion and mental state. In contrast, in Old French a verbal form determines word order preferences: non-finite form predicts OV, and finite form is a preferred context for VO. However, it is unclear whether these four factors, namely verb semantics, verb form, animacy and abstractness, are language-specific: they have not been evaluated cross-linguistically to my knowledge. Furthermore, as Gries (2001) points out, this type of variable has several shortcomings. First of all, most variables are based on frequency analysis, without any statistical support. Second, some focus on verb position, e.g. verb-initial vs. verb-final, whereas others are relevant to OV and VO. Finally, these factors are traditionally analyzed in isolation. However, such a mono-factorial analysis fails to encompass all levels of language, namely semantics, pragmatics, syntax and sociolinguistics.

The next review is based on the multi-factorial analysis conducted in this study. Since there are various linguistic levels involved, the evaluation will be performed separately for each group, namely, sociolinguistic, syntactic, pragmatic and semantic groups as well as heaviness and frequency.

### 8.1.1 Social Factors

I will start by comparing the sociolinguistic factors assessed in this study. The factors include *period*, *genre*, *metrics* and *theme*. As strong evidence for word order change, both languages show a shift from OV to VO across time. Furthermore, the rate of change in both languages appears to be nearly the same (1.6 - 1.7), as illustrated in Figure 8.1. Although the initial value in Old French is smaller than it is in Latin, this could be attributed to specific qualities of Early Old French texts, which are sparse and predominately in verse.

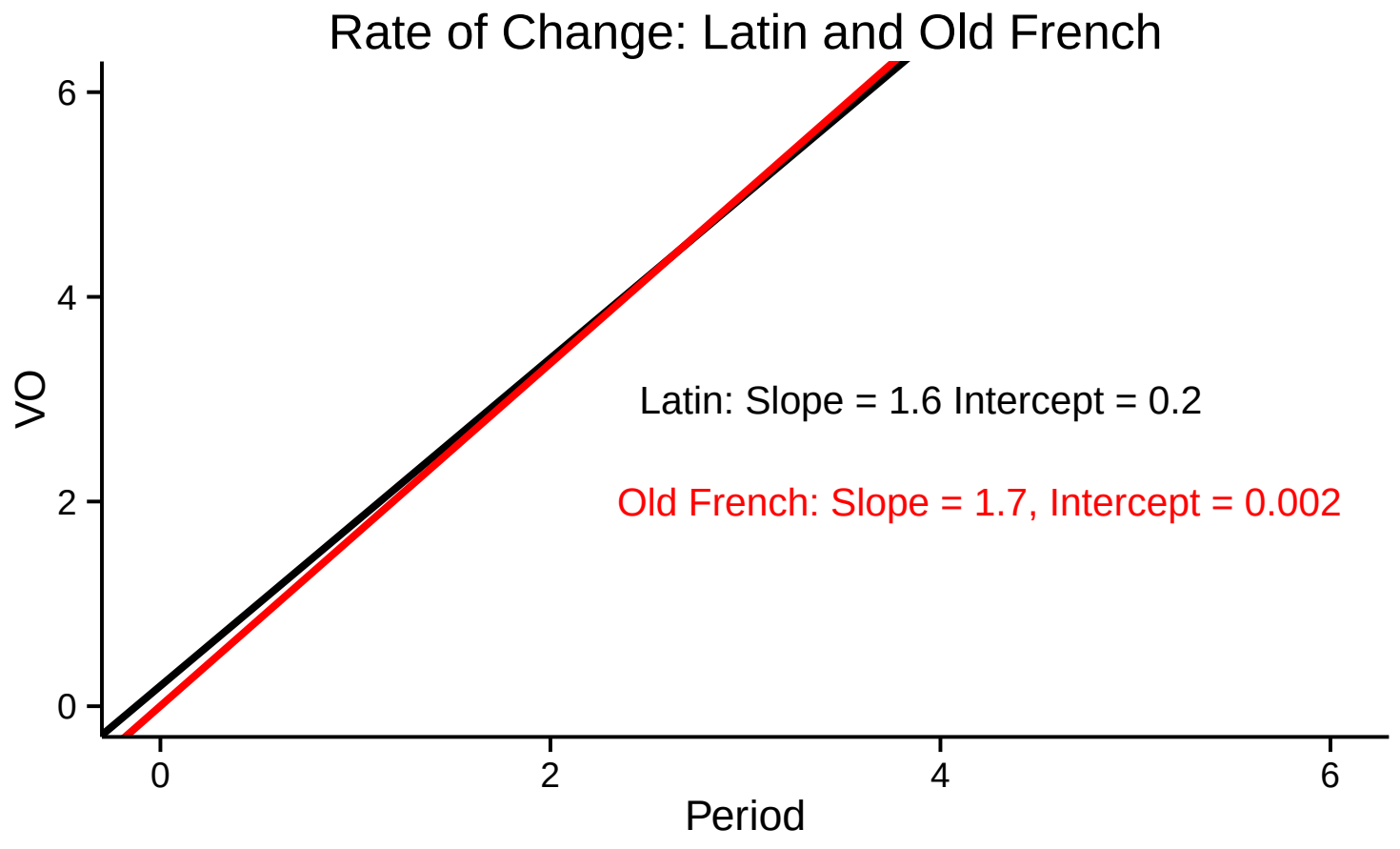

Figure 8.1: Comparison of Rate of Change for VO order in Latin and Old French Infinitival Clauses

Having compared the rate of change in both languages, the next evaluation examines the OV/VO distribution chronologically. A Conditional tree analysis statistically identifies the following periods to be significant ($p < 0.001$): Classical Latin, Late Imperial Latin, Late Latin and the 12th century in Old French. Figure 8.2 illustrates the following benchmarks and their proportions: i) Classical Latin - OV (80%), ii) Imperial Latin - OV (70%), iii) Early Late Latin - VO (56%) and Late Latin - VO (∼50%) and vi) 13th century - VO (80%).

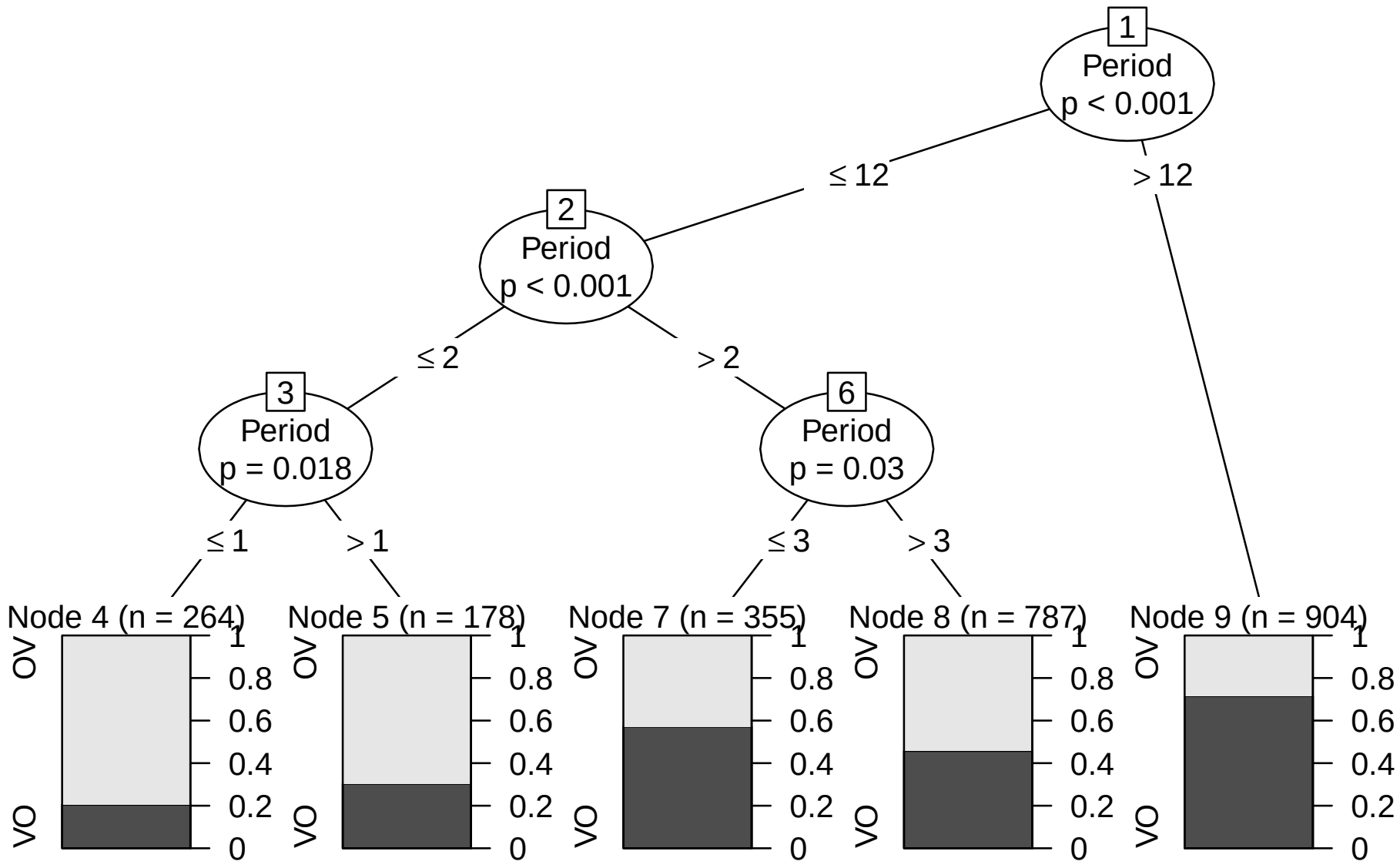

Figure 8.2: Chronological Benchmarks

The next social factors to compare are *genre*, *theme* and *metrics*. It has often been claimed that in Latin *genre* plays 'a decisive role in determining the degree to which OV was preferred' (Halla-aho, 2008, 122). Our results show that *theme* is the best predictor for Latin word order (see Figure 6.19 and Figure 8.3). Furthermore, section 6.2 showed that stylistic word order variation may affect texts written by the same author. For example, Jerome's personal letters (4th century) illustrate a ratio of 1.4:1 with a higher rate for OV, while his Vulgar Bible translation has a ratio of 1:2.6 with high VO rate. In Old French *metrics* is identified as the best predictor for word order. It is traditionally argued that that verse allows for more variation due to the metric demands of poetry (Hirschbühler, 1990; Marchello-Nizia, 1995). Indeed, the data show a great variation in word order in verse (∼50%). However, we also see that there is variation in *prose*, specifically in the *treatise* genre. Not only does this study confirm the importance of metrics, it also shows that other sociolinguistic factors play important role in word order variation.

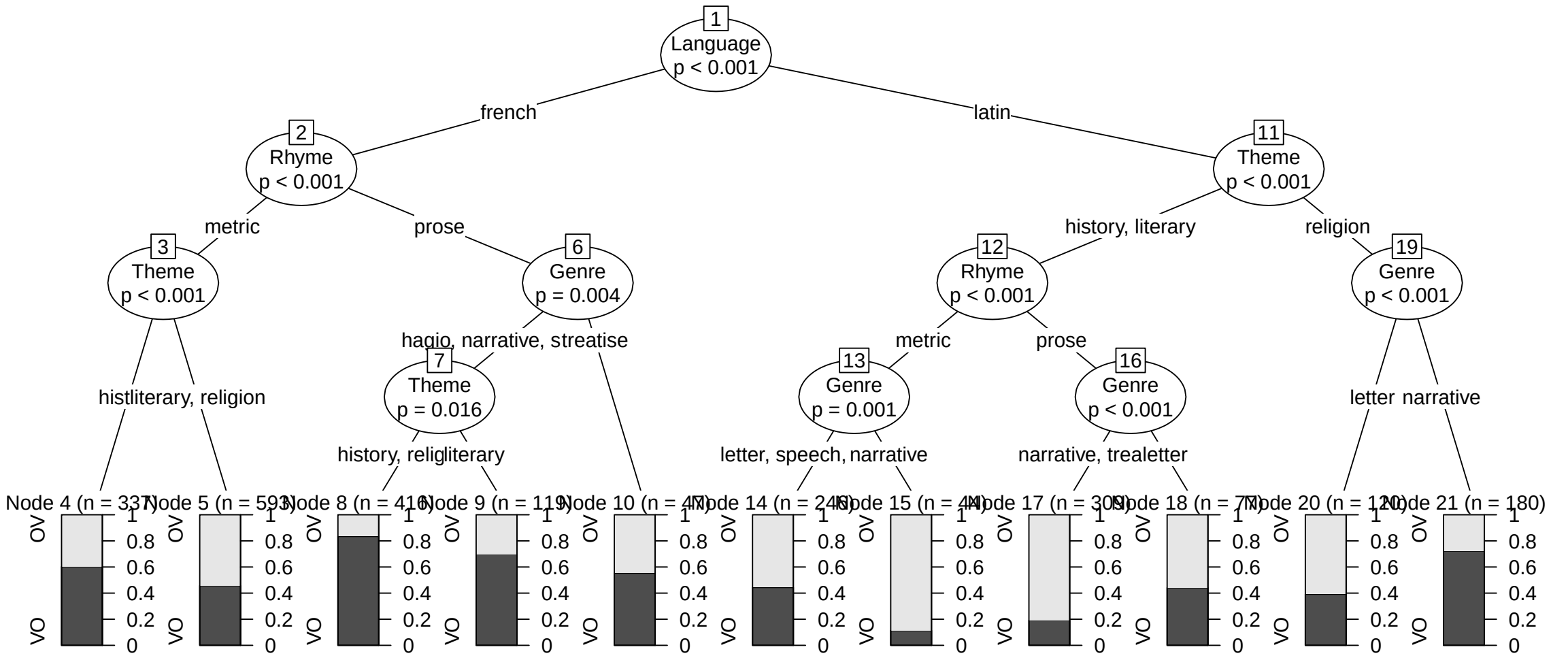

Figure 8.3: Sociolinguistic Factors in Latin and Old French Infinitival Clauses

### 8.1.2 Syntactic Factors

A cross-examination of word order patterns reveals the following chronological patterns of change:

(97) a. Classical Latin **SOV** → Late Latin **SVO**

b. Classical Latin **OV** → Late Latin **VO**

c. 11-12th **(X)O(X)V** → 13th **OVX/VXO** → 14th **VO(X)**

Based on these facts, I have argued that we have evidence to consider Late Latin as a VO language. On the other hand, Old French manifests strong evidence for VO only in the 14th century, with Early French as a transitional stage, namely V2. In fact, our data delay VO fixation until the 14th century, whereas Marchello-Nizia (2007) establishes the 13th century as the date for VO fixation. One may argue that these results should not be applied to the whole language system, as the present data represent only a specialized area of language, namely infinitival clauses. I provide here several reasons as to why this is not the case, supporting the extension of these results to language change in general. First, most studies examine main declarative sentences to determine word order change. It is well known that this type of clause presents various challenges to diachronic studies. For example, not having access to speakers and their prosody makes it difficult to determine whether a given pattern is a basic order or a deviation triggered by emphasis, focus or other constraints. While this is also true for infinitival clauses, their reduced character minimizes possible ambiguity. Second, most previous approaches are based on a simple frequency analysis that is subject to great variation based on the nature of historical data. In contrast, the present analysis is based on a probabilistic approach, that is, the results show the likelihood of a given pattern's association with a given period.

Let us turn now to the syntactic factors considered in this dissertation, namely *position of infinitives*, *structure of main verb*, *subject*, *split* and *intervening material*. First of all, *position* is identified as the best syntactic predictor for word order in both languages. More specifically, postposed infinitives are more likely to predict VO order. Furthermore, several ongoing changes have been observed with respect to the position of infinitives: i) the disappearance of independent infinitival clauses by the 6th century, ii) the increase in postposed infinitives and iii) the disappearance of preposed infinitives in Old French (Figure 6.16 and Figure 7.12). Curiously, the rate of VO change for each of these types is very different for Latin (Figure 6.17) but equal for Old French (Figure 7.13). As discussed earlier, this fact does not contradict the Constant Rate Hypothesis. In contrast, the disharmony in the rate demonstrates that these structures are highly influenced by some external contexts. Indeed, the highest slope (2.8) is found with independent infinitives that are often considered full-fledged sentences, allowing for a full subject and a tense. Thus, this clause is more likely to diverge from the basic order. Another interesting observation can be made with respect to preposed infinitives in Old French. It appears that this type almost exclusively demonstrates OV order, suggesting its archaic use.

The second predictor for word order is type of main verb. The results show that infinitival structures are not homogeneous and that word order alternation can be predicted according to infinitival type. Restructuring verbs have, by far, the highest frequency rate for VO as compared to other verbal categories (see Figure 8.4). On the other hand, if we look at the distribution of OV order, it remains dominant until the 13th century even among restructuring verbs (see Figure 8.5).

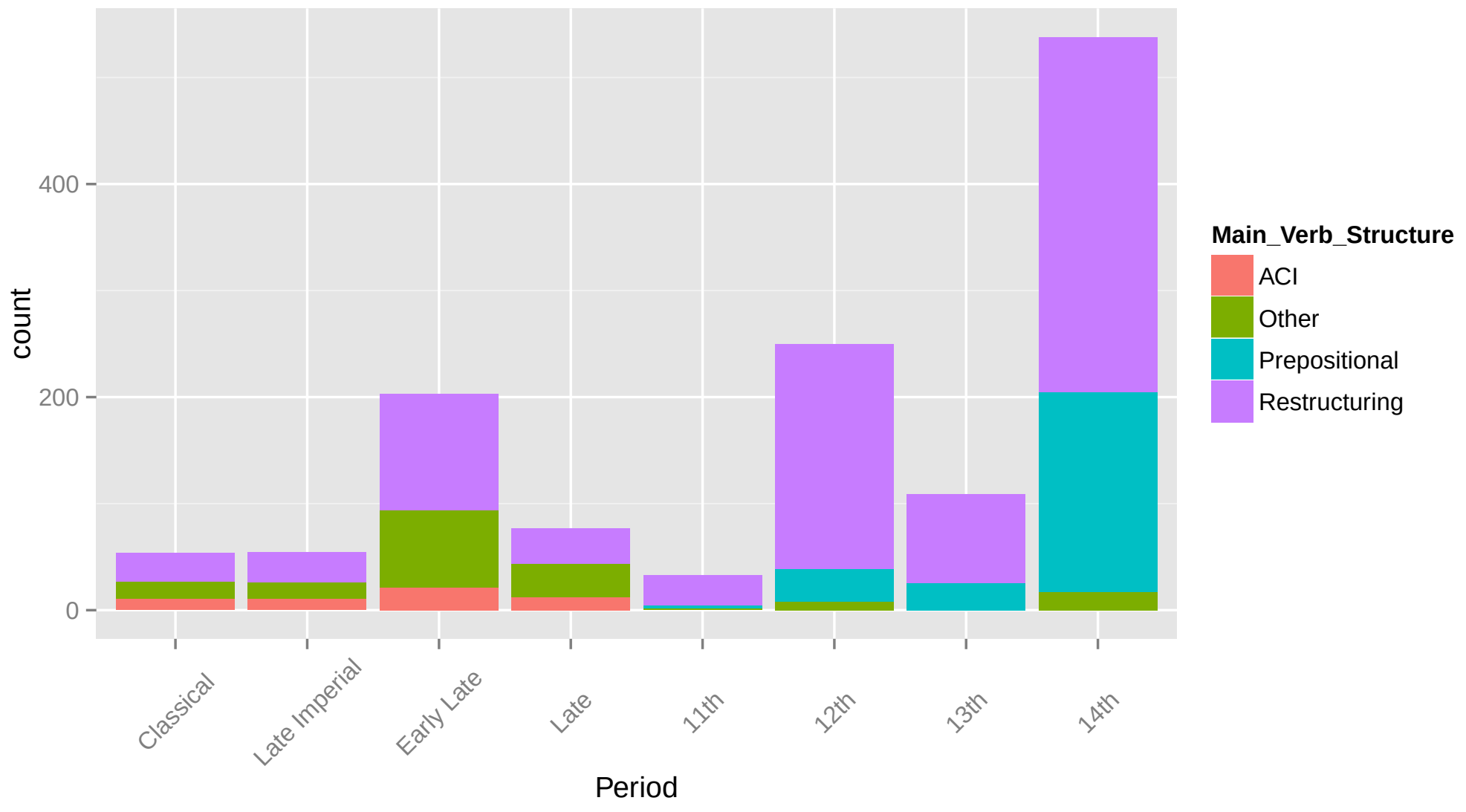

Figure 8.4: VO Order and Main Verb Categories in Infinitival Clauses in Latin and Old French

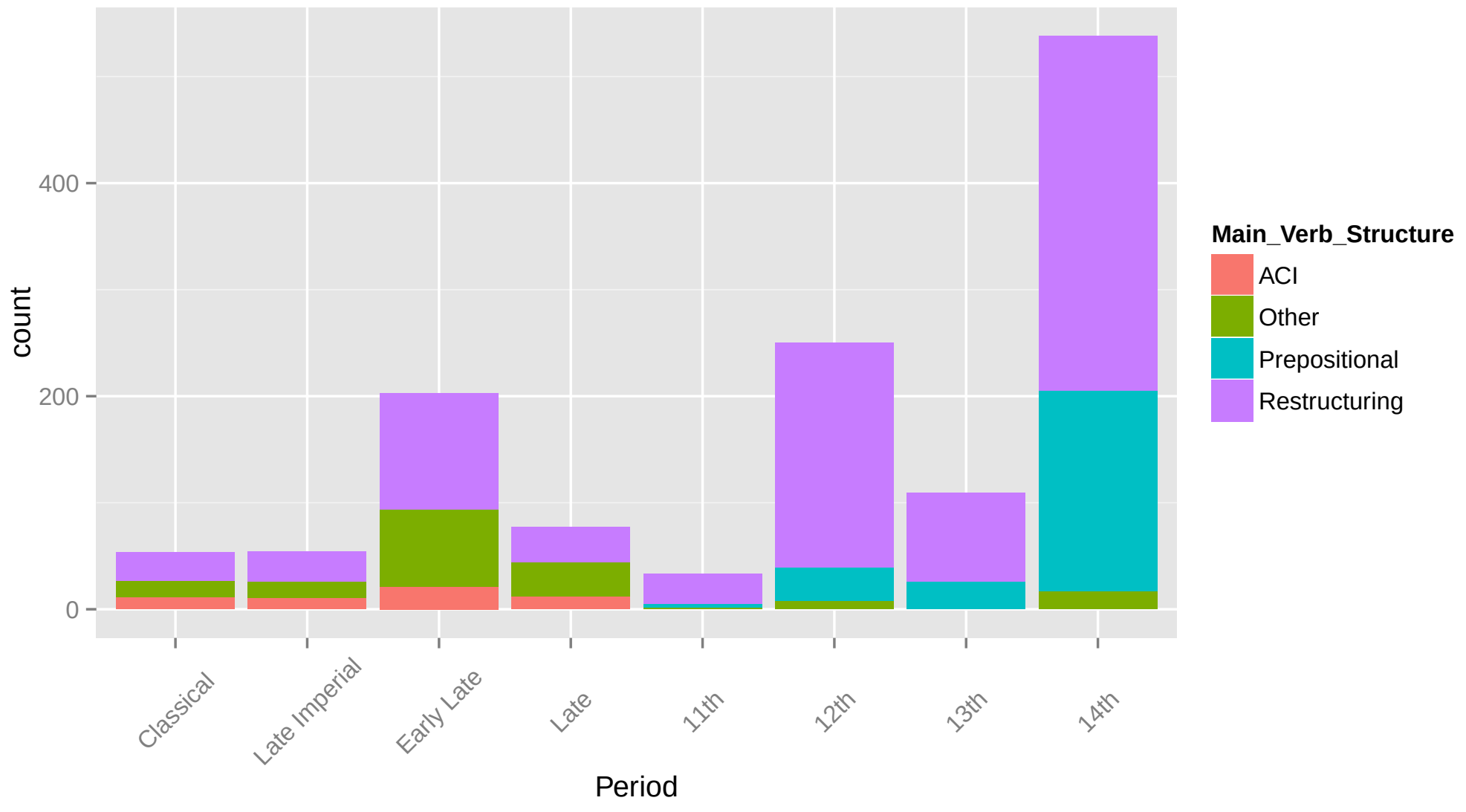

Figure 8.5: OV Order and Main Verb Categories in Infinitival Clauses in Latin and Old French

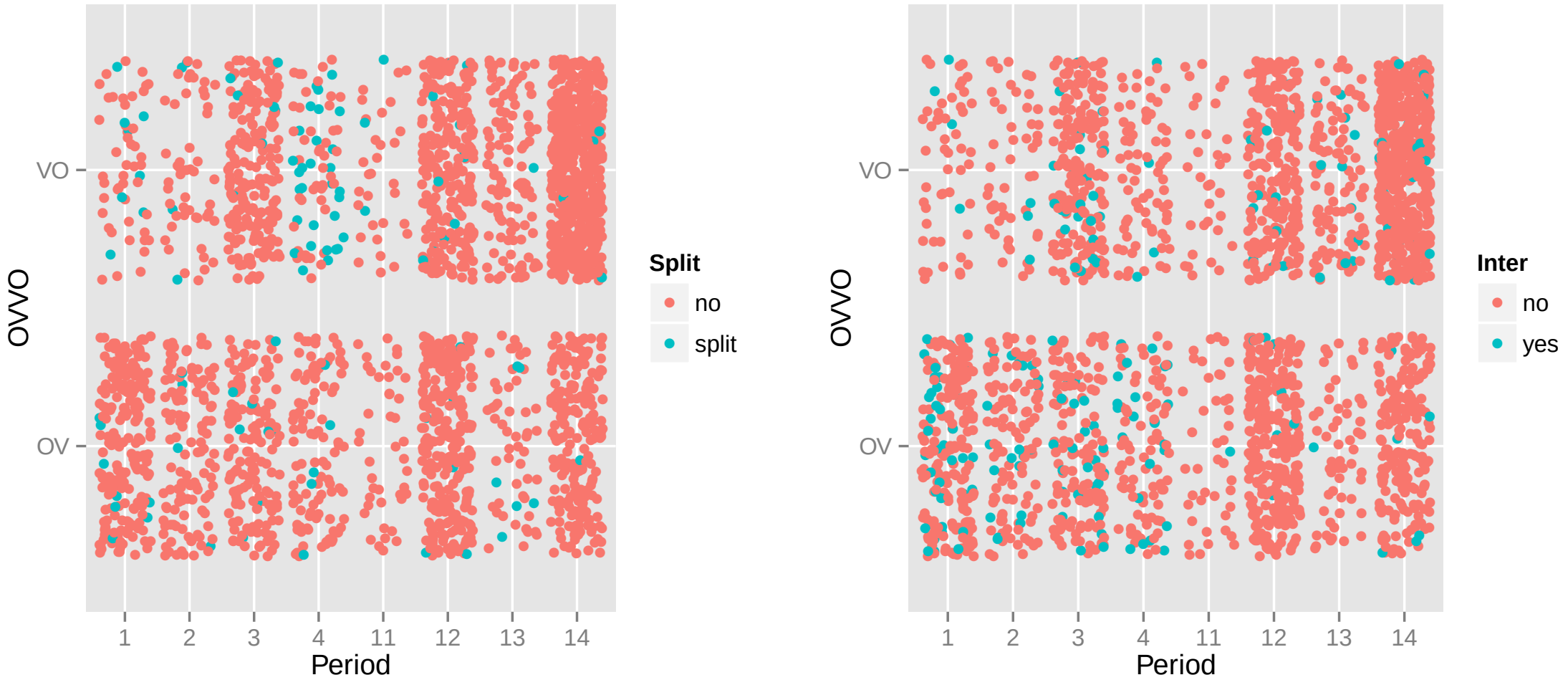

(a) Factor: Split (b) Factor: Intervening Material

Figure 8.6: Split and Intervening Material: Latin and Old French

Two important structural changes are further observed in infinitival clauses: i) the decrease of AcI by the 6th century and ii) the emergence of prepositional clauses in Old French. It is possible that these two changes have some relevance for word order distribution, as both structures show the highest rate for OV order. This tendency is reversed only in the 14th century, where VO becomes dominant in the prepositional clause. In fact, prepositional clauses also show deviation in their rate of change, suggesting a contextual influence on word order (Figure 7.20). That is, their syntactic structure is more likely to have additional functional projections, in contrast to restructuring verbs with reduced syntactic structures. While restructuring verbs lead VO diffusion in both languages (Figure 6.14 and Figure 7.15), their statistical significance is only shown for Latin data. It is noticeable from Figure 7.15, however, that prepositional infinitival clauses not only increase their frequency across time, but also show a drastic increase in VO by the 14th century.

The third factor, *split NP*, shows a strong influence on word order alternation only in Latin. Furthermore, the data show that in Classical Latin and Late Imperial Latin discontinuous split nouns allow for both XVO and OVX orders, whereas in Late Latin split nouns are predominantly XVO. In contrast, in Old French, both orders are equally represented (see Figure 8.7). Immediately, it is clear that discontinuous nouns have reversed their direction in the 13-14th centuries:[1] i) Before the 13th century, they tend toward VO; ii) After the 12th century, they tend toward OV.

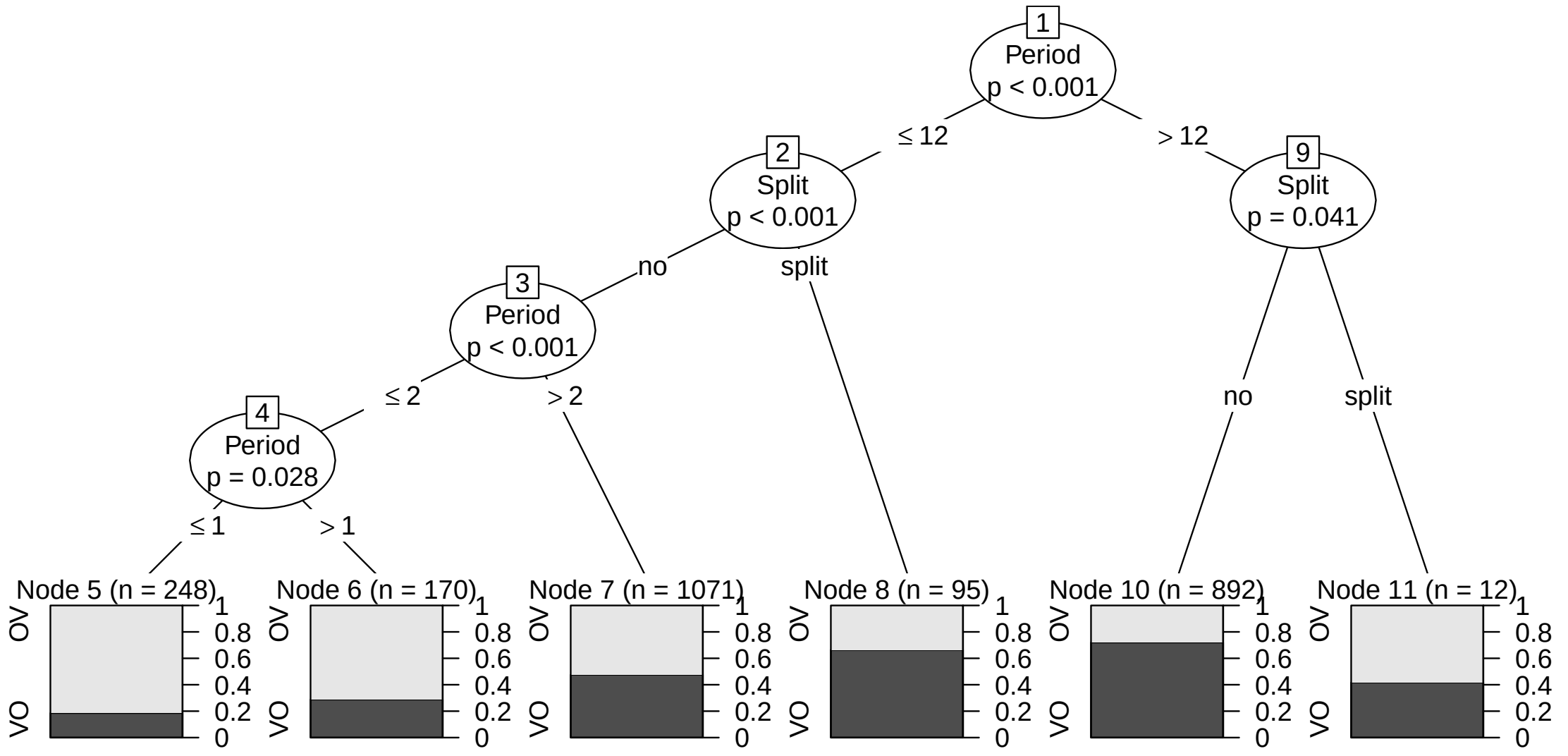

Figure 8.7: Split in Latin and Old French

The remaining factors do not show a strong significant influence on word order alternation. Figure 8.8 illustrates *intervening material* in both languages. The effect of intervening material is very small and shows a slight preference for OV in Classical/Imperial Latin.

[1]It should be noted that there are only 106 cases of split nouns in the corpus.

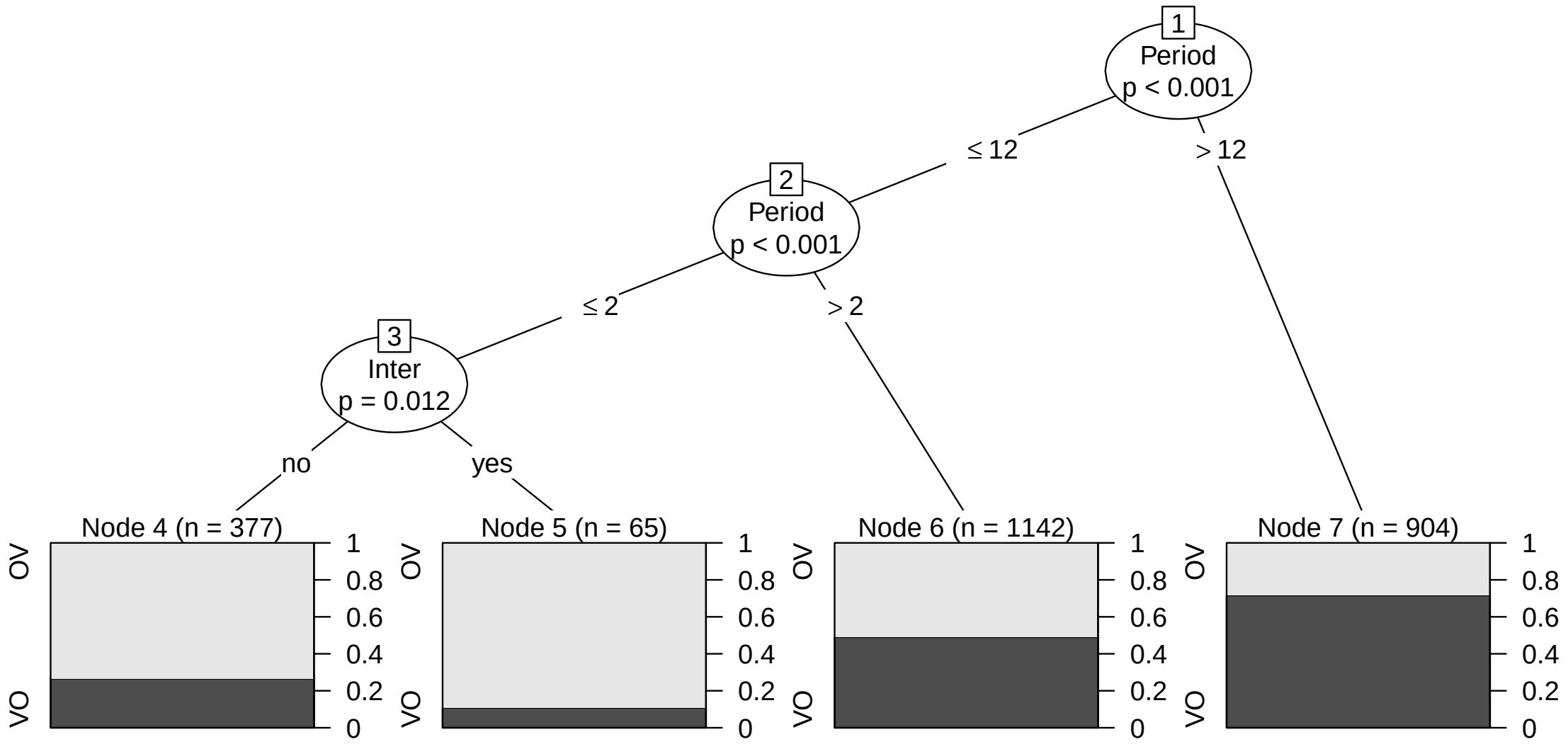

Figure 8.8: Intervening Material in Latin and Old French

Finally, presence of subject does not appear to have a significant role in word order alternation. Figure 8.9 confirms that the influence of subjects is very small, affecting only Classical Latin.

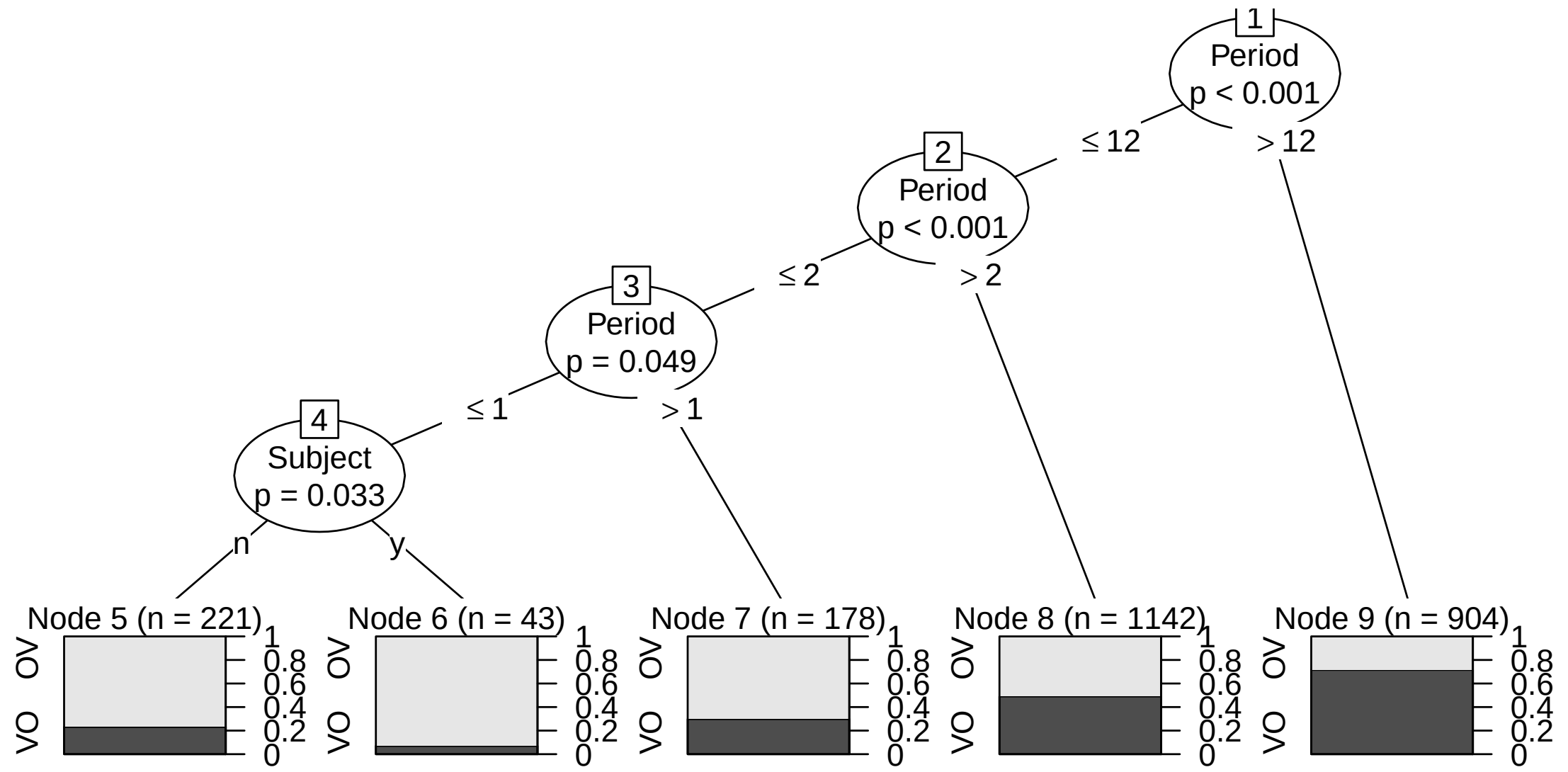

Figure 8.9: Subject in Latin and Old French

### 8.1.3 Pragmatic Factors

This study examines two pragmatic factors, namely *focus* and *old/new information.* Both factors are found to be strong predictors of word order alternation (Figure 6.19 and Figure 7.21). However, they exhibit different patterns. Contrastive focus predicts OV order in both Latin and Old French. Since the effect of focus is the same cross-linguistically, this indicates a stable word order alternation. In contrast, information status shows instability. In Classical and Imperial Latin, tokens with new/accessible information display a higher rate of OV than those with old information. In Late Latin, however, there is a change of direction, showing a high rate of VO for new information in comparison to old and accessible information. This fact suggests that this factor is not stable across time and strongly indicates word order change in Late Latin. That is, the word order change from OV to VO has occurred in Late Latin. Figure 8.10 presents a summary of these factors in one hierarchical sequencing. This view allows for the examination of chronological divisions for these factors.

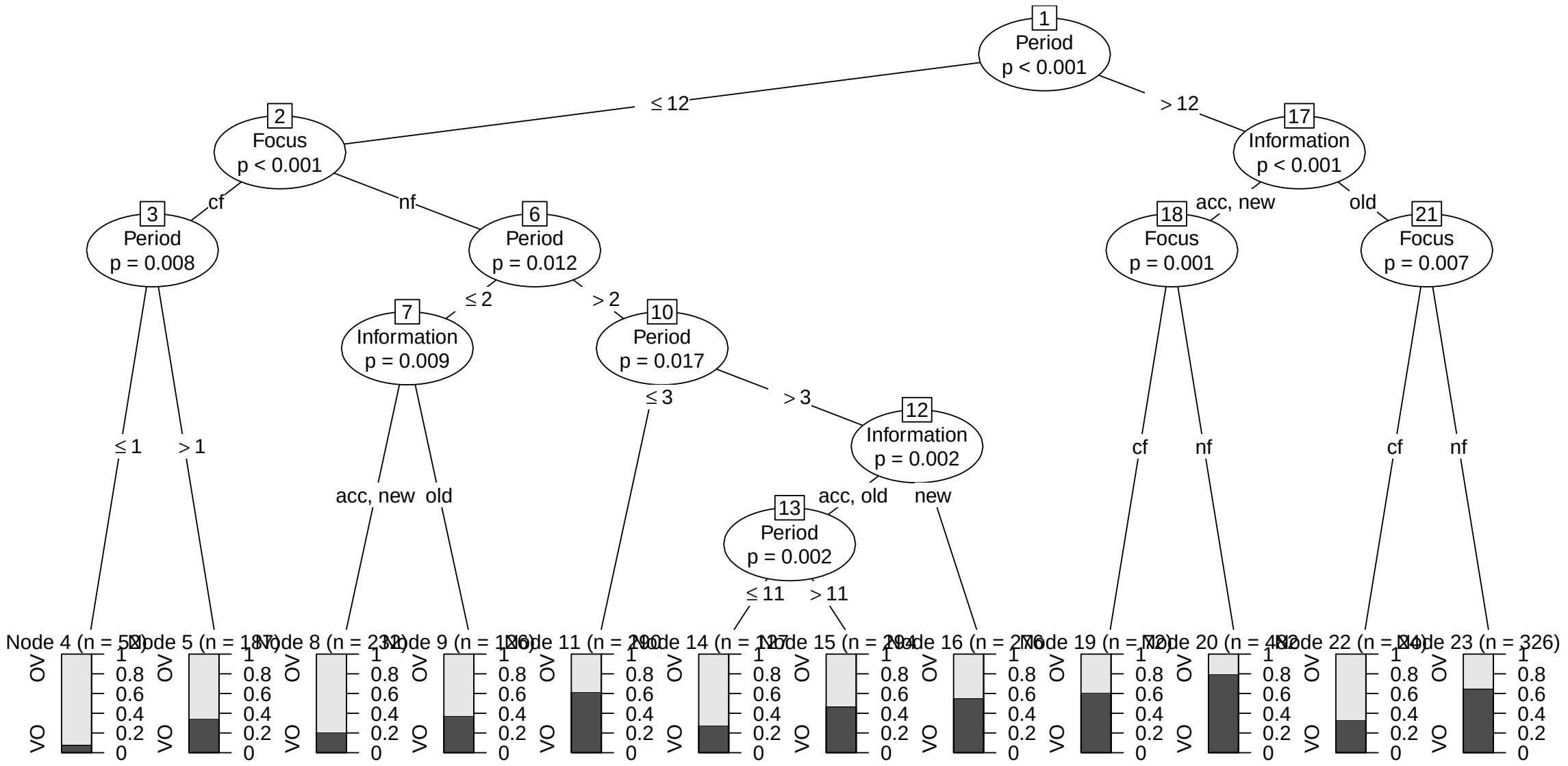

Figure 8.10: Pragmatic factors in Latin and Old French

From the inferential tree in Figure 8.10, it is obvious that the data are divided into two sections, namely before the 13th century and after the 13th century. Let us look at each side of the boundary in greater detail to determine which specific factor causes this change. First, before the 13th century, there is a clear split by focus. Recall that focus is shown to be a stable factor across time according to the statistical results. Contrastive focus displays a high rate of OV, whereas non-contrastive focus is further split between Classical/Imperial Latin and Late Latin.[2] It is noticeable that new and accessible information have the highest rate of OV as compared to old information. As mentioned earlier, this is an indication that the basic word order is OV. In Late Latin, there is another division between Early (4th century) and Late (6th century) Latin. In the 4th century, non-contrastive focus reaches about 60% of VO, suggesting that VO becomes a basic word order. In the 6th century, new information also represents about 60% of VO order, whereas old and accessible information show about 80% of OV order. However, there is an increase of VO to almost 50% in the 12th century. Thus, the data show a progressive change from new information being predominantly OV to a 50% variation by the 12th century. From the 13th century onward, VO is a dominant word order with new information, while focus still shows OV prevalence, specifically with contrastive tokens.

### 8.1.4 Semantic Factors

The semantic factor examined in this study is *animacy*. This variable does not show significant influence in fixed and mixed models, suggesting that there is no stable effect across time. However, the condition tree model in Figure 8.11 makes it possible to examine whether the influence of this factor has changed.

[2]Legend: 1 - Classical Latin, 2 - Imperial Latin, 3- Early Late Latin, 4 - Late Latin, cf - contrastive focus, nf - non-contrastive focus, acc - accessible information, new - new information and old - old information.

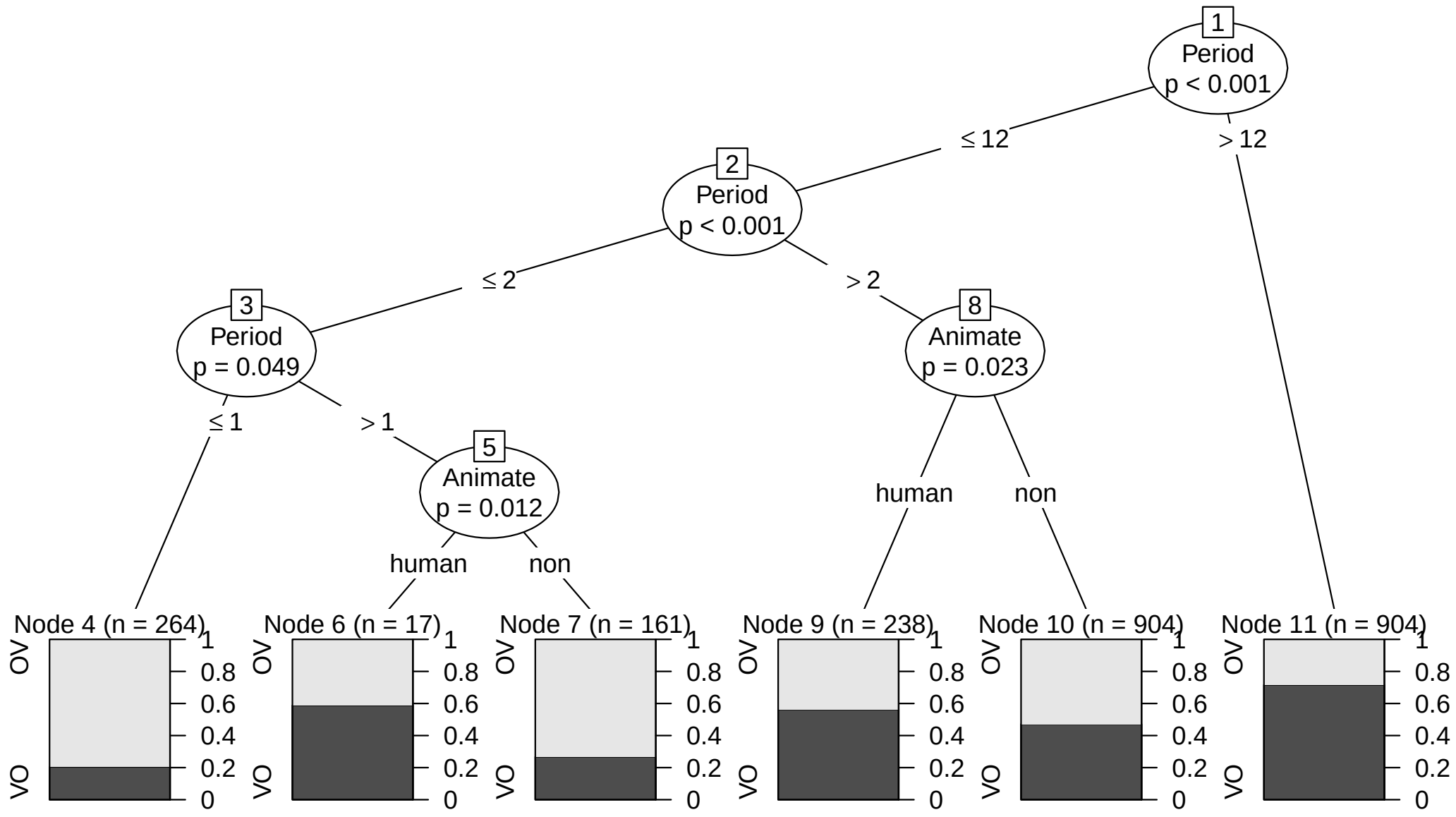

Figure 8.11: Animate NPs in Latin and Old French

Indeed, the change occurs in Late Latin. In Imperial Latin, nouns referring to humans show a high rate of OV (60%), whereas non-human referents represent only 20% of VO. Interestingly, in Late Latin, the rate of OV for animate nouns remains the same (60%), while the rate of VO for non-animate nouns increases to 50%. While these findings confirm the effect of animate nouns on word order, as suggested by Devine and Stephens (2006) with respect to Classical Latin, they also show that this effect seems to discontinue its influence in Late Latin.

### 8.1.5 Heaviness and Frequency

The present data suggest that lengthy constituents become significant only in the Old French period. That is, in Latin increases in length do not increase the probability of VO until Late Latin. Similarly, frequency is not identified as significant by Latin fixed models, whereas in Old French frequency has positive values, that is, the more frequent the verb, the higher the probability of VO order. This fact suggests that the effect of heaviness has changed over time. While it is traditionally assumed that heaviness triggers VO order, the present data demonstrate that this effect only concerns Late Latin and Old French. This supports the earlier assumption that Classical/Imperial Latin is an OV language.

### 8.1.6 Preliminary Summary

Having completed the assessment of conditioning factors for infinitival clauses as a whole, I summarize the findings in Table 8.3:

| Group | Variable | Period | OV | VO |
|---|---|---|---|---|
| Sociolinguistic | **Period** | | Classical/Imperial | Late Latin/ Old French |
| | **Metrics** | Late Latin | non-metric prose | metric prose |
| | | 10-12th centuries | verse | prose |
| | **Genre/Theme** | Latin | treatise, narrative (history, literature) | narrative (religion) |
| | **Genre** | Old French | treatise (50%) | narrative, speech, hagio (80%) |
| Syntactic | **Position** | Latin | preposed, independent | postposed (Late Latin) |
| | | Old French | preposed | postposed |
| | **Main Verb** | | all others | Restructuring |
| | **Split** | Latin | no | yes |
| | | Old French | yes | no |
| Pragmatic | **Focus** | Latin/Old French | contrastive | non-contrastive |
| | **Information** | Classical/Imperial | new (80%) | old (60%) |
| | | Late Latin - 11th | old (70%) | new (50%) |
| | | 12th century | old (50%) | new (50%) |
| | | 13-14th centuries | old (60%) | new (80%) |
| Semantic | **Animacy** | Imperial Latin | non-human (80%) | human (60%) |
| | | Late Latin | non-human (50%) | human (∼55%) |
| | **Length** | Old French | light | heavy |
| | **Frequency** | Old French | less frequent | frequent |

Table 8.3: Summary

A comparison of infinitival clauses in Latin and Old French demonstrates the extent to which linguistic and sociolinguistic factors are embedded into word order alternation. Given the assumption that Latin and Old French are just two stages of the same continuum, it should be possible to investigate language change from a different angle. That is, instead of identifying general tendencies for each separate language, the model will treat language as a binary variable.

## 8.2 Language Model

It is time to recall that our infinitival clauses are not uniform structures. I have mentioned earlier that restructuring verbs favor the VO order. It is traditionally assumed that these verbs form a mono-clausal structure with an infinitive. The raw frequencies from bi-clausal structures, e.g., prepositional and AcI, are more likely to be subjects of external distortion, stylistic or pragmatic. This distortion might also interact with the results of statistical analysis. By narrowing the investigation to restructuring verbs it is possible to also reduce ambiguity in linear word order patterns and obtain more accurate results. The results are illustrated in Figure 8.12.

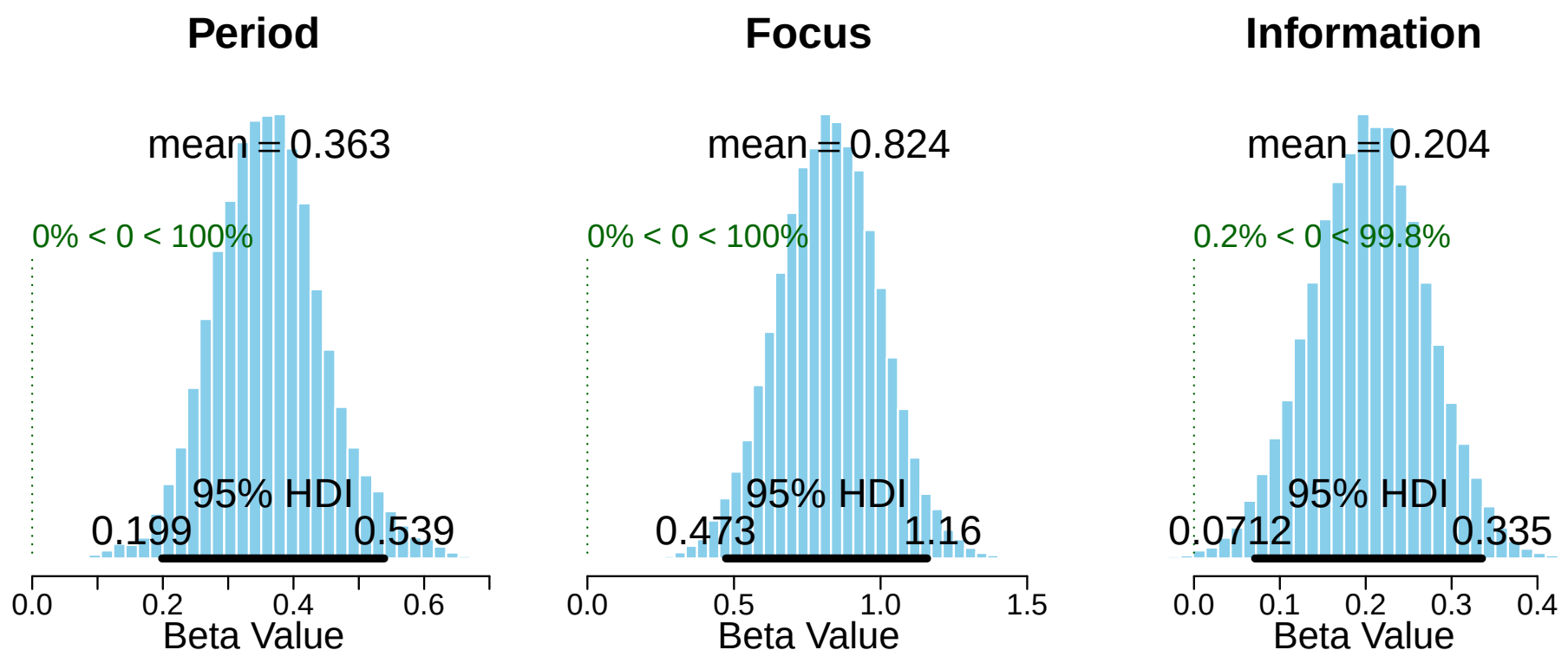

Figure 8.12: Posterior Distribution in Reduced Mono-clausal Model

From Figure 8.12, it is clear that pragmatic factors, namely given information and contrastive focus, are strongly associated with OV in mono-clausal restructuring infinitives. throughout the Latin and OF periods. These facts suggest that the preverbal position is more likely to host NPs with pragmatic features, such as contrastive focus and old information. In contrast, the postverbal position is associated with new non-contrastive information. If the assumption that new non-contrastive NPs reflect a basic word order is true, this study can claim that VO is a basic order. Figure 8.13 illustrates a conditional tree for a subset of restructuring verbs with *new non-contrastive information.* The predictors in this model are *Period* and *Language.*[3] First, the *Language* factor is not selected by this model, suggesting that there is a continuity in language change from Latin to Old French. Second, the *Period* factor is split into two segments: Classical/Late Imperial Latin and Late Latin/Old French. That is, the results suggest that Early Late Latin is the turning point for the change.

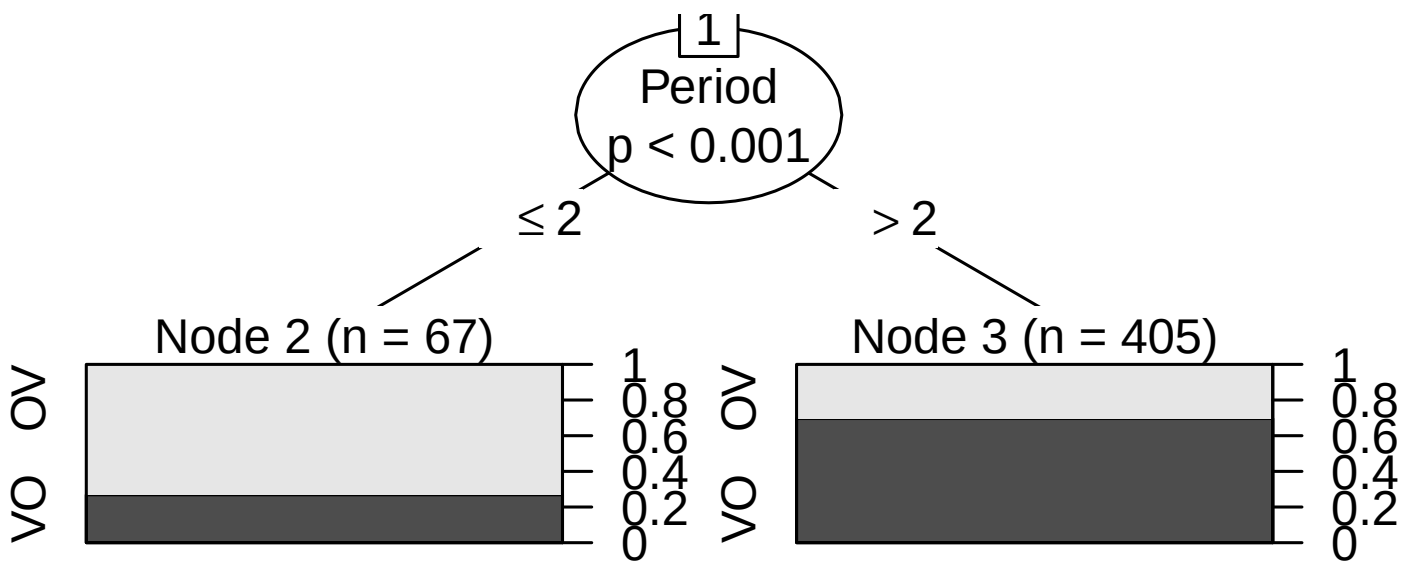

Figure 8.13: Restructuring Verbs with New Information in Infinitival Clauses

These findings also suggest that the VO change had already started in Latin. The evidence for this change was simply masked in finite clauses and bi-clausal infinitives (prepositional and AcI) through various stylistic factors, displaying OV order.

## 8.3 Summary

In this chapter, I have cross-evaluated the results from the previous chapters. While virtually every factor contributes to word order alternation, each varies in its significance. In addition, I have identified three general trends in variables, describing their impact and direction. Some factors are stable across time, and they play a part in a synchronic variation. For example, the effect of focus remains the same for Latin and Old French. Other factors are language-specific: lengthy constituents and frequency predict VO order in Old French, but in Latin they are not significant. Finally, the third group changes direction by switching

[3] mytree = ctree(OVVO Period + Language, data=restructuring) - R package “partykit”.

the value of prediction. For example, new information is OV in Classical Latin and VO in Late Latin. I have argued that these variables indicate a diachronic variation, namely a language change.

Furthermore, on the basis of the pragmatic factors cross-evaluated in this chapter, I have made an attempt to outline a chronological evolution of VO diffusion. Three benchmarks have been established: Classical/Imperial Latin (VO - 20%), Late Latin (VO - 50%) and 13th century (VO - 70%). In fact, one construction in the present study follows a constant increase in the VO probability - restructuring verbs. I have argued that examining this reduced construction in both languages as a continuum would reflect a more accurate image of language change. The pragmatic pattern identified through this model mirrors precisely the schema of VO: new information and non-contrastive nouns predict VO. That is, this structure illustrates a stable synchronic variation, suggesting that the actuation of this change has already occurred before Classical Latin and that the most favorable context for VO diffusion is through reduced mono-clausal structures, namely aspectual and functional verbs.

# Chapter 9

# Conclusions

*I have tried to show here the dramatic increase*
*in power and perception which the new tools of probability theory*
*can give to linguistic analysis.*
(Labov, 1975:30)

## 9.1 Aim and Contributions

The present thesis has dealt with the question of word order variation and word order change in the history of Latin and Old French. During the passage from Latin to Old French, word order undergoes a radical shift from Object-Verb to Verb-Object order. On the other hand, there is stable word order variation during each period. This variation often reflects stylistic, pragmatic or syntactic nuances imposed by author or by other language constraints. Thus, any given word order pattern may signal change or variation, making it difficult to investigate word order evolution. In this thesis, I have proposed a methodology that allows for the teasing apart of word order variation and word order change. First, I have used infinitival clauses as a device for minimizing ambiguity in linear word order strings. It is well known that main clauses illustrate great variation in word order patterns. However, these patterns do not always present textual clues, indicating their pragmatic or stylistic features. On the other hand, infinitival clauses are viewed as reduced clauses with fewer functional projections than main clauses (Cinque, 2004). Second, I have used *information* *structure* annotation as a device to determine a common, pragmatically neutral word order. Without a clear consensus on basic word order and its periodization in Latin and Old French, a historical linguist must rely on available textual clues. Using these clues, I have sought to investigate the following key aspects of the OV/VO change: its chronology, character, diffusion and constraints.

The second methodological contribution of this thesis regards the sociolinguistic and quantitative approaches to the study of word order. The first approach is the incorporation of multiple sociolinguistic and linguistic factors into Latin and Old French data. While such multi-factorial analyses are common in sociolinguistic studies, the application of this method to diachronic word order studies in Latin and Old French is new in comparison with traditional mono-factorial analyses. The second approach includes the introduction of advanced statistical tools into diachronic word order studies. These tools are capable of modeling, ranking and clustering multiple factors, allowing for a more subtle understanding of language variation. In addition, their visual representation makes it possible to identify patterns otherwise hidden in the traditional data representation.

Finally, this thesis has demonstrated the usability of corpus linguistic methods in diachronic studies. Syntactically annotated corpora offer an effective alternative way of collecting data as compared to traditional methods in historical linguistics, which consists of reading through written records and manually extracting linguistic data. Such annotated corpora make it possible to access and query data collection in Classical Latin, Late Latin and Old French. However, there exist certain limitations with existing annotated corpora: each corpus is limited to a certain chronological period and provides access to a limited number of textual materials. For example, Latin resources are represented by the following periods: 1st BC, 1st AD and 4th century, whereas Old French annotated resources cover texts starting from the 12th century. On the other hand, recent advances in computational linguistics offer state-of-the-art methods for creating additional annotated resources. By using pre-existing annotated resources, I was able to add additional corpora in Latin and Early Old French. In fact, the evaluation of annotation accuracy in Latin and Old French demonstrated that such methods are feasible in historical linguistics.

## 9.2 Syntactic Change

From the statistical estimates obtained from the data, the following picture of change has emerged: The change from OV to VO begins during the Classical period. Using *information structure* clues the following benchmarks have been established: 1) The initial stage - Classical/Imperial Latin (VO - 20%), 2) the transitional stage - Late Latin (VO - 50%) and 3) the final stage - the 13th-14th centuries (VO - 70%).

### 9.2.1 Stage 1

The initial phase in Classical Latin is OV order with a VO variant that occurs at a low frequency in limited contexts, specifically with *old information* nouns. The second category of information structure, namely *information relevance*, is not found to be significant. At this period, however, we have some sociolinguistic evidence suggesting the emergence of a new VO form. The data present evidence for a slightly higher VO rate in *letter* and *speech* genres that are traditionally considered less literary than, for example, philosophical treatises. Without a direct access to informal spoken Latin, these two genres allow to detect some language patterns that are closer to spoken language than other types of genre.

### 9.2.2 Stage 2

The second stage is Late Latin (4th-6th centuries). First of all, this stage predicts a different effect of information structure on word order variation: i) *Information relevance* becomes significant at this stage with *contrastive focus* for preverbal nouns and *information focus* for postverbal nouns and ii) *information status* displays an opposite tendency, where postverbal nouns are more likely to have *new information* and preverbal nouns are more likely to have *old information*, as compared to Classical Latin, which exhibits the reverse order. Second, new constraints emerge during this period. The data show the effect of heaviness on postverbal nouns. In addition, there is an interplay between syntactic factors and word order variation: i) *AcI*, *independent* and *preposed* infinitives are more likely to predict OV order and ii) *postposed* infinitives, *restructuring* and *other* verb types predict VO order. Furthermore, the close examination of these infinitival structures reveals an on-going change concerning AcI constructions, namely their decline. However, this change is found to be statistically independent from OV/VO change. If we compare the present data with Salvi's three-stage model, the Late Latin stage corresponds to the Phase I (Stage 2) (Salvi, 2005). Similar to that model, few infrequent verbs are found in the initial position in Classical Latin; similarly few perception and ditransitive verbs are found in Late Imperial Latin. In contrast, in Early Late Latin (4th century), there is a considerable expansion of various verbs into the initial position, including very frequent verbs. Early Old French continues this transitional OV/VO stage with a VO rate no higher than about 50% of cases. It displays effects similar to Late Latin: i) *Heavy* nouns favor postverbal position, ii) *old information* favors OV order and iii) restructuring verbs favor postverbal nouns. On the other hand, there are certain differences between the transitional stage in Late Latin and the transitional stage in Early Old French. These dissimilarities concern syntactic and pragmatic factors. On the syntactic level, the data show that AcI constructions have disappeared in Old French. In addition, while preposed infinitives remain, they almost exclusively display preverbal nouns in Old French. Finally, there is an emergence of a new construction, namely a prepositional clause, which favors OV order. Statistically, there is no correlation between this on-going change in infinitival structure and OV/VO change. On the information structure level, the difference concerns the ratio between *old information* and *new information.* While in Late Latin, this ratio reaches 1:3 (old preverbal nouns versus new postverbal nouns), in Early Old French this ratio is about 1:1.3 (old preverbal nouns versus new postverbal nouns).

### 9.2.3 Stage 3

The final VO stage is the period of Old French in the 13th-14th centuries. At this stage the effect of *information relevance* is found to be significant with *contrastive focus* on preverbal nouns and *information focus* on postverbal nouns. *Restructuring* verbs and *new information* continue to predict VO order, whereas *prepositional* verbs and *old information* are more likely to have OV order.

### 9.2.4 Summary

Thus, in the passage from Latin to Old French, word order undergoes certain changes related to information structure and syntactic constraints. While the aforementioned parallel syntactic changes in infinitival clauses, namely the decline of AcI, the rise of prepositional clauses and the decline of preposed clauses (and their archaic OV character), are not directly related to OV/VO *change*, these changes play an important role in word order *variation*, as they influence OV order. Second, there are several important findings with respect to information structure effects. First, the non-significance of *contrastive focus* and *information focus* for word order variation in Classical/Imperial Latin suggests that both types of foci are preverbal. While they may have a different syntactic position in the sentence, e.g. *FocP* for *contrastive focus* and *FocVP* for *information focus* (see Figure 3.2), their linear order remains the same - OV. Second, in Stage 2 two facts point to a structural change: i) the significance of *information relevance* for word order variation and ii) the reverse effect of *information status*. From the 4th century onward *contrastive focus* favors preverbal objects, whereas *information focus* favors postverbal objects. That is, these two foci now map to two different positions with respect to the verb - *FocVP* for *contrastive focus* must be preverbal and *FocVP* for *new information focus* must be postverbal. This fact supports earlier statements that there is a process of verb-raising (Ledgeway, 2012b; Salvi, 2005). Similarly, this change affects *information status*. First, in Classical Latin, there is only a small number of postverbal *old* nouns, which are more likely to correspond to what Devine and Stephens (2006) call *tail* nouns, namely nouns that usually serve to refresh a hearer's memory about some old or inferable referent. The majority of *old/new information* nouns are preverbal. In contrast, in Late Latin, there is a clear distinction between postverbal *new* nouns and preverbal *old* nouns. If, following the convention from Devine and Stephens (2006), we represent *old information* as *TopVP* and *new information* as *FocVP*,[1] we will see that *TopVP* remains on the left but *FocVP* is on the right of the verb. Once again, this fact suggests a process of verb-raising in Late Latin. Since Early Old French follows the same mapping patterns between information structure and syntactic structure, we can assume that the same process is in place, namely verb-raising. However, there is a small but noticeable difference between the rates of information status categories: in Early Old French postverbal nouns display a higher rate of *old* information; however by the 12th century the difference in rates between *old* and new information becomes smaller: the rate of *old* information slightly decreases, while the rate of *new* information slightly increases (see Figure 7.7). The examination of infinitival clauses reveals that this small change occurs with preverbal nouns in prepositional structures in the 13th century (see Figure 7.16) and restructuring verbs in the 13th century (see Figure 7.18).

## 9.3 A Merger between Historical Linguistics, Corpus Linguistics and Quantitative Data Analysis

The recent surge in digital collections, computational linguistic methods and data mining applications has inevitably led to a change in the way we engage in research analysis. Digital collections are replacing library catalogues, the automatization of data retrieval is allowing for increased efficiency, and advanced statistical tools are becoming a common methodological instrument in various linguistic disciplines, e.g. sociolinguistics and psycholinguistics. These advancements, however, have not been fully recognized in historical linguistics. First, working with paper-based manuscripts has been a long tradition in historical linguistics. Second, while many historical collections have been recently digitized, they are mostly available in a scanned image format. As a result, manual data extraction remains a dominant methodological tool in diachronic studies. Due to such extensive manual work, language change studies have often been based on the examination of only a few texts, and comparative analysis has been usually limited to a few centuries. Consequently, the use of robust quantitative analysis has not been fully exploited in diachronic research. Some of these challenges, however, can be solved by an interdisciplinary merger of historical linguistics with corpus linguistics, sociolinguistics and digital humanities. Such a merger allows for collaborative sharing of tools and methods. For example, digital humanities can offer a number of visualization tools and state-of-the-art methods for text digitization and codification, whereas computational linguistics provides methods for automatic corpora creation. Sociolinguistics and psycholinguistics can supply us with advanced statistical toolkits, while corpus linguistics introduces efficient methods for data query and retrieval. Furthermore, introducing computational linguistics methods into a field traditionally related to humanities provides an opportunity to move research in historical linguistics towards new approaches, where automatic annotation and linguistic knowledge can help accessing larger datasets, which will allow for new insights. There have already been several successful attempts to bridge the gap between historical linguistics, corpus linguistics and digital humanities (see Section 1.4 and Section 3.2). The present thesis has also contributed to Digital Humanities by merging statistics, corpus linguistics and computational linguistics. Finally, such cross-disciplinary methods have made it possible to look at language change from a different angle, which would not have been possible with traditional methods of historical linguistics.

[1]This representation coincides with Information Focus.

## 9.4 Conclusion

The findings in this thesis make clear that word order change and word order variation are just two dimensions of language variation: word order change is a diachronic variation, and word order variation is a synchronic variation. Furthermore, it has been shown that word order is a complex phenomenon that incorporates many levels of language, e.g. pragmatic, syntactic and sociolinguistic. In addition, this study has shown that the comparative investigation of Latin and Old French provides a valuable source for learning about language change. While Old French and early Old Occitan are the only Romance languages treated here, the same methodology can be applied to other Romance languages, thus contributing to Latin and Romance Linguistics. Finally, this thesis has shown that historical linguistics can benefit from corpus linguistic and computational linguistic methods. Computational linguistics provides models for creating additional annotated corpora that enable a researcher to conduct investigations by using effective corpus linguistic methods.

# Appendices

## Appendix A Latin Verbs

Glosses: Initial Verbs in Latin Infinitival Clauses with Light NPs (VO, VOX and VXO)

| Period | Verb Types |
|---|---|
| Classical Latin | spargere (scatter) parare (prepare) exsugere (dry) |
| Late Imperial | accipere (take)<br>coniungere (connect) effundere (pour, loosen) operire (cover)<br>extrahere (remove) numerare (count) ornare (adorn)<br>relinquere (abandon) impertire (bestow) exprobare (reproach)<br>perspicere (examine) |
| Early Late | adhibere (summon) prodere (create) referre (bring) facere (make)<br>figere (establish) imponere (impose) accipere (take) conducere (draw)<br>diffamare (slander) diligere (select) dimittere (send) eicere (perform)<br>ferre (bring) habere (have) haurire (draw) laudare (praise)<br>mittere (send) parare (prepare) percutere (strike) praedicare (declare)<br>reddere (return) respondere (reply) sanare (cure) separare (separate)<br>sustinere (support) temptare (test) thesaurizare (collect) videre (see)<br>manducare (eat) dare (give) subire (move) vocare (call) |
| Late Latin | concedere (concede) confluere (assemble) definire (mark) facere (make)<br>movere (move) necare (kill) occulere (cover) praestare (keep)<br>replere (complete) retinere (restrain) scire (know) **scribere** (write)<br>praeterire (neglect) soluere (free) spernere (despise) |

## Appendix B List of Latin Texts (Chronological Order)

## Appendix C List of Old French Texts (Chronological Order)

**La Cantilène de Sainte Eulalia**. Altfranzösisches Übungsbuch, ed. W. Braune and K. Helm, Halle, 1952.

**Saint Léger**. Etude de la langue du manuscrit de Clermont-Ferrand suivie d'une édition critique du texte avec commentaire et glossaire, éd. par Joseph Linskill, Paris : Droz, 1937.

**Evangile de Saint Jean**. C. Hofmann dans Gelehrte Anzeigen des K. Bayer. Academie des Wissenschaften, 369-376. In Chrestomathie provençale accompagnée d'une grammaire et d'un glossaire, éd. Karl Bartsch, Berlin: Wiegandt & Schotte, 1892.

**Deux Sermons**. Jarhbuch für Romanische und Englishe Literatur VII, 81-84, publié par M. Paul Meyer. In Chrestomathie provençale accompagnée d'une grammaire et d'un glossaire, éd. Karl Bartsch, Berlin: Wiegandt & Schotte, 1892.

**Le martyre de Saint Etienne**. Raynouard, choix des poésies originales des troubadours, tome II, 146-151. In Chrestomathie provençale accompagnée d'une grammaire et d'un glossaire, éd. Karl Bartsch, Berlin: Wiegandt & Schotte, 1892.

**Poëme sur Boece**. Sprachdenkmale berichtigt und erklärt von Friedrich Diez, Bonn 1846, 39-72. In Chrestomathie provençale accompagnée d'une grammaire et d'un glossaire, éd. Karl Bartsch, Berlin: Wiegandt & Schotte, 1892.

**La Chanson de Sainte Foi d'Agen** : poème provençal du XIe siècle, édition d'après le manuscrit de Leyde avec fac-similé, traduction, notes et glossaire par Antoine Thomas, Champion, 1925 ; réédition 1974.

**Gormond et Isembart**. Reproduction photocopique du manuscrit unique, II. 181 de la Bibliothèque royale de Belgique avec une transcription littérale par Alphonse Bayot. Brixelles : Misch & Chron. 1906.

**La vie de saint Alexis**, éd. par Christopher Storey, Genève : Droz, 1968.

**Die Passion Christi**. Altfranzösisches Übungsbuch, ed. W. Förster and E. Koschwitz, Heilbronn: Henninger, 1884.

**La Chanson de Roland**, éd. par Grard Moignet, texte établi d'après le manuscrit d'Oxford ; traduction, notes et commentaires, 3e édition revue et corrigée, Paris : Bordas (Bibliothèque Bordas), 1969, MCVF.

**Le Roman de Tristan**, Béroul, éd. E. Muret. Paris, 1913, NCA.

**Chrétien de Troyes**, Le Chevalier au lion (Yvain), éd. par Mario Roques, Paris : Champion, 1960, NCA.

Marie de France. **Les Lais**, éd. par Jean Rychner, Paris : Honoré Champion, 1981 [1971], MCVF.

**La Queste del Saint Graal**, roman en prose du XIIIe siècle, éd. par Albert Pauphilet, Paris : Champion, 1923, MCVF.

**Le Livre Roisin** : coutumier lillois de la fin du 13e siècle, publié avec une introduction et un glossaire par Raymond Monier, préface d'Alexandre de Saint-Léger, Paris : Domat-Montchrestien, 1932, MCVF.

**Joinville**, Jean sire de, Vie de saint Louis, éd. bilingue par Jacques Monfrin, Paris : Dunod (Classiques Garnier), 1995, MCVF.

# OLGA SCRIVNER

## PERSONAL INFORMATION

*email* obscrivn@indiana.edu

*website* http://nlp.indiana.edu/~obscrivn/

## EDUCATION

*PhD* — *2015 - Indiana University*
Dual PhD in French Linguistics and Computational Linguistics

*MA* — *2009 - Indiana University* French Linguistics

*MA* — *1998 - State University of St-Petersburg, Russia*
French Literature and Language

## WORK EXPERIENCE

*Indiana University* — *2007–2015* *Associate Instructor*
Taught French for undergraduate and graduate students; taught Spanish for undergraduate students; led discussion for Introduction to the Study of Language; designed several promotional videos for French Courses and French Conversation Club; managed clinical scheduling and Web page for EMT-basic course

*Honors Program* — *Summer 2009* *Instructor and Financial Coordinator*
Managed daily financing and scheduling; developed grammar syllabus; supervised and advised high school students in Foreign Language Program in France

*Norton Healthcare* — *2004-2007* *Contract Interpreter — Louisville, KY*
Assisted with oral and written communication in medical and court settings

*Kosair Children Hospital* — *2004-2007* *Volunteer — Louisville, KY*
Helped to improve relationship between patients and healthcare providers; provided translation assistance to parents

## PUBLICATIONS

*2015* — *Tools for Digital Humanities: Enabling Access to the Old Occitan Romance of Flamenca.* In Proceedings of the Fourth Workshop on Computational Linguistics for Literature (with Sandra Kübler)

*2014* — *Metrical Annotation For a Verse Treebank*. In Proceeding of The 13th International Workshop on Treebanks and Linguistic Theories (TLT13) (with T.M. Rainsford)

*2014* — *Vowel Variation in the Context of /s/: A Study of a Caracas Corpus*. New Directions in Hispanic Linguistics, ed. Rafael Orozco

*2013* — *Le Roman de Flamenca: An Annotated Corpus of Old Occitan*. In Proceedings of The Third Workshop on Annotation of Corpora for Research in the Humanities (with Sandra Kübler, Eric Beurlein, Barbara Vance)

*2013* — *SWIFT Aligner: A Tool for the Visualization and Correction of Word Alignment and for Cross Language Transfer*. In Proceedings of Corpus-Linguistics 2013, Saint-Petersburg, Russia (with Tim Gilmanov)

*2012* — *Building an Old Occitan Corpus via Cross-Language Transfer*. In Proceedings of KONVENS, Austria (with Sandra Kübler)

## SKILLS AND CERTIFICATIONS

*Computer* — JAVA, PYTHON, HTML, XML, LaTeX, Unix

*Technology* — Audio and Video Editing, Game developing, Web Publishing

*Certificates* — EMT-Basic certification, CPR Healthcare provider

## AWARDS

*2015* — Travel grant (Montclair State University)

*2015* — Conference Travel Grant (Linguistics Department Indiana University)

*2014/15* — HASTAC Scholar (Indiana University)

*2014* — Recognized with Distinction Associate Instructor in Spanish

*2013* — Graduate Student Conference Travel Award (Indiana University)

*2011* — Grace P.Young Graduate Award - Excellent Achievement in French Studies

*2010* — Teaching Award - Outstanding Performance as Associate Instructor in French

*2005* — President Volunteer Award (Kosair Children's Hospital)